\pdfoutput=1 

\documentclass[twocolumn]{article}

\usepackage[export]{adjustbox} 
\usepackage{array} 
\usepackage{amsmath}
\usepackage{amssymb}
\usepackage{booktabs}
\usepackage{caption}
\usepackage{csquotes} 
\usepackage{graphicx}
\usepackage[greek, british]{babel}
\usepackage{makecell}
\usepackage{multirow} 
\usepackage{rotating} 
\usepackage{siunitx}
\usepackage{subcaption}
\usepackage{tikz}
\usepackage{xcolor}

\definecolor{fhgDark}{RGB}{17,117,94}
\definecolor{fhg}{RGB}{23,156,125}
\definecolor{fhgLight}{RGB}{82,229,195}
\definecolor{fhgBright}{RGB}{197,246,235}
\usepackage[colorlinks=true, pdfstartview=FitV, allcolors=fhg, pagebackref=true]{hyperref}
\usepackage{cleveref} 

\usetikzlibrary{intersections,3d,decorations.text,shapes.arrows,positioning,fit,backgrounds,trees,positioning,fit,arrows,decorations.pathreplacing, calc,math,through,arrows.meta }
\definecolor{tikzBoxColourTruePositive}{RGB}{90,105,0}
\definecolor{tikzBoxColourFalsePositive}{RGB}{214,122,62}
\definecolor{tikzBoxColourFalseNegative}{RGB}{117,147,176}
\newcommand{\tikzBoxSize}{0.15cm}
\DeclareRobustCommand\tikzBoxTruePositive{\tikz\node[rectangle,fill=tikzBoxColourTruePositive,minimum width=\tikzBoxSize,minimum height = \tikzBoxSize,] (r) at (0,0) {};}
\DeclareRobustCommand\tikzBoxFalsePositive{\tikz\node[rectangle,fill=tikzBoxColourFalsePositive,minimum width=\tikzBoxSize,minimum height = \tikzBoxSize,] (r) at (0,0) {};}
\DeclareRobustCommand\tikzBoxFalseNegative{\tikz\node[rectangle,fill=tikzBoxColourFalseNegative,minimum width=\tikzBoxSize,minimum height = \tikzBoxSize,] (r) at (0,0) {};}

\usepackage{enumitem}
\usepackage[colorinlistoftodos]{todonotes}
\usepackage{comment}

\usepackage{mwe} 

\DeclareRobustCommand{\fineTunedModel}{vit\_b\textsubscript{\MakeUppercase{cvb} }}

\title{Adapting SAM for Volumetric X-Ray Data-sets of Arbitrary Sizes}
\author{
	\href{https://orcid.org/0000-0002-3429-732X}{\includegraphics[scale=0.06]{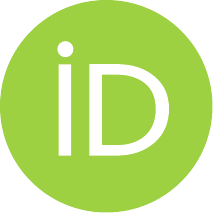}}Roland Gruber$^{1,2}$
	Steffen Rüger$^{1}$
	\href{https://orcid.org/0000-0003-0840-8695}{\includegraphics[scale=0.06]{./data/full/orcid.pdf}}Thomas Wittenberg$^{1,2}$\\ 
	\\
	$^{1}$ \small Fraunhofer IIS, Fraunhofer Institute for Integrated Circuits IIS,  \\ \small Division Development Center X-Ray Technology, Fürth, Germany \\
	$^{2}$ \small Friedrich-Alexander-Universität Erlangen-Nürnberg, \\ \small Chair for Visual Computing, Erlangen, Germany \\
}

\date{2024-02-09}

\begin{document}
	\maketitle
	
	\begin{abstract}	
		Objective: We propose a new approach for volumetric instance segmentation in X-ray Computed Tomography (CT) data for Non-Destructive Testing (NDT) by combining the Segment Anything Model (SAM) with tile-based Flood Filling Networks (FFN). Our work evaluates the performance of SAM on volumetric NDT data-sets and demonstrates its effectiveness to segment instances in challenging imaging scenarios. 
		Methods: We implemented and evaluated techniques to extend the image-based SAM algorithm fo the use with volumetric data-sets, enabling the segmentation of three-dimensional objects using FFN's spatially adaptability. The tile-based approach for SAM leverages FFN's capabilities to segment objects of any size. We also explore the use of dense prompts to guide SAM in combining segmented tiles for improved segmentation accuracy. 
		Results: Our research indicates the potential of combining SAM with FFN for volumetric instance segmentation tasks, particularly in NDT scenarios and segmenting large entities and objects. 
		Conclusion: While acknowledging remaining limitations, our study provides insights and establishes a foundation for advancements in instance segmentation in NDT scenarios.	
	\end{abstract}
	
	\section{Introduction}
	
		In the field of Non-Destructive Testing (NDT), of large scale components and assemblies, such as cars \cite{IntroCar}, shipping containers \cite{IntroContainerMobile, IntroContainerHigh}, or even airplanes \cite{Gruber2020, Gruber2022, fraunhoferMe163InstanceSegmentationDataset, Gruber2024Challenge} are often captured using large scale 3D X-ray computed tomography (CT) and are subsequently subjected to automated analysis and evaluation. In this context an important step of the analysis process consists of instance segmentation, where an attempt is made to assign a unique semantic identifier or label to each entity in a data-set. For example, all voxels belonging to a specific screw are hereby assigned the same unique identifier, while voxels belonging to another component are assigned a different unique identifier. 
				
\begin{figure*}
			\centering		
						
			\begin{subfigure}[b]{0,48\textwidth}
				\centering
				\includegraphics[width=\linewidth, keepaspectratio]{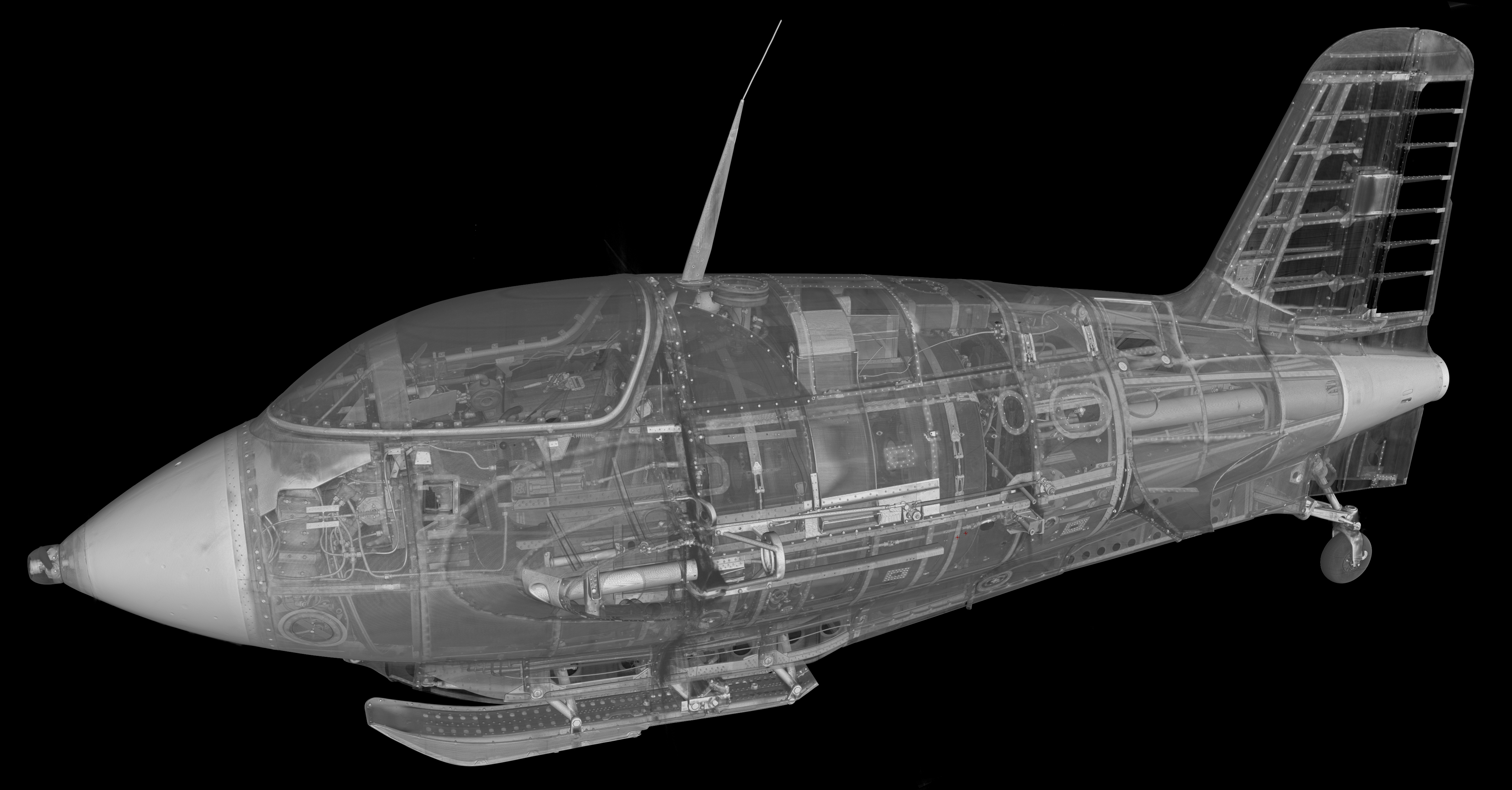} 
				\caption{Reconstruction}
				\label{fig:introduction-task-reko}
			\end{subfigure}~	
			\begin{subfigure}[b]{0.25\textwidth}
				\centering
				\includegraphics[width=\linewidth, keepaspectratio]{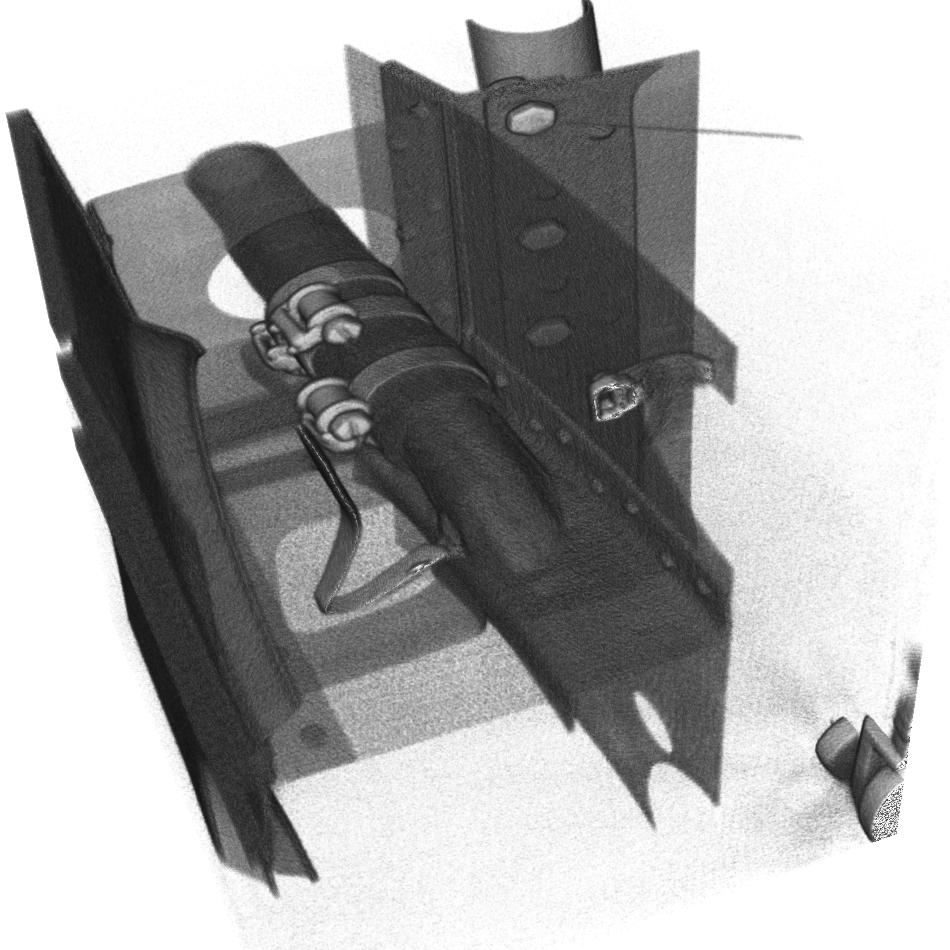}
				\caption{Input sub-volume}
			\label{fig:introduction-task-subvolume}
			\end{subfigure}
			\begin{subfigure}[b]{0.25\textwidth}
				\centering
				\includegraphics[width=\linewidth, keepaspectratio]{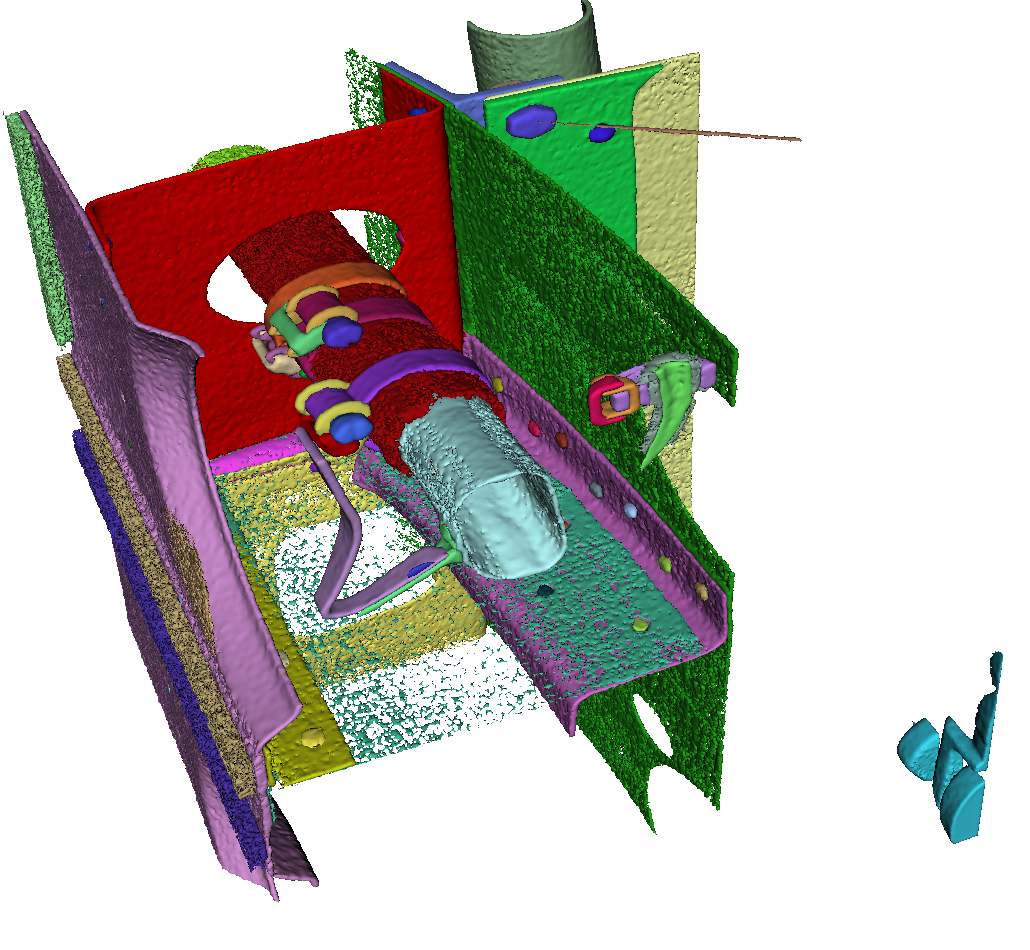}
				\caption{Reference sub-volume }
				\label{fig:introduction-task-segmentation}
			\end{subfigure}
		\caption{Rendered example of instance segmentation (Figure \ref{fig:introduction-task-segmentation}) of a sub-volume of size $512\times512\times512$\,voxels (Figure \ref{fig:introduction-task-subvolume}) from the XXL-CT Me\,163 data-set with a data resolution of $\num{10000}\times\num{10000}\times\num{8000}$\,voxels (Figure \ref{fig:introduction-task-reko}).}
		\label{fig:introduction-task}
		\end{figure*}

		The process shown in Figure \ref{fig:introduction-task} exemplifies this attempt using an XXL-CT data-set of a historic aircraft \cite{Gruber2022}. It begins with acquisition of data from the specimen, in this case the airplane, and proceeds with the reconstruction of a volumetric voxel data-set (Figure \ref{fig:introduction-task-reko}). Figure \ref{fig:introduction-task-subvolume} and Figure \ref{fig:introduction-task-segmentation} show a sub-volume of size $512 \times 512 \times 512$\,voxels of the reconstruction and the instance segmentation. In Figure  \ref{fig:introduction-task-segmentation} each semantic entity within the sub-volume has been assigned a unique identifier. The classes of these entities (primarily screws and metal plates) are not considered, as the classification of the entities is not performed and is focus of future work.
		
		Instance segmentation is essential for automated image processing and data exploration in NDT and medical \cite{Thomas4} applications. By segmenting a large scale volumetric image data-set into its semantic instances, it becomes easier to extract valuable information and to analyze complex component geometries. This is particularly important in cases where the data-set contains various acquisition and reconstruction  that can make interpretation difficult for both experts and non-experts.
		
		Instance segmentation is a critical task in computer vision, leading to the proposal and development of numerous methods that leverage both classical image processing and neural networks. These approaches, however, are not without their limitations. Some methods necessitate manual intervention and corrections \cite{Seg2Link, NeuTu}, and others are tailored specifically to predefined component classes \cite{x76}. Challenges associated with data quality, particularly in data-sets with a high incidence of artefacts, can significantly hinder the effectiveness of segmentation algorithms \cite{Gruber2024Challenge},

		\subsection{Segment Anything Model}\label{sec:introdution-SAM}
			The Segment Anything Model (SAM) \cite{72} is an instance segmentation model based on the vision transformer architecture \cite{x61}. It is an advanced model for segmenting arbitrary entities out of photographs. It stands out primarily for its high quality, robustness, and minimal required user input. One of its notable features is the ability to be queried using a variety of prompts, allowing it to segment a RGB input image with a spatial resolution up to $1024 \times 1024$ pixels into multiple segments in one inference call. SAM supports prompts in various forms, such as seed points (point prompts), bounding boxes, brush masks (dense prompts) and text prompts.
			
			Furthermore, SAM allows the generation of multiple output masks for each input prompt hence enabling image segmentations at varying hierarchical levels of granularity. Another advancement presented by the SAM is the extensive training data-set SA-1B, which has been iteratively collected and refined through prior versions of SAM during its own training process.		

		\subsection{Combination with Tile-based FFN}\label{sec:introduction-FFN}
			This work aims to evaluate the applicability of SAM for segmenting volumetric NDT data-sets and to examine its potential enhancement through the integration of Flood Filling Networks (FFN), initially proposed by Januszewski et al. \cite{14-Januszewski2018}. FFNs are instance segmentation methods originally based on convolutional networks \cite{2-1-72-Zeiler2013-LeCun1989, 2-1-72-Zeiler2013-LeCun1989-LeCun1989}, which are able to segment arbitrarily large data-sets based on tiles. Originally FFN have been developed for the  segmentation of organic objects, but have in the past been extended to other applications, including the delineation of large scale XXL-CT data \cite{Gruber2020}.
									
			The FFN approach maintains the current state of segmentation within an accumulator volume, which is sized to match the dimensions of the input volume. During each segmentation step, a sub-volume or tile of the input volume and the corresponding partially computed tile of the accumulator are passed to the model (in our case, a volumetric variant of SAM). The segmentation proposal of the tile is then updated and written back to the corresponding tile position within the accumulator. 
			
			Candidates for neighbouring tile positions with significant overlap, which could extend the current segment, are determined using the updated accumulator state and added to a queue of tiles pending processing. In the subsequent iteration, the next unprocessed tile is removed from the front of the queue for processing. Starting from a seed point, the FFN then processes all tiles which potentially belong to the current segment. The processing of the current segment is completed when the queue of potentially belonging tiles is depleted. The algorithm then proceeds with the next segment starting from another seed point. 

			The seed points of the segments can be manually specified or computed automatically by a reasonable algorithm.
			
		\subsection{Contributions}
			In this work, we propose a novel approach for volumetric instance segmentation in NDT by combining SAM with FFN. Our contributions include:			
			
			\begin{description}[style=nextline, leftmargin=0pt, labelwidth=0pt, wide] 
				\item [{Evaluation of SAM on NDT data-sets:\hfill}] We assess the performance of SAM on non-destructive testing data-sets and demonstrate its effectiveness in accurately segmenting instances in challenging CT imaging scenarios.
				
				\item [{Implementation and evaluation of various methods to combine image-based SAM for the application with volumetric data-sets:}] We implement and evaluate different techniques to integrate output of the image-based SAM approach for the application of volumetric data-sets, enabling the segmentation of three-dimensional objects using FFN's spatially adaptive capabilities.
				
				\item [{Extending SAM for objects of arbitrary size through tile-based approaches:}] We propose a tile-based approach that leverages FFN's capabilities to segment objects of arbitrarily size. By initially dividing the input volumes into tiles and then applying SAM on each tile individually, we achieve accurate and efficient segmentation results for objects of any size.
				
				\item [{Utilizing dense prompts for SAM to combine tiles in an accumulator:}] To further improve the accuracy of the proposed tiled-based approach of SAM, we use dense prompts to guide SAM in combining the segmented tiles into a cohesive instance segmentation result. By leveraging the accumulated information from neighbouring tiles, we try to achieve more robust and accurate instance segmentation results.
			\end{description}
			
	\section{Methods}
		This section presents the methodology and the experimental setup used, including the introduction of the data-sets (Section \ref{sec:methods-datasets}) used for the evaluation of the proposed methods. We furthermore describe a technique to improve the image segmentation performance of SAM with respect to the Me\,163 airplane XXL-CT data-set by fine-tuning it specifically for this task (Section \ref{sec:methods-fineTune}). Additionally, we detail our inference workflow in Section \ref{sec:methods-inference}, which adapts the top-performing SAM model for volumetric data-sets. This process includes tile-based segmentation, accumulator-based dense prompts, and postprocessing. The workflow aims to integrate the best model into a cohesive volumetric inference approach.
			
		\subsection{Data-sets and Data Processing}\label{sec:methods-datasets}
\begin{figure}
				\centering		
				
				Me\,163
				\vspace{0.5\baselineskip}
				
				\begin{subfigure}[b]{0.3\columnwidth}
					\centering
					\includegraphics[height=68px, width=\linewidth, keepaspectratio]{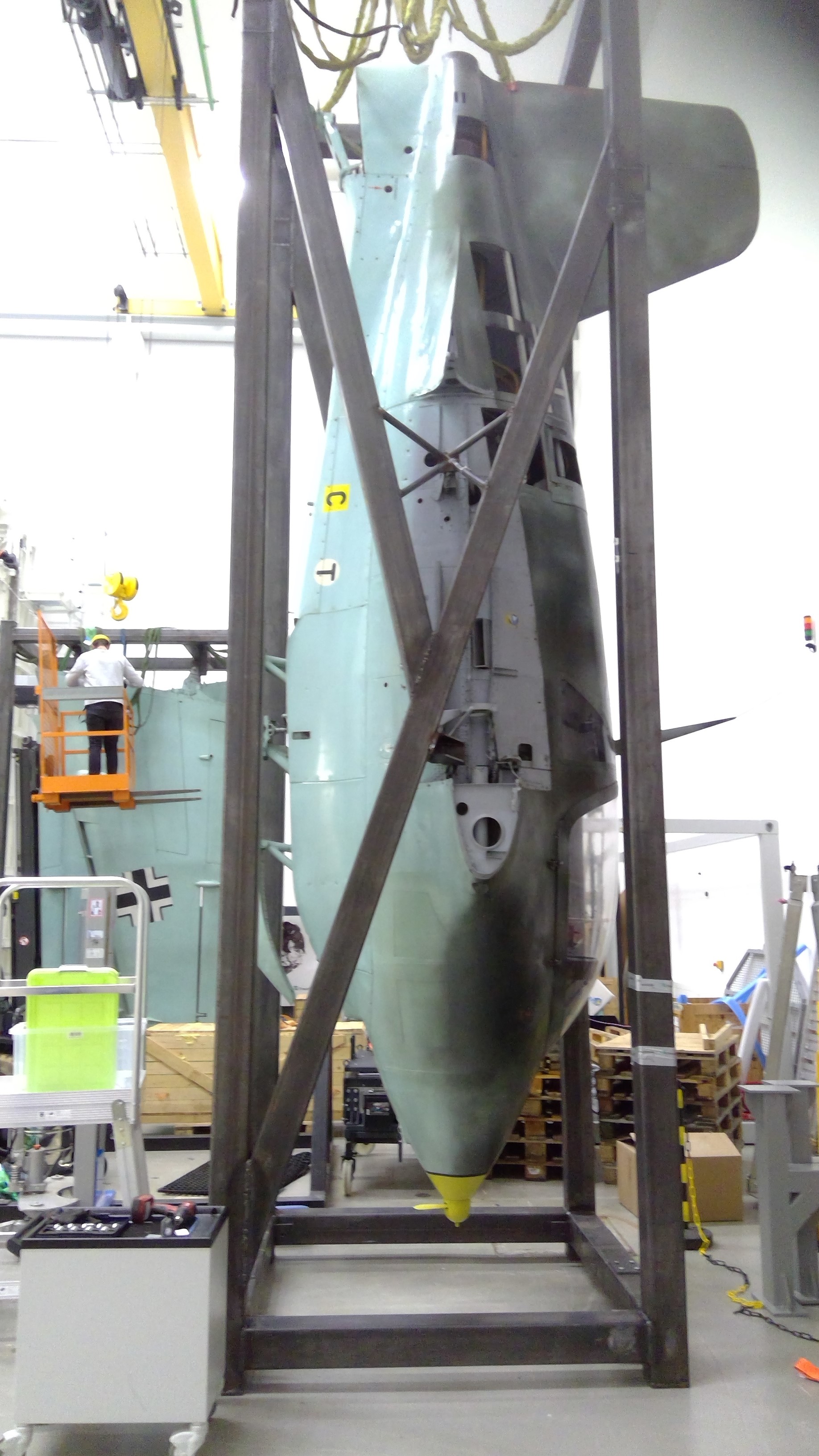} 
					\caption{Specimen}
					\label{fig:methods-datasets-me163-photo}
				\end{subfigure}
				\begin{subfigure}[b]{0.3\columnwidth}
					\centering
					\includegraphics[width=\linewidth, interpolate=false]{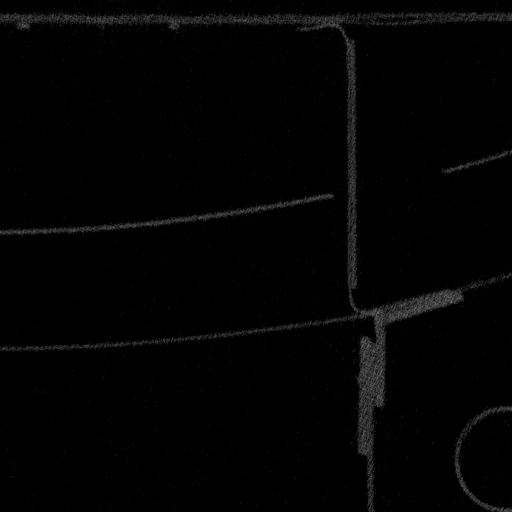}
					\caption{Input}
					\label{fig:methods-datasets-me163-reco}
				\end{subfigure}
				\begin{subfigure}[b]{0.3\columnwidth}
					\centering
					\includegraphics[width=\linewidth, interpolate=false]{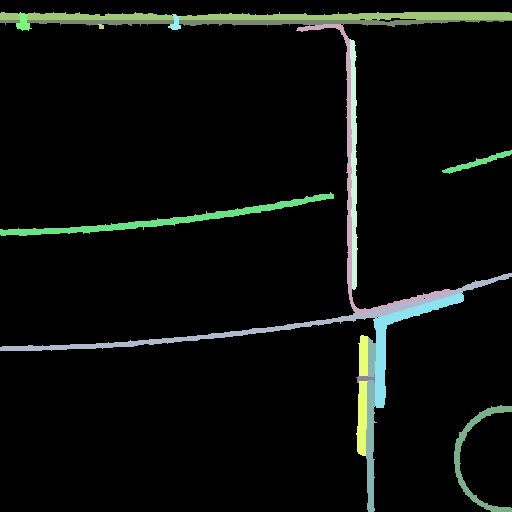}
					\caption{Reference}
					\label{fig:methods-datasets-me163-reference}
				\end{subfigure}			
				
				\vspace{\baselineskip}
				Corn
				\vspace{0.5\baselineskip}
								
				\begin{subfigure}[b]{0.3\columnwidth}
					\centering
					\includegraphics[width=\linewidth]{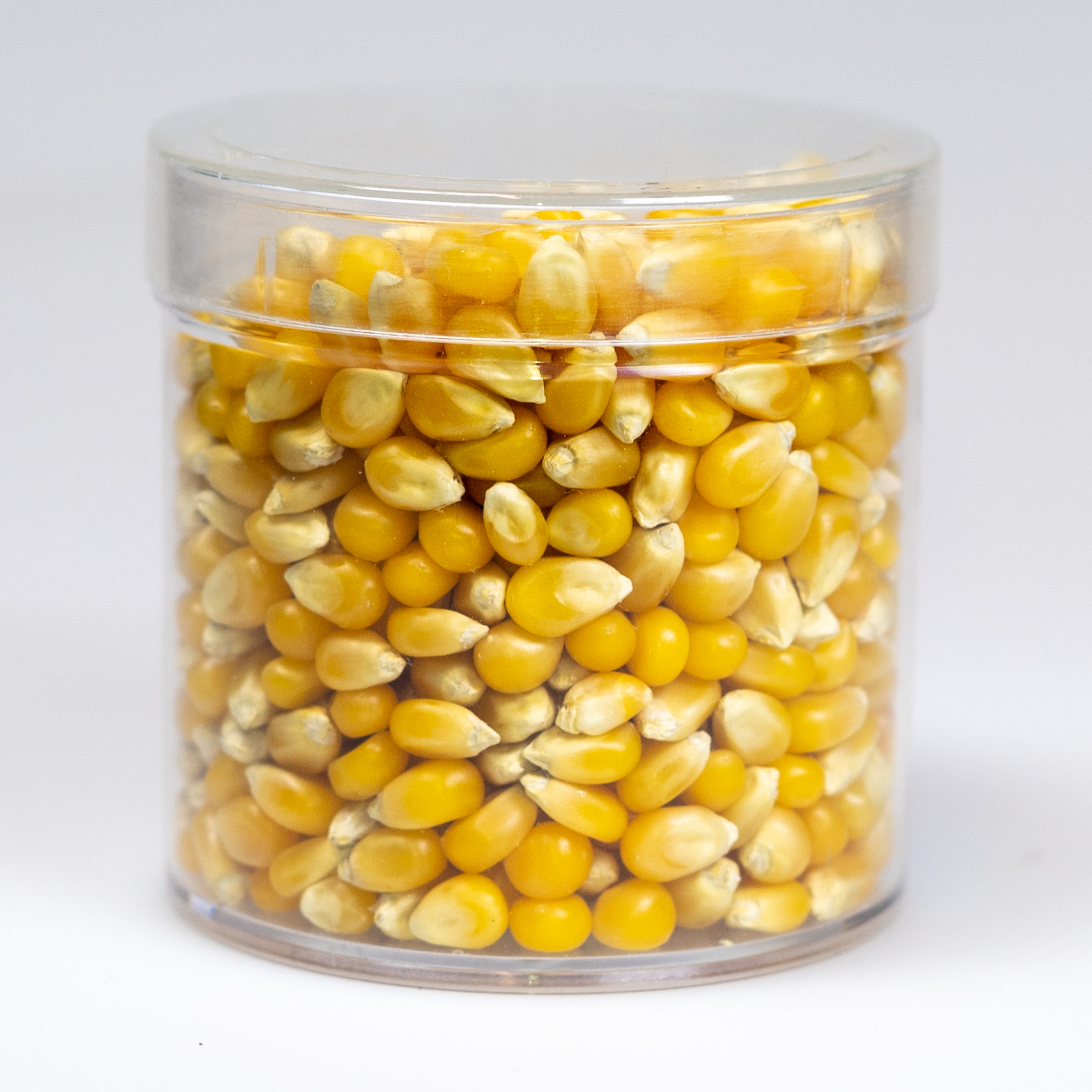}
					\caption{Specimen}
					\label{fig:methods-datasets-corn-photo}
				\end{subfigure}
				\begin{subfigure}[b]{0.3\columnwidth}
					\centering
					\includegraphics[width=\linewidth, interpolate=false]{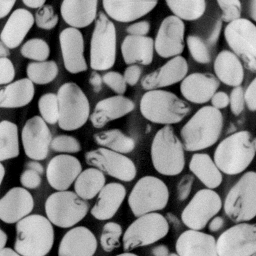}
					\caption{Input}
					\label{fig:methods-datasets-corn-reco}
				\end{subfigure}
				\begin{subfigure}[b]{0.3\columnwidth}
					\centering
					\includegraphics[width=\linewidth, interpolate=false]{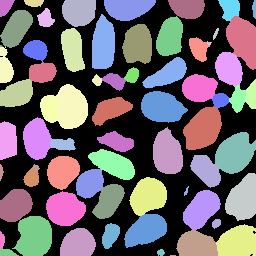}
					\caption{Reference}
					\label{fig:methods-datasets-corn-reference}
				\end{subfigure}								
				
				\vspace{\baselineskip}
				Marbles	
				\vspace{0.5\baselineskip}
											
				\begin{subfigure}[b]{0.3\columnwidth}
					\centering
					\includegraphics[width=\linewidth]{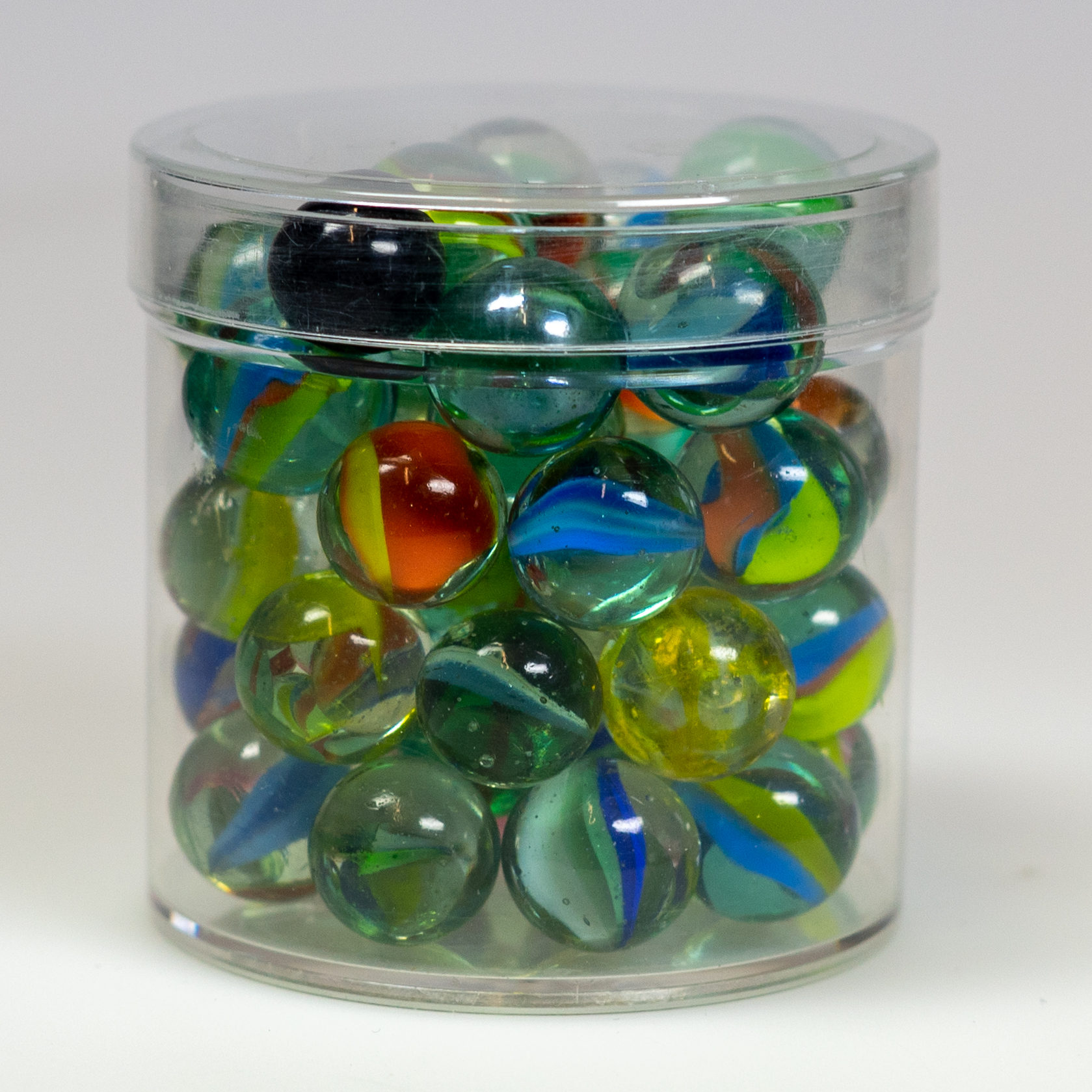}
					\caption{Specimen}
					\label{fig:methods-datasets-marbles-photo}
				\end{subfigure}
				\begin{subfigure}[b]{0.3\columnwidth}
					\centering
					\includegraphics[width=\linewidth, interpolate=false]{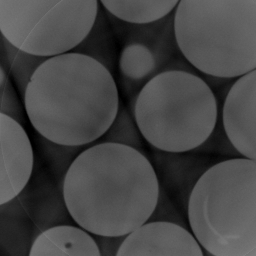}
					\caption{Input}
					\label{fig:methods-datasets-marbles-reco}
				\end{subfigure}
				\begin{subfigure}[b]{0.3\columnwidth}
					\centering
					\includegraphics[width=\linewidth, interpolate=false]{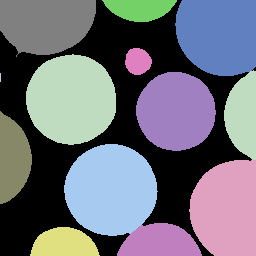}
					\caption{Reference}
					\label{fig:methods-datasets-marbles-reference}
				\end{subfigure}
								
				\caption{Photographs, exemplary CT slices, and reference segmentation of the Me\,163 (Figures \ref{fig:methods-datasets-me163-photo}, \ref{fig:methods-datasets-me163-reco}, \ref{fig:methods-datasets-me163-reference}), marbles (Figures \ref{fig:methods-datasets-marbles-photo}, \ref{fig:methods-datasets-marbles-reco}, \ref{fig:methods-datasets-marbles-reference}), and corn (Figures \ref{fig:methods-datasets-corn-photo}, \ref{fig:methods-datasets-corn-reco}, \ref{fig:methods-datasets-corn-reference}) data-sets respectively.}
				\label{fig:methods-datasets}
			\end{figure}

			To demonstrate, exemplify, and evaluate our achievements, we make use of three distinct data-sets: A specific sub-volume of the Me\,163 data-set of a Second World War fighter airplane \cite{Gruber2022}, as well as two bulk material data-sets depicting entities of glass marbles and corn kernels \cite{Gruber2020}. Figure~\ref{fig:methods-datasets} shows a photograph of each specimen, along with one typical slice from the reconstructed volume and a corresponding reference segmentation.
			
			The Me\,163 data-set utilized in this study consists of a volumetric subset and manually obtained reference segmentation XXL-CT data-set obtained from a historic airplane \cite{fraunhoferMe163InstanceSegmentationDataset}, which itself was extracted from an XXL-CT reconstruction. The reference segmentation sub-volumes of the Me\,163 data-set were manually annotated and underwent morphological postprocessing to clean up the edges. The acquisition process involved addressing challenging aspects such as noisy data, low contrast, and limited spatial resolution. A detailed description of the data-set creation, including the annotation and postprocessing process, can be found in \cite{Gruber2022}.
		
			The data-set consists of eight sets of sub-volume pairs, each sub-volume having the spatial dimensions of $512 \times 512 \times 512$\,voxels. For training, six sub-volume pairs of the data-set are used, while one sub-volume pair is used for validation and one for testing, respectively. Each sub-volume pair consists of a reconstructed sub-volume (see Figure \ref{fig:methods-datasets-me163-reco}) and its corresponding reference segmentation sub-volume (see Figure \ref{fig:methods-datasets-me163-reference}).		
			
			The reconstruction sub-volume is a small volumetric region which is extracted  from the reconstructed Me\,163 XXL-CT data. To ensure compatibility with SAM, both the reconstruction or input sub-volumes and the corresponding reference segmentation sub-volumes are extended with zero-padded 512\,voxels in every direction. This results in an embedded version of the sub-volumes with working dimensions of $1536 \times 1536 \times 1536$\,voxels. This arrangement allows for the extraction of a slice, centred on any arbitrary voxel within the original sub-volume, with the resolution of \(1024 \times 1024 \times 1\) voxels, matching the native input dimensions required by SAM.	
			
\begin{figure*}
			\centering		
			\begin{tabular}{@{} m{.05\linewidth} @{\hspace{-8pt}} m{0.22\linewidth} @{\hspace{5pt}} m{0.22\linewidth} @{\hspace{5pt}} m{0.22\linewidth} @{\hspace{5pt}} m{0.22\linewidth} @{}}
				& \multicolumn{2}{c}{Available Data} & \multicolumn{2}{c}{Used Data} \\
				& \multicolumn{1}{c}{Input} & \multicolumn{1}{c}{Reference} & \multicolumn{1}{c}{Input} & \multicolumn{1}{c}{Reference} \\
				\rotatebox[origin=c]{90}{Me\,163}			
				& \fbox{\includegraphics[width=\hsize]{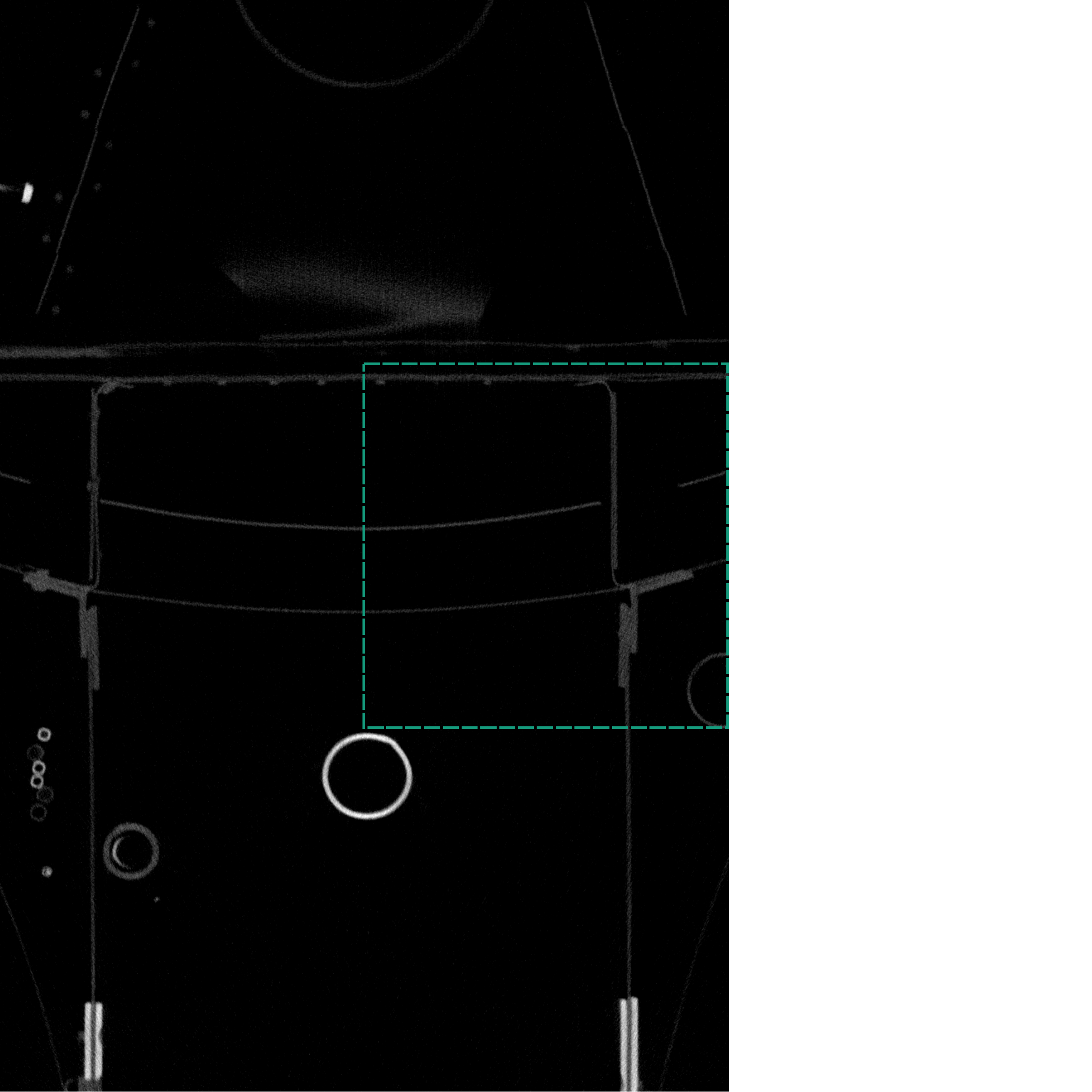}}
				& \fbox{\includegraphics[width=\hsize]{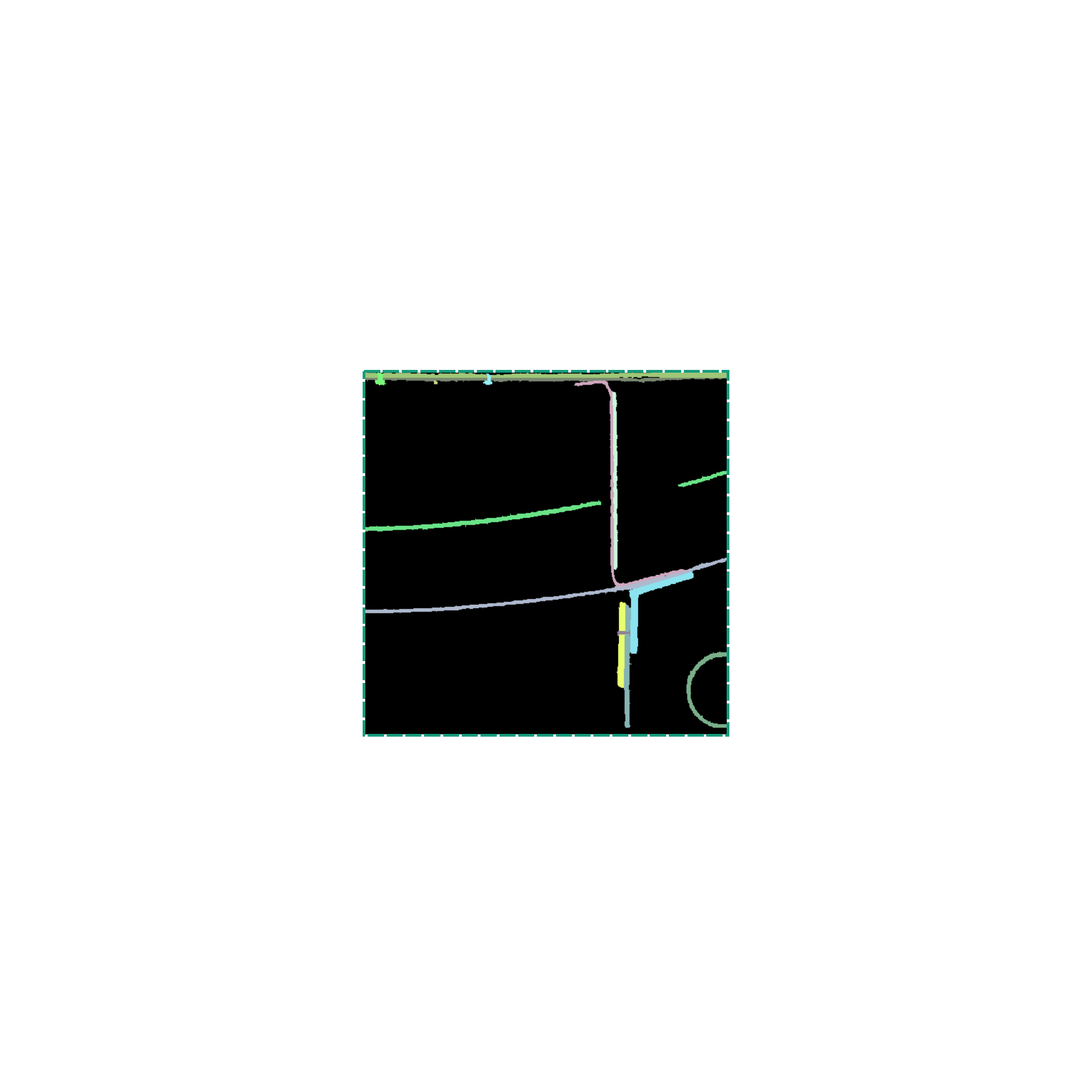}}
				& \fbox{\includegraphics[width=\hsize]{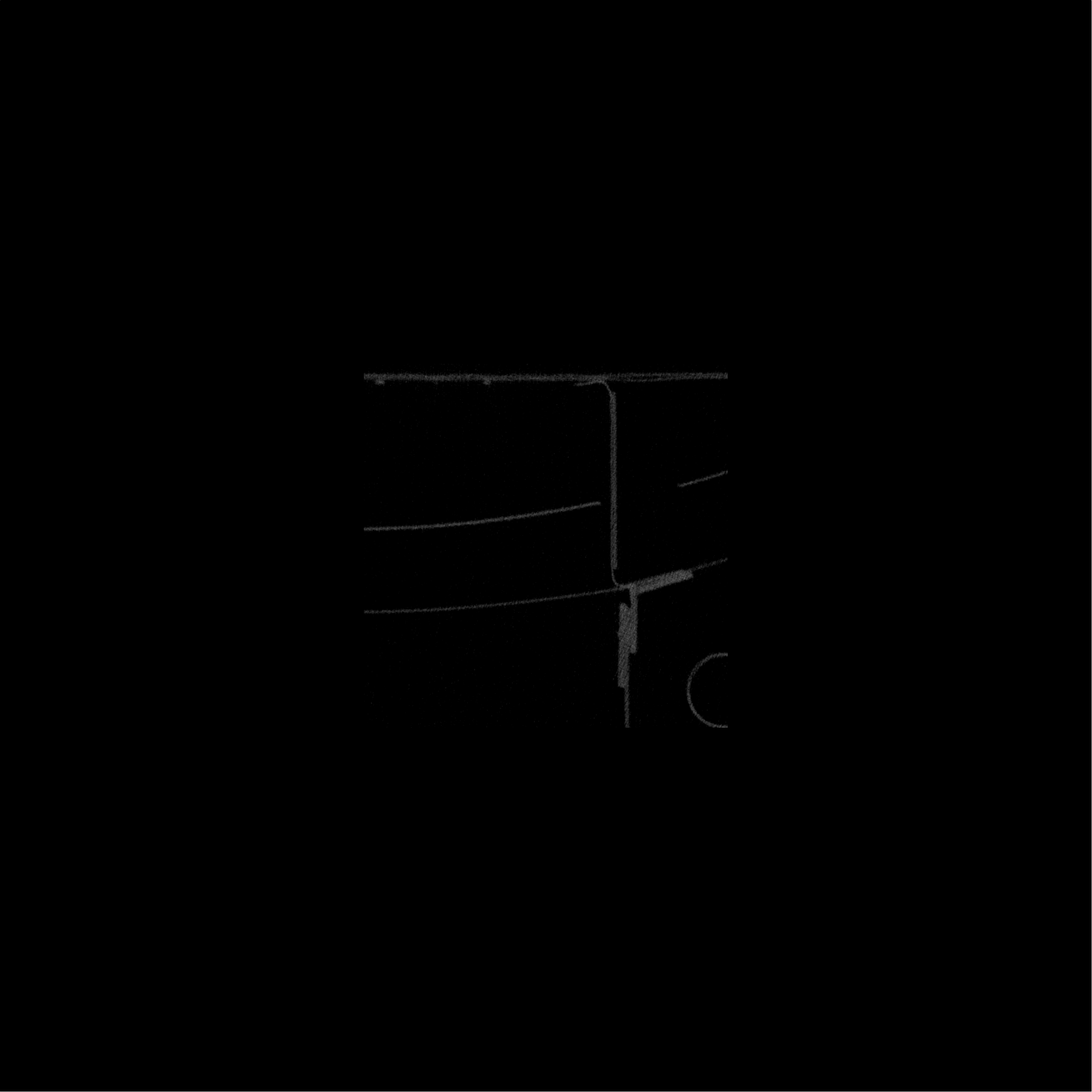}}
				& \fbox{\includegraphics[width=\hsize]{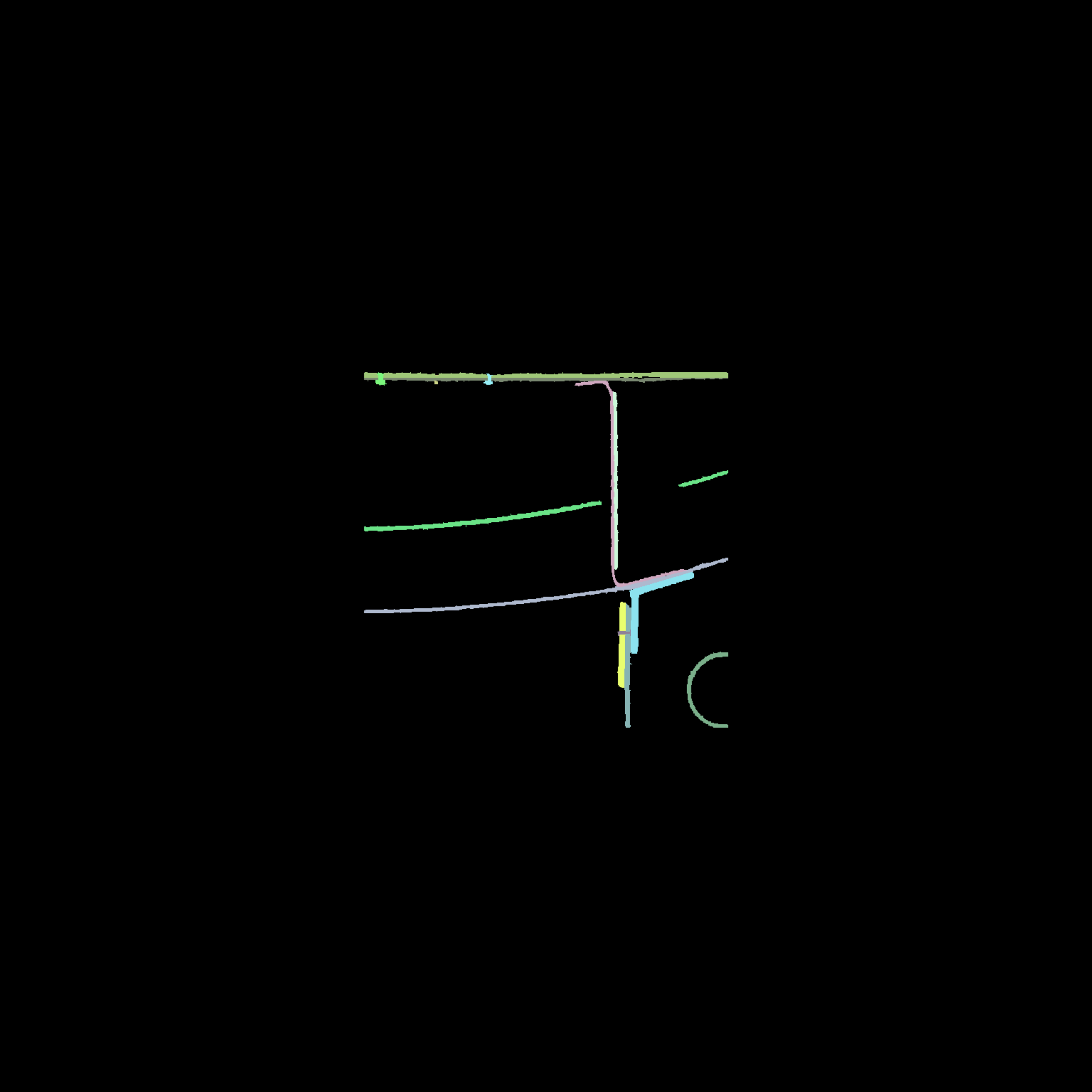}} \\
				\addlinespace[2pt]			
				\rotatebox[origin=c]{90}{Corn} 
				& \fbox{\includegraphics[width=\hsize]{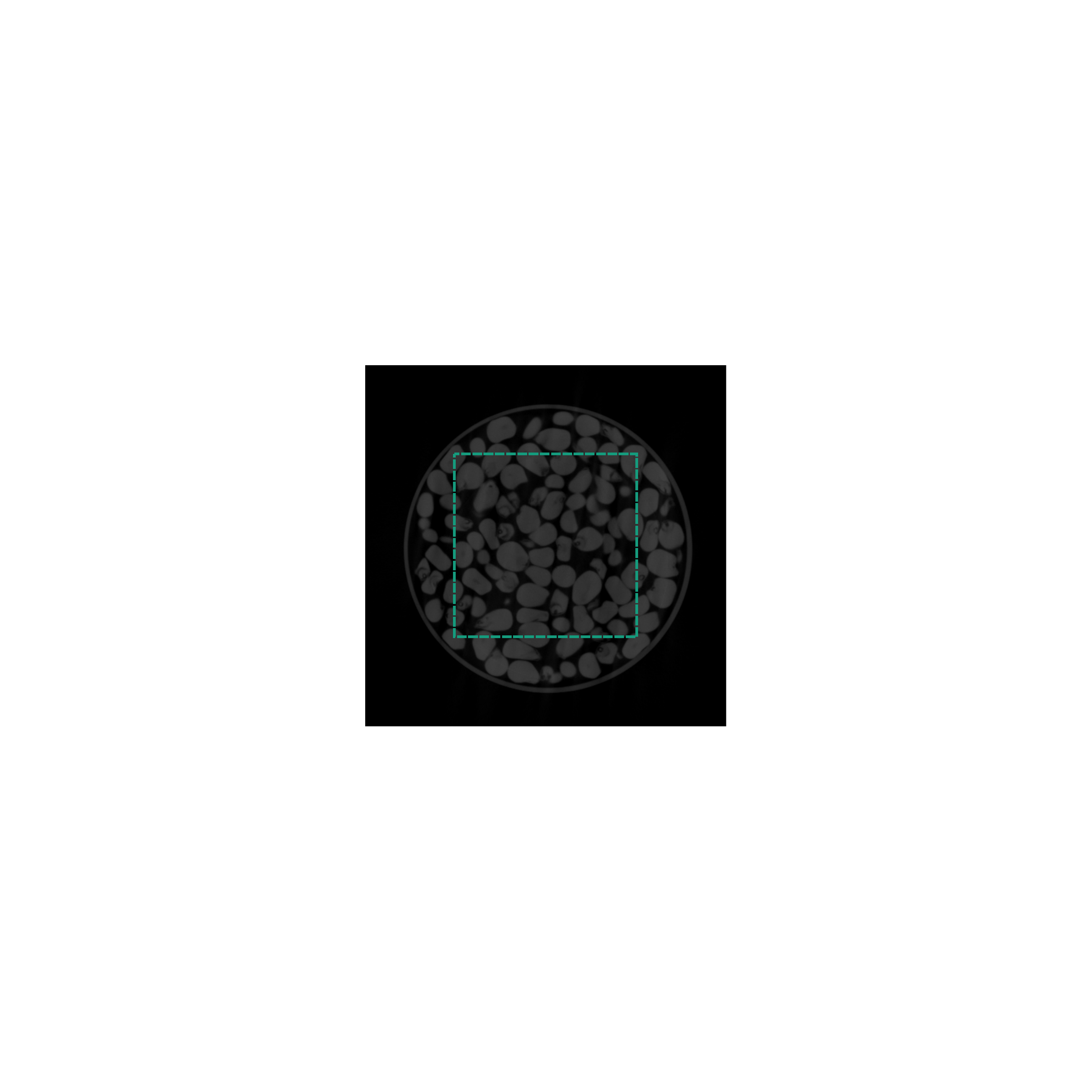}}
				& \fbox{\includegraphics[width=\hsize]{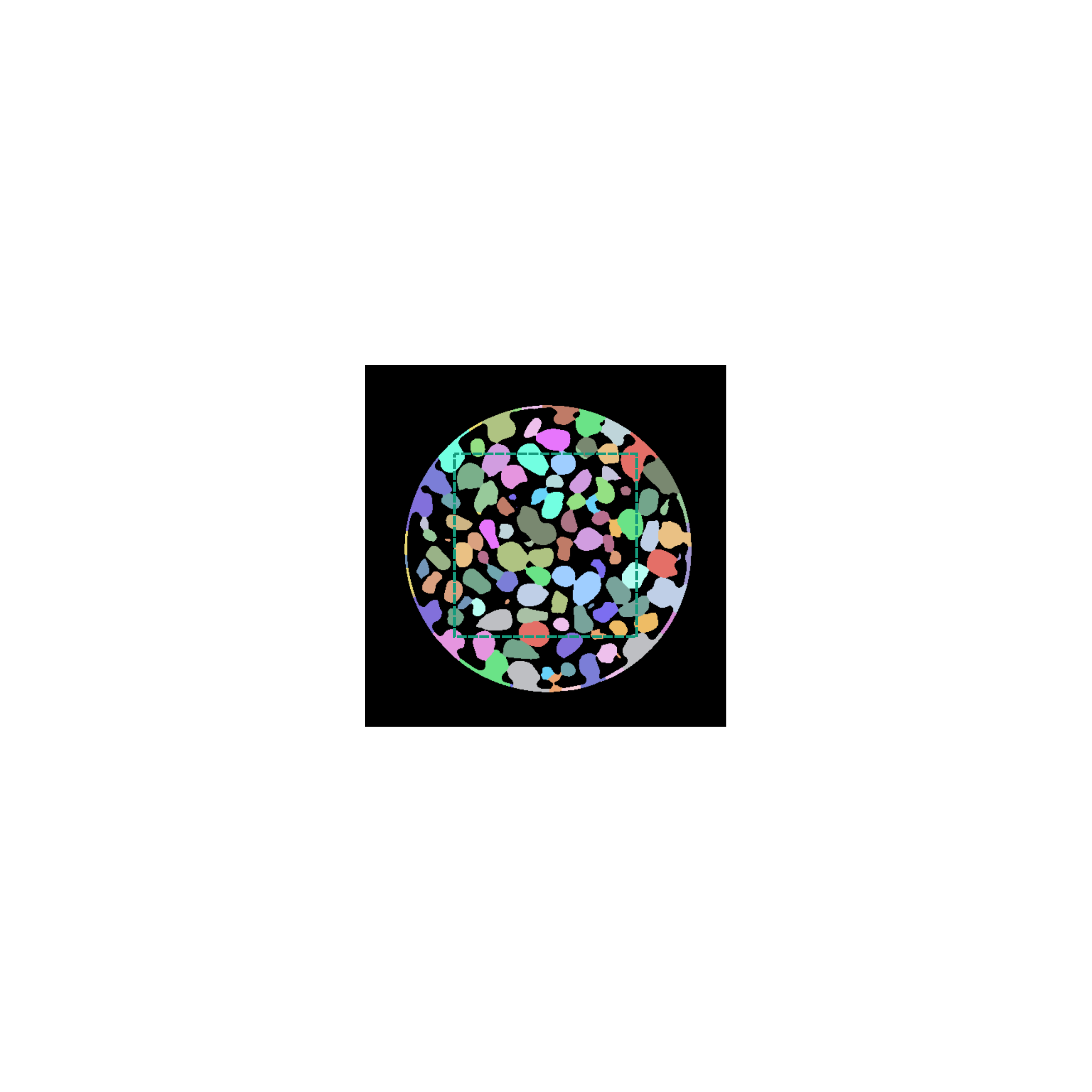}}
				& \fbox{\includegraphics[width=\hsize]{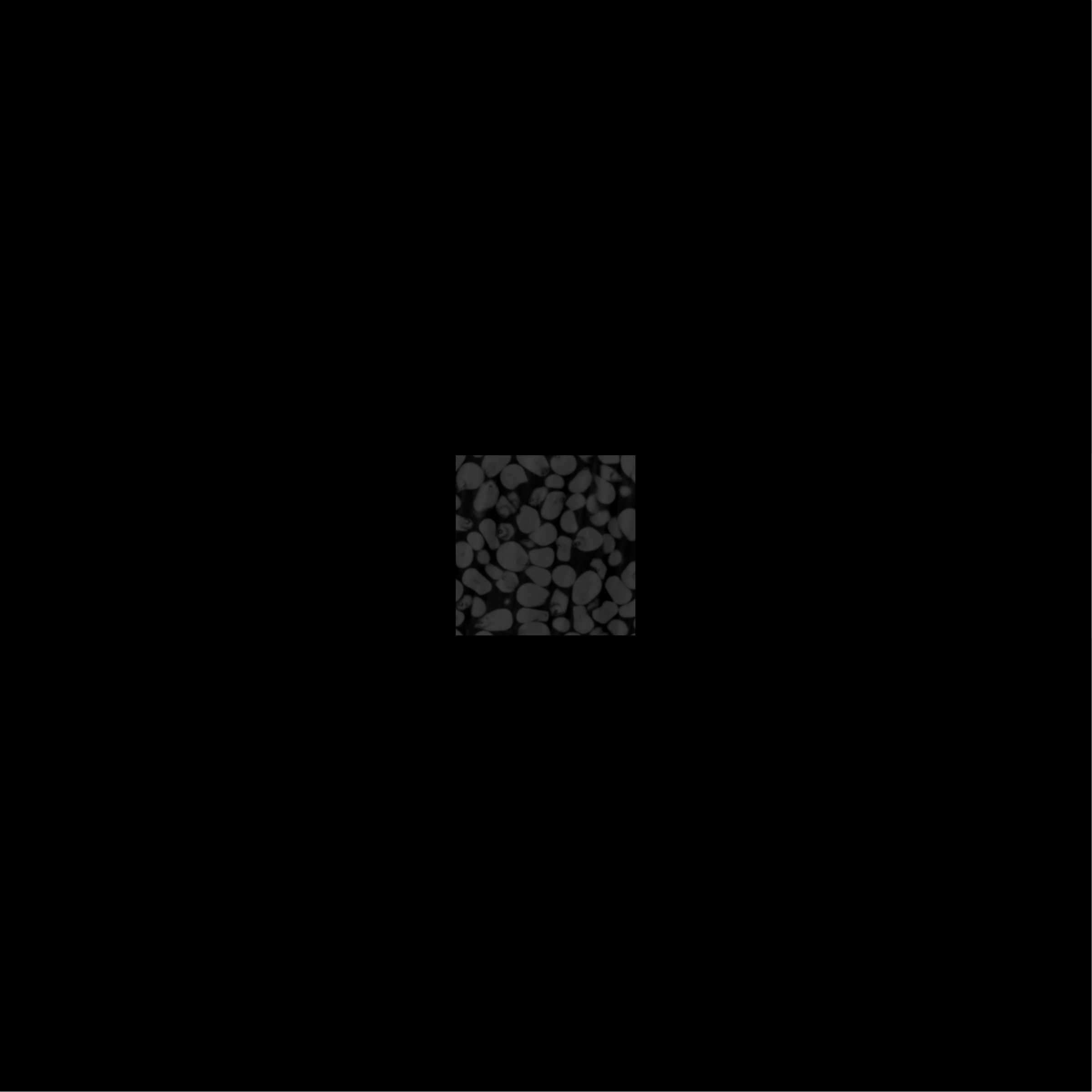}}
				& \fbox{\includegraphics[width=\hsize]{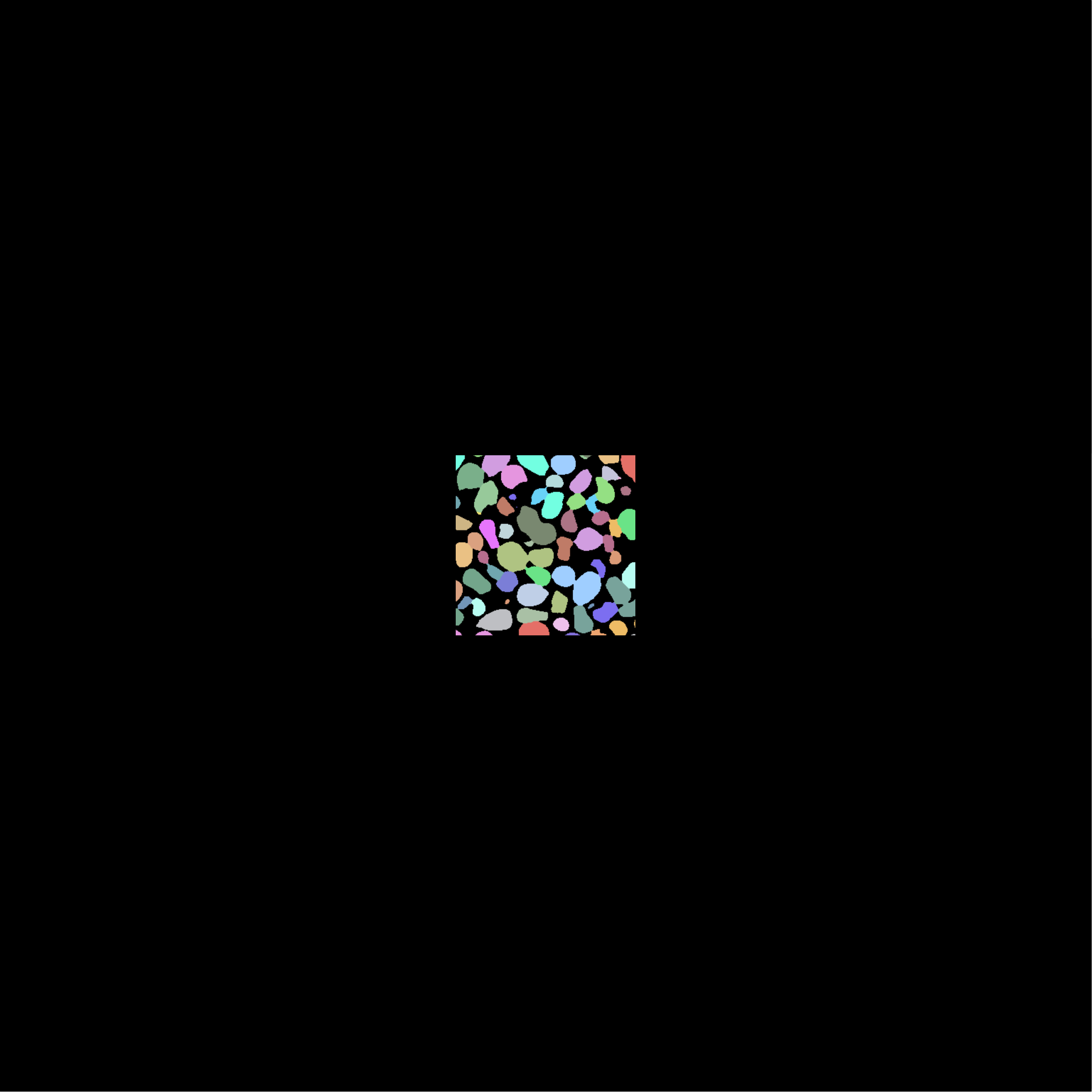}} \\	
				\addlinespace[2pt]				
				\rotatebox[origin=c]{90}{Marbles} 
				& \fbox{\includegraphics[width=\hsize]{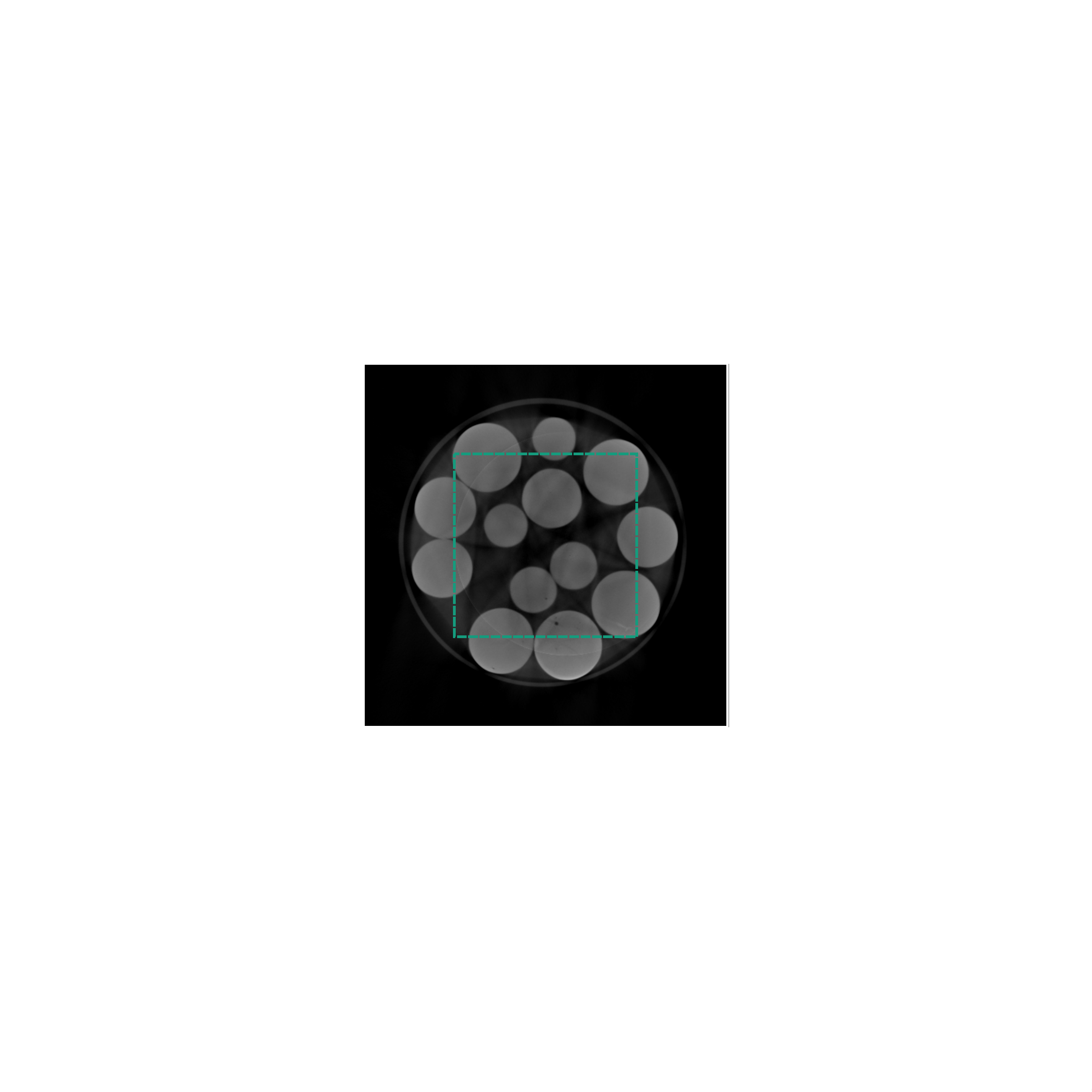}}
				& \fbox{\includegraphics[width=\hsize]{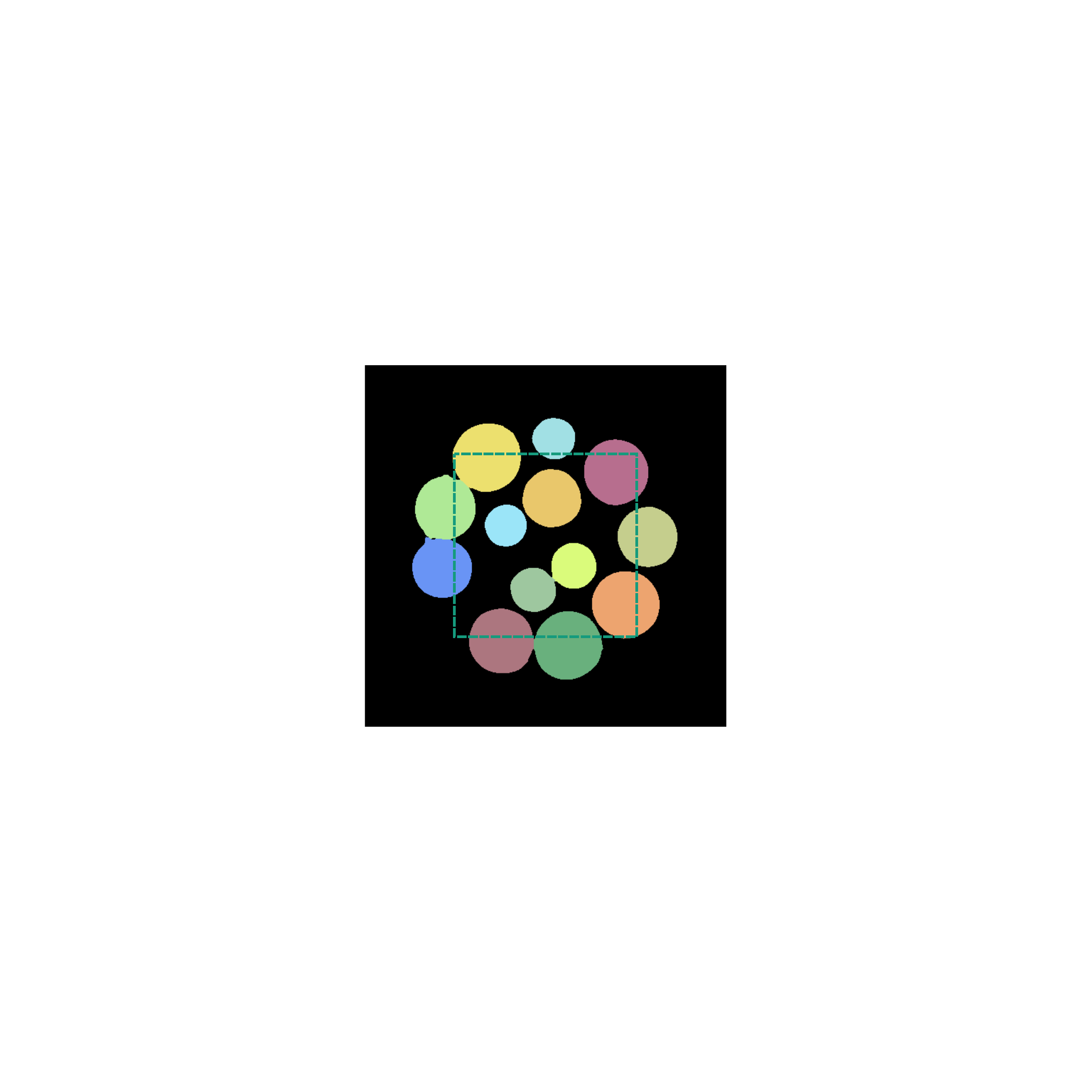}}
				& \fbox{\includegraphics[width=\hsize]{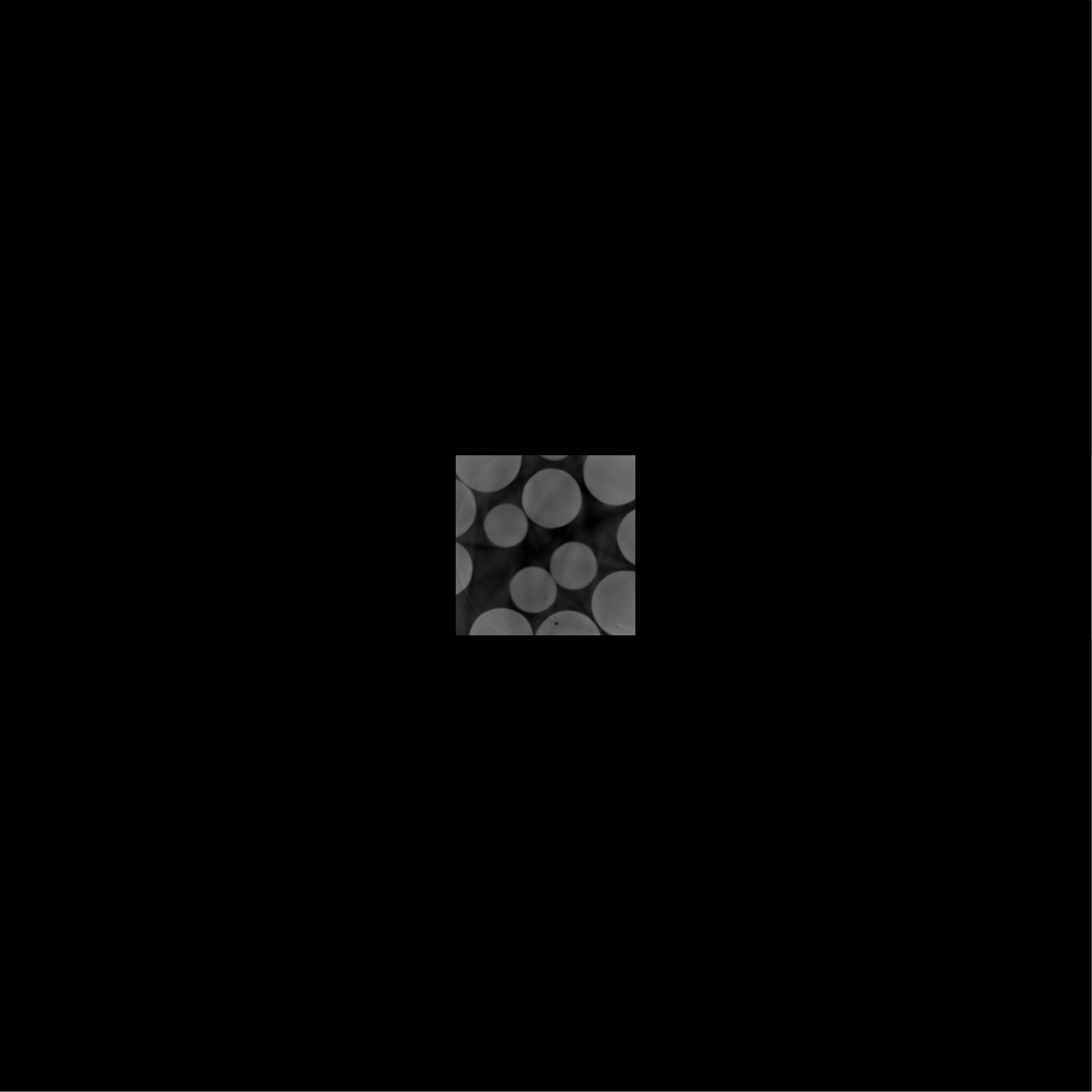}}
				& \fbox{\includegraphics[width=\hsize]{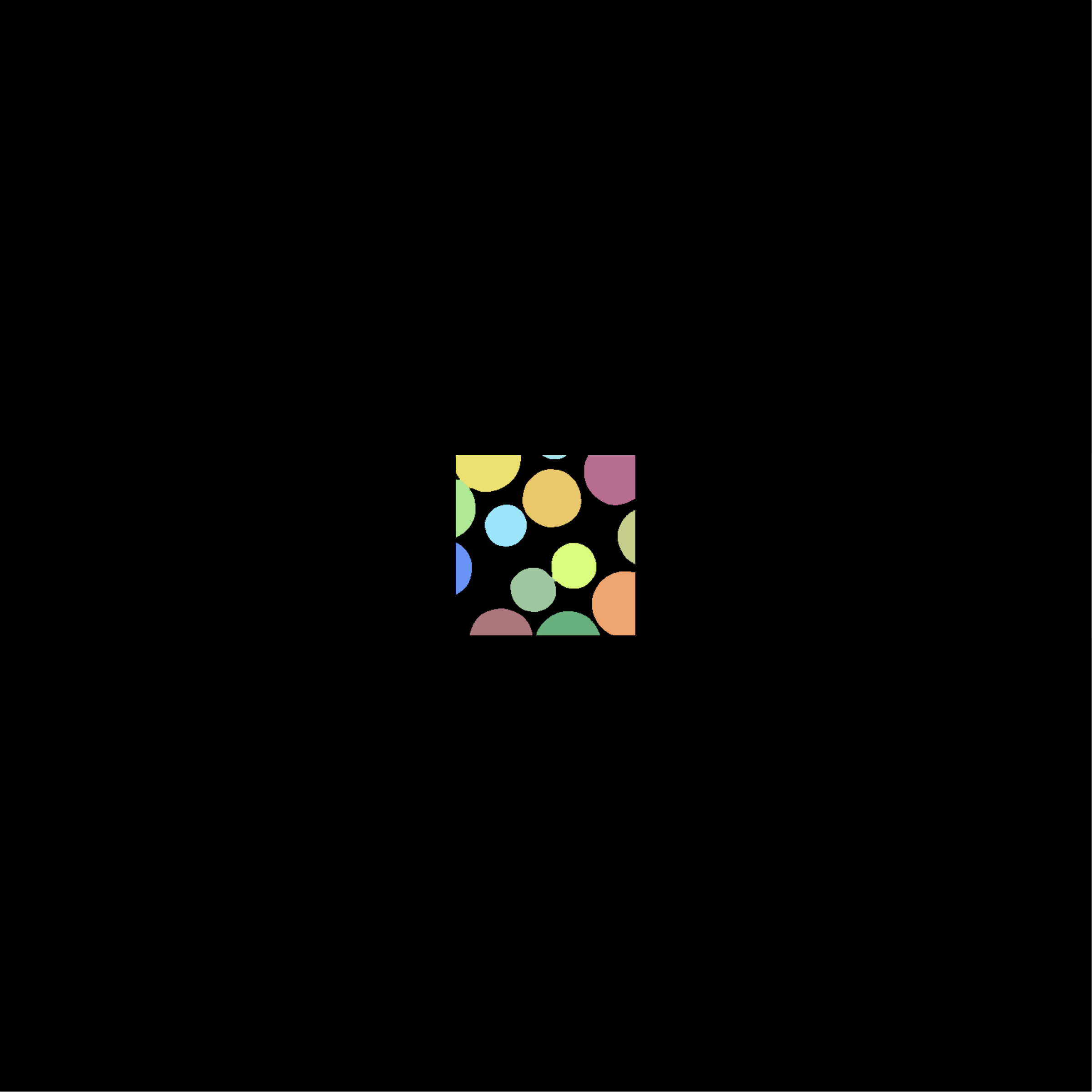}} \\
			\end{tabular}		
			
			\caption{Zero-padding preparation steps were performed on the input and reference slices of the different data-sets to create slices of size $1024\times1024$ pixels centred around each possible seed point. The white border regions in the available input and reference slices were filled with constant values of zero.} \label{fig:methods-datasets-enframed}
		\end{figure*}

			The first row of Figure \ref{fig:methods-datasets-enframed} illustrates the described enframing process for the Me\,163 data-set. The green rectangles in the first two columns indicate the unembedded region with $512 \times 512 \times 512$\,voxels and their manually annotated references. Due to the fact that the input sub-volumes of this data-set are located directly at the edge of the XXL-CT volume, it was not possible to fill the border of the sub-volumes with actual reconstruction values. Instead, we decided to use a border with a constant value of zero in all directions. The last two columns of Figure \ref{fig:methods-datasets-enframed} display the prepared input and reference slices used in the subsequent processing.
			
			The other two data-sets, which consist of CT scans of jars filled with marbles and corn, each also contain two sub-volumes: one for the input CT reconstruction sub-volume and one for its reference segmentation sub-volume. The segmentation process to yield the reference volumes of the bulk material data-set involved semi-automatic segmentation using threshold binarization with a threshold obtained from Otsu's method \cite{Otsu}, followed by a distance transform, watershed transform, and label-wise morphological closing, as described in more detail in \cite{Gruber2020}. As this traditional computer-vision process resulted in some erroneous segmentations in the contact regions between the jar and the bulk material, we only used a correctly segmented sub-volume in the centre of the jar, having a spatial dimension of $256 \times 256 \times 256$\,voxels (denoted by the green rectangle in Figure \ref{fig:methods-datasets-enframed}). Also, the sub-volumes of the bulk material were enframed by a border of 512\,voxels thickness with a constant value of zero.
			
		\subsection{Fine-Tuning on NDT Data-set}\label{sec:methods-fineTune}
			The SA-1B training data-set published by the authors of the SAM \cite{72} contains predominantly coloured natural photographs, such as street scenes or still life compositions of semantically well-known objects from daily life. In contrast volumetric data-sets obtained from the NDT field, and particularly the slices extracted from the volumes, are frequently of a rather abstract nature and do not depict recognizable objects. Hence, these NDT images deviate quite much from the familiar photographic data-set used by SAM, and this deviation poses several challenges in achieving sufficient segmentation quality (see Section \ref{sec:results-sam-slices}). This, within the CT imaging domain even familiar objects can be difficult to recognize for non-experts, as they exhibit unusual structures or non-orthogonal sections due to the specimen's imaging geometry, or they may contain strong imaging and reconstruction artefacts.			
			
			Ma et al. \cite{73} showcased a potential improvement in segmentation quality by fine-tuning SAM on the problem domain, which inspired us to adopt a similar fine-tuning approach. 
			
			In this study, we opted to perform fine-tuning on a certain part of the SAM, specifically the \emph{Mask Decoder}. For this purpose, we utilized, extracted, and pre-processed slices from the ME\,163 training data-set. Our approach adhered to the guidelines outlined in \cite{73}, which have previously been employed for fine-tuning on medical volume CT data-sets. 
			
			The Me\,163 data-set was chosen due to its distinct level of complexity, setting it apart from the bulk material data-sets also being investigated. In contrast, the marble and corn data-sets can be segmented relatively easily using conventional image processing techniques. 
	
			For the fine-tuning process we randomly selected voxel positions from the Me\,163 training data-set. If the chosen voxel was a foreground voxel belonging to an known labelled entity, three orthogonal slices centred around its position were extracted. These slices were used as training examples, with the data range of the input slice normalized to $[0.0,255.0]$. For the target slice all voxels of the entities belonging to the centre voxel were one-hot encoded. 
			
\begin{figure}
				\centering
				
				\begin{subfigure}[t]{0.235\columnwidth}
					\centering
					\fbox{\includegraphics[width=\linewidth]{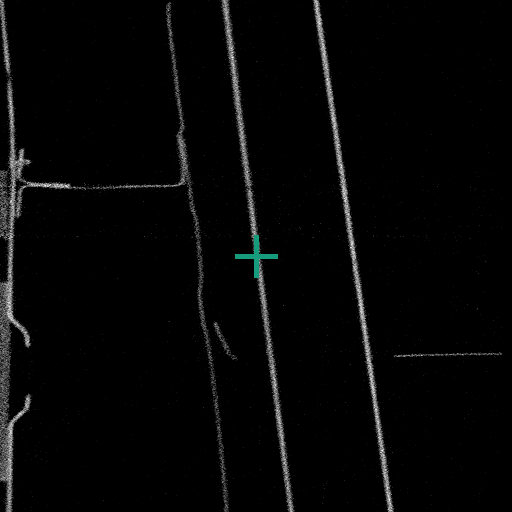}}
					\caption{Input}
					\label{fig:methods-fineTune-dataset-foreground-input}
				\end{subfigure}
				\begin{subfigure}[t]{0.235\columnwidth}
					\centering
					\fbox{\includegraphics[width=\linewidth]{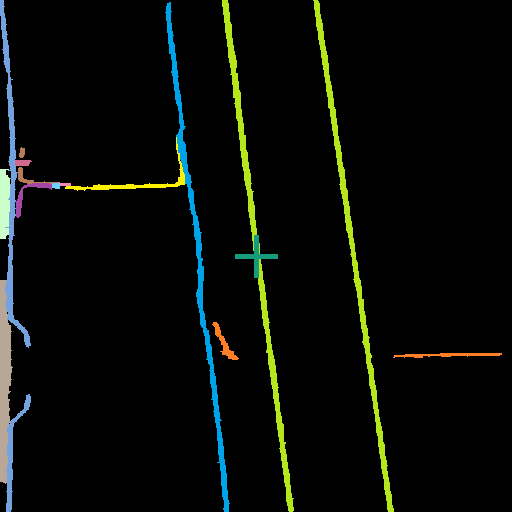}}
					\caption{Reference}
					\label{fig:methods-fineTune-dataset-foreground-reference}
				\end{subfigure}
				\begin{subfigure}[t]{0.235\columnwidth}
					\centering
					\fbox{\includegraphics[width=\linewidth]{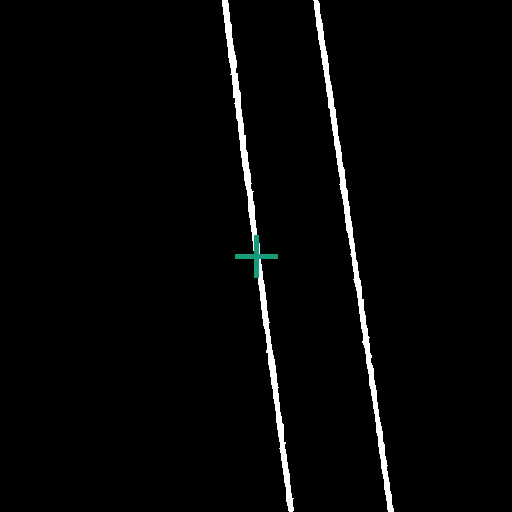}}
					\caption{One-hot \\encoded foreground target}
					\label{fig:methods-fineTune-dataset-foreground-oneHot}
				\end{subfigure}
				\begin{subfigure}[t]{0.235\columnwidth}
					\centering
					\fbox{\includegraphics[width=\linewidth]{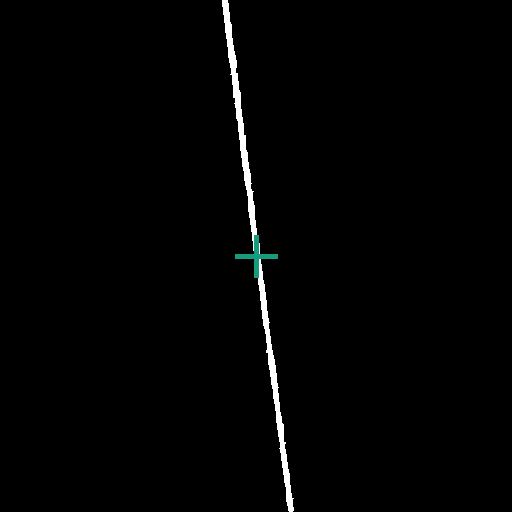}}
					\caption{Connected component target}					
					\label{fig:methods-fineTune-dataset-forground-ConnectedComponentTarget}
				\end{subfigure}					
				\caption{\label{fig:methods-fineTune-dataset-forground}Processing of an example foreground slice used for fine tuning SAM. Consisting of reconstruction slice (Figure  \ref{fig:methods-fineTune-dataset-foreground-input}), reference slice (Figure \ref{fig:methods-fineTune-dataset-foreground-reference}), one-hot encoded (Figure \ref{fig:methods-fineTune-dataset-foreground-oneHot}), and connected component training target (Figure \ref{fig:methods-fineTune-dataset-forground-ConnectedComponentTarget}). The green cross marks the centre of the slice.}
			\end{figure}

			SAM operates on images, while our attempted input is a single slice from a volumetric data-set. To ensure that a three-dimensional connected object was represented by a single segment in the two-dimensional slices, a connected component analysis (CCA) was performed on the one-hot encoded target slice. Only the segment connected to the centre of the target slice was used as the target for training (see Figure \ref{fig:methods-fineTune-dataset-forground}). The surrounding image does not provide sufficient information to distinguish if neighbouring non-touching segments are belonging to the same segment. Thus we performed a CCA and treat the parts of segments not connected in the current slice as separate segments.
						
\begin{figure}
				\centering
				\begin{subfigure}[t]{0.235\columnwidth}
					\centering
					\includegraphics[width=\linewidth, interpolate=false]{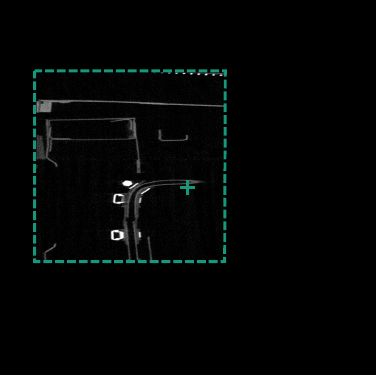}
					\caption{Input}
					\label{fig:methods-fineTune-dataset-background-input}
				\end{subfigure}
				\begin{subfigure}[t]{0.235\columnwidth}
					\centering
					\includegraphics[width=\linewidth]{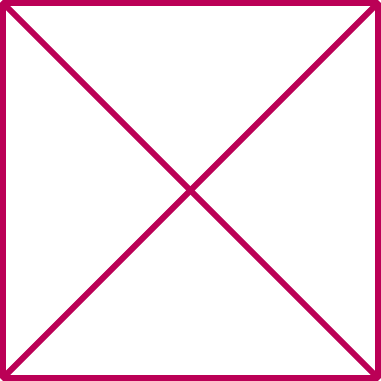}
					\caption{No background}
					\label{fig:methods-fineTune-dataset-background-noBackground}
				\end{subfigure}
				\begin{subfigure}[t]{0.235\columnwidth}
					\centering
					\includegraphics[width=\linewidth]{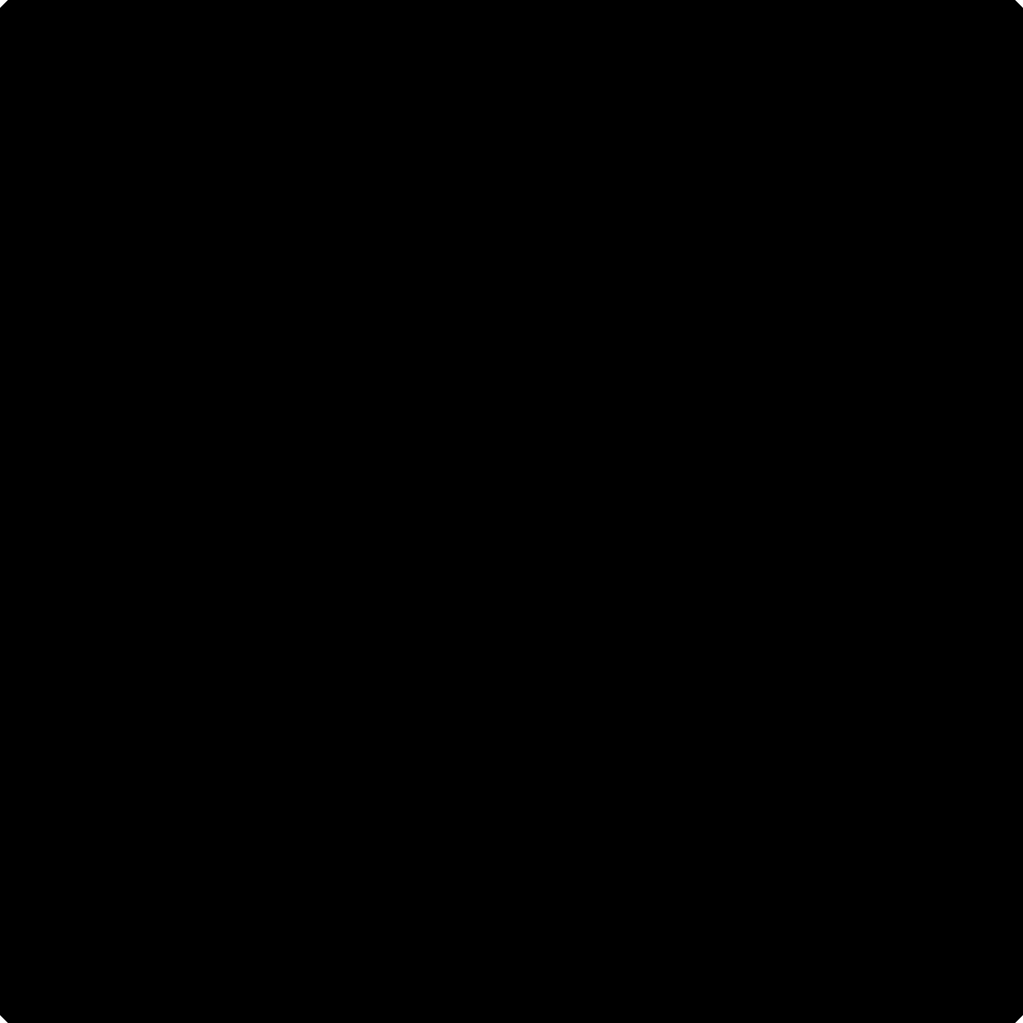}
					\caption{Constant value background}
					\label{fig:methods-fineTune-dataset-background-constantValueBackground}
				\end{subfigure}
				\begin{subfigure}[t]{0.235\columnwidth}
					\centering
					\fbox{\includegraphics[width=\linewidth, interpolate=false]{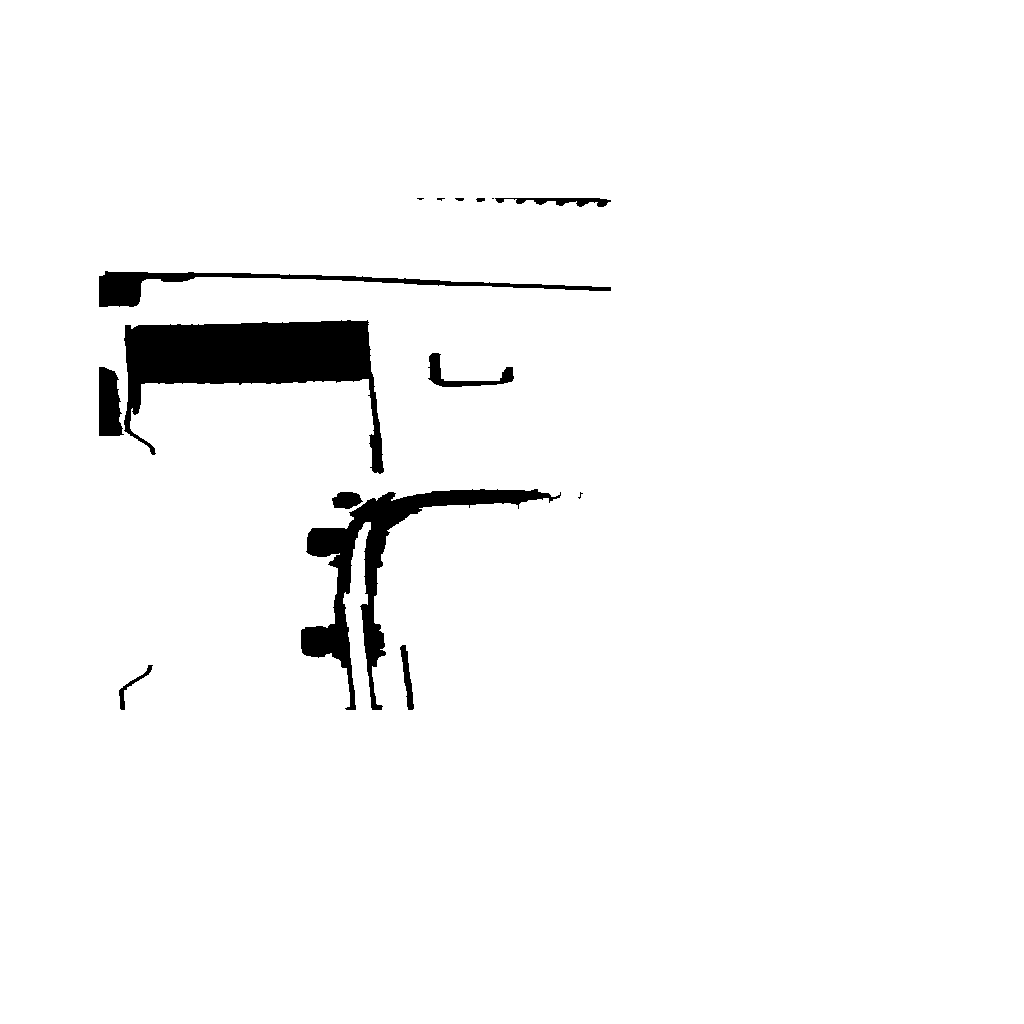}}
					\caption{Connected component background}					
					\label{fig:methods-fineTune-dataset-background-connectedComponentBackground}
				\end{subfigure}				
				\caption{\label{fig:methods-fineTune-dataset-background}Processing of an example background slice used for fine tuning SAM. The green cross marks the centre of the slice, which is located in the background of the reconstruction. The green border around the reconstruction slice in Figure \ref{fig:methods-fineTune-dataset-background-input} depicts the original volume size, which then was enframed with an constant value border. The other sub-figures show the tested possibilities for target slices for the fine-tuning: \emph{ForegroundOnly} (Figure \ref{fig:methods-fineTune-dataset-background-noBackground}), \emph{ConstantValueBackground} (Figure  \ref{fig:methods-fineTune-dataset-background-constantValueBackground}), and \emph{ConnectedComponentBackground}  (Figure\ref{fig:methods-fineTune-dataset-background-connectedComponentBackground}).}
			\end{figure}

			If the voxel at the centre of a slice represented the background, we generated three orthogonal background examples, each containing an normalized input slice and an target slice. We evaluated three versions: \emph{ForegroundOnly}, which included only foreground input slices; \emph{ConstantValueBackground}, where we provided both background and foreground input and target slices for training, but expected SAM to produce a completely empty response for background slices; and \emph{ConnectedComponentBackground}, where we identified all background voxels connected to the centre voxel of the slice as the target segment. This was achieved through CCA on the data-set's background, formed by also enframing the reference segmentation with a zero-padded boundary. Consequently, the network was prompted to consider all voxels connected to the air space in the slice's centre as part of that segment. Figure \ref{fig:methods-fineTune-dataset-background-connectedComponentBackground} provides an illustrative example of the different target versions.
			
			Due to the significantly lower count of foreground voxels (0.1-9.4\%) compared to background voxels in the Me\,163 data-set, we included all foreground examples while randomly selecting a subset of background examples of the same size. This approach ensured a balanced representation of both classes. To prevent batches from containing closely located examples, the selected examples were shuffled and grouped into batches, with each batch containing sixteen foreground examples and 16 background examples. Additionally, to further diversify the examples within each batch, we employed a relatively large stride during the example extraction process. This ensured that the examples originated from different sub-volumes within the data-set. In each iteration over the data-sets, a new random initial position offset was chosen, employing a non-repetitive selection process to extract different examples.		

			We chose a single point prompt in the exact centre of each slice as the input for SAM during training. This choice aligns with the input for our validation application as well as the tile-based SAM integration for volume data-sets (see Section \ref{sec:methods-inference}).
		
			The batch size was set to 64. We initiated the training with an learning rate of $8 \mathrm{e}^{-4}$, which was linearly increased over the first 250\,iterations. For optimization, we utilized the AdamW optimizer \cite{68} with $\beta_1=0.9$ and $\beta_2=0.999$, along with a weight decay of 0.1. Our loss function consisted of a combination of Dice Loss (sigmoid=true, squared-pred=true, and mean reduction) and binary cross-entropy loss (mean reduction). We let the training run until overfitting for 10 to 25\,days. We selected the model with the lowest validation loss, determined at moving window intervals of 128\,iterations. 
		
		\subsection{Inference Workflow for Volumetric Data-sets} \label{sec:methods-inference}
\begin{figure*}
				\definecolor{tikzBoxColourInferenceGreyBox}{RGB}{230,230,230}
				\DeclareRobustCommand{\tikzBoxInferenceGreyBox}{\tikz\node[rectangle,fill=tikzBoxColourInferenceGreyBox,minimum width=\tikzBoxSize,minimum height = \tikzBoxSize,] (r) at (0,0) {};}	
				
				\centering
				\includegraphics[width=1.00\textwidth,height=1.0\textheight,keepaspectratio]{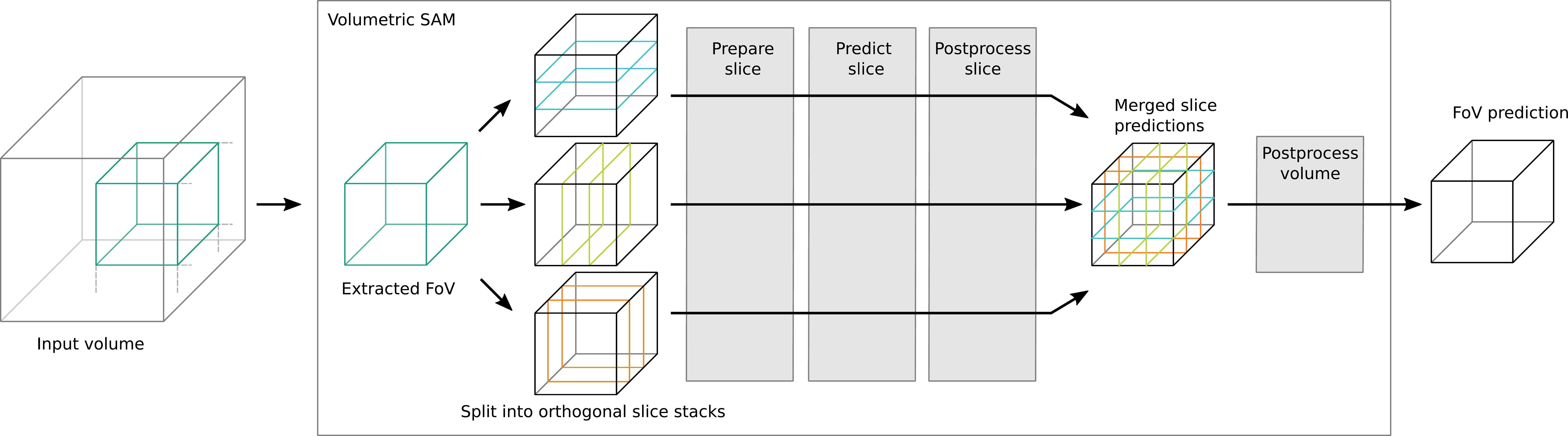}
				\caption{\label{fig:methods-inference-overview}Schematic workflow of the volumetric data inference segmentation using SAM. Algorithm options and steps for the configurable stages (grey boxes (\tikzBoxInferenceGreyBox)) are listed in Table \ref{tab:methods-inference-overview}.}
			\end{figure*}

			Since SAM works only on RGB image data-sets but we wanted to segment volumetric data-sets, we had to incorporate a adequate workflow to translate between these two spatial domains. Since our goal was to evaluate SAM for volumentric data-sets and not necessarily to implement a complete new volumetric version, we refered to simple operators. 
			Figure \ref{fig:methods-inference-overview} shows an overview of the approximate workflow for a volumetric data inference of SAM. In short, we extract a sub-volume tile from the input volume and pass it to the volumetric SAM adaption, which transforms it into three orthogonal slice stacks. 
	
			For each slice stack, we perform slice preparations (such as normalization and zero-padding), a forward pass through SAM, selection of the corresponding outputs, and slice postprocessing. The output slice stacks are then merged and undergo further volumetric postprocessing to generate segmentation proposals which are returned from the volumetric SAM adaption into the inference algorithm. The evaluated algorithms are listed and compared in Table \ref{tab:methods-inference-overview}
		
			\begin{table*}
				\centering
				\begin{adjustbox}{max width=\textwidth}
					\begin{tabular}{llp{0.8\textwidth}}
						\toprule
						\textbf{Stage} & \textbf{Algorithm} & \textbf{Description (\emph{Options})} \\
						\midrule
						\textbf{Preprocess Slice}&& \\
						\textbf{Algorithms}&& \\
						& Slice Normalization & Normalization of pixel values in each slice to the minimum and maximum range of the slice. \\
						& Outlier and Empty Slice Detection & Identification and handling of outlier and empty slices. \\		
						& RGB Conversion & Conversion of grey values to RGB colour in order to comply with SAM interface requirements. \\
						& Enframing & Adds a zero-padded border to each slice to centre the seed point to comply with SAM interface requirements. \\
						& Estimated Foreground Volume & Utilizes different \emph{binarization strategies} and \emph{thresholds} to estimate the foreground volume. \\
						\midrule
						\textbf{Predict Slice}&& \\			
						\textbf{Algorithms}&& \\
						& Prompt Type & Type of prompt is used for invoking SAM: point prompt for tile centre; point prompt for tile centre and dense prompt from accumulator.\\
						& Multimask Output Selector & Select mask from multiple disambiguating instance output channels predicted by SAM: maximum predicted IoU: fixed index of channel ; maximum IoU with estimated foreground to avoid segmenting background; minimum count of voxels to reduce under segmentation. \\
						& Mask Output Selector & Selected output format of SAM: binary full resolution mask; quarter resolution logits with subsequent \emph{threshold} and \emph{upscaling algorithm}. \\		
						\midrule
						\textbf{Postprocess Slice}&& \\			
						\textbf{Algorithms}&& \\
						& Seed Point Filter & Aborts or continues prediction based on the seed point's classification as background or foreground (\emph{count of slices}). \\
						& Merge Slice Rule & Rule which should be used to decide if an how to merge slices to stacks and when to abort an computation stack: \emph{Always}; \emph{BreakOnEmptySlice}; \emph{MinimumIOUToLastSlice} (\emph{threshold}); \emph{MinimumIOUToForeground} (\emph{threshold}). \\
						& Slice Median & Apply median filter to each slice (\emph{enabled or disabled}).\\
						& Connected Component Analysis & Analyse connected components and keep only segment connected to seed point (\emph{enabled or disabled}).\\				
						\midrule
						\textbf{Postprocess Volume}&& \\			
						\textbf{Algorithms}&& \\
						& Merge Slice Predictions & Merge orthogonal slice stack predictions based on \emph{count} of foreground voxels.\\
						& Volume Median & Apply median filter to merged volume (\emph{enabled or disabled}).\\
						\bottomrule
					\end{tabular}
				\end{adjustbox}
				\caption{Overview of algorithm choices and options for different stages of the volumetric SAM adaption seen in Figure \ref{fig:methods-inference-overview}.}
				\label{tab:methods-inference-overview}
			\end{table*}
				
			\subsubsection{Adapting SAM for Volumetric Data-sets}\label{sec:methods-volumetricSAM}	
				Adapting SAM, which was originally designed for segmenting image data-sets, to our volumetric CT data-sets required certain modifications and the implementation of appropriate postprocessing steps. In this section, we explore various possibilities for this transition and subsequently outline the approach  we finally selected. 
				
				Several 2D to 3D techniques can be utilized to facilitate this transformation \cite{76}. For example in \cite{76-Xia2018} a Volumetric Fusion Net (VFN) was employed to merge multiple 2D segmentation predictions into a comprehensive 3D prediction volume.	In a related work, \cite{77} adopted a similar methodology for pancreas segmentation, albeit utilizing a different VFN. According to \cite{76} other approaches involve incorporating neighbouring 2D slices as additional channel information or utilizing specialized topologies to extract and merge features in both the 2D and 3D domains. However, the effectiveness of these methods for improving segmentation results heavily depends on the specific data-sets at hand.
				
				Due to reports on the segmentation performance of SAM  on volumetric medical data sets, such as those in \cite{73-6}, and our own preliminary experiments, which suggested that the segmentation quality of SAM was likely to be mixed, we opted for a simple majority voting approach to merge the 2D predictions into 3D volumes.
				
				During the slice merging process, we experimented with different rules to determine when to terminate the slice-wise merging. We either combined all slice within the current field of view regardless of their content or stopped at the first empty slice, i.e., a slice without foreground voxels. We also tested various rules based on different thresholds of overlap or Intersection over Union (IoU) between the proposed segmentation of the current slice and the preceding slice or a foreground volume obtained through global Otsu thresholding followed by a morphological closing step. 
				
				As an optimization strategy, the slice-wise prediction was performed in an alternating manner, starting from the centre of the current sub-volume and moving outward slice-wise in both directions. This approach was implemented to save computational time and prevent the segmentation of unconnected segments, ensuring that only cohesive regions were accurately identified.
				
				In situations where the segmentation results in an identification of unconnected segments, the algorithm may inadvertently continue segmenting entire regions composed of non-cohesive segments. This phenomenon occurs when the segmentation quality is significantly compromised. During the subsequent hyperparameter search, we also permitted segmentations without applying these rules. However, it appears that these deviations have only minimal impact on the output quality.				
				
				Subsequently, a new target volume is constructed. Voxels are included in the output volume if they are segmented as foreground in at least one, and depending on the configuration, up to three slice-wise predictions.					
				
				Additionally, we employed postprocessing techniques such as slice-wise and volume-based median filtering and CCA prior and after merging the slices into volumes to smooth scattered and miss segmented voxels. 
				
				We also conducted experiments with different variants of SAM's outputs. Since SAM has the ability to generate multiple outputs per prompt, such as e.g. separating a backpack from a person wearing it, we investigated whether selecting any of these outputs could improve the segmentation quality. Specifically, we examined whether it is better for volumetric segmentation to use the segmentation proposal provided by SAM with the highest probable IoU or the one with the maximum IoU of the approximated foreground volume. Additionally, as SAM often tends to under-segment and include background or neighboring segments as part of the foreground, we investigated whether selecting the output with the smallest count of voxels among the multiple outputs would improve the segmentation quality.
				
				In this context, experiments were conducted using both, the binarized output of SAM and the raw probability values, which are available at a lower resolution than the binary mask. After upscaling, different threshold values can be applied to the probability outputs for further processing and experimentation.
				
			\subsubsection{Tile based segmentation for data-sets of arbitrary size}\label{sec:methods-Tile-based-segmentation-for-datasets-of-arbitrary-size}
\begin{figure}
					\definecolor{tikzBoxColourSliceBlue}{RGB}{57,193,205}
					\definecolor{tikzBoxColourSliceGreen}{RGB}{185,202,50}
					\definecolor{tikzBoxColourSliceOrange}{RGB}{245,130,32}
					\DeclareRobustCommand{\tikzBoxSliceBlue}{\tikz\node[rectangle,fill=tikzBoxColourSliceBlue,minimum width=\tikzBoxSize,minimum height = \tikzBoxSize,] (r) at (0,0) {};}					
					\DeclareRobustCommand{\tikzBoxSliceGreen}{\tikz\node[rectangle,fill=tikzBoxColourSliceGreen,minimum width=\tikzBoxSize,minimum height = \tikzBoxSize,] (r) at (0,0) {};}					
					\DeclareRobustCommand{\tikzBoxSliceOrange}{\tikz\node[rectangle,fill=tikzBoxColourSliceOrange,minimum width=\tikzBoxSize,minimum height = \tikzBoxSize,] (r) at (0,0) {};}
					\centering
					\includegraphics[width=0.40\textwidth,height=1.0\textheight,keepaspectratio]{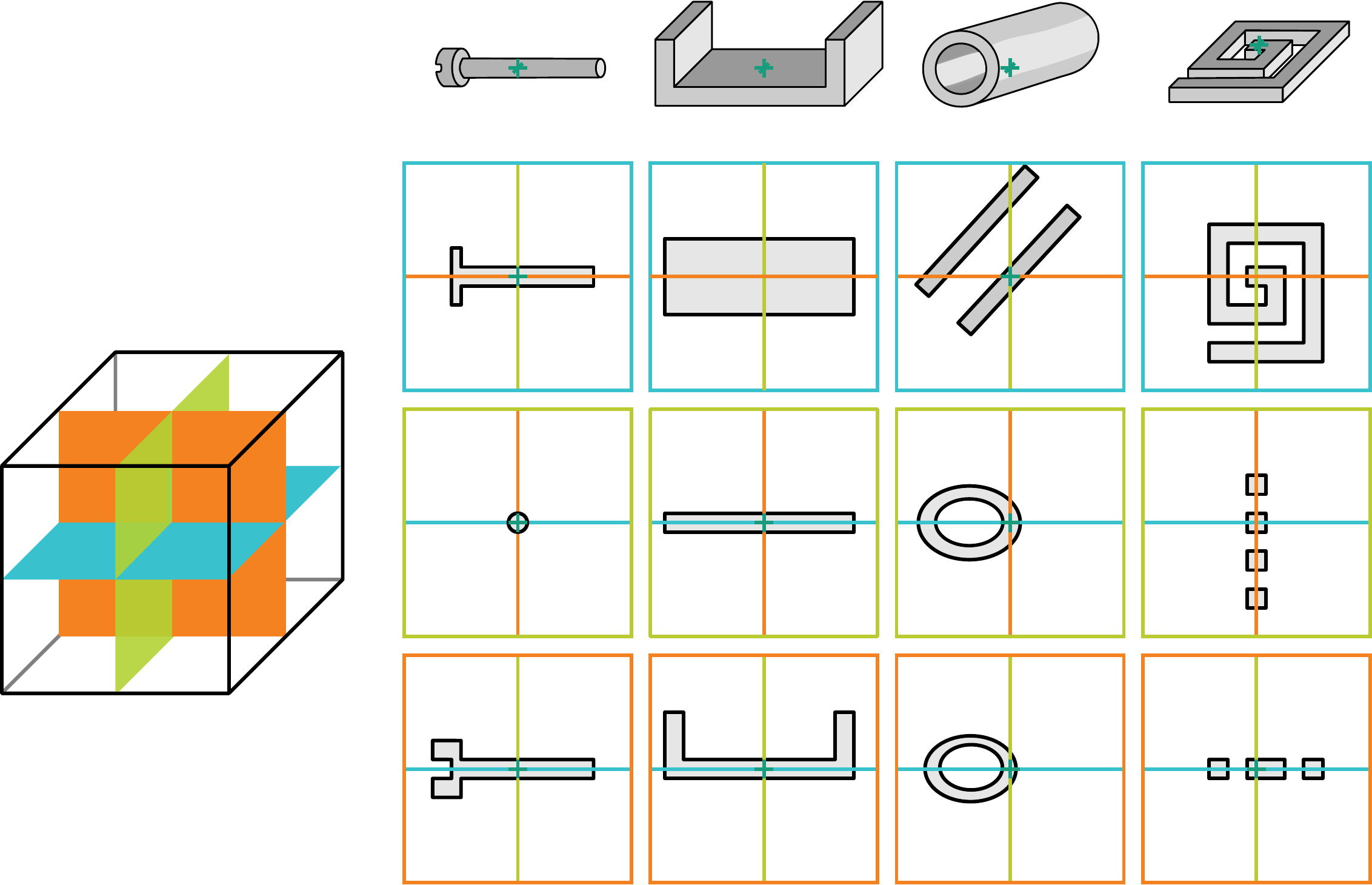}
					\caption{\label{fig:methods-volumetricSAM-obstacles}Schematic views of multiple simple volumetric objects (bolt, U-profile, pipe, and spiral spring) and cross-sectional slices along their central axes in three orthogonal directions marked by three respective colours (\tikzBoxSliceBlue, \tikzBoxSliceGreen, \tikzBoxSliceOrange). The disjunction of simple objects into multiple components if processed slice-wise poses a challenge, as there are no straightforward rules for merging them without a step-by-step traversal of the object.}
				\end{figure}

				Due to SAM's image-based nature, we encounter segmentation challenges when dealing with topologically complex objects depicted by volumentric CT NDT data-sets. These volumes may contain holes or inclusions, complex folds, are spatially sparse, or may extend beyond the boundaries of the currently processed tile. Figure \ref{fig:methods-volumetricSAM-obstacles} displays several schematic examples of different complexity. In volumetric data-sets, such complex segments are easier to understand, but when segmenting them slice by slice there is a risk of mistakenly delineating them as multiple segments. This effect also occurs when the tile is smaller than the entity's size.
				
				To overcome these challenges, we utilize the volume-based SAM inference (see Section \ref{sec:methods-volumetricSAM}) within the FFN framework (see Section \ref{sec:introduction-FFN}). The inference process starts with a single seed point and is applied to a small sub-volume tile. The resulting segmentation proposal is then stored in an result buffer, the accumulator volume. If a segment intersects the outer boundaries of a tile, the intersection position is added to a queue. In subsequent iterations, corresponding slightly shifted tiles aimed at theses intersection points are processed by the volume-based SAM inference. This iterative process generates segmentation proposals which are incorporated into the accumulator. This process repeats until the intersection points queue is empty and the segmentation proposal in the accumulator is no longer constrained by the boundary of the processed tiles.
				
				As an optimization step, the proposed additional intersection positions are filtered based on the approximated foreground volume. They are added to the intersection points queue only if the corresponding voxels have a high probability to be foreground voxels.
				
				The proposed combination of SAM and FFN allows to compute segments and input volumes of arbitrary size by combining multiple overlapping tiles using an temporary accumulator temporary volume. Nevertheless this approach also increases the runtime due to the recomputation of the overlapping tiles. 		
				
				The choice of using 48\,voxels per tile side was made heuristically based on the original FFN algorithm, which also uses this tile size. However, the algorithm can be adjusted by changing the tile size up to 1024\,voxels in each dimension, the maximum dimension SAM can handle without resizing the input. When the tile size is below this threshold, no resizing of tiles is required as we add a constant value border around the tile. Additionally, the step width between tiles and the overlap of the tiles can be adjusted to mitigate artefacts caused by the tile-based algorithm. Tile-based algorithms are capable of assembling entities with complex topologies. These algorithms can follow or trace the segment itself over multiple tiles and steps, even if it forms highly complex shapes. But tile-based algorithms may introduce additional artefacts. The segmentation result of the combined algorithms is heavily dependent on the performance of the SAM segmentation.
				
			\subsubsection{Prompt Selection and Accumulator Integration}
				As mentioned above SAM allows queries using various prompts such as point prompts (seed points, bounding boxes) and dense prompts (masks, brushes). Multiple studies \cite{73, 75} have shown that, depending on the input data, higher segmentation quality can be achieved by using multiple prompts, such as point prompts distributed evenly over the segment region or negative point prompts, which are not considered part of the segment. Additionally, the use of rectangular prompts consisting of two anchor points often leads to adequate segmentation results. 
				
				Given that the main objective of this study is to evaluate the applicability of SAM in the automated NDT domain, we have opted to solely assess single point prompts and dense prompts, as they can be easily automated.
				
				We placed a single point prompt at the exact centre of the tile. The centre point of a tile was either chosen by a seed point or deemed highly likely to belong to the current segment, due to the iterative processing of the tiles.
				
\begin{figure*}				
					\DeclareRobustCommand{\tikzBoxFovGreen}{\tikz\node[rectangle,fill=fhg,minimum width=\tikzBoxSize,minimum height = \tikzBoxSize,] (r) at (0,0) {};}	
					
					\definecolor{tikzBoxColourFovBlue}{RGB}{57,193,205}				
					\DeclareRobustCommand{\tikzBoxFovBlue}{\tikz\node[rectangle,fill=tikzBoxColourFovBlue,minimum width=\tikzBoxSize,minimum height = \tikzBoxSize,] (r) at (0,0) {};}	
					
					\centering
					\includegraphics[width=1.00\textwidth,height=1.0\textheight,keepaspectratio]{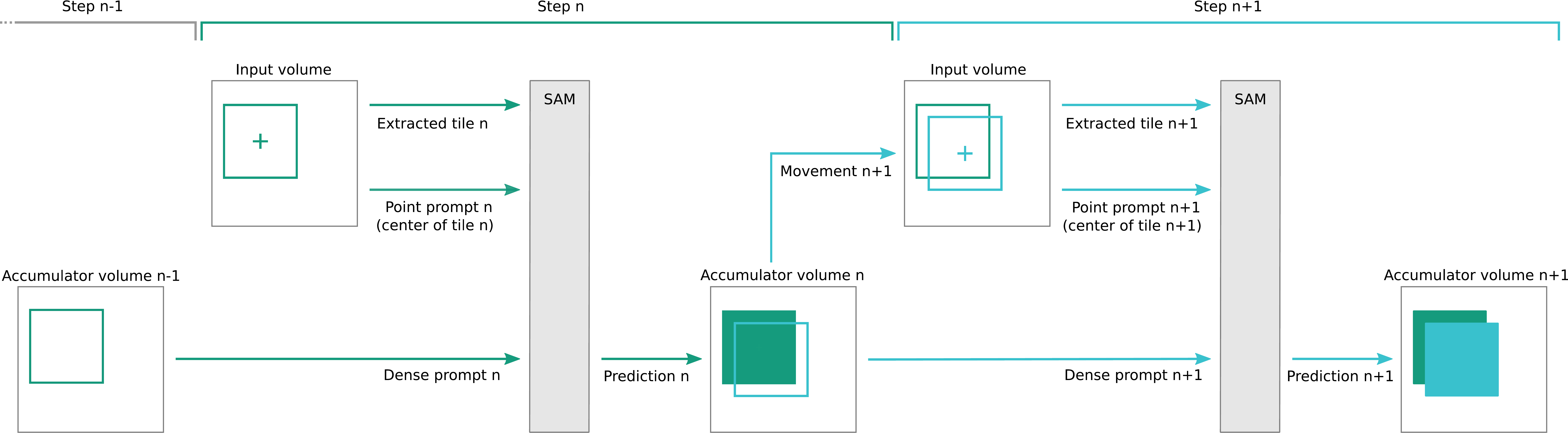}
					\caption{\label{fig:methods-inference-dense}
					Schematic view of two subsequent inference steps, denoted as $n$ (represented by \tikzBoxFovGreen) and $n+1$ (represented by \tikzBoxFovBlue), which use the modified accumulator volume from the previous step to create a dense SAM prompt.	
					In step $n$, the content of the accumulator volume of the previous step $n-1$ is used to generate a dense SAM prompt $n+1$. This prompt, along with the point prompt $n$ and the extracted input volume tile $n$, is used by SAM to compute prediction $n$. Subsequently, the accumulator volume is updated to the state $n$ based on this prediction.	
					In the subsequent step $n+1$, the accumulator volume $n$ is used to determine the movement $n+1$ to the tile $n+1$. Tile $n+1$ significantly overlaps with tile $n$. SAM is parametrized with the extracted input volume tile $n+1$, point prompt $n+1$, and dense prompt $n+1$ to compute prediction $n+1$. Which is used to update the accumulator volume $n+1$.}
				\end{figure*}

				For dense prompts, we utilized the SAM output stored in the accumulator, which was shifted by the relative position of the current point prompt. This requires SAM to complete the segmentation proposal at the edge of the current tile. Since our tile step size was $[1,20]$ voxels, the overlap between the tiles and the dense prompt with the expected segmentation proposal was high, allowing SAM to only predict a relative slim border of new voxels. Figure \ref{fig:methods-inference-dense} illustrates an idealized schematic of such an operation. In the case of dense prompts, we also include a corresponding point prompt at the centre of the tile as more prompts tend to increase segmentation performance \cite{73}.

	\section{Results}
		\subsection{Evaluation of SAM Segmentation Quality in NDT Slice Data-sets}\label{sec:results-sam-slices}
\begin{figure}
				\centering
			
				\begin{subfigure}[t]{0.235\columnwidth}
					\centering
					\fbox{\includegraphics[width=\linewidth, interpolate=false]{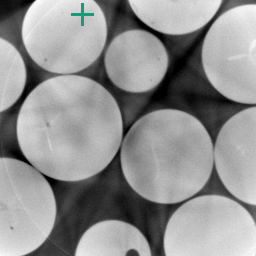}}
					\caption{Input}
					\label{fig:results-evalLoss-marbles-102-input-cross}
				\end{subfigure}
				\begin{subfigure}[t]{0.235\columnwidth}
					\centering
					\fbox{\includegraphics[width=\linewidth, interpolate=false]{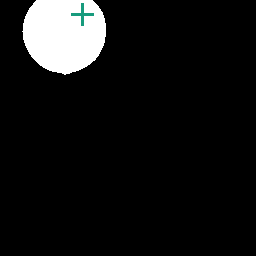}}
					\caption{Reference}
					\label{fig:results-evalLoss-marbles-102-reference-mask-cross}
				\end{subfigure}
				\begin{subfigure}[t]{0.235\columnwidth}
					\centering
					\fbox{\includegraphics[width=\linewidth, interpolate=false]{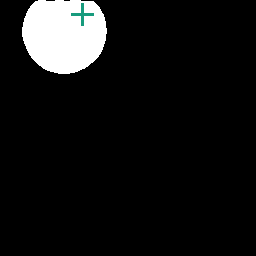}}
					\caption{vit\_b \\loss=0.02} 
					\label{fig:results-evalLoss-marbles-102-vit_b-mask-cross}
				\end{subfigure}
				\begin{subfigure}[t]{0.235\columnwidth}
					\centering
					\fbox{\includegraphics[width=\linewidth, interpolate=false]{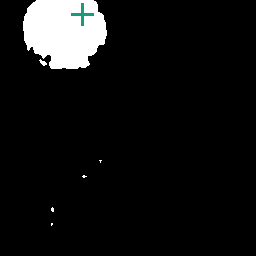}}
					\caption{\fineTunedModel \\loss=0.04} 
					\label{fig:results-evalLoss-marbles-102-vit_b-439-mask-cross}
				\end{subfigure}\\				
				\begin{subfigure}[t]{0.235\columnwidth}
					\centering
					\fbox{\includegraphics[width=\linewidth, interpolate=false]{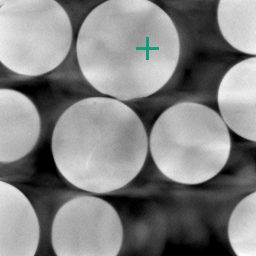}}
					\caption{Input}
					\label{fig:results-evalLoss-marbles-077-input-cross}
				\end{subfigure}
				\begin{subfigure}[t]{0.235\columnwidth}
					\centering
					\fbox{\includegraphics[width=\linewidth, interpolate=false]{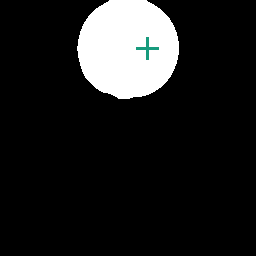}}
					\caption{Reference}
					\label{fig:results-evalLoss-marbles-077-reference-mask-cross}
				\end{subfigure}
				\begin{subfigure}[t]{0.235\columnwidth}
					\centering
					\fbox{\includegraphics[width=\linewidth, interpolate=false]{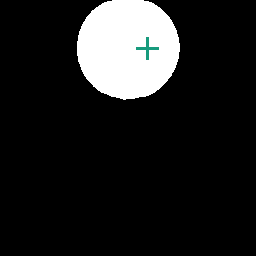}}
					\caption{vit\_b \\loss=0.01} 
					\label{fig:results-evalLoss-marbles-077-vit_b-mask-cross}
				\end{subfigure}
				\begin{subfigure}[t]{0.235\columnwidth}
					\centering
					\fbox{\includegraphics[width=\linewidth, interpolate=false]{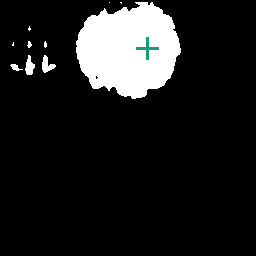}}
					\caption{\fineTunedModel \\loss=0.05} 
					\label{fig:results-evalLoss-marbles-077-vit_b-439-mask-cross}
				\end{subfigure}\\
				\begin{subfigure}[t]{0.235\columnwidth}
					\centering
					\fbox{\includegraphics[width=\linewidth, interpolate=false]{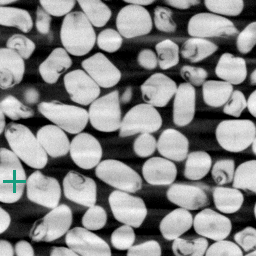}}
					\caption{Input}
					\label{fig:results-evalLoss-corn-052-input-cross}
				\end{subfigure}
				\begin{subfigure}[t]{0.235\columnwidth}
					\centering
					\fbox{\includegraphics[width=\linewidth, interpolate=false]{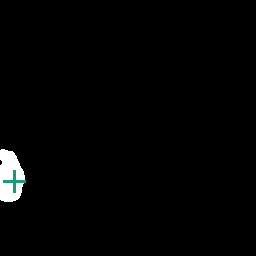}}
					\caption{Reference}
					\label{fig:results-evalLoss-corn-052-reference-mask-cross}
				\end{subfigure}
				\begin{subfigure}[t]{0.235\columnwidth}
					\centering
					\fbox{\includegraphics[width=\linewidth, interpolate=false]{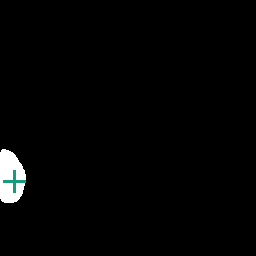}}
					\caption{vit\_b \\loss=0.04} 
					\label{fig:results-evalLoss-corn-052-vit_b-mask-cross}
				\end{subfigure}
				\begin{subfigure}[t]{0.235\columnwidth}
					\centering
					\fbox{\includegraphics[width=\linewidth, interpolate=false]{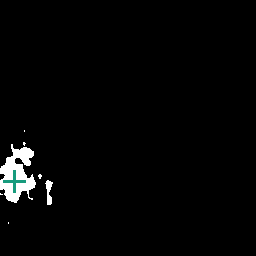}}
					\caption{\fineTunedModel \\loss=0.29} 
					\label{fig:results-evalLoss-corn-052-vit_b-439-mask-cross}
				\end{subfigure}\\							
				\begin{subfigure}[t]{0.235\columnwidth}
					\centering
					\fbox{\includegraphics[width=\linewidth, interpolate=false]{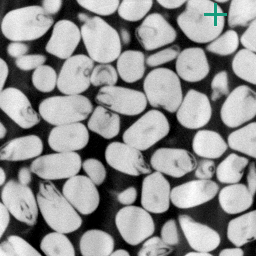}}
					\caption{Input}
					\label{fig:results-evalLoss-corn-102-input-cross}
				\end{subfigure}
				\begin{subfigure}[t]{0.235\columnwidth}
					\centering
					\fbox{\includegraphics[width=\linewidth, interpolate=false]{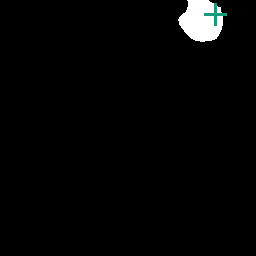}}
					\caption{Reference}
					\label{fig:results-evalLoss-corn-102-reference-mask-cross}
				\end{subfigure}
				\begin{subfigure}[t]{0.235\columnwidth}
					\centering
					\fbox{\includegraphics[width=\linewidth, interpolate=false]{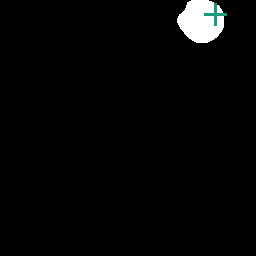}}
					\caption{vit\_b \\loss=0.04} 
					\label{fig:results-evalLoss-corn-102-vit_b-mask-cross}
				\end{subfigure}
				\begin{subfigure}[t]{0.235\columnwidth}
					\centering
					\fbox{\includegraphics[width=\linewidth, interpolate=false]{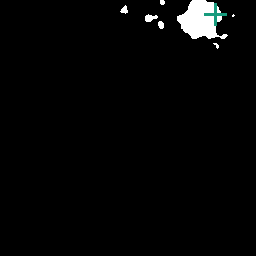}}
					\caption{\fineTunedModel \\loss=0.12} 
					\label{fig:results-evalLoss-corn-102-vit_b-439-mask-cross}
				\end{subfigure}\\
				\begin{subfigure}[t]{0.235\columnwidth}
					\centering
					\fbox{\includegraphics[width=\linewidth, interpolate=false]{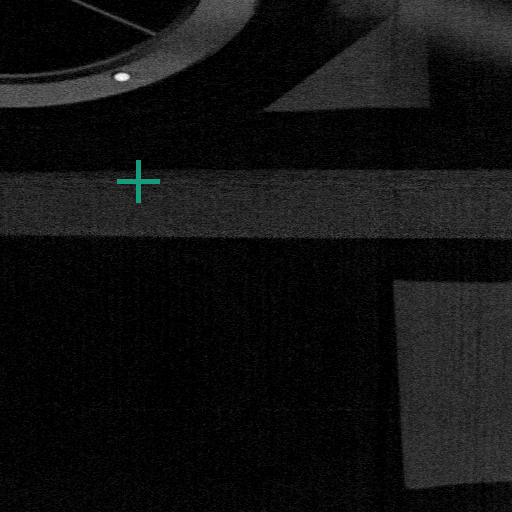}}
					\caption{Input}
					\label{fig:results-evalLoss-me163-011-input-cross}
				\end{subfigure}
				\begin{subfigure}[t]{0.235\columnwidth}
					\centering
					\fbox{\includegraphics[width=\linewidth, interpolate=false]{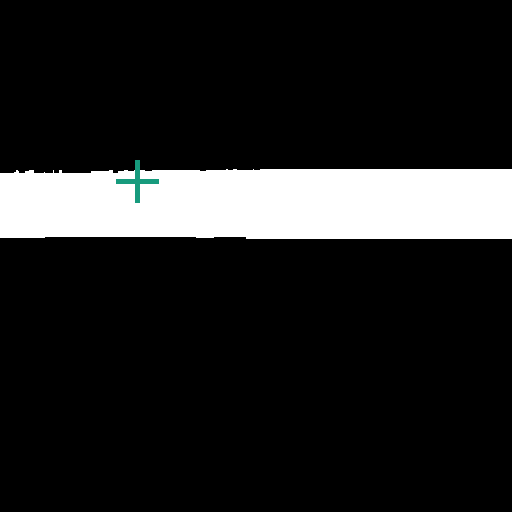}}
					\caption{Reference}
					\label{fig:results-evalLoss-me163-011-reference-mask-cross}
				\end{subfigure}
				\begin{subfigure}[t]{0.235\columnwidth}
					\centering
					\fbox{\includegraphics[width=\linewidth, interpolate=false]{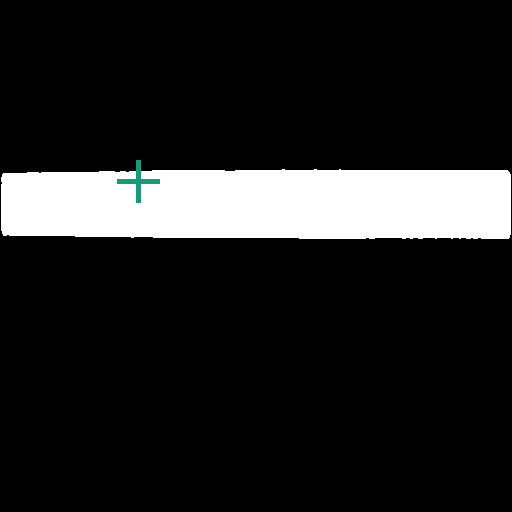}}
					\caption{vit\_b \\loss=0.01} 
					\label{fig:results-evalLoss-me163-011-vit_b-mask-cross}
				\end{subfigure}
				\begin{subfigure}[t]{0.235\columnwidth}
					\centering
					\fbox{\includegraphics[width=\linewidth, interpolate=false]{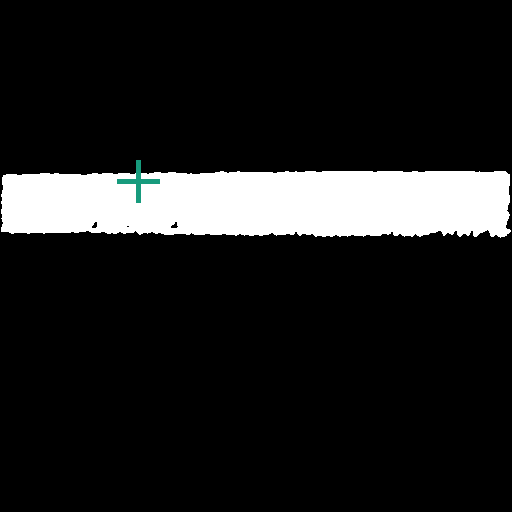}}
					\caption{\fineTunedModel \\loss=0.44} 
					\label{fig:results-evalLoss-me163-011-vit_b-439-mask-cross}
				\end{subfigure}\\				
				\begin{subfigure}[t]{0.235\columnwidth}
					\centering
					\fbox{\includegraphics[width=\linewidth, interpolate=false]{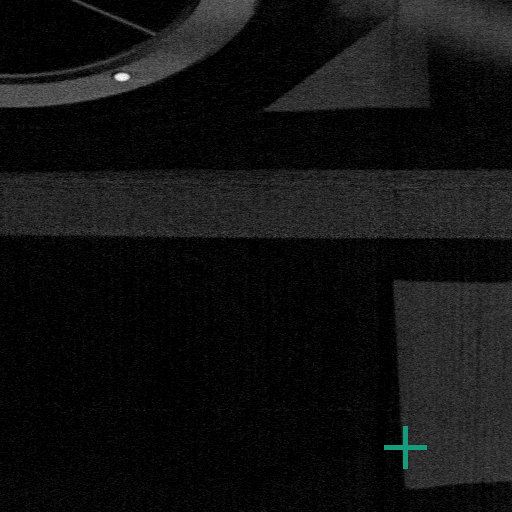}}
					\caption{Input}
					\label{fig:results-evalLoss-me163-080-input-cross}
				\end{subfigure}
				\begin{subfigure}[t]{0.235\columnwidth}
					\centering
					\fbox{\includegraphics[width=\linewidth, interpolate=false]{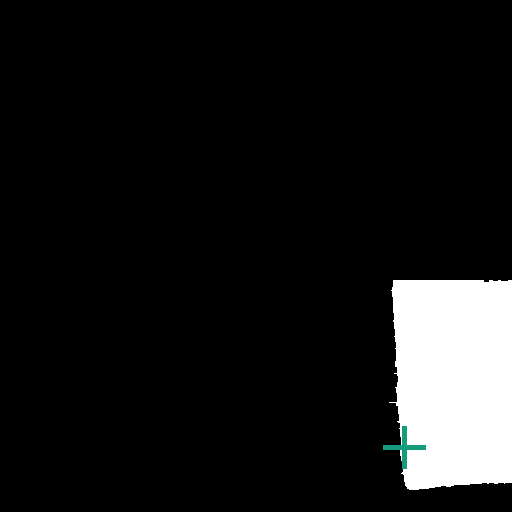}}
					\caption{Reference}
					\label{fig:results-evalLoss-me163-080-reference-mask-cross}
				\end{subfigure}
				\begin{subfigure}[t]{0.235\columnwidth}
					\centering
					\fbox{\includegraphics[width=\linewidth, interpolate=false]{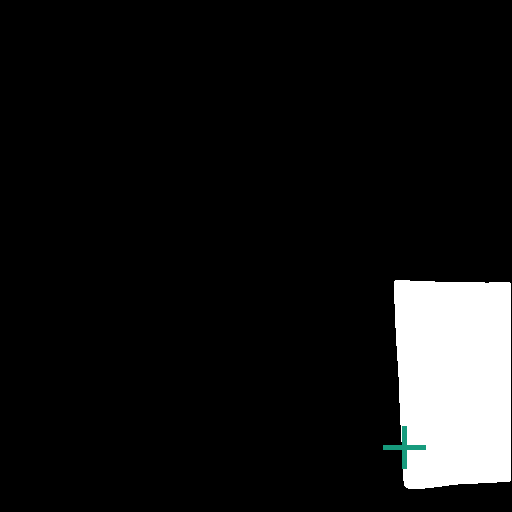}}
					\caption{vit\_b \\loss=0.01} 
					\label{fig:results-evalLoss-me163-080-vit_b-mask-cross}
				\end{subfigure}
				\begin{subfigure}[t]{0.235\columnwidth}
					\centering
					\fbox{\includegraphics[width=\linewidth, interpolate=false]{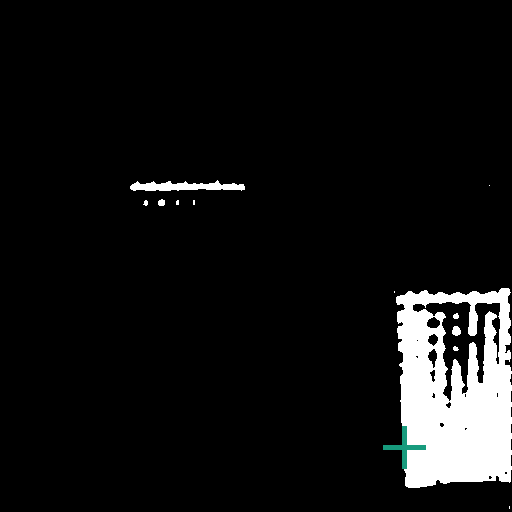}}
					\caption{\fineTunedModel \\loss=0.13} 
					\label{fig:results-evalLoss-me163-080-vit_b-439-mask-cross}
				\end{subfigure}			
				\caption{\label{fig:results-evalLoss-goodExamples}Segmented examples of the corn and marbles data-set. The green crosses mark the position of the currently used point prompt. The last column depicts the result of the vit\_b model which was fine-tuned on the Me\,163 data-set.}
			\end{figure}

			In an initial test of SAM's segmentation quality for CT NDT data, we applied SAM to segment individual slices from NDT volumetric data-sets. For each of the three data-sets introduced in Section \ref{sec:methods-datasets} randomly selected slices were selected and segmented, which accounted for approximately 0.5\% of all available validation data-sets. Each example underwent the preparation steps outlined in Section \ref{sec:methods-fineTune} before being processed by SAM. SAM then tried to segment the entity located at the exact centre of each slice using point prompts. Examples of typical segments can be seen in Figure \ref{fig:results-evalLoss-goodExamples}. Notably, SAM demonstrated good segmentation performance for the marbles and corn kernels data-sets, while the segmentation quality was significantly inferior for the individual segments of the Me\,163 data-set. Table \ref{tab:results-evalLoss-LossOnDifferentDatasetsAndModels} provides a summary of the statistics obtained from the conducted experiments, categorized by data-set and the model used.
				
			\begin{figure}[t]
				\centering
				\captionsetup[subfigure]{justification=justified,singlelinecheck=false}
				\begin{subfigure}[t]{0.235\columnwidth}
					\centering
					\fbox{\includegraphics[width=\linewidth, interpolate=false]{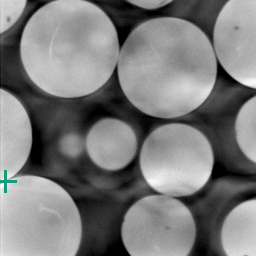}}
					\caption{Input}
					\label{fig:results-evalLoss-errorCase-marbles-008-input-cross}
				\end{subfigure}
				\begin{subfigure}[t]{0.235\columnwidth}
					\centering
					\fbox{\includegraphics[width=\linewidth, interpolate=false]{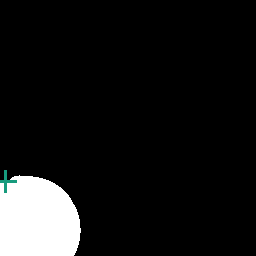}}
					\caption{Reference}
					\label{fig:results-evalLoss-errorCase-marbles-008-reference-mask-cross}
				\end{subfigure}
				\begin{subfigure}[t]{0.235\columnwidth}
					\centering
					\fbox{\includegraphics[width=\linewidth, interpolate=false]{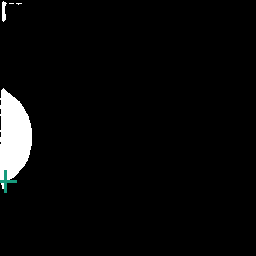}}
					\caption{vit\_b \\loss=0.93} 
					\label{fig:results-evalLoss-errorCase-marbles-008-vit_b-mask-cross}
				\end{subfigure}
				\begin{subfigure}[t]{0.235\columnwidth}
					\centering
					\fbox{\includegraphics[width=\linewidth, interpolate=false]{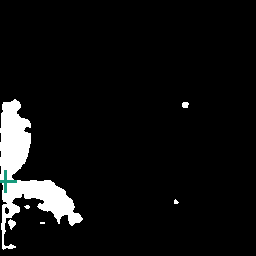}}
					\caption{\fineTunedModel \\loss=0.50} 
					\label{fig:results-evalLoss-errorCase-marbles-008-vit_b-439-mask-cross}
				\end{subfigure}	
				\vspace{\baselineskip}					
				\begin{subfigure}[t]{0.235\columnwidth}
					\centering
					\fbox{\includegraphics[width=\linewidth, interpolate=false]{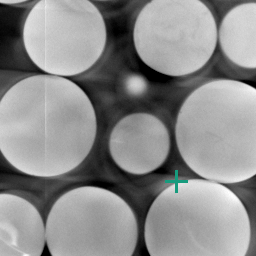}}
					\caption{Input}
					\label{fig:results-evalLoss-errorCase-marbles-047-input-cross}
				\end{subfigure}
				\begin{subfigure}[t]{0.235\columnwidth}
					\centering
					\fbox{\includegraphics[width=\linewidth, interpolate=false]{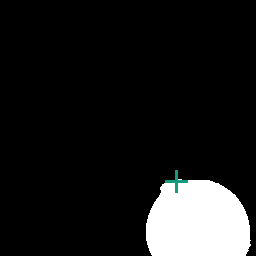}}
					\caption{Reference}
					\label{fig:results-evalLoss-errorCase-marbles-047-reference-mask-cross}
				\end{subfigure}
				\begin{subfigure}[t]{0.235\columnwidth}
					\centering
					\fbox{\includegraphics[width=\linewidth, interpolate=false]{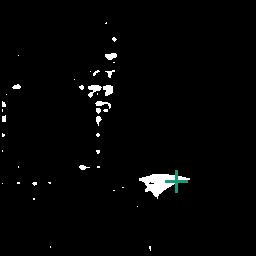}}
					\caption{vit\_b \\loss=0.50}  
					\label{fig:results-evalLoss-errorCase-corn-047-vit_b-mask-cross}
				\end{subfigure}
				\begin{subfigure}[t]{0.235\columnwidth}
					\centering
					\fbox{\includegraphics[width=\linewidth, interpolate=false]{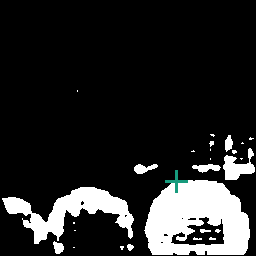}}
					\caption{\fineTunedModel \\loss=0.26} 
					\label{fig:results-evalLoss-errorCase-marbles-047-vit_b-439-mask-cross}
				\end{subfigure}
				\caption{\label{fig:results-evalLoss-errorCase-marbles}Error cases for the marble data-set. Here the reference segmentation which generated by an connected component analysis is erroneous. In Figure \ref{fig:results-evalLoss-errorCase-marbles-008-reference-mask-cross} the point prompt (marked with an green cross) lies on the boundary of two marbles and vit\_b segments the upper marble instead of the lower marble. While in Figure \ref{fig:results-evalLoss-errorCase-marbles-047-reference-mask-cross} the point prompt lies inside an artefact region.}
			\end{figure}
			
			\begin{figure}[t]
				\centering					
				\begin{subfigure}[t]{0.235\columnwidth}
					\centering
					\fbox{\includegraphics[width=\linewidth, interpolate=false]{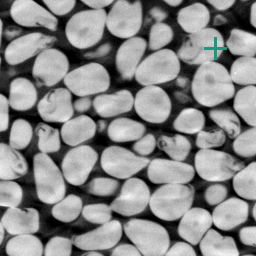}}
					\caption{Input}
					\label{fig:results-evalLoss-errorCase-corn-083-input-cross}
				\end{subfigure}
				\begin{subfigure}[t]{0.235\columnwidth}
					\centering
					\fbox{\includegraphics[width=\linewidth, interpolate=false]{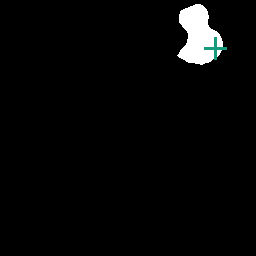}}
					\caption{Reference}
					\label{fig:results-evalLoss-errorCase-corn-083-reference-mask-cross}
				\end{subfigure}
				\begin{subfigure}[t]{0.235\columnwidth}
					\centering
					\fbox{\includegraphics[width=\linewidth, interpolate=false]{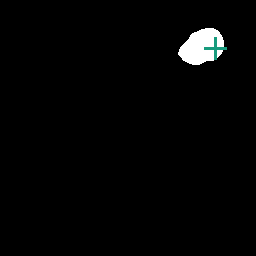}}
					\caption{vit\_b \\loss=0.26} 
					\label{fig:results-evalLoss-errorCase-corn-083-vit_b-mask-cross}
				\end{subfigure}
				\begin{subfigure}[t]{0.235\columnwidth}
					\centering
					\fbox{\includegraphics[width=\linewidth, interpolate=false]{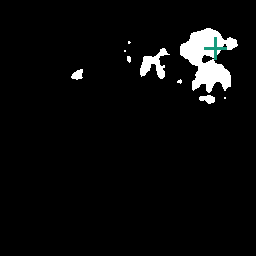}}
					\caption{\fineTunedModel \\loss=0.50} 
					\label{fig:results-evalLoss-errorCase-corn-083-vit_b-439-mask-cross}
				\end{subfigure}	
				\vspace{\baselineskip}			
				\begin{subfigure}[t]{0.235\columnwidth}
					\centering
					\fbox{\includegraphics[width=\linewidth, interpolate=false]{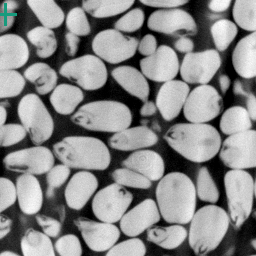}}
					\caption{Input}
					\label{fig:results-evalLoss-errorCase-corn-022-input-cross}
				\end{subfigure}
				\begin{subfigure}[t]{0.235\columnwidth}
					\centering
					\fbox{\includegraphics[width=\linewidth, interpolate=false]{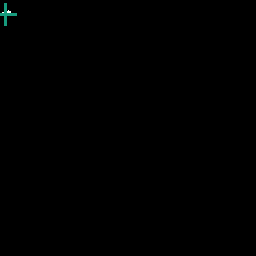}}
					\caption{Reference}
					\label{fig:results-evalLoss-errorCase-corn-022-reference-mask-cross}
				\end{subfigure}
				\begin{subfigure}[t]{0.235\columnwidth}
					\centering
					\fbox{\includegraphics[width=\linewidth, interpolate=false]{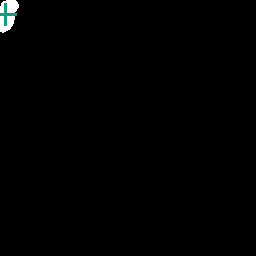}}
					\caption{vit\_b \\loss=0.90} 
					\label{fig:results-evalLoss-errorCase-corn-022-vit_b-mask-cross}
				\end{subfigure}
				\begin{subfigure}[t]{0.235\columnwidth}
					\centering
					\fbox{\includegraphics[width=\linewidth, interpolate=false]{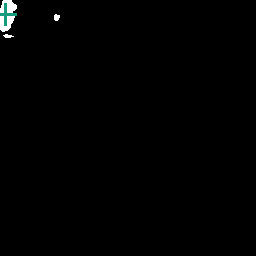}}
					\caption{\fineTunedModel \\loss=0.88} 
					\label{fig:results-evalLoss-errorCase-corn-022-vit_b-439-mask-cross}
				\end{subfigure}			
				\caption{\label{fig:results-evalLoss-errorCase-corn}Error cases of the corn data-set. In the first case in Figure \ref{fig:results-evalLoss-errorCase-corn-083-reference-mask-cross} two kernels have been erroneously segmented together in the reference segmentation. In contrast in Figure \ref{fig:results-evalLoss-errorCase-corn-022-reference-mask-cross} the reference segmentation only appears erroneous as the current slice only depicts one voxel. The next slice in the input volume contains the kernel this voxel belongs to. The green crosses mark the position of the currently used point prompt.}
			\end{figure}				

			\begin{table*}
				\centering	
				\begin{tabular}{lccc}
					\toprule
					  										& Marbles 				& Corn					& Me\,163 \\
					\midrule
					vit\_h   								& \textbf{0.03} (0.06) 	& 0.11 (0.10) 			& 0.49 (0.34) \\
					vit\_l   								& \textbf{0.03} (0.06) 	& 0.11 (0.10)			& 0.46 (0.34) \\
					vit\_b   								& \textbf{0.03} (0.07) 	& \textbf{0.10} (0.10)	& 0.44 (0.32) \\
					vit\_b ForegroundOnly   				& 0.41 (0.27)  			& 0.66 (0.24) 			& 0.49 (0.26) \\
					vit\_b ConstantValueBackground  		& 0.15 (0.10)  			& 0.44 (0.16)			& \textbf{0.36} (0.25) \\
					vit\_b ConnectedComponentBackground   	& 0.51 (0.24)  			& 0.51 (0.19)			& 0.57 (0.23) \\
					\bottomrule
				\end{tabular}
				\caption{\label{tab:results-evalLoss-LossOnDifferentDatasetsAndModels} Mean loss value (and standard deviation) over all slice-wise predictions on the validation data-sets by multiple models for the graphs in Figure~\ref{fig:result-evalLoss-plots}}
			\end{table*}	
		
\begin{figure}
				\centering
				
				\begin{subfigure}[b]{\columnwidth}
					\centering
					\includegraphics[width=\linewidth]{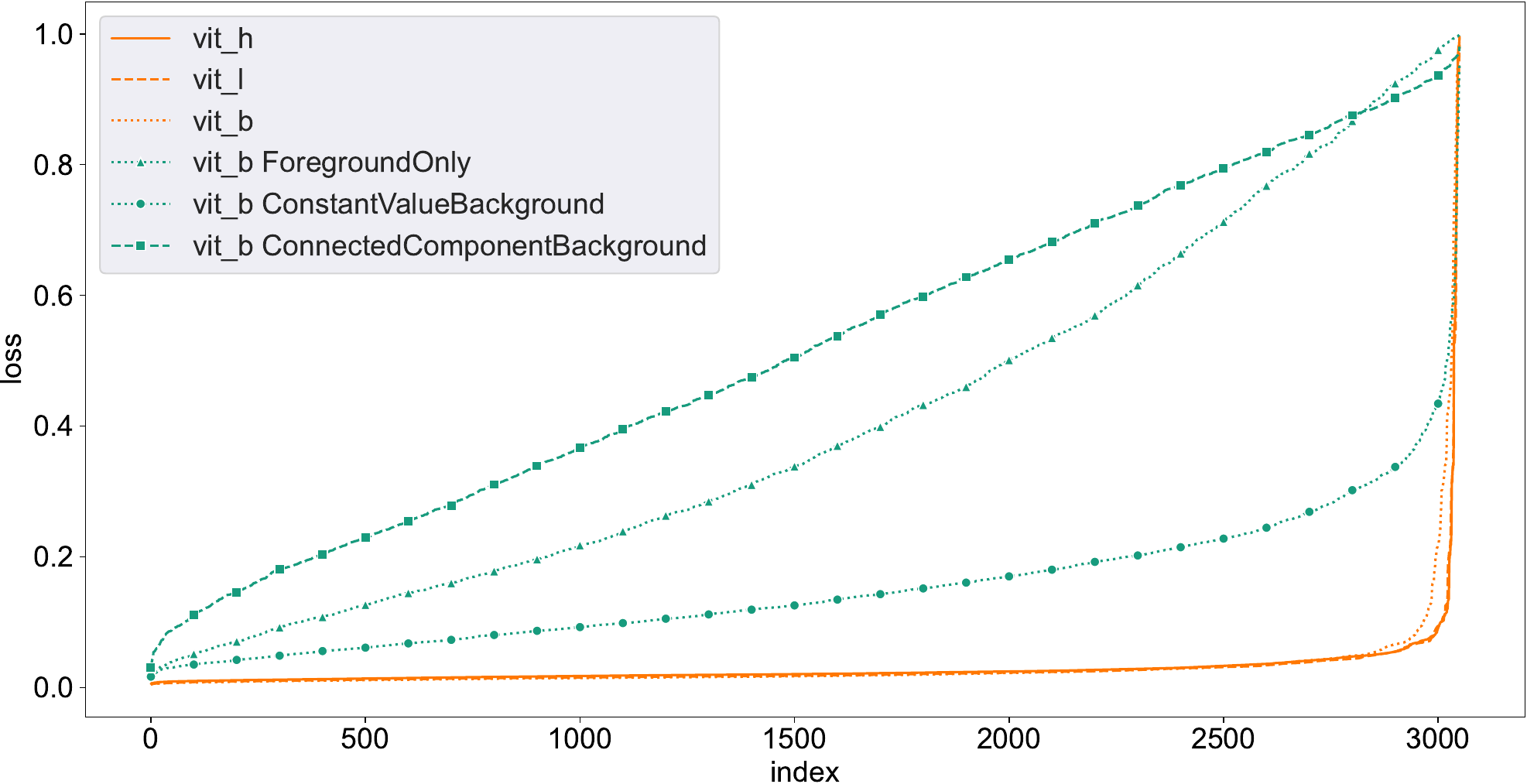}
					\caption{Marbles}
					\label{fig:result-evalLoss-plotMarble}
				\end{subfigure}
				\vspace{1pt} \ 
				
				\begin{subfigure}[b]{\columnwidth}
					\centering
					\includegraphics[width=\linewidth]{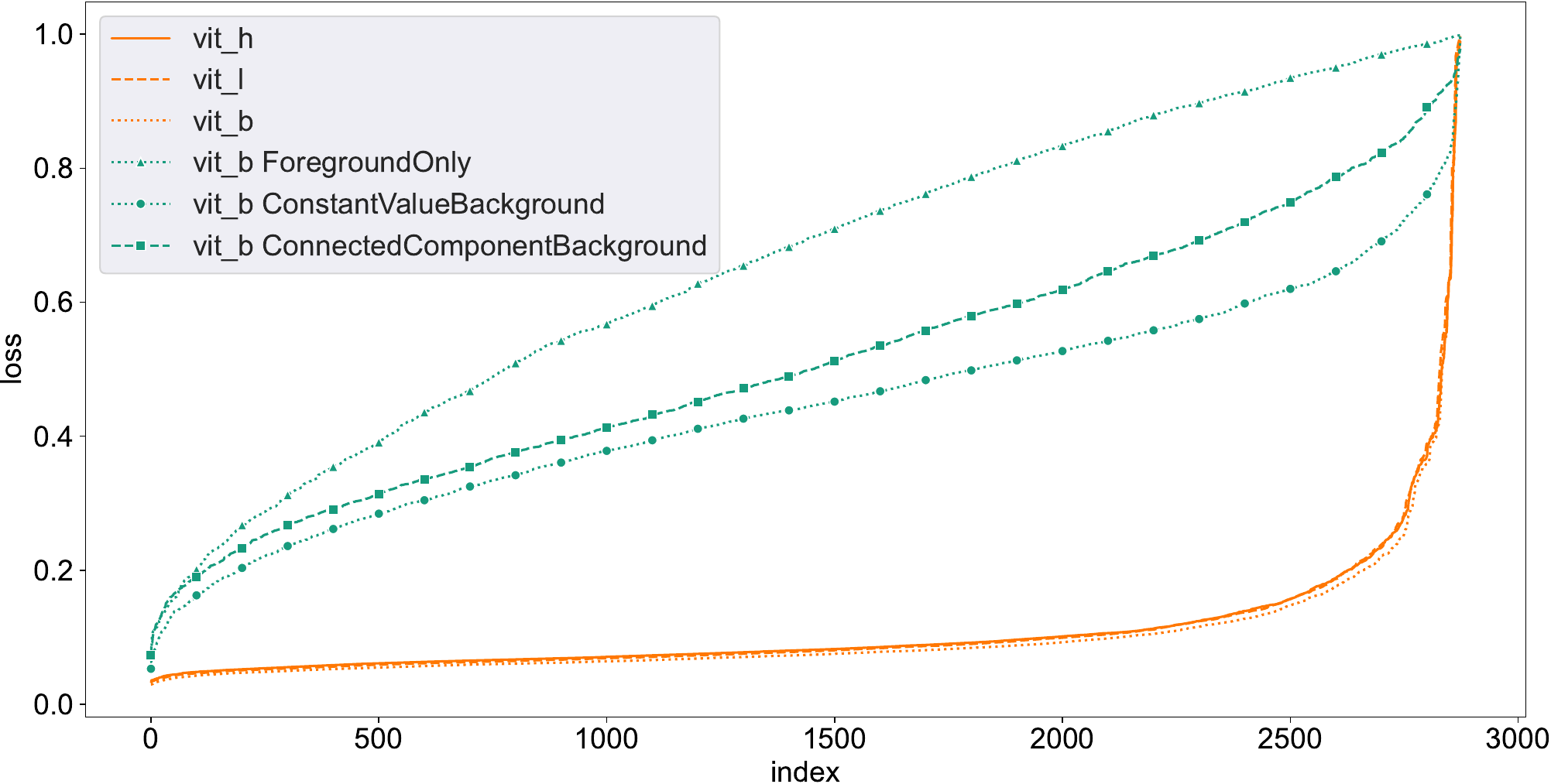}
					\caption{Corn}
					\label{fig:result-evalLoss-plotCorn}
				\end{subfigure}
				\vspace{1pt} \
				
				\begin{subfigure}[b]{\columnwidth}
					\centering
					\includegraphics[width=\linewidth]{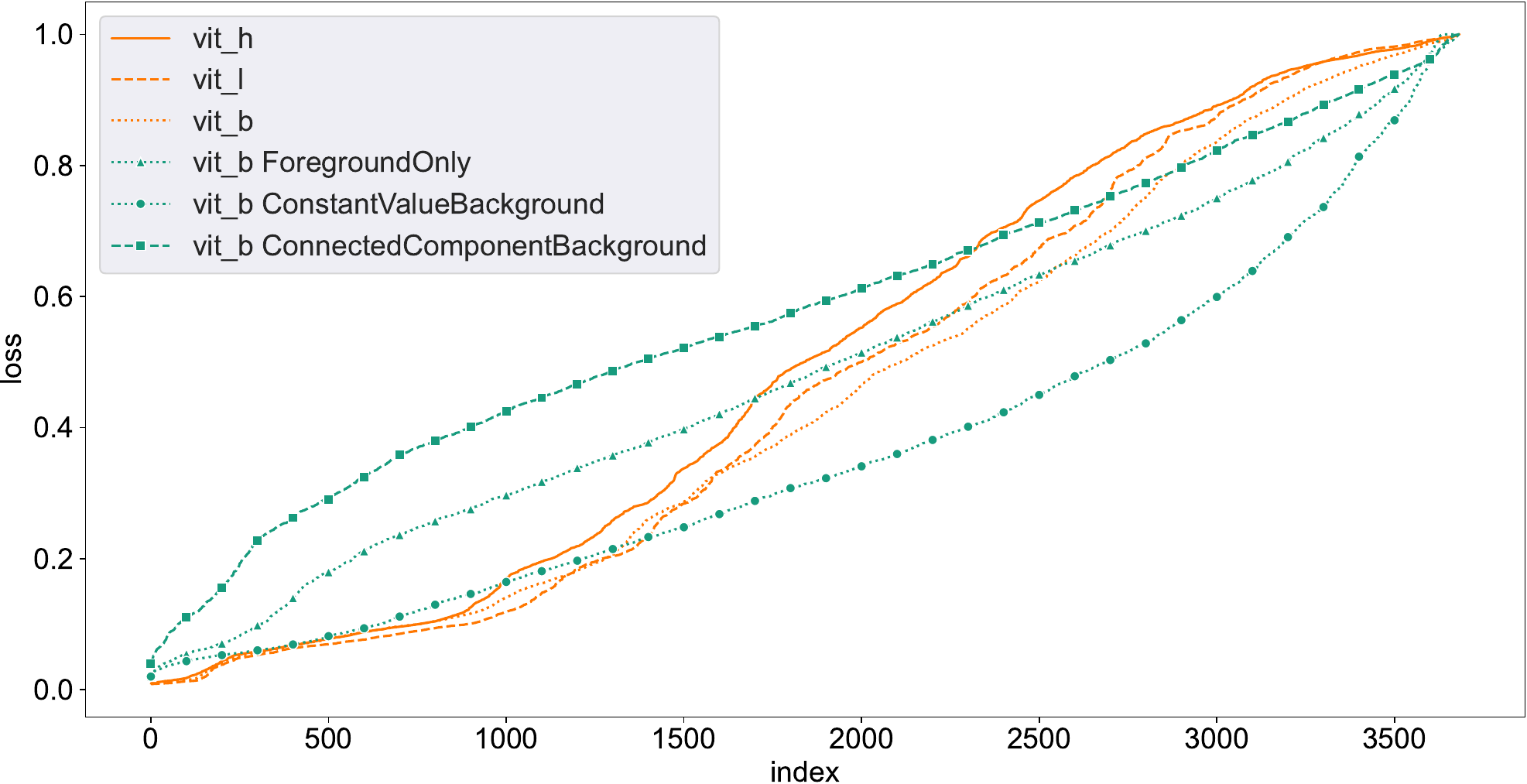}
					\caption{Me\,163}
					\label{fig:result-evalLoss-plotMe163}
				\end{subfigure}
				
				\caption{Graphs depicting the slice segmentation performance of the six evaluated SAM models on the three different testing data-sets. From left to right the index of each segmented slice sorted by their loss value. In an ideal case only an horizontal line close to the loss value of 0 would be visible.}
				\label{fig:result-evalLoss-plots}
			\end{figure}

			Figures \ref{fig:result-evalLoss-plotMarble}, \ref{fig:result-evalLoss-plotCorn}, and \ref{fig:result-evalLoss-plotMe163} demonstrate the segmentation dynamics of the individual models on the different data-sets: These plots represent the loss of the segmentation proposals generated by SAM for the entities at the centre of each layer of the corresponding validation data-set. The loss values are determined with respect to the reference data-set. From left to right the loss values are sorted in ascending order, so that the nearly correctly segmented segments are on the left side of the graph, while the difficult and often incorrectly segmented segments are on the right side. The seed points of the segments were chosen in such a way that each of them corresponds to an foreground voxel, so the networks are not tasked with segmenting the background. The different colours in the plots correspond to different networks.
			
			It can be observed that the unchanged SAM networks perform very well in segmenting the marble and corn data-sets. The few entities which exhibit lower segmentation quality in these data-sets, and are located on the right edge, are often due to insufficient quality in the reference segmentation data-set, as illustrated in Figures \ref{fig:results-evalLoss-errorCase-marbles} and \ref{fig:results-evalLoss-errorCase-corn}. A slightly lower segmentation quality can be observed for the corn data-set, which consists of a higher count of entities that are also not as homogeneous in colour compared to the marble data-set.
		    
\begin{figure}
				\centering
				\begin{subfigure}[t]{0.235\columnwidth}
					\centering
					\fbox{\includegraphics[width=\linewidth, interpolate=false]{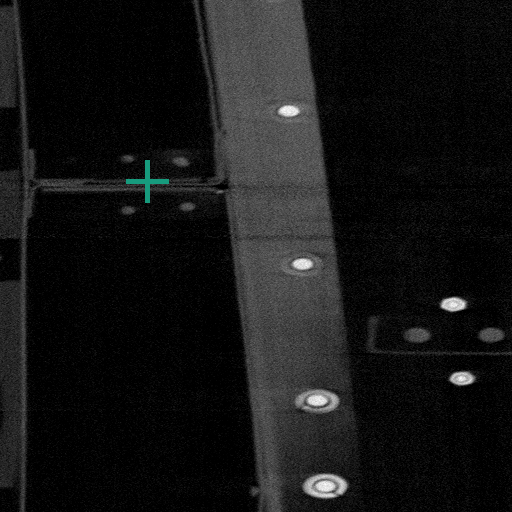}}
					\caption{Input}
					\label{fig:results-evalLoss-errorCase-me163-060-input-cross}
				\end{subfigure}
				\begin{subfigure}[t]{0.235\columnwidth}
					\centering
					\fbox{\includegraphics[width=\linewidth, interpolate=false]{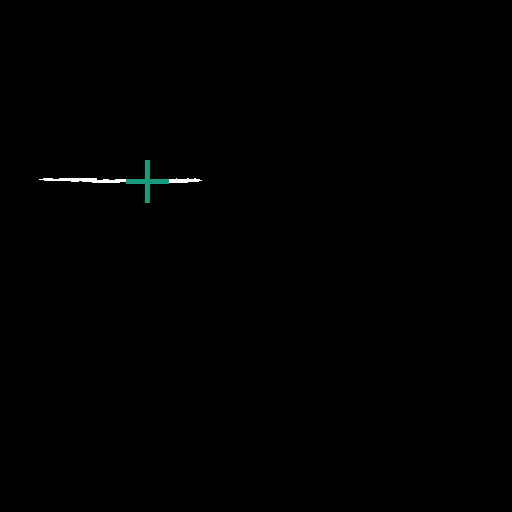}}
					\caption{Reference}
					\label{fig:results-evalLoss-errorCase-me163-060-reference-mask-cross}
				\end{subfigure}
				\begin{subfigure}[t]{0.235\columnwidth}
					\centering
					\fbox{\includegraphics[width=\linewidth, interpolate=false]{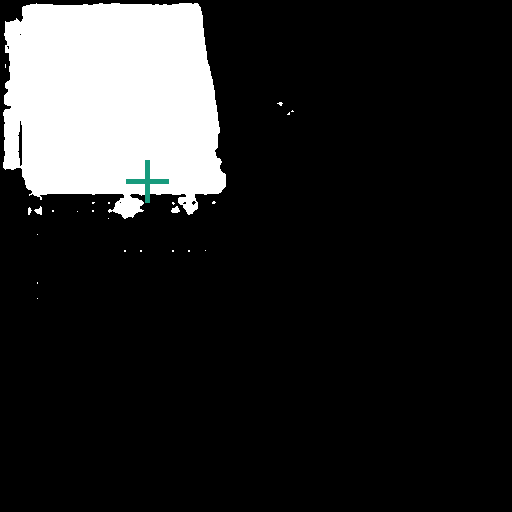}}
					\caption{vit\_b \\loss=0.99} 
					\label{fig:results-evalLoss-errorCase-me163-060-vit_b-mask-cross}
				\end{subfigure}
				\begin{subfigure}[t]{0.235\columnwidth}
					\centering
					\fbox{\includegraphics[width=\linewidth, interpolate=false]{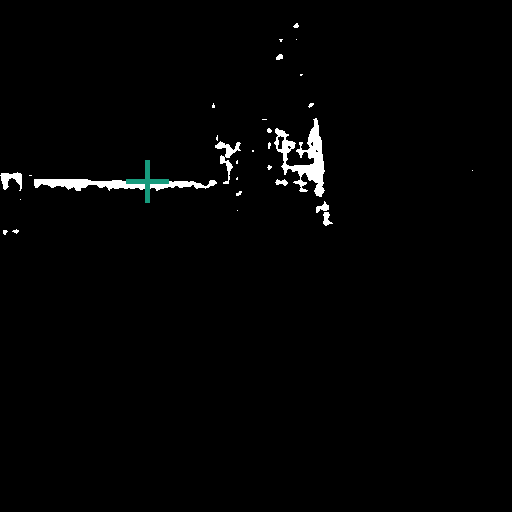}}
					\caption{\fineTunedModel \\loss=0.86}	
					\label{fig:results-evalLoss-errorCase-me163-060-vit_b-439-mask-cross}
				\end{subfigure}		
				\vspace{\baselineskip}				
				\begin{subfigure}[t]{0.235\columnwidth}
					\centering
					\fbox{\includegraphics[width=\linewidth, interpolate=false]{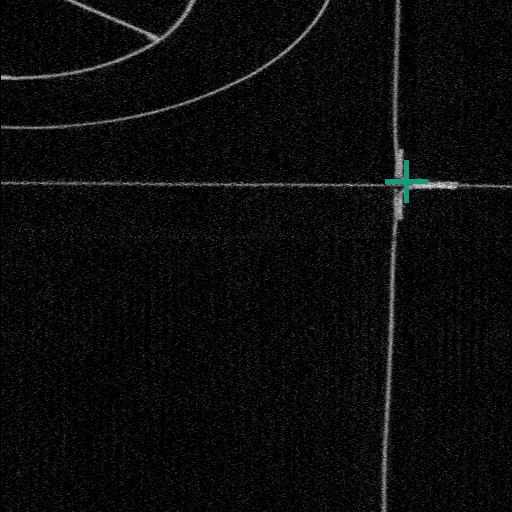}}
					\caption{Input}
					\label{fig:results-evalLoss-errorCase-me163-046-input-cross}
				\end{subfigure}
				\begin{subfigure}[t]{0.235\columnwidth}
					\centering
					\fbox{\includegraphics[width=\linewidth, interpolate=false]{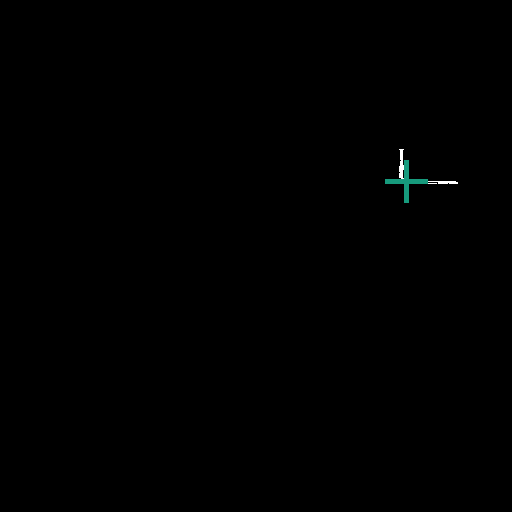}}
					\caption{Reference}
					\label{fig:results-evalLoss-errorCase-me163-046-reference-mask-cross}
				\end{subfigure}
				\begin{subfigure}[t]{0.235\columnwidth}
					\centering
					\fbox{\includegraphics[width=\linewidth, interpolate=false]{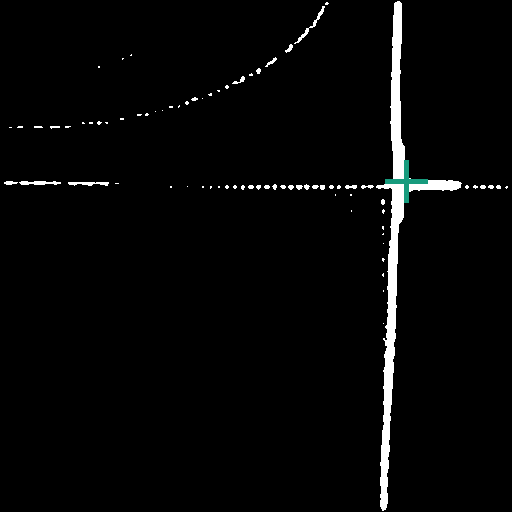}}
					\caption{vit\_b \\loss=0.99} 
					\label{fig:results-evalLoss-errorCase-me163-046-vit_b-mask-cross}
				\end{subfigure}
				\begin{subfigure}[t]{0.235\columnwidth}
					\centering
					\fbox{\includegraphics[width=\linewidth, interpolate=false]{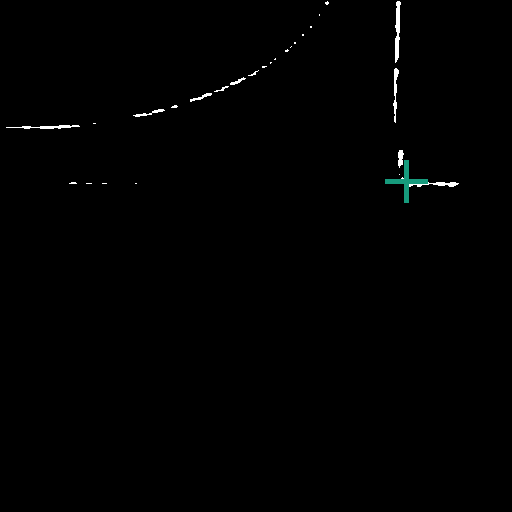}}
					\caption{\fineTunedModel \\loss=0.76} 
					\label{fig:results-evalLoss-errorCase-me163-046-vit_b-439-mask-cross}
				\end{subfigure}	
			
				\caption{\label{fig:results-evalLoss-errorCase-Me163-SAM}Poorly performing cases for SAM vit\_b segmenting thin metal sheets in the Me\,163 data-set, as well as the better but still not optimal segmentation results achieved by the model fine-tuned on the Me\,163 data-set.}
			\end{figure}

			Figure \ref{fig:result-evalLoss-plotMe163} demonstrates that the segmentation quality for the Me163 data-set is notably lower compared to the previously mentioned data-sets. Figure \ref{fig:results-evalLoss-errorCase-Me163-SAM} displays some typical error patterns in the original trained SAM images. Both under-segmentation and over-segmentation occur, and segments are sometimes partially or not recognized at all. 
			
			Among the different not fine-tuned SAM models, the smallest model vit\_b showed the most promising results. While it was sometimes outperformed by the other two original SAM models, vit\_l and vit\_h, in the well-segmented slices, it still had a higher segmentation quality in the moderately segmented slices. Therefore, we decided to use vit\_b as the base model for fine-tuning and volumetric segmentation experiments.
			
			Among the subsequently trained networks, \fineTunedModel exhibits the highest quality in Figure \ref{fig:result-evalLoss-plotMe163}. It is based on vit\_b and uses \emph{ConstantValueBackground} ($CV\!B$) (see Section \ref{sec:methods-fineTune}) for background examples. In simple cases, it matches the segmentation quality of non fine-tuned SAM variants. A considerable improvement in segmentation quality on the challenging entities could be achieved through training, although not to a satisfactory level. This model was chosen as the representative of our fine-tuned model for further tests on our data.
			
		\subsection{Tile-Based Algorithms and Artefact Mitigation}
\begin{figure}
				\centering		
				\begin{subfigure}[t]{0.3\columnwidth}
					\centering
					\fbox{\includegraphics[width=\linewidth, interpolate=false, trim=15px 8px 15px 22px, clip]{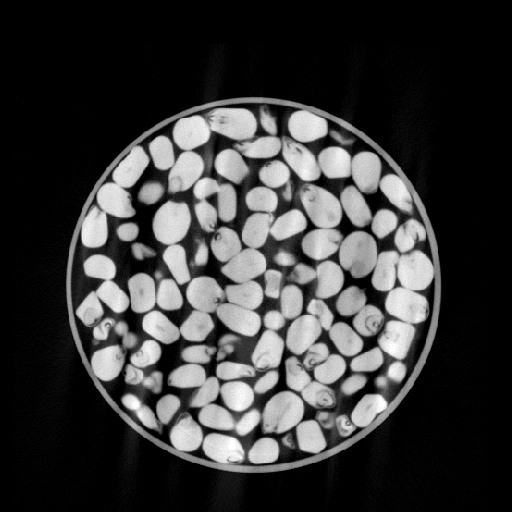}}
					\caption{Input} 
					\label{fig:results-inference-volumetric-bulkMaterial-corn-input}
				\end{subfigure}
				\begin{subfigure}[t]{0.3\columnwidth}
					\centering
					\fbox{\includegraphics[width=\linewidth, interpolate=false, trim=15px 8px 15px 22px, clip]{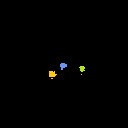}}
					\caption{vit\_b 48}
					\label{fig:results-inference-volumetric-bulkMaterial-corn-48}
				\end{subfigure}
				\begin{subfigure}[t]{0.3\columnwidth}
					\centering
					\fbox{\includegraphics[width=\linewidth, interpolate=false, trim=15px 8px 15px 22px, clip]{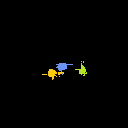}}
					\caption{vit\_b 1024}
					\label{fig:results-inference-volumetric-bulkMaterial-corn-1024}
				\end{subfigure}
				\vspace{\baselineskip}	
				
				\begin{subfigure}[t]{0.3\columnwidth}
					\centering
					\fbox{\includegraphics[width=\linewidth, interpolate=false, trim=40px 21px 40px 59px, clip]{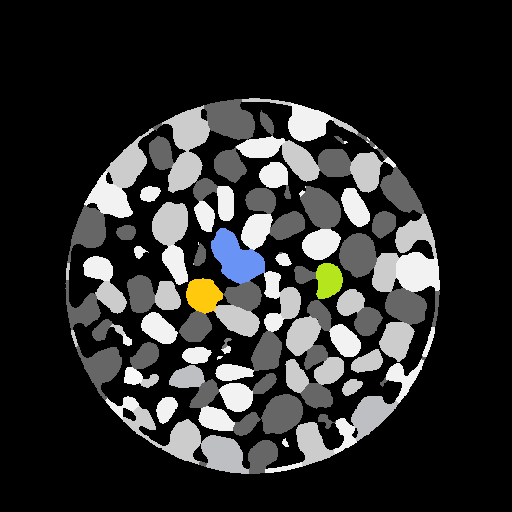}}
					\caption{Reference}
					\label{fig:results-inference-volumetric-bulkMaterial-corn-reference}
				\end{subfigure}	
				\begin{subfigure}[t]{0.3\columnwidth}
					\centering
					\fbox{\includegraphics[width=\linewidth, interpolate=false, trim=15px 8px 15px 22px, clip]{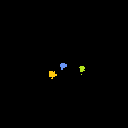}}
					\caption{vit\_b 48 \\postprocessed}
					\label{fig:results-inference-volumetric-bulkMaterial-corn-48-closed}
				\end{subfigure}
				\begin{subfigure}[t]{0.3\columnwidth}
					\centering
					\fbox{\includegraphics[width=\linewidth, interpolate=false, trim=15px 8px 15px 22px, clip]{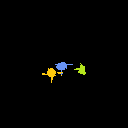}}
					\caption{vit\_b 1024 \\postprocessed}
					\label{fig:results-inference-volumetric-bulkMaterial-corn-1024-closed}
				\end{subfigure}			
				\caption{\label{fig:results-inference-volumetric-bulkMaterial-corn}Slices from a volumetric inference run on three corn kernels of the corn data-set. The input volume (Figure \ref{fig:results-inference-volumetric-bulkMaterial-corn-input}), reference volume (Figure \ref{fig:results-inference-volumetric-bulkMaterial-corn-reference}), and the proposed segmentations generated by the proposed algorithm using the two tile sizes: $48\times48\times48$\,voxels (Figure \ref{fig:results-inference-volumetric-bulkMaterial-corn-48}) and $1024\times1024\times1024$\,voxels (Figure \ref{fig:results-inference-volumetric-bulkMaterial-corn-1024}). Additionally, the postprocessed volumes are depicted in Figures \ref{fig:results-inference-volumetric-bulkMaterial-corn-48-closed} and \ref{fig:results-inference-volumetric-bulkMaterial-corn-1024-closed}.}
			\end{figure}

\begin{figure}
				\centering
				\begin{subfigure}[t]{0.3\columnwidth}
					\centering
					\fbox{\includegraphics[width=\linewidth, interpolate=false]{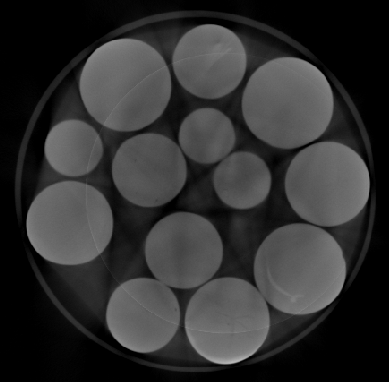}}
					\caption{Input} 
					\label{fig:results-inference-volumetric-bulkMaterial-marbles-input}
				\end{subfigure}
				\begin{subfigure}[t]{0.3\columnwidth}
					\centering
					\fbox{\includegraphics[width=\linewidth, interpolate=false]{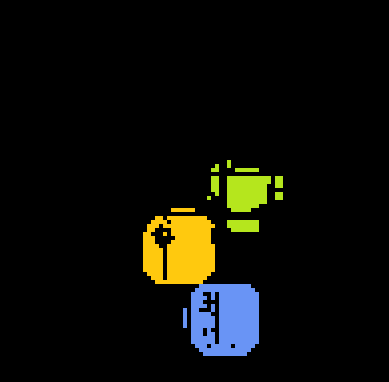}}
					\caption{vit\_b 48}
					\label{fig:results-inference-volumetric-bulkMaterial-marbles-48}
				\end{subfigure}
				\begin{subfigure}[t]{0.3\columnwidth}
					\centering
					\fbox{\includegraphics[width=\linewidth, interpolate=false]{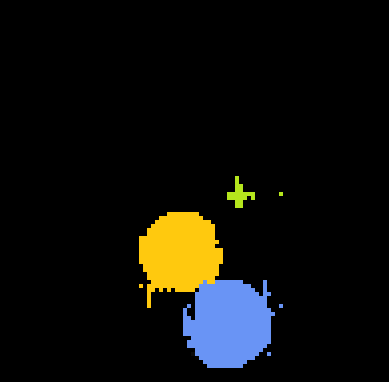}}
					\caption{vit\_b 1024}
					\label{fig:results-inference-volumetric-bulkMaterial-marbles-1024}
				\end{subfigure}
				\vspace{\baselineskip}	
				
				\begin{subfigure}[t]{0.3\columnwidth}
					\centering
					\fbox{\includegraphics[width=\linewidth, interpolate=false]{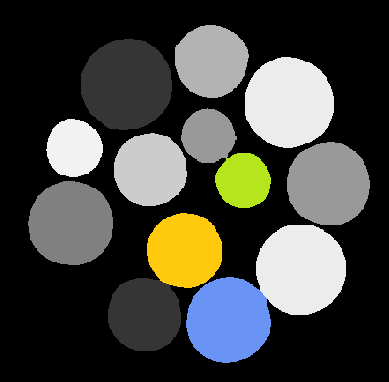}}
					\caption{Reference}
					\label{fig:results-inference-volumetric-bulkMaterial-marbles-reference}
				\end{subfigure}	
				\begin{subfigure}[t]{0.3\columnwidth}
					\centering
					\fbox{\includegraphics[width=\linewidth, interpolate=false]{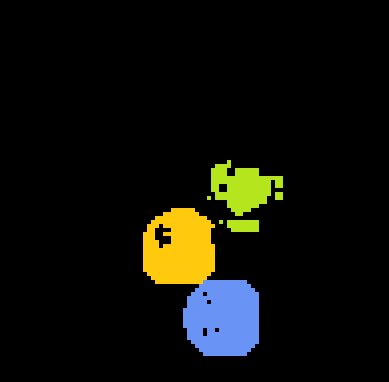}}
					\caption{vit\_b 48 \\postprocessed}
					\label{fig:results-inference-volumetric-bulkMaterial-marbles-48-closed}
				\end{subfigure}
				\begin{subfigure}[t]{0.3\columnwidth}
					\centering
					\fbox{\includegraphics[width=\linewidth, interpolate=false]{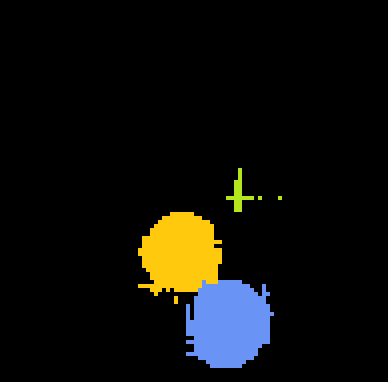}}
					\caption{vit\_b 1024 \\postprocessed}
					\label{fig:results-inference-volumetric-bulkMaterial-marbles-1024-closed}
				\end{subfigure}			
				\caption{\label{fig:results-inference-volumetric-bulkMaterial-marbles}Slices from a volumetric inference run on three marbles of the marbles data-set. The input volume (Figure \ref{fig:results-inference-volumetric-bulkMaterial-marbles-input}), reference volume (Figure \ref{fig:results-inference-volumetric-bulkMaterial-marbles-reference}), and the proposed segmentations generated by the proposed algorithm using the two tile sizes: $48\times48\times48$\,voxels (Figure \ref{fig:results-inference-volumetric-bulkMaterial-marbles-48}) and $1024\times1024\times1024$\,voxels (Figure \ref{fig:results-inference-volumetric-bulkMaterial-marbles-1024}). Additionally, the postprocessed volumes are depicted in Figures \ref{fig:results-inference-volumetric-bulkMaterial-marbles-48-closed} and \ref{fig:results-inference-volumetric-bulkMaterial-marbles-1024-closed}.}
			\end{figure}

			Figures \ref{fig:results-inference-volumetric-bulkMaterial-corn} and \ref{fig:results-inference-volumetric-bulkMaterial-marbles} showcase the segmentation results of a volumetric inference run using the proposed SAM algorithm on a small subset of the marble and corn data-sets for the two tile sizes $48\times48\times48$\,voxels and $1024\times1024\times1024$\,voxels.	These results exhibit segmentation errors in the form of erroneous segmented edges as well as tiling artefacts, resulting in a textured appearance of the segment with noticeable gaps. 
			
			Notably for a tile size of $48\times48\times48$\,voxels, the marble example in Figure \ref{fig:results-inference-volumetric-bulkMaterial-marbles-48} demonstrates tiling artefacts. Since the volumetric inference algorithm with the small tile size cannot segment the entire marble in a single step, it must combine multiple steps, which can introduce and propagate errors. These artefacts can be cleaned up using a morphological closing operation as postprocessing step. 
			
			In contrast, segmentations using a larger tile size of $1024\times1024\times1024$\,voxels exhibit fewer of these textured artefacts. However, segmentations may extend beyond the actual segment due to segmentation errors, as illustrated in Figure \ref{fig:results-inference-volumetric-bulkMaterial-corn-1024}, where thin segments protrude vertically and horizontally beyond the intended boundaries. These protrusions often occur within the initially segmented slices that include the seed point of the current segment. In the green upper right marble of the example in Figure \ref{fig:results-inference-volumetric-bulkMaterial-marbles-1024}, the adjacent slices directly connected to the seed point were misclassified as not belonging to the marble, resulting in an early termination of the slice-wise segmentation process.
			
			The inference algorithm with a tile size of $1024\times1024\times1024$\,voxels can only attempt to segment the segment once as due to its high field of view it performs a single volumetric step per seed point. In contrast, the inference algorithm with a tile size of $48\times48\times48$\,voxels iterates over the volume in multiple steps, providing the ability to compensate for weak and erroneous segmentations in subsequent steps. However, this approach tends to under-segment when a neighbouring segment has already been partially segmented in a previous step.			
			
\begin{figure*}
				\centering
				\begin{minipage}[t]{0.45\linewidth}
				\begin{subfigure}[t]{0.49\columnwidth}
					\centering
					\includegraphics[width=\linewidth]{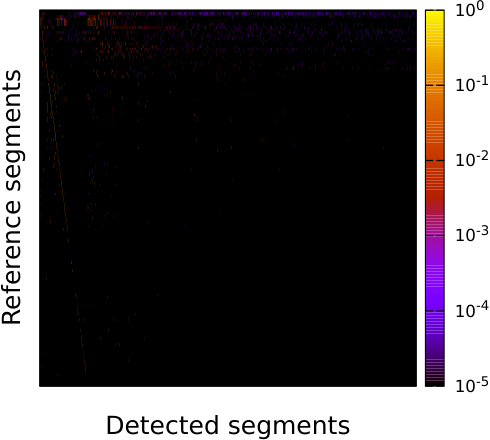}
					\caption{vit\_b 48}
					\label{fig:result-inference-volumetric-me163-vit_b-48}
				\end{subfigure}
				\begin{subfigure}[t]{0.49\columnwidth}
					\centering
					\includegraphics[width=\linewidth]{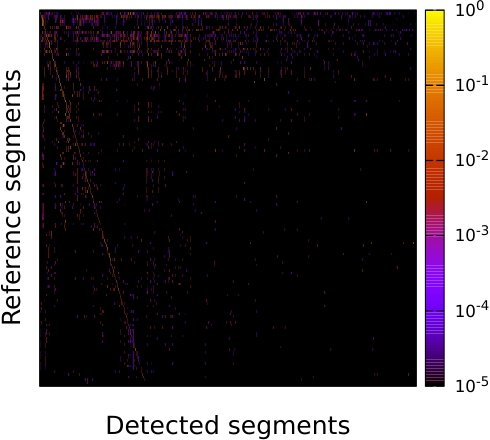}
					\caption{vit\_b 1024}
					\label{fig:result-inference-volumetric-me163-vit_b-1024}
				\end{subfigure}
				\begin{subfigure}[t]{0.49\columnwidth}
					\centering
					\includegraphics[width=\linewidth]{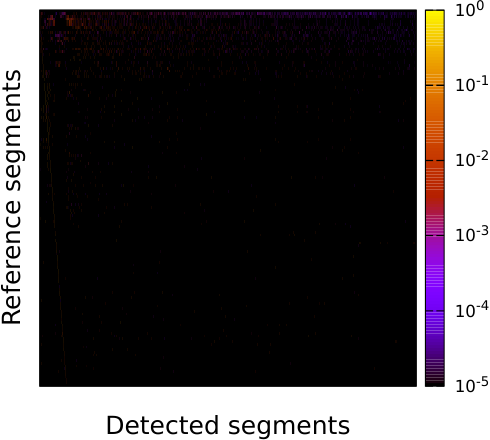}
					\caption{\fineTunedModel 48}
					\label{fig:result-inference-volumetric-me163-vit_b-tTF493-48}
				\end{subfigure}
				\begin{subfigure}[t]{0.49\columnwidth}
					\centering
					\includegraphics[width=\linewidth]{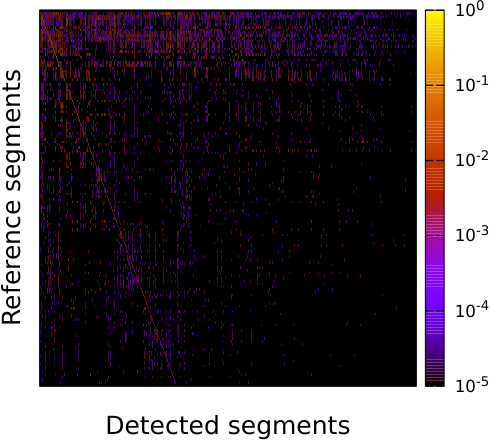}
					\caption{\fineTunedModel 1024}			
					\label{fig:result-inference-volumetric-me163-vit_b-tTF493-1024}
				\end{subfigure}	
				
				\caption{\label{fig:result-inference-volumetric-me163}Correlation matrix of default and fine-tuned volumetric SAM with multiple tiles of size $48\times48\times48$\,voxels or a single tile of size $1024\times1024\times1024$\,voxels of the Me\,163 testing data-set.}
			\end{minipage}
			\hfill 
			\begin{minipage}[t]{0.45\linewidth}
				\centering
				\begin{subfigure}[t]{0.49\columnwidth}
					\centering
					\includegraphics[width=\linewidth, interpolate=false]{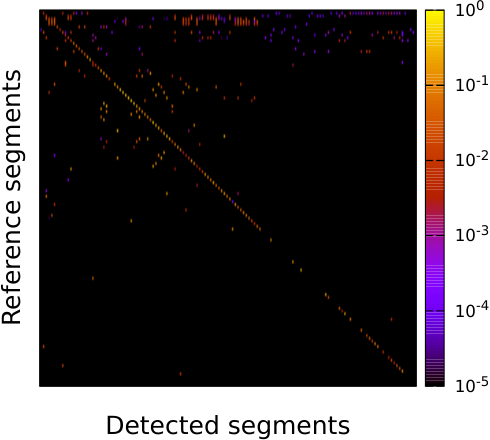}
					\caption{vit\_b 48}
					\label{fig:result-inference-volumetric-me163-mainDiagonalvit_b-48}
				\end{subfigure}
				\begin{subfigure}[t]{0.49\columnwidth}
					\centering
					\includegraphics[width=\linewidth, interpolate=false]{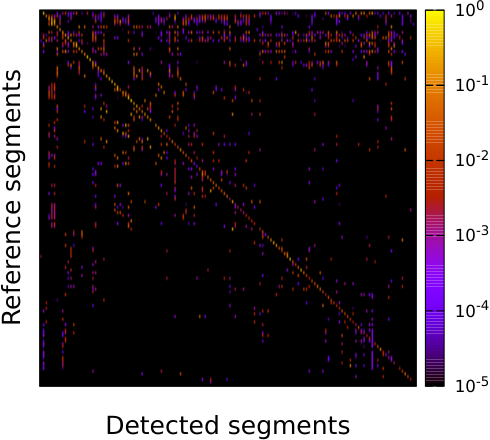}
					\caption{vit\_b 1024}
					\label{fig:result-inference-volumetric-me163-mainDiagonalvit_b-1024}
				\end{subfigure}
				\begin{subfigure}[t]{0.49\columnwidth}
					\centering
					\includegraphics[width=\linewidth, interpolate=false]{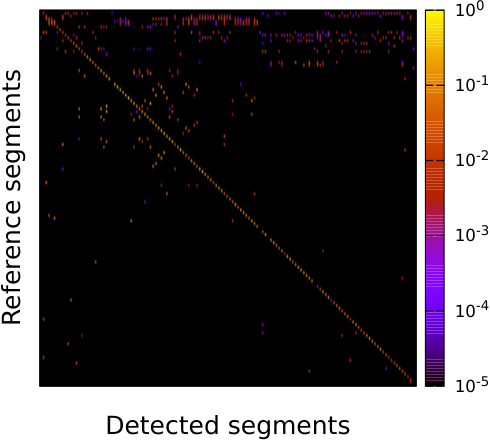}
					\caption{\fineTunedModel 48}
					\label{fig:result-inference-volumetric-me163-mainDiagonalvit_b-tTF493-48}
				\end{subfigure}
				\begin{subfigure}[t]{0.49\columnwidth}
					\centering
					\includegraphics[width=\linewidth, interpolate=false]{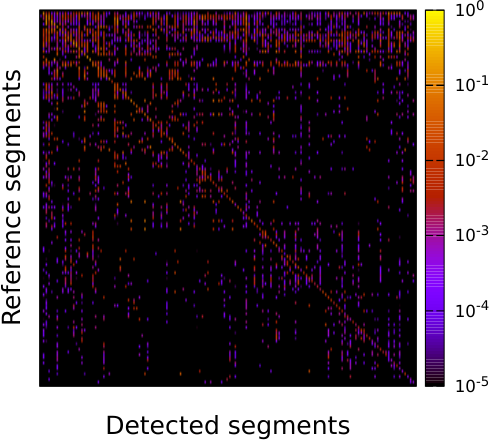}
					\caption{\fineTunedModel 1024}			
					\label{fig:result-inference-volumetric-me163-mainDiagonalvit_b-tTF493-1024}
				\end{subfigure}	
				
				\caption{\label{fig:result-inference-volumetric-me163-mainDiagonal}Correlation matrix of default and find-tuned volumetric SAM with multiple tiles of size $48\times48\times48$\,voxels or a single tile of size $1024\times1024\times1024$\,voxels of the Me\,163 testing data-set. Detected segments have been limited to the best matches for each reference segment.}
			\end{minipage}
			\end{figure*}

			Figure \ref{fig:result-inference-volumetric-me163} shows the correlation matrices for the result of four inference runs on the Me\,163 testing data-sets. Two of the inference runs were performed using the default SAM model vit\_b, while the other two were performed using the fine-tuned model \fineTunedModel. Two of the four experiments used a tile size of $48\times48\times48$\,voxels, and the other two used a tile size of $1024\times1024\times1024$\,voxels. Each experiment was fine-tuned on the validation data-set using \cite{optuna_2019}. 
			
			The correlation matrices show the IoU of each reference segment in relation to each detected segment. The reference segments are sorted from top to bottom based on their voxel count, with the segment having the largest voxel count at the top. Similarly, the columns representing the detected segments, are sorted so that the segment with the highest IoU if compared with the largest reference segment is on the left side. The segment with the highest IoU if compared with the second largest reference segment is then placed in the second column, and so on. Each detected segment can only be linked to one refernence segment once. In an ideal case, we would see a bright diagonal line from the upper left corner to the lower right corner of the matrix, indicating a perfect match between the reference and detected segments. Segments outside this diagonal indicate segmentation errors. Vertical lines indicate under-segmentation, where reference segments extend over multiple detected segments. Horizontal lines indicate over-segmentation, where reference segments are falsely split into multiple detected segments. 			
						
			The individual parameters of the four inference runs can be found in Table \ref{tab:result-infernece-columentric-optuna}. Figure \ref{fig:result-inference-volumetric-me163-mainDiagonal} displays correlation matrices from Figure  \ref{fig:result-inference-volumetric-me163} but constrained to the detected segments with the highest IoU.
			
				\begin{table*}
				\centering	
				\begin{tabular}{lcccc}
					\toprule
					& vit\_b 48 				& vit\_b 1024 				& \fineTunedModel 48 		& \fineTunedModel 1024 \\
					\midrule
					best IoU			& 0.15						& \textbf{0.17}				& 0.07						& 0.09 \\
					&&&&\\				
					movement step\textsuperscript{*} 				& 1 						&  --						& 1							& -- \\
					seed FG count\textsuperscript{*}   				& 2 						&  2						& 1 						& 1 \\
					slice FG count\textsuperscript{*}  				& 3  						&  1						& 1 						& 1 \\
					FG threshold\textsuperscript{*}  				& 0.3  						& 0.2						& 0.2 						& 0.5 \\
					prompt type\textsuperscript{*}   					& centre and dense  		& centre					& centre and dense  		& centre and dense \\
					SAM output channel\textsuperscript{*}			& index 1					& max IoU					& max IoU with FG			& max IoU with FG \\
					\shortstack{slice merge Rule\textsuperscript{*} \\ \phantom{spacer}}	& \makecell{IoU to previous\\ slice \textgreater 0.5}	& \makecell{IoU to previous\\ slice \textgreater 0.25} & \makecell{IoU to previous\\ slice \textgreater 0.5} & always \\
					slice median\textsuperscript{*}					& $\checkmark$				& $\times$					& $\times$					& $\times$\\
					CCA\textsuperscript{*}							& $\checkmark$				& $\checkmark$				& $\times$					& $\times$ \\
					volume median\textsuperscript{*}				& $\times$					& $\checkmark$				& $\times$					& $\checkmark$ \\	
					check step width\textsuperscript{$\times$}   	& 13						&  --						& 19						& -- \\
					accumulator update\textsuperscript{$\times$} 	& FG						& FG						& always					& always \\
					restrict movement\textsuperscript{$\times$}		& FG (128\,steps)			& eroded FG 				& eroded FG	(128\,steps)		& FG \\
									
					\bottomrule
				\end{tabular}
				\caption{\label{tab:result-infernece-columentric-optuna} Parameters optimized on the Me\,163 validation data-set for the default vit\_b and fine-tuned \fineTunedModel SAM model for the tile sizes of $48\times48\times48$\,voxels and  $1024\times1024\times1024$\,voxels. ($FG = \text{foreground}; \text{--} = \text{not applicable}$; Options marked with * indicate volumetric SAM parameters as seen in Table \ref{tab:methods-inference-overview}; Options marked with $\times$ indicate FFN related parameters)}
			\end{table*}			
			
\begin{figure*}
				\centering						
				\begin{subfigure}[t]{0.18\textwidth}
					\centering
					\includegraphics[width=\linewidth, interpolate=false]{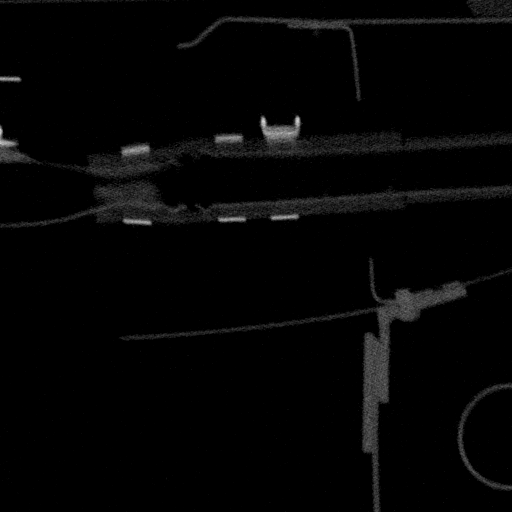}
					\caption{Input}
					\label{fig:results-volumetric-me163-slices-reco}
				\end{subfigure}
				\begin{subfigure}[t]{0.18\textwidth}
					\centering
					\includegraphics[width=\linewidth, interpolate=false]{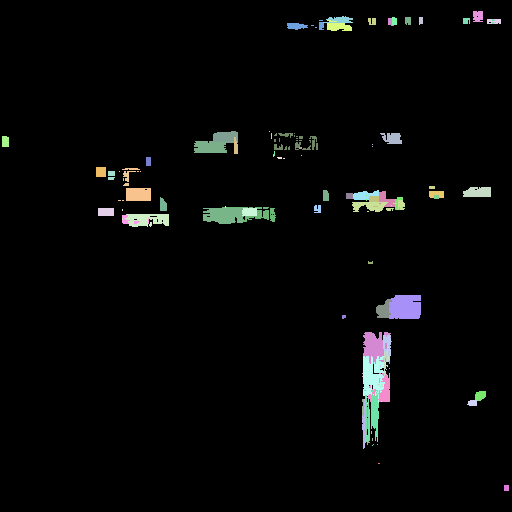}
					\caption{vit\_b 48}
					\label{fig:results-volumetric-me163-slices-vit_b-48}
				\end{subfigure}
				\begin{subfigure}[t]{0.18\textwidth}
					\centering
					\includegraphics[width=\linewidth, interpolate=false]{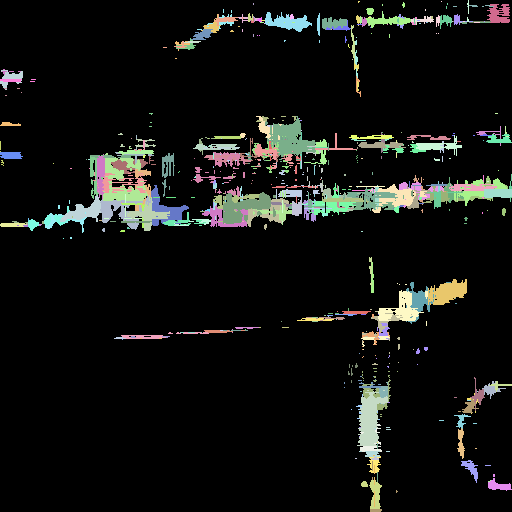}
					\caption{\fineTunedModel 48}
					\label{fig:results-volumetric-me163-slices-vit_b-tTF439-48}
				\end{subfigure}
				\begin{subfigure}[t]{0.18\textwidth}
					\centering
					\includegraphics[width=\linewidth, interpolate=false]{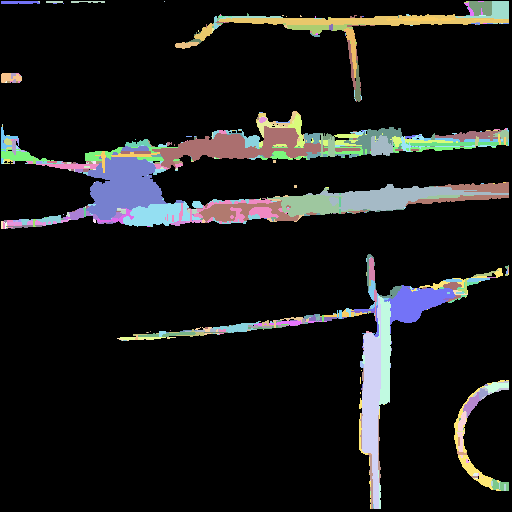}
					\caption{vit\_b 1024}
					\label{fig:results-volumetric-me163-slices-vit_b-1024}
				\end{subfigure}
				\begin{subfigure}[t]{0.18\textwidth}
					\centering
					\includegraphics[width=\linewidth, interpolate=false]{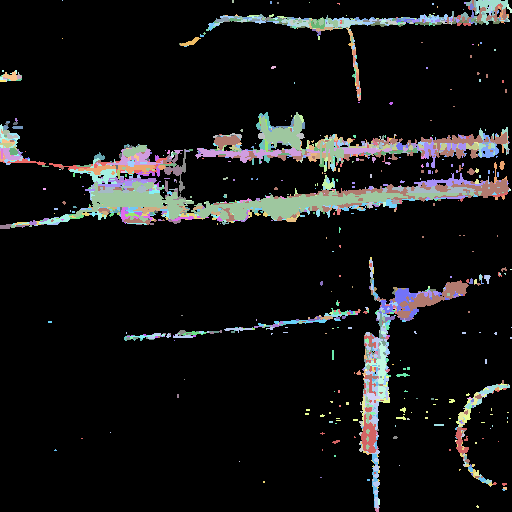}
					\caption{\fineTunedModel 1024}
					\label{fig:results-volumetric-me163-slices-vit_b-tTF439-1024}
				\end{subfigure}				

			\vspace{\baselineskip}
			
				\begin{subfigure}[t]{0.18\textwidth}
					\centering
					\includegraphics[width=\linewidth, interpolate=false]{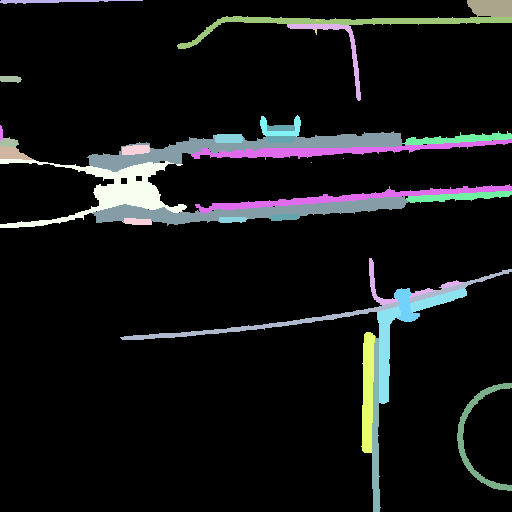}
					\caption{Reference}	
					\label{fig:results-volumetric-me163-slices-reference}
				\end{subfigure}
				\begin{subfigure}[t]{0.18\textwidth}
					\centering
					\includegraphics[width=\linewidth, interpolate=false]{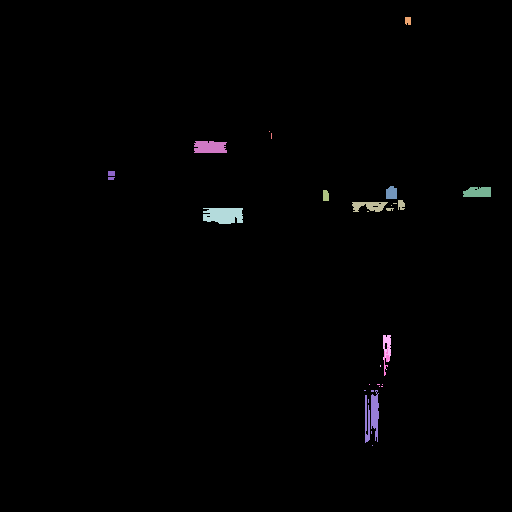}
					\caption{vit\_b 48 \\(main diagonal)}
					\label{fig:results-volumetric-me163-slices--vit_b-48-mainDiagonal}
				\end{subfigure}
				\begin{subfigure}[t]{0.18\textwidth}
					\centering
					\includegraphics[width=\linewidth, interpolate=false]{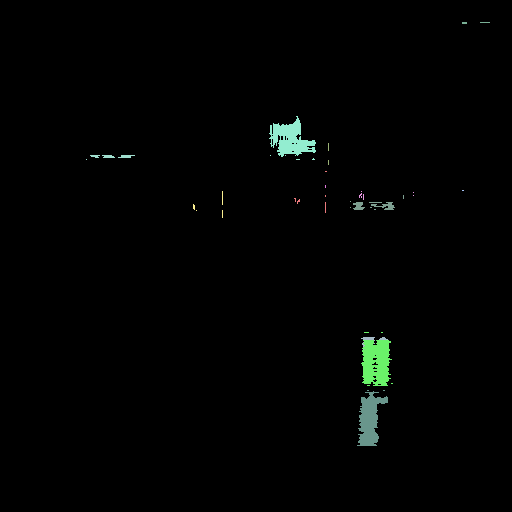}
					\caption{\fineTunedModel 48 \\(main diagonal)}
					\label{fig:results-volumetric-me163-slices-vit_b-tTF439-48-mainDiagonal}
				\end{subfigure}
				\begin{subfigure}[t]{0.18\textwidth}
					\centering
					\includegraphics[width=\linewidth, interpolate=false]{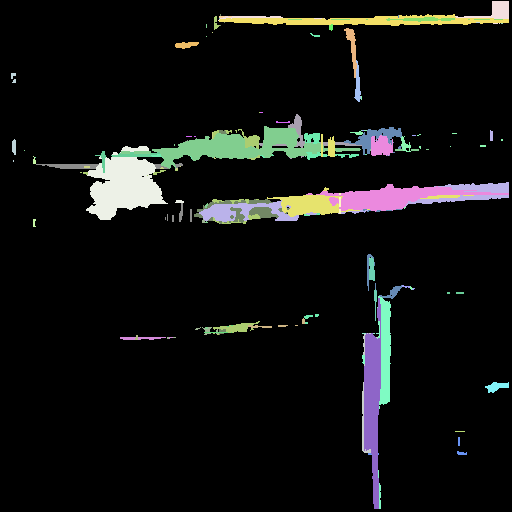}
					\caption{vit\_b 1024 \\(main diagonal)}
					\label{fig:results-volumetric-me163-slices-vit_b-1024-mainDiagonal}
				\end{subfigure}
				\begin{subfigure}[t]{0.18\textwidth}
					\centering
					\includegraphics[width=\linewidth, interpolate=false]{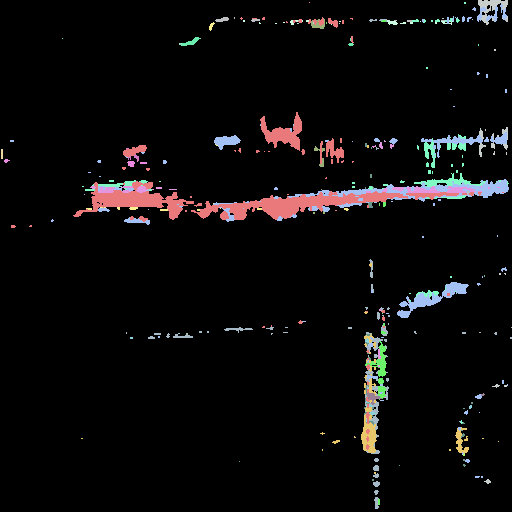}
					\caption{\fineTunedModel 1024 \\(main diagonal)}
					\label{fig:results-volumetric-me163-slices-vit_b-tTF439-1024-mainDiagonal}
				\end{subfigure}
			
				\caption{Exemplary slices of the proposed volumetric inference output performed by default and fine-tuned SAM models on the Me\,163 reference data-set (Figures \ref{fig:results-volumetric-me163-slices-reco} and  \ref{fig:results-volumetric-me163-slices-reference}). For the remaining figures the top row (Figures \ref{fig:results-volumetric-me163-slices-vit_b-48} -- \ref{fig:results-volumetric-me163-slices-vit_b-tTF439-1024}) shows all segments depicted in Figure \ref{fig:result-inference-volumetric-me163} while the bottom row (Figures \ref{fig:results-volumetric-me163-slices--vit_b-48-mainDiagonal} -- \ref{fig:results-volumetric-me163-slices-vit_b-tTF439-1024-mainDiagonal}) only shows the segments corresponding to the main diagonal in Figure \ref{fig:result-inference-volumetric-me163-mainDiagonal}.}
			\end{figure*}

			As can be seen, the \fineTunedModel models tends to generate more noise outside the main diagonal. Especially Figure \ref{fig:result-inference-volumetric-me163-mainDiagonalvit_b-tTF493-1024} depicts many over- and under-segmented segments. This can also be observed in the corresponding segmentation volume slice shown in Figure \ref{fig:results-volumetric-me163-slices-vit_b-tTF439-1024} and \ref{fig:results-volumetric-me163-slices-vit_b-tTF439-1024-mainDiagonal}. The correlation matrix of the fine-tuned \fineTunedModel model with tile size $48\times48\times48$\,voxels in Figure \ref{fig:result-inference-volumetric-me163-mainDiagonalvit_b-tTF493-48} seems to perform best with respect to diagonal segments. But comparing the corresponding segmentation volume slice in Figure \ref{fig:results-volumetric-me163-slices-vit_b-tTF439-48-mainDiagonal} shows, that this model, tile, and parameter combination tends to miss most of the foreground segments.
			It seems that the default vit\_b model with tile size $1024\times1024\times1024$\,voxels produces the visually best results, followed by the fine-tuned \fineTunedModel model with tile size $48\times48\times48$ shown in Figure \ref{fig:results-volumetric-me163-slices-vit_b-tTF439-48}.
		
			\begin{figure*}[ht!]
				\setlength\tabcolsep{3pt} 
				\centering
				\begin{tabular}{@{} r m{0.13\linewidth} m{0.13\linewidth} m{0.13\linewidth} m{0.13\linewidth} m{0.13\linewidth}  m{0.13\linewidth} m{0.13\linewidth} @{}}
					
					\multirow{2}{*}{\rotatebox[origin=c]{90}{vit\_b 48\hspace{2.0em}}}
					\rotatebox[origin=c]{90}{pred}
					& \includegraphics[width=\hsize]{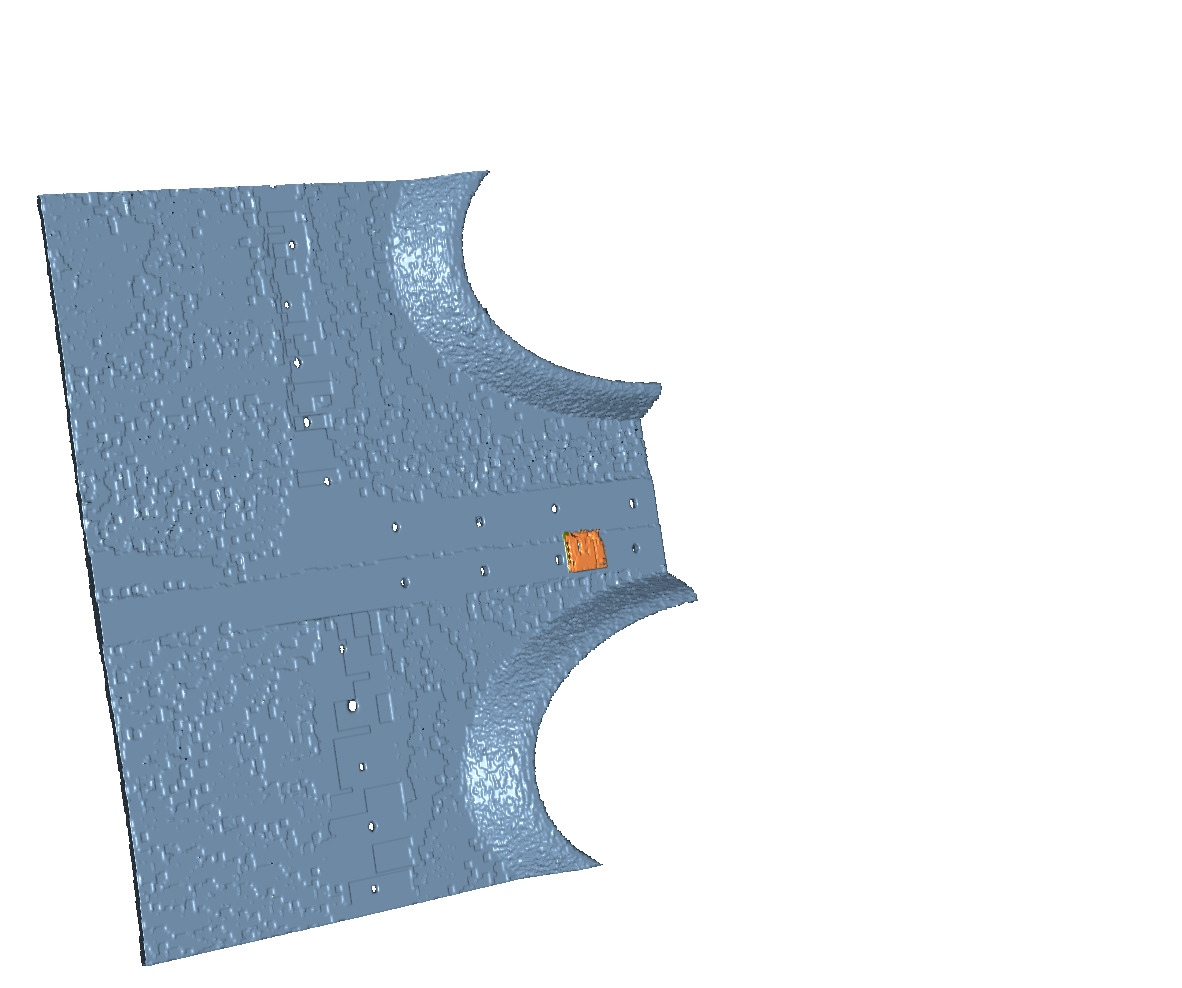} 
					& \includegraphics[width=\hsize]{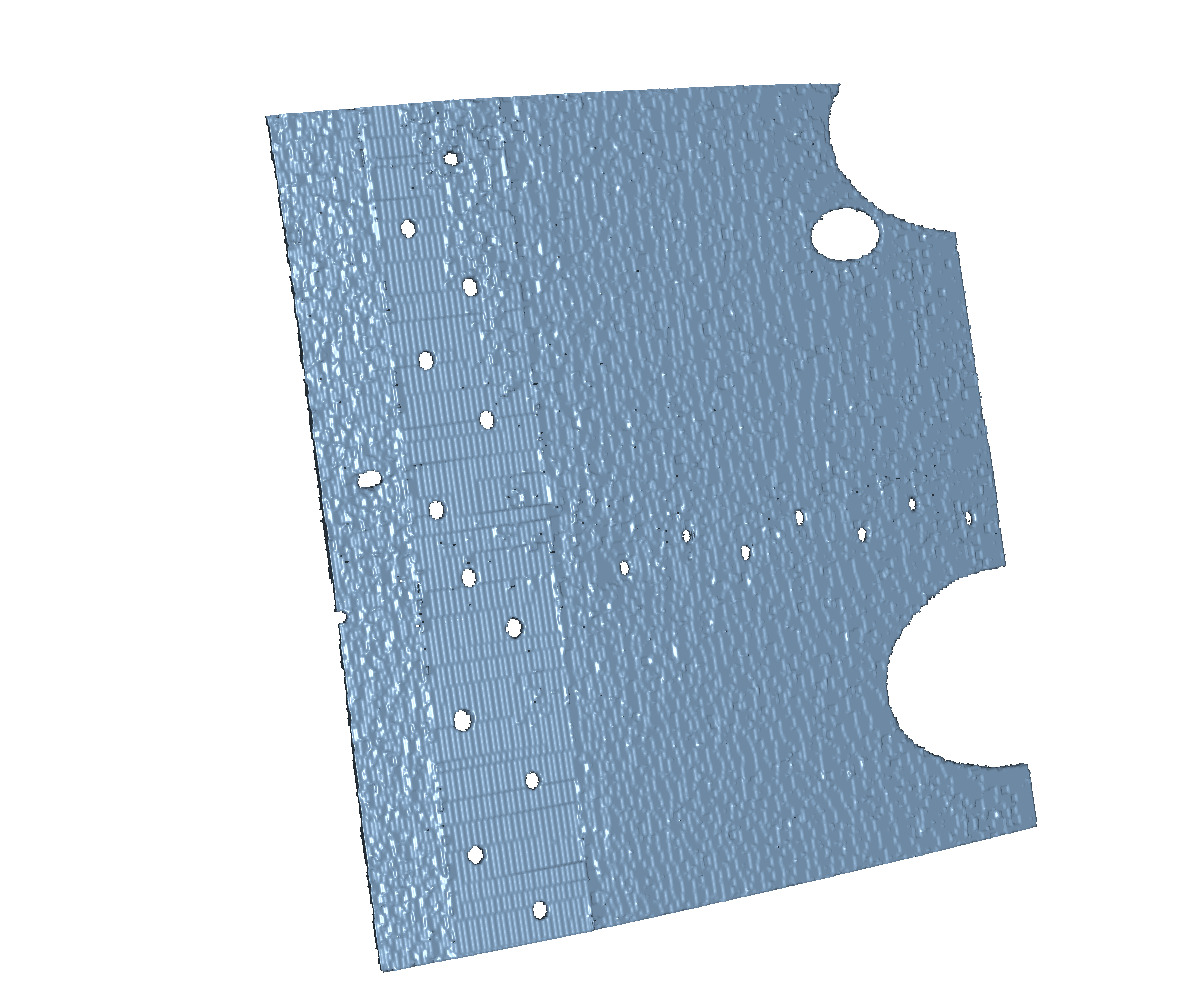} 
					& \includegraphics[width=\hsize]{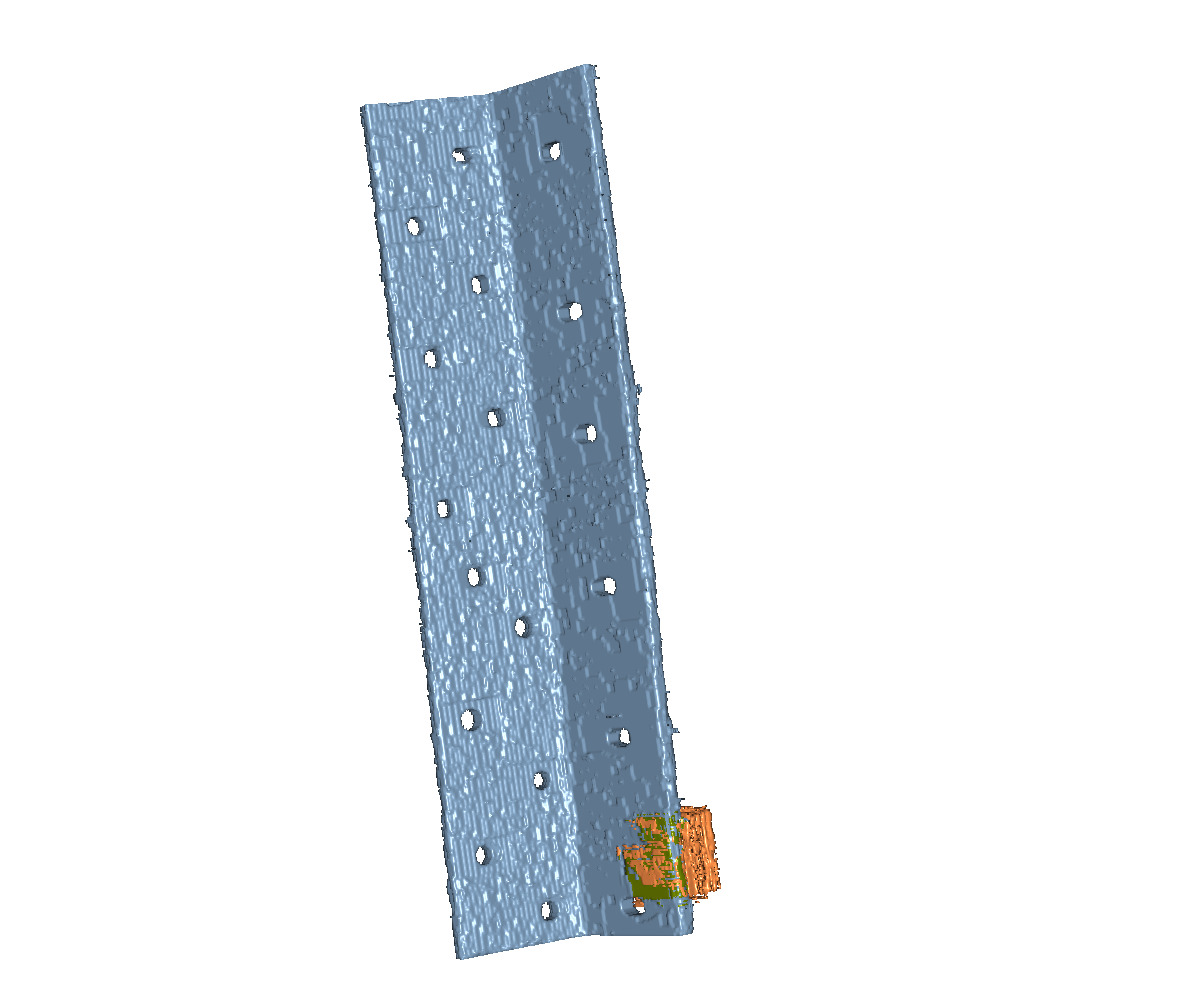} 
					& \includegraphics[width=\hsize]{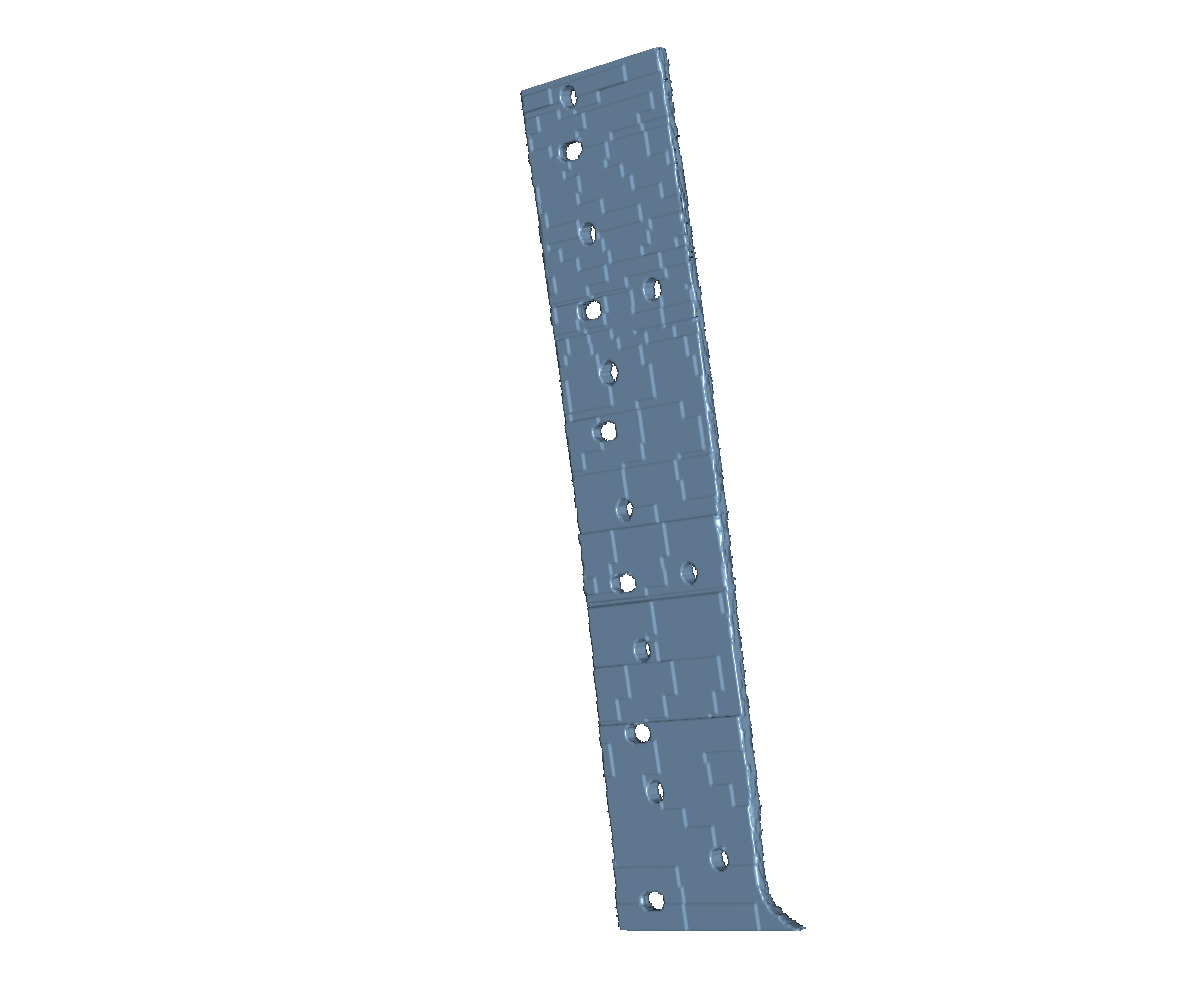} 
					& \includegraphics[width=\hsize]{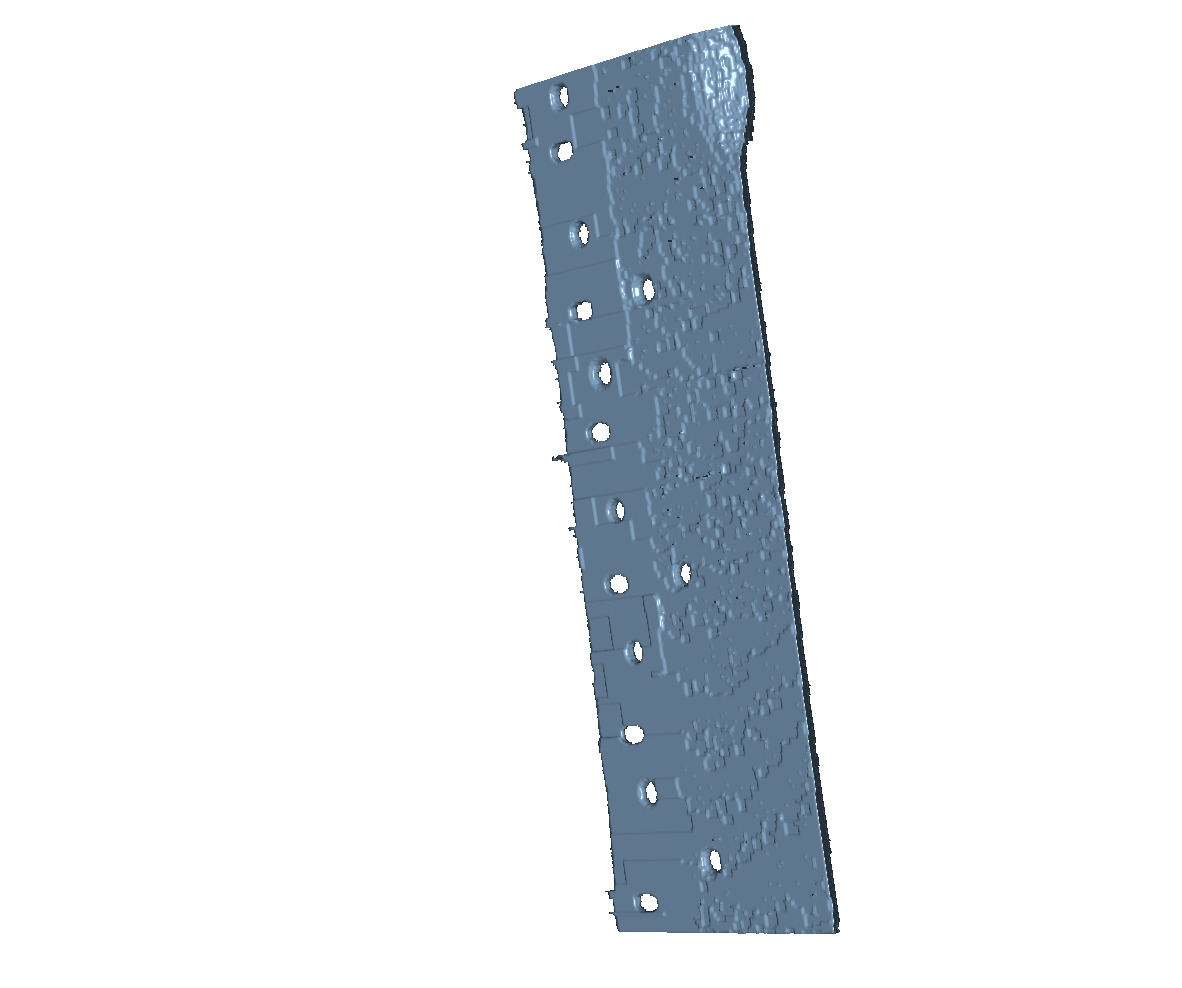} 
					& \includegraphics[width=\hsize]{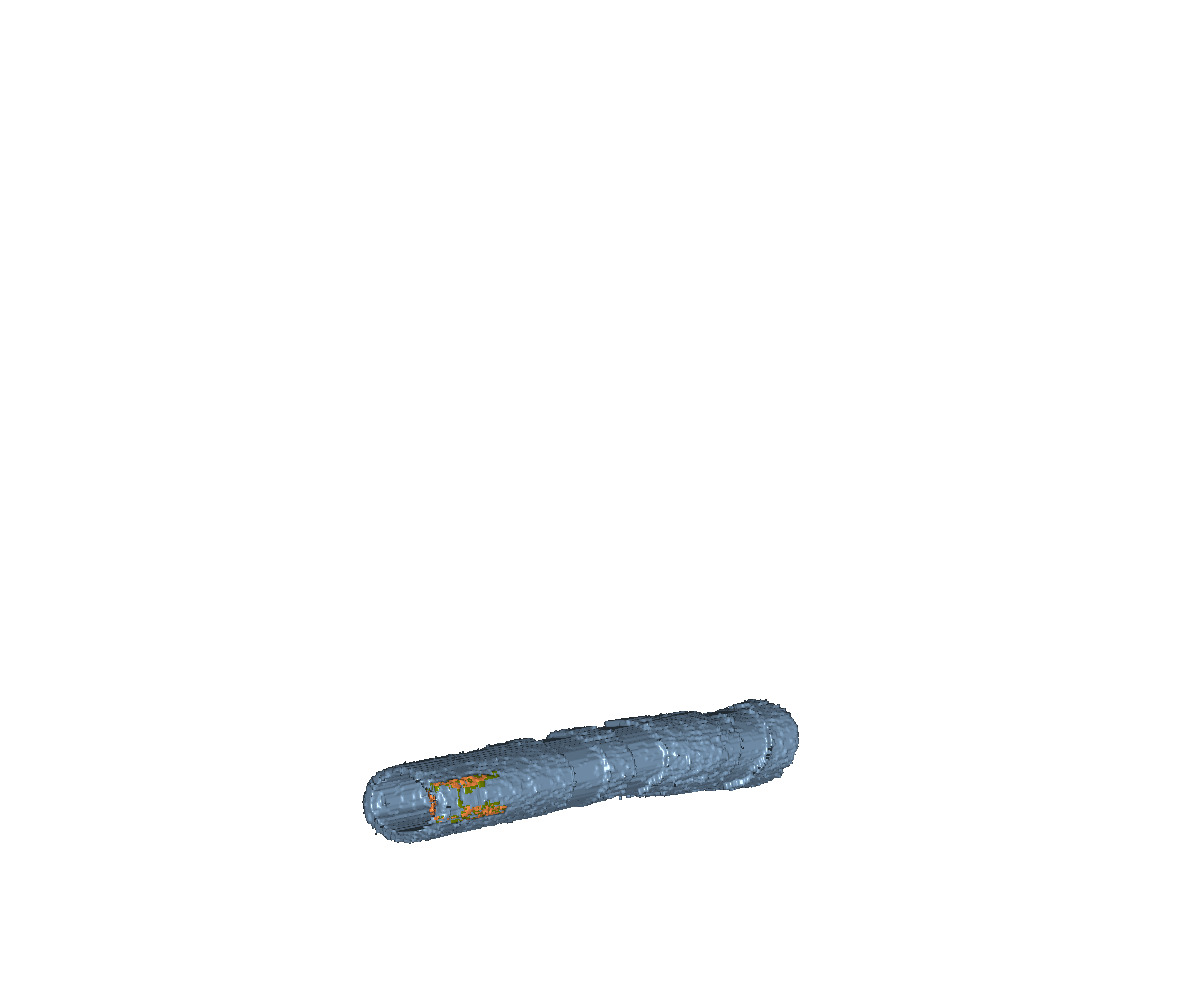} 
					& \includegraphics[width=\hsize]{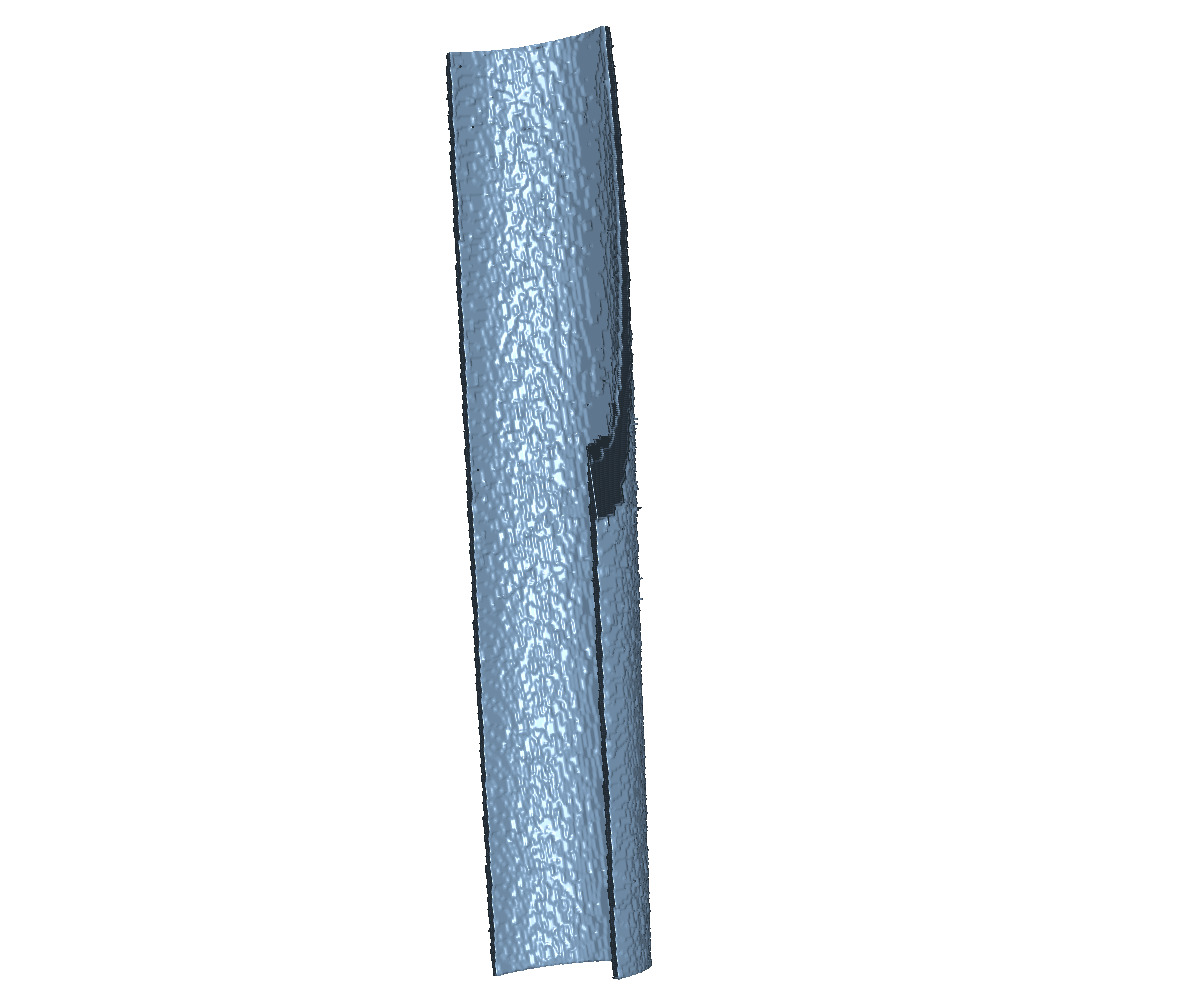} 
					\\ \addlinespace					
					\rotatebox[origin=c]{90}{TP}
					& \includegraphics[width=\hsize]{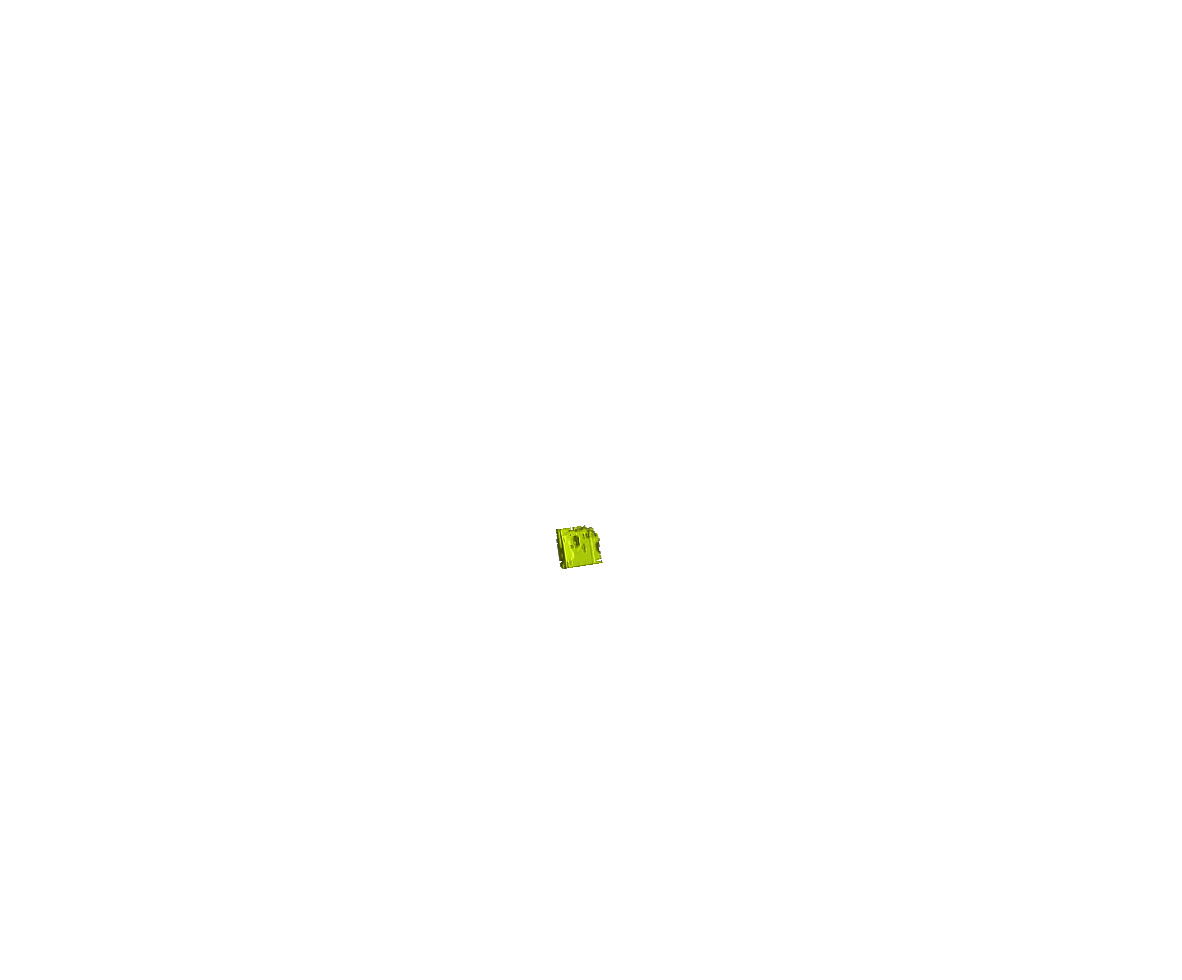} 
					& \includegraphics[width=\hsize]{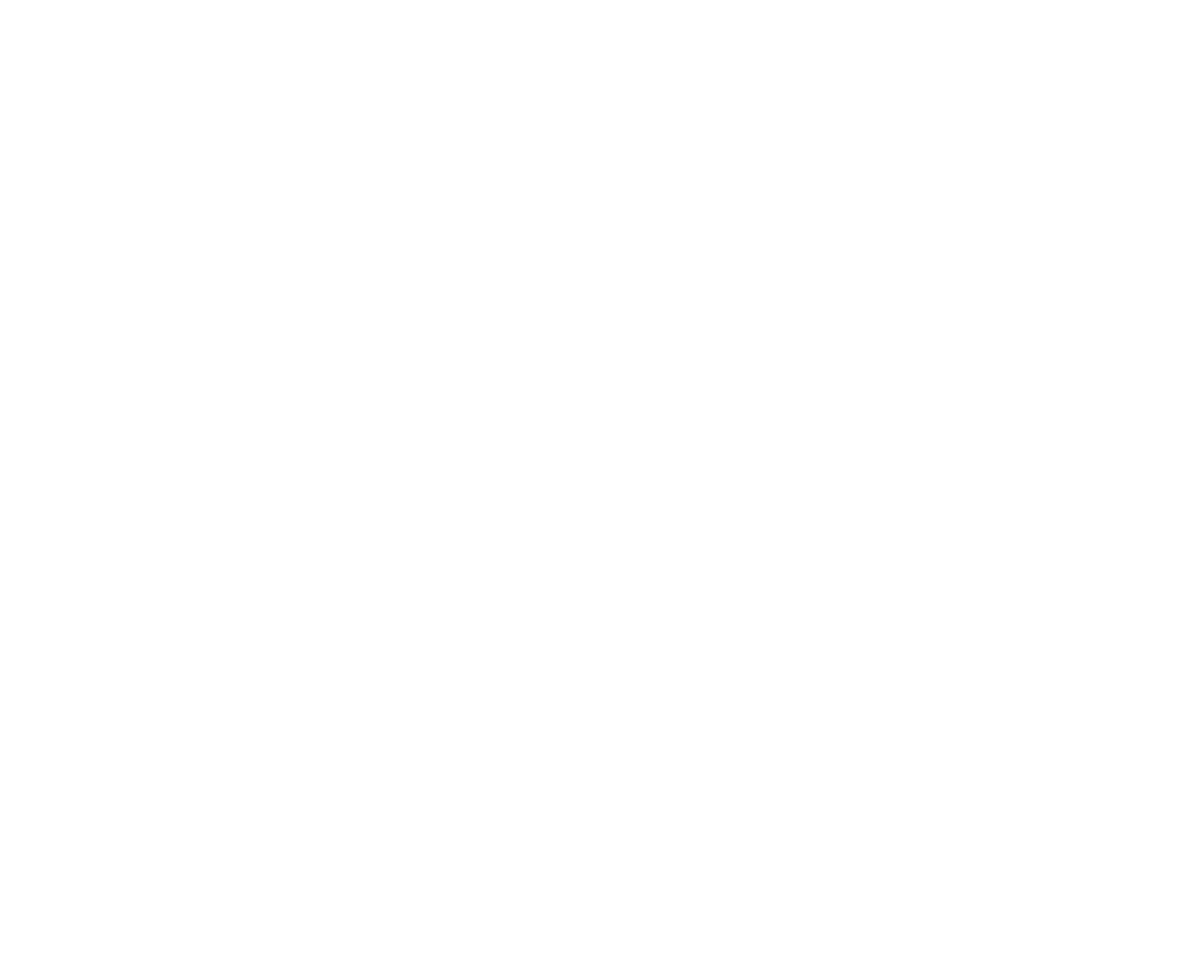} 
					& \includegraphics[width=\hsize]{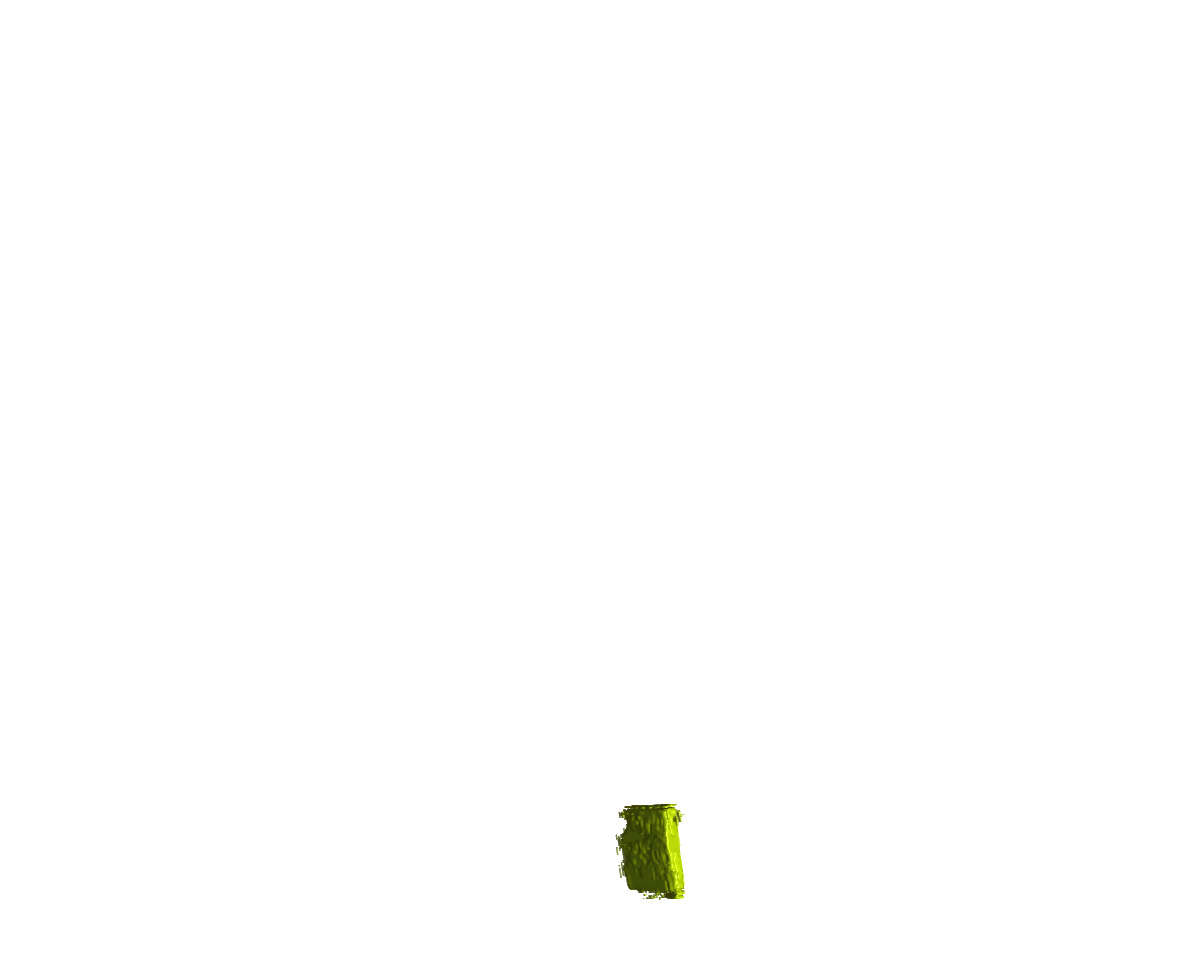} 
					& \includegraphics[width=\hsize]{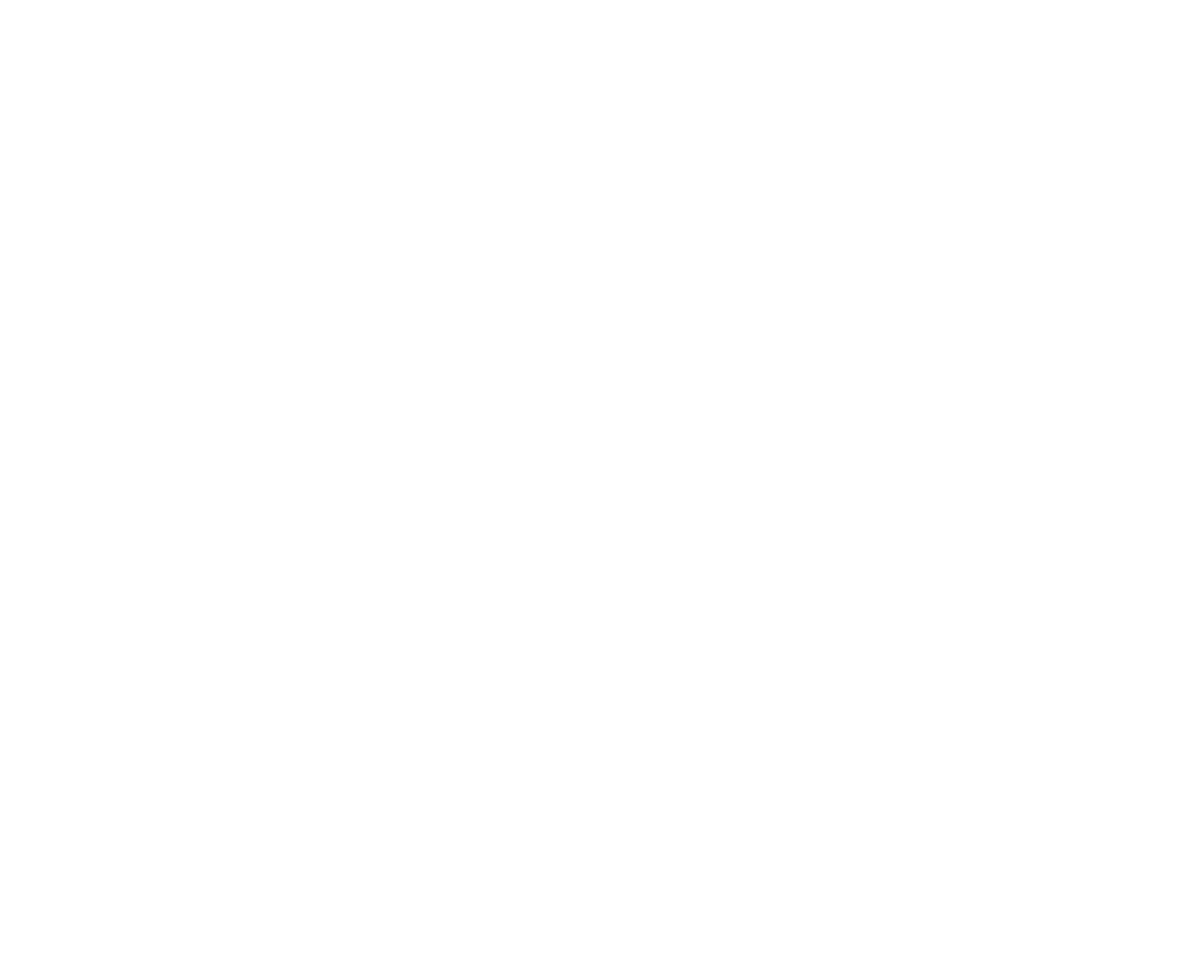} 
					& \includegraphics[width=\hsize]{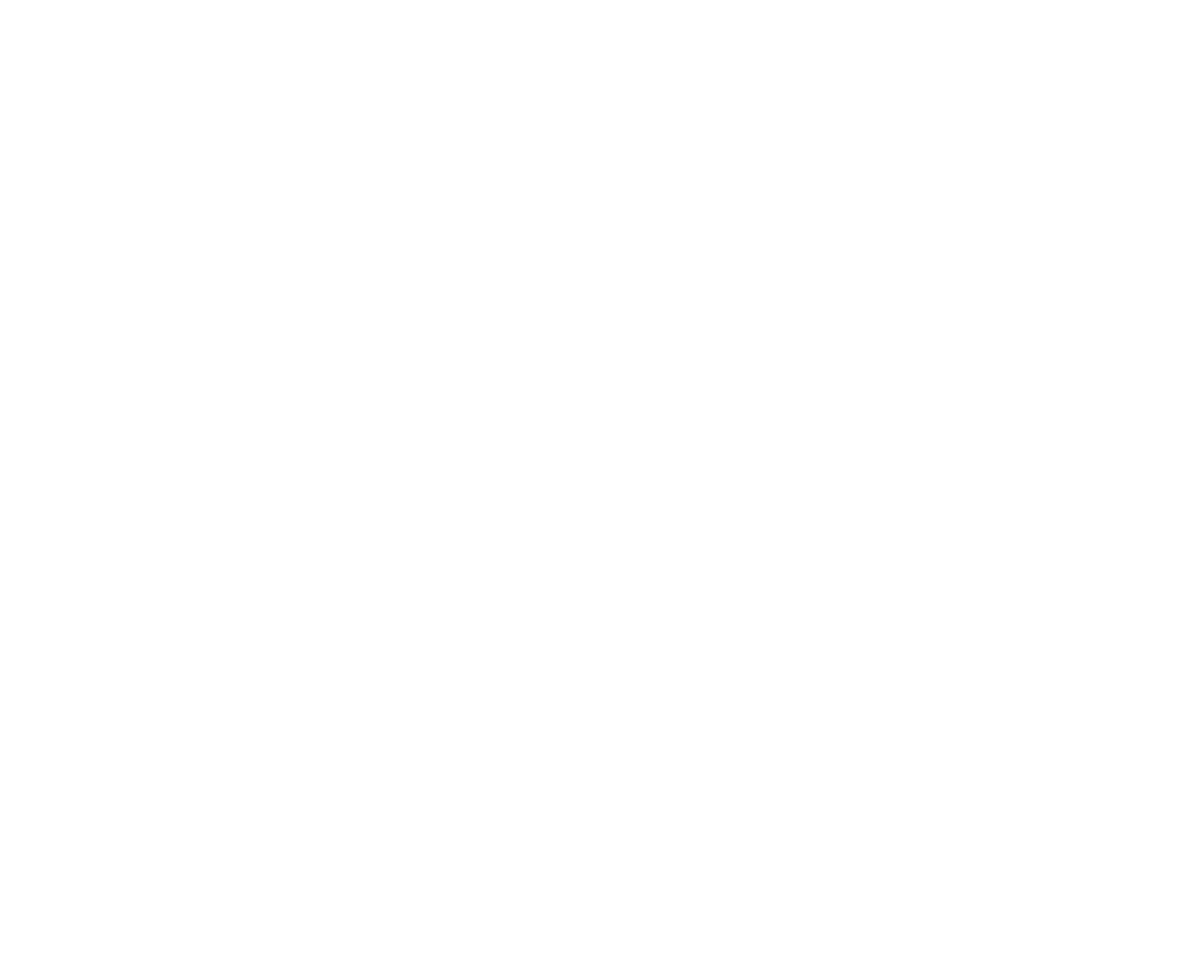} 
					& \includegraphics[width=\hsize]{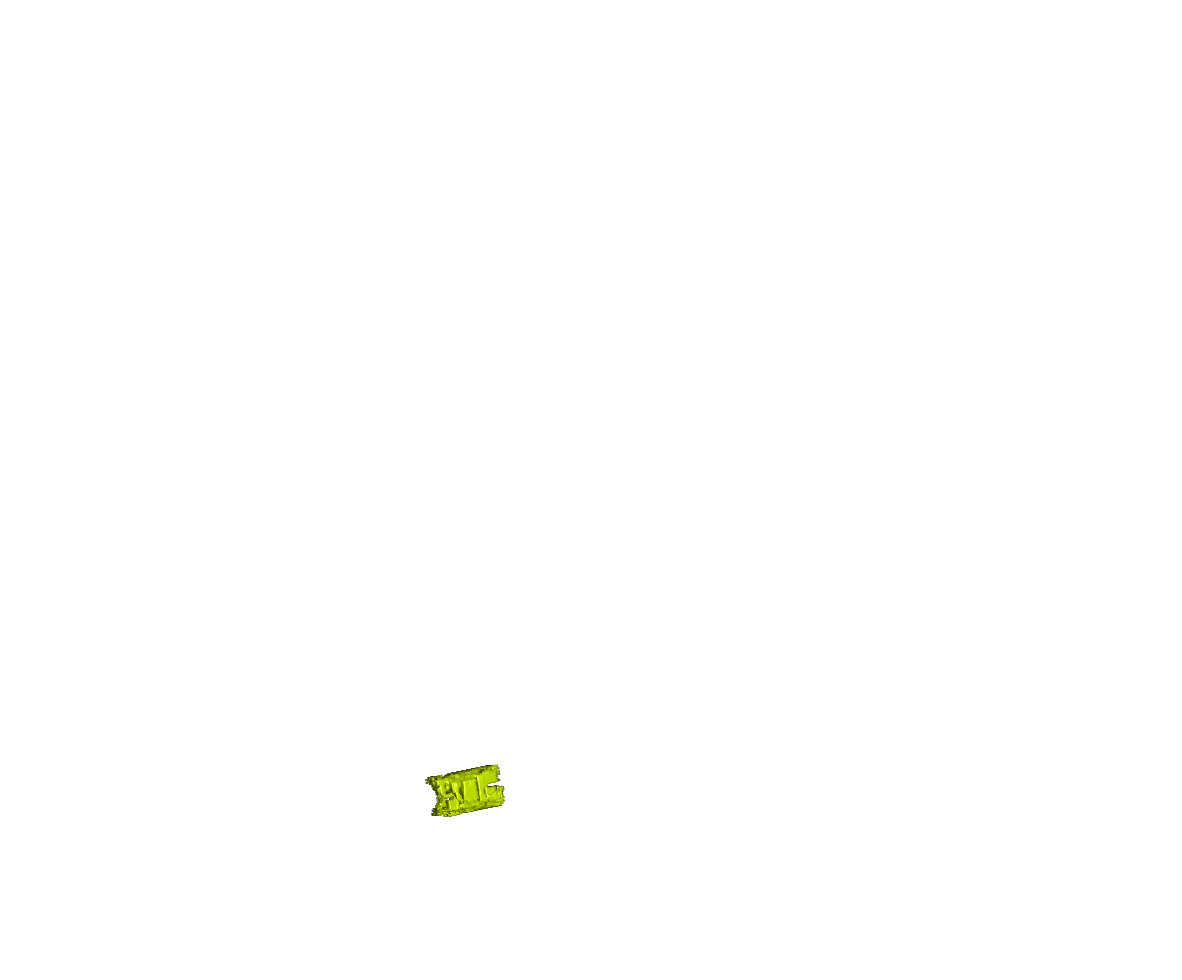} 
					& \includegraphics[width=\hsize]{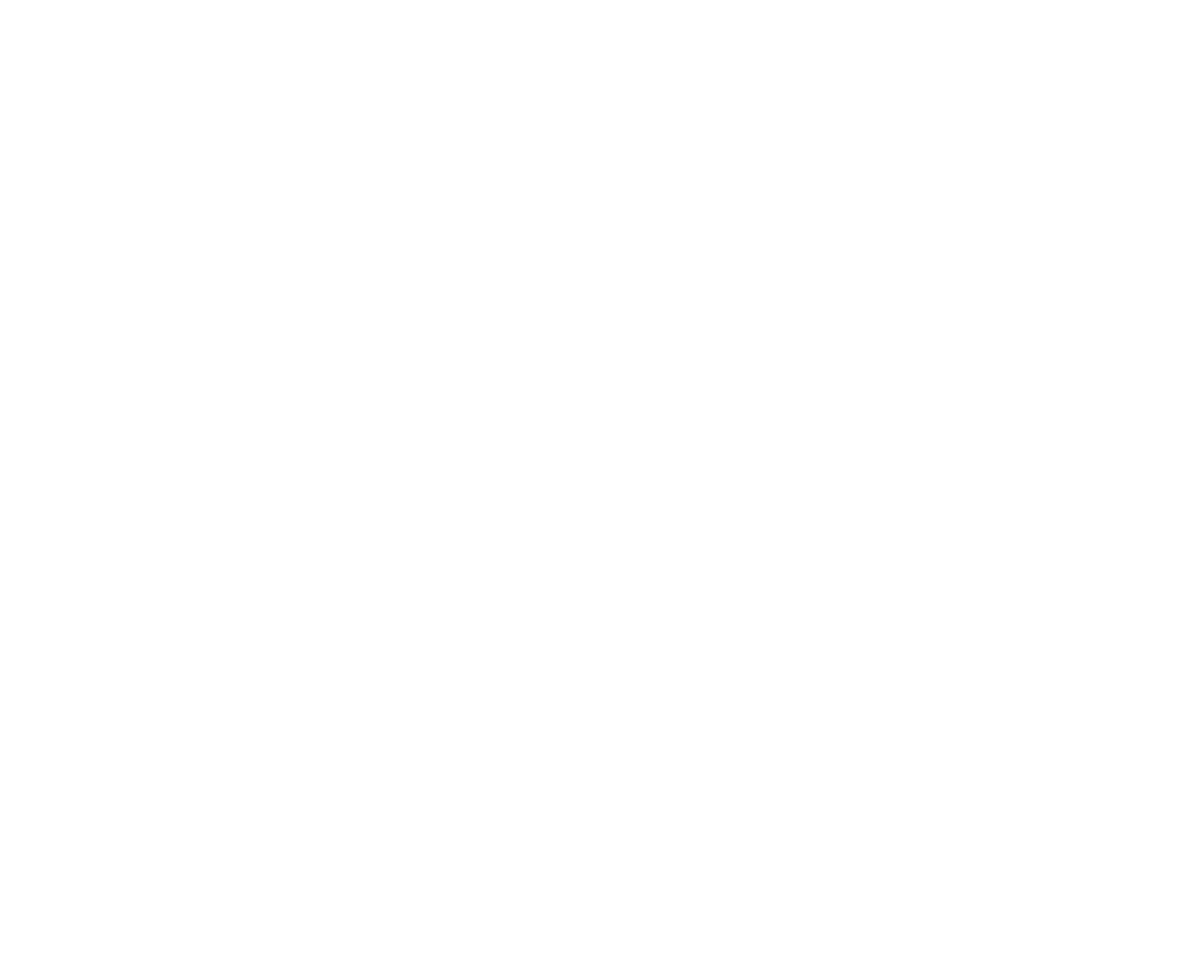} 
					\\ \addlinespace \hline\addlinespace
					
					\multirow{2}{*}{\rotatebox[origin=c]{90}{\fineTunedModel 48\hspace{1.0em}}}
					\rotatebox[origin=c]{90}{pred}
					& \includegraphics[width=\hsize]{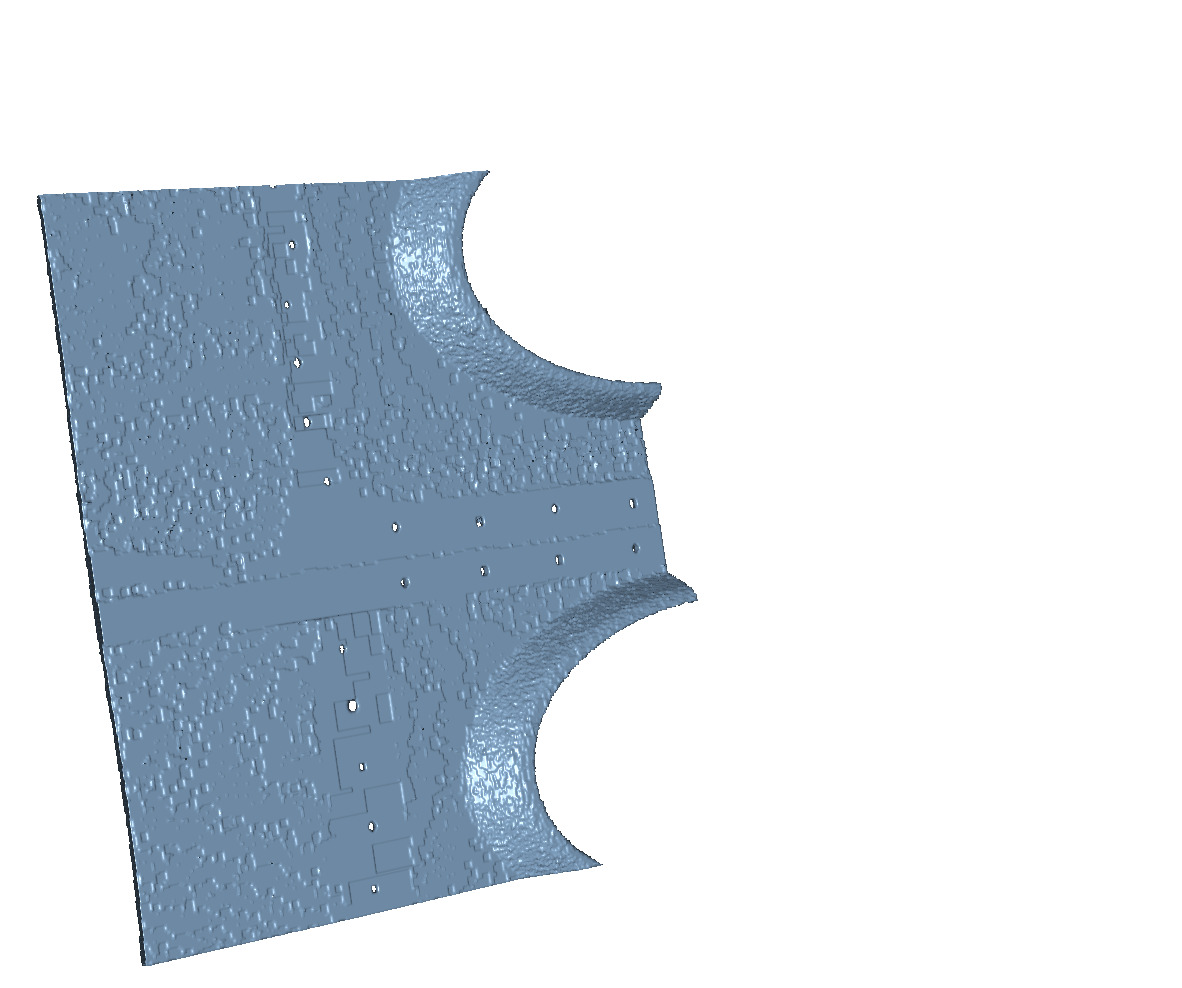} 
					& \includegraphics[width=\hsize]{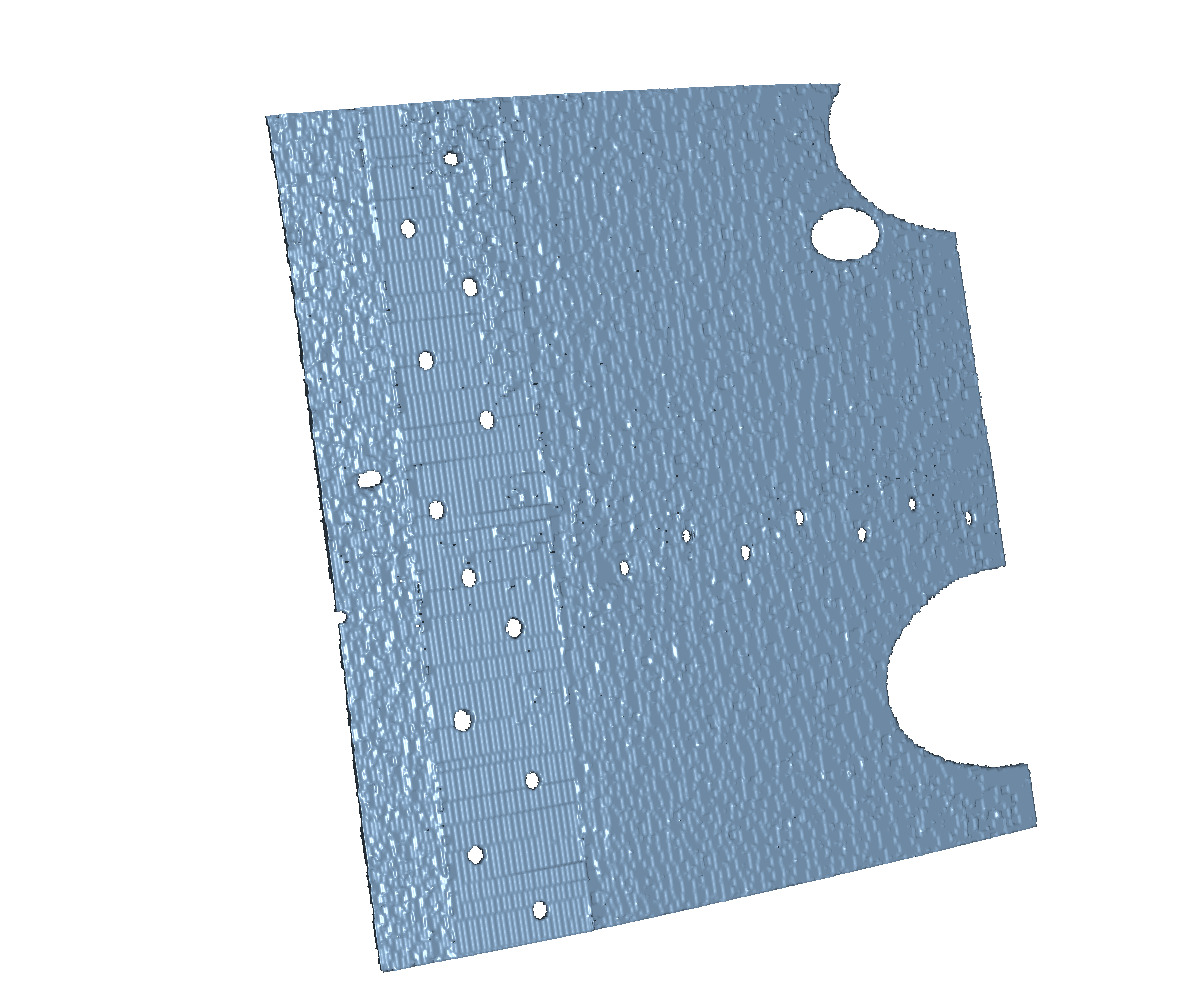} 
					& \includegraphics[width=\hsize]{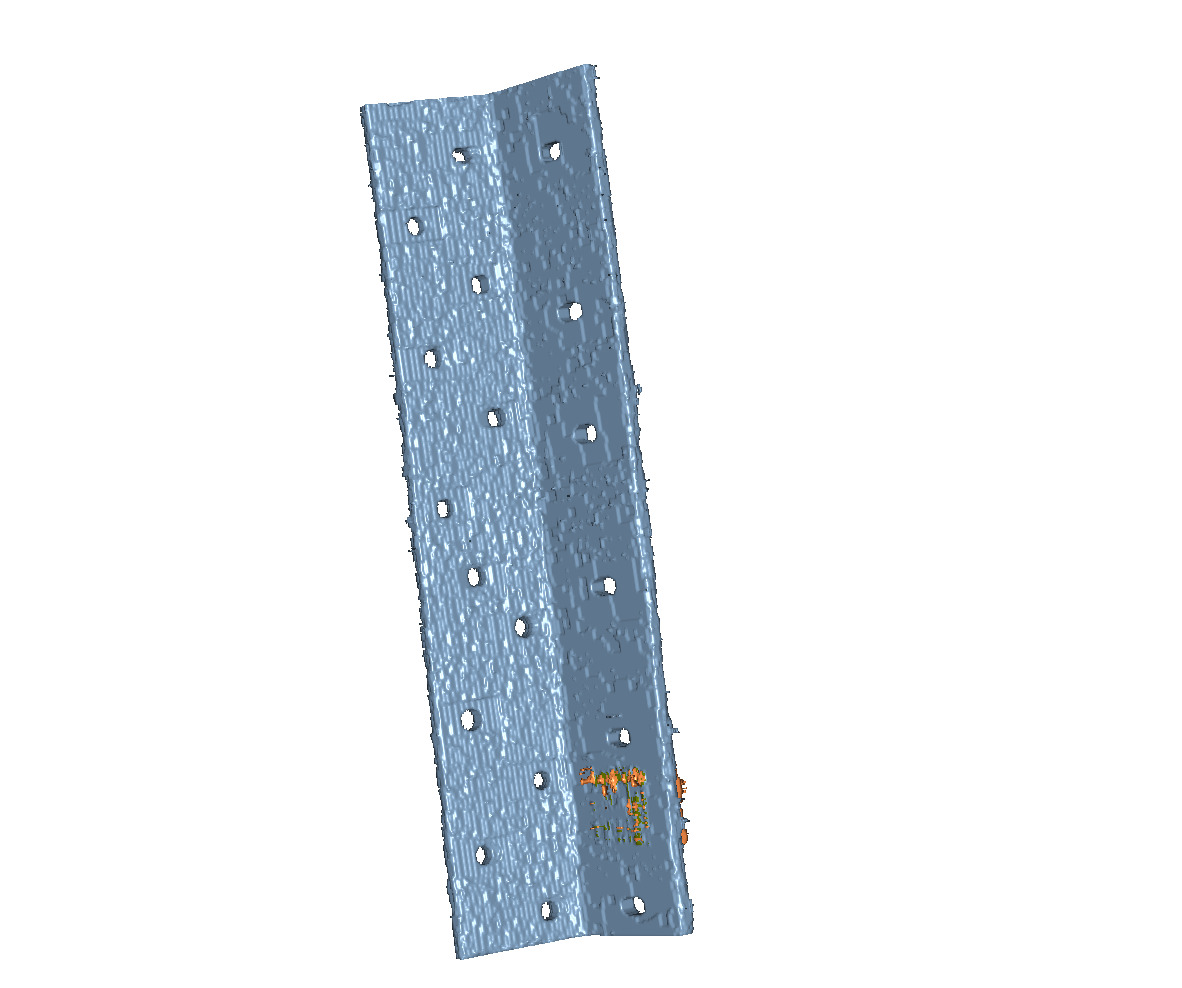} 
					& \includegraphics[width=\hsize]{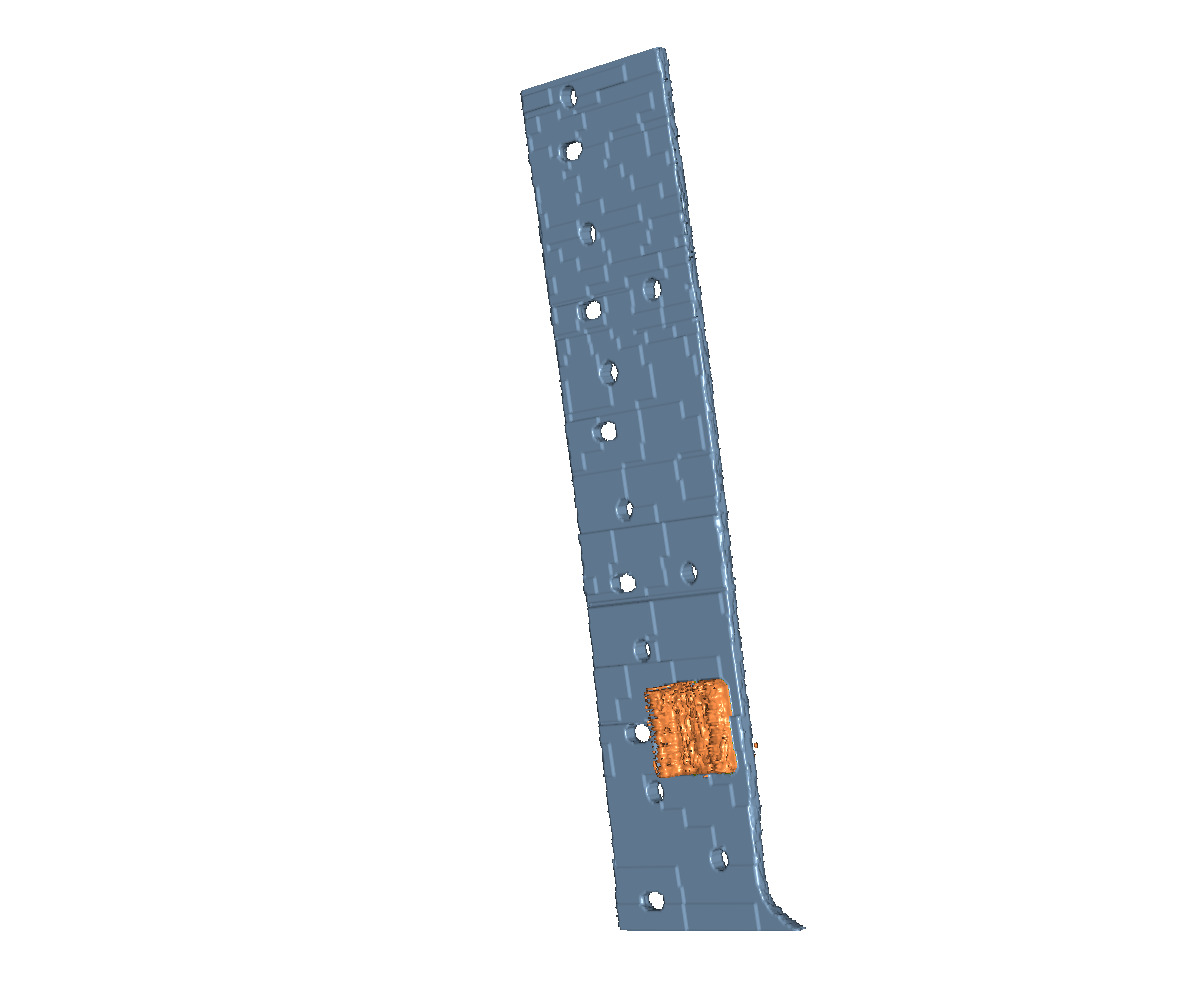} 
					& \includegraphics[width=\hsize]{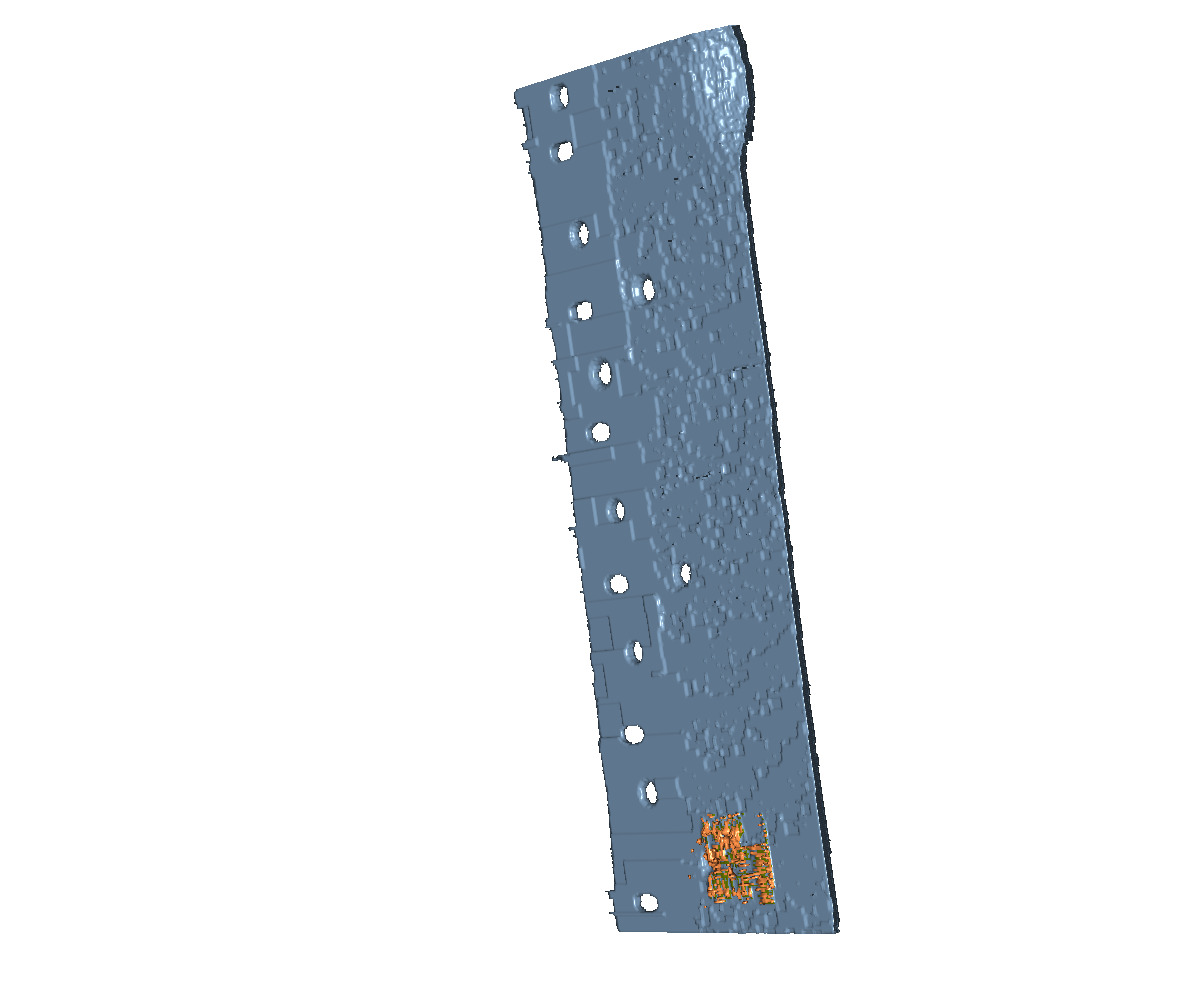} 
					& \includegraphics[width=\hsize]{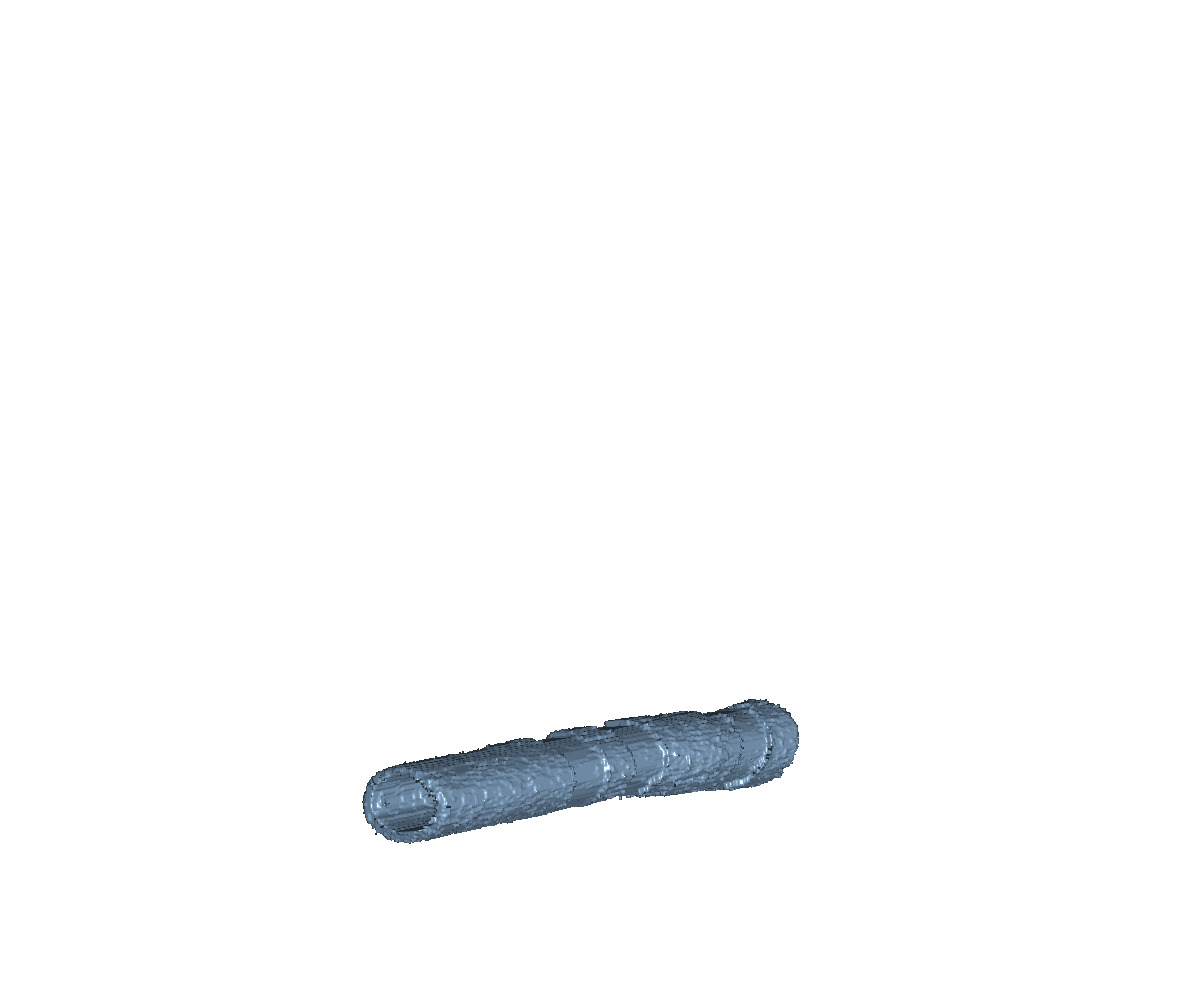} 
					& \includegraphics[width=\hsize]{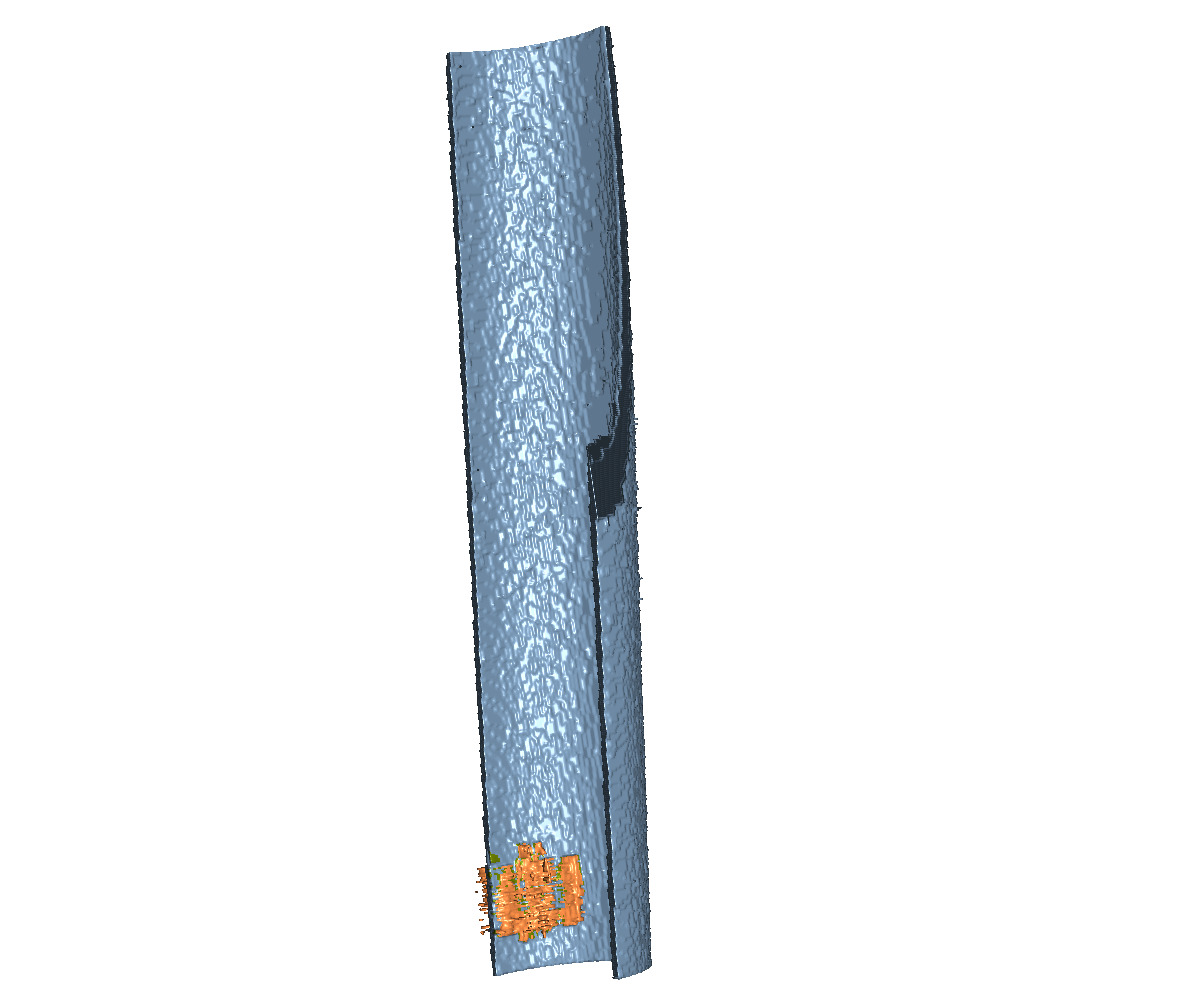} 
					\\ \addlinespace					
					\rotatebox[origin=c]{90}{TP}
					& \includegraphics[width=\hsize]{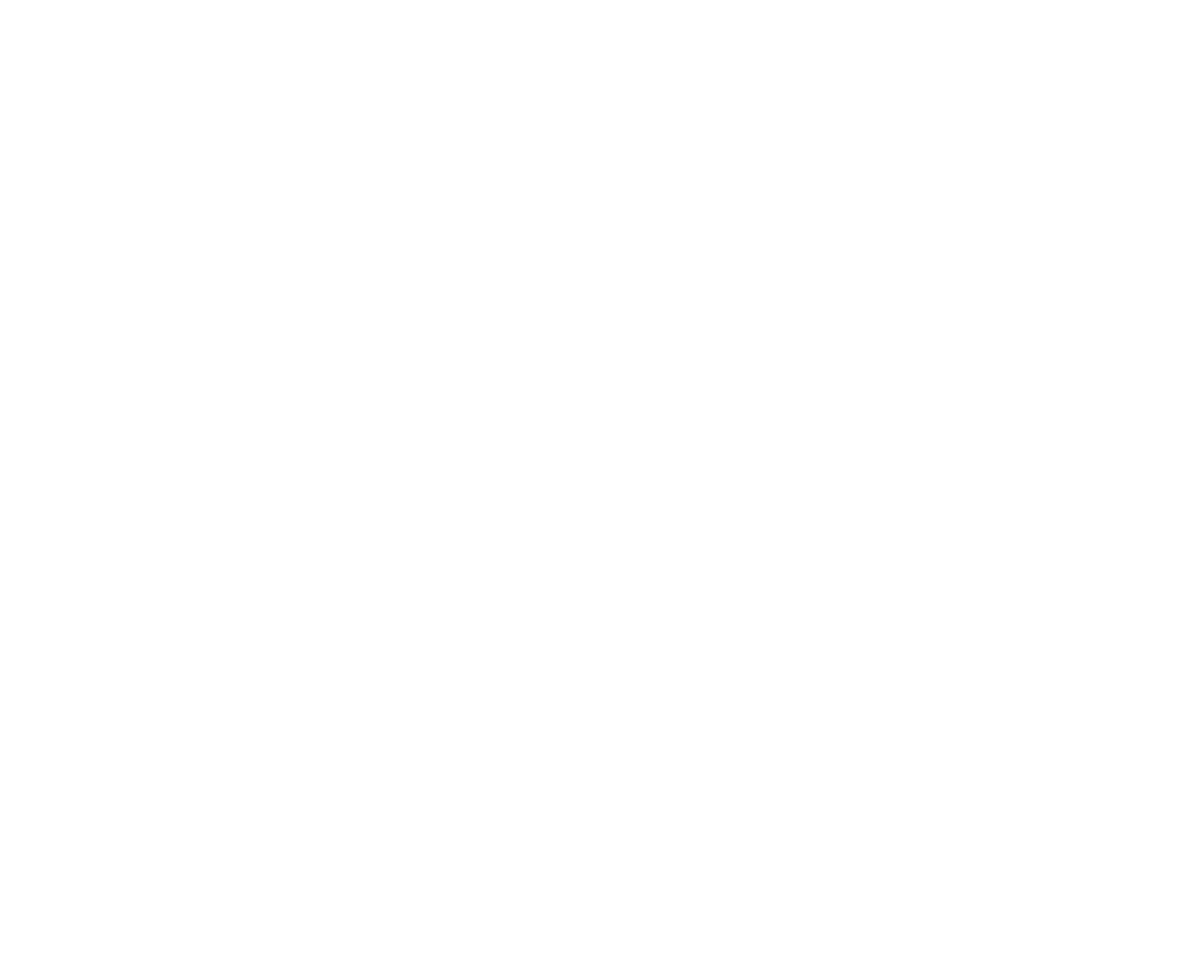} 
					& \includegraphics[width=\hsize]{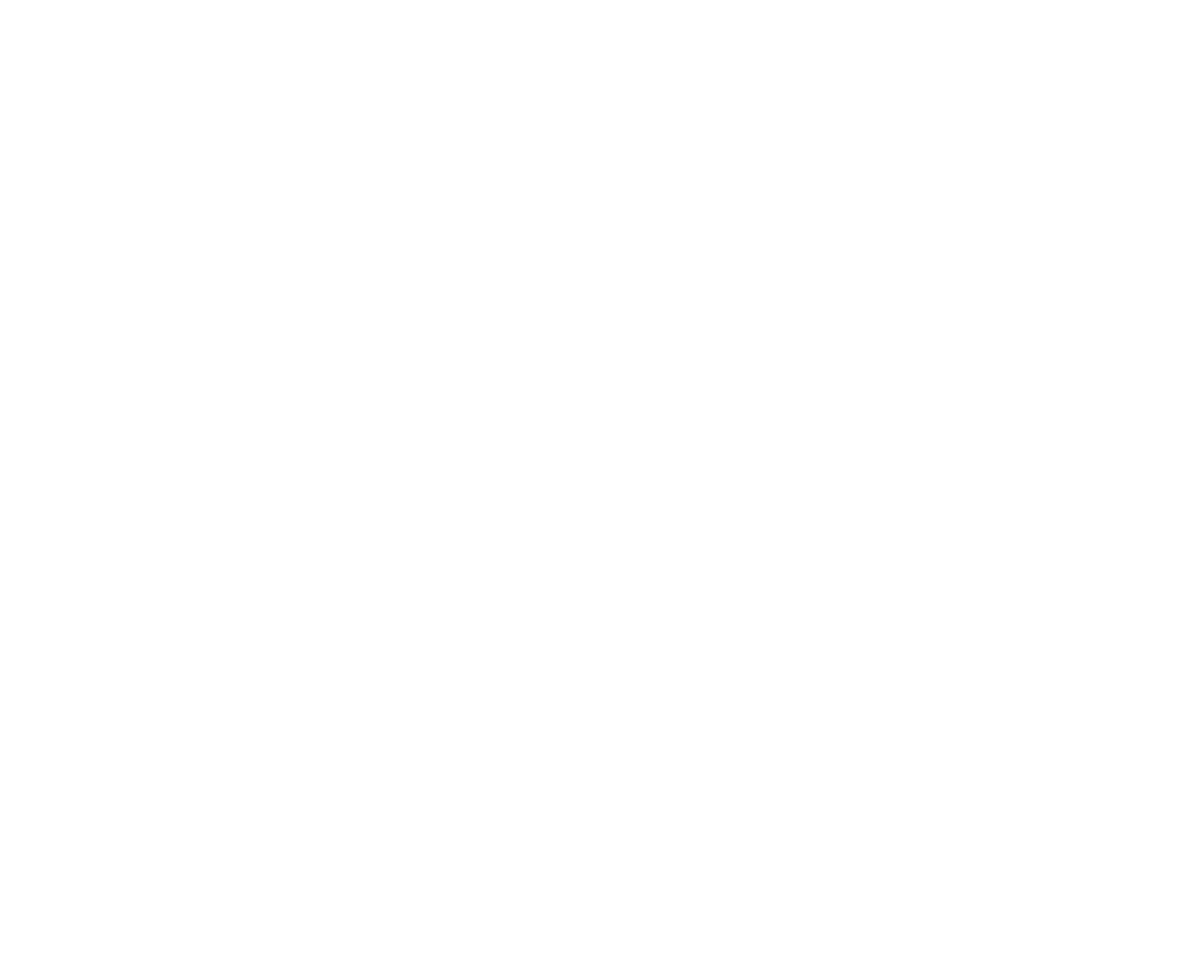} 
					& \includegraphics[width=\hsize]{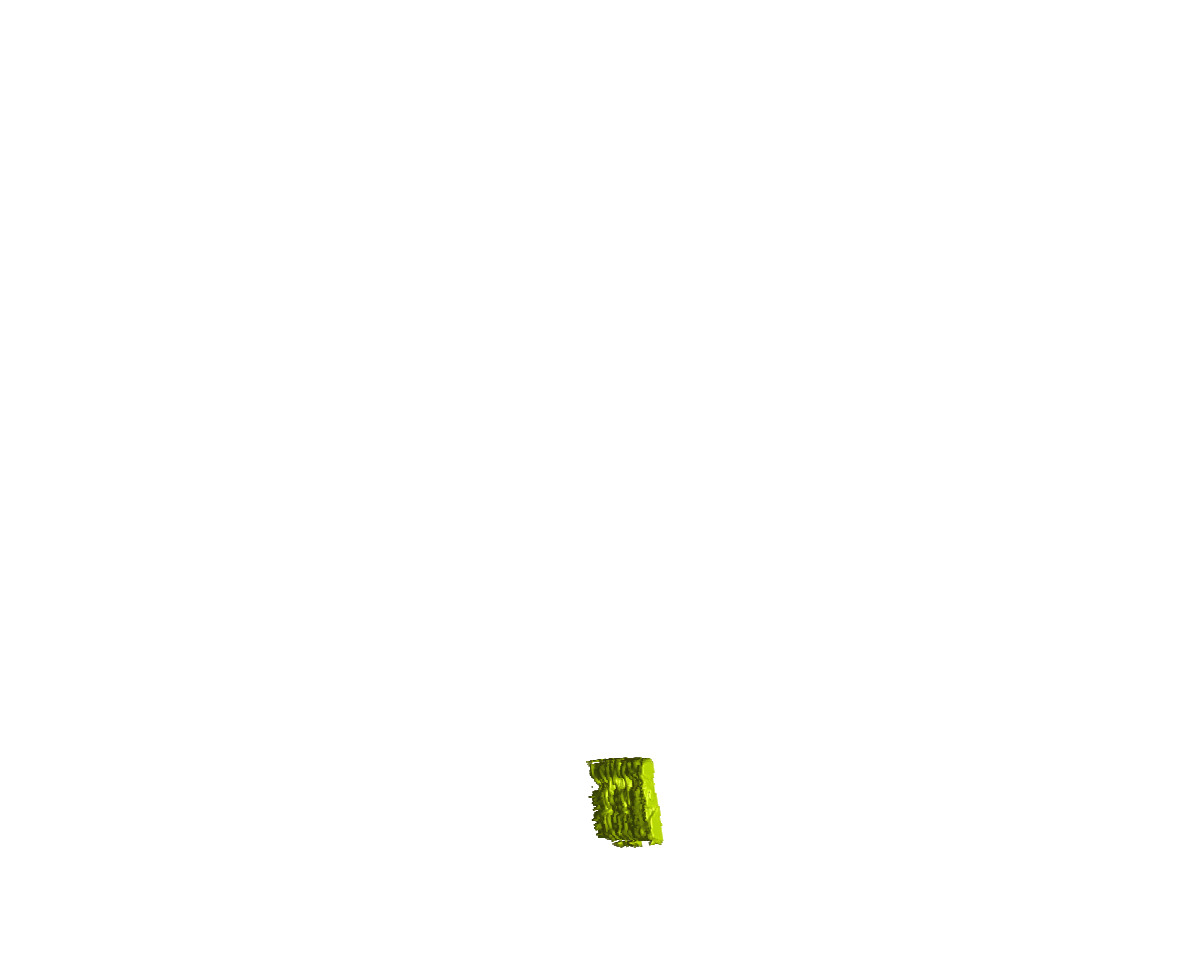} 
					& \includegraphics[width=\hsize]{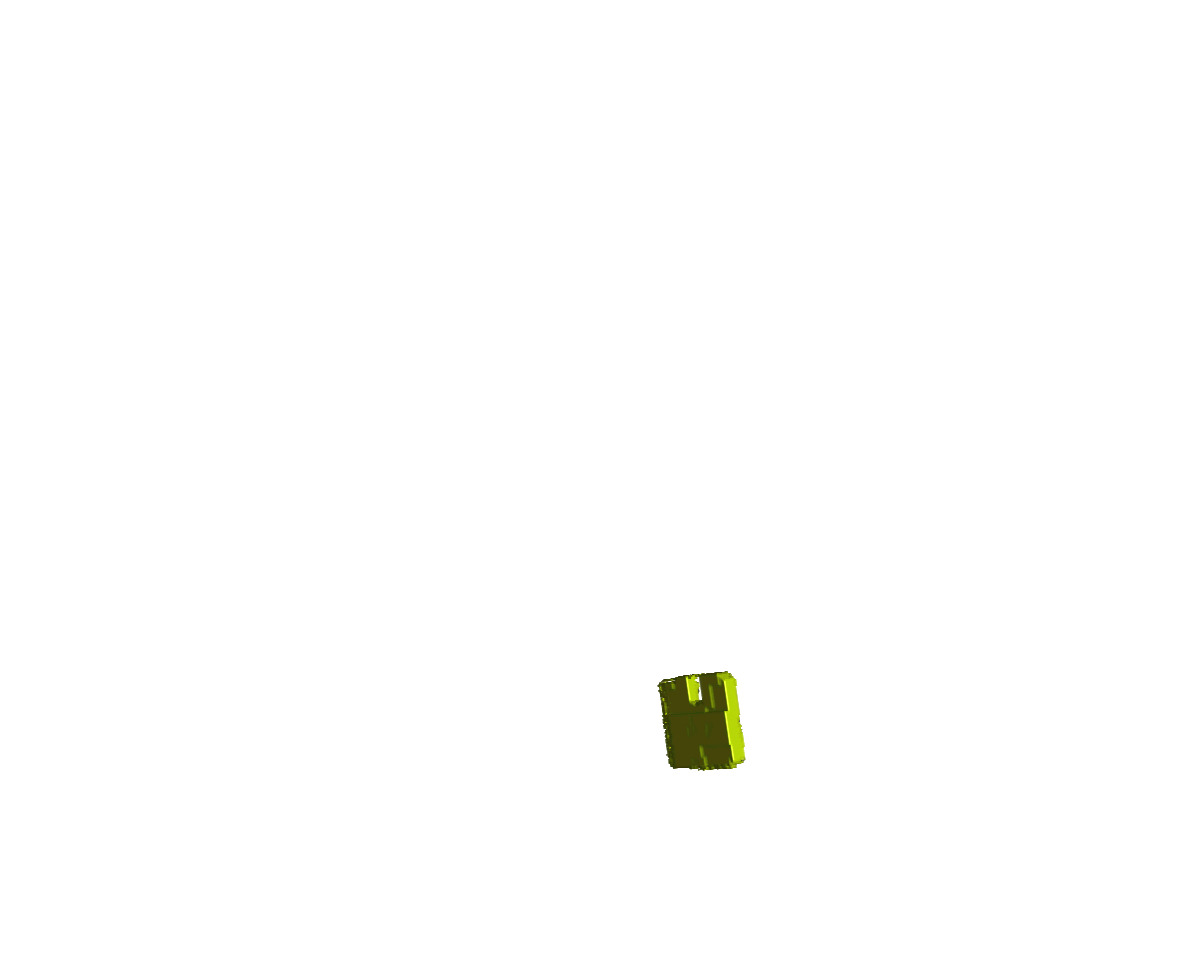} 
					& \includegraphics[width=\hsize]{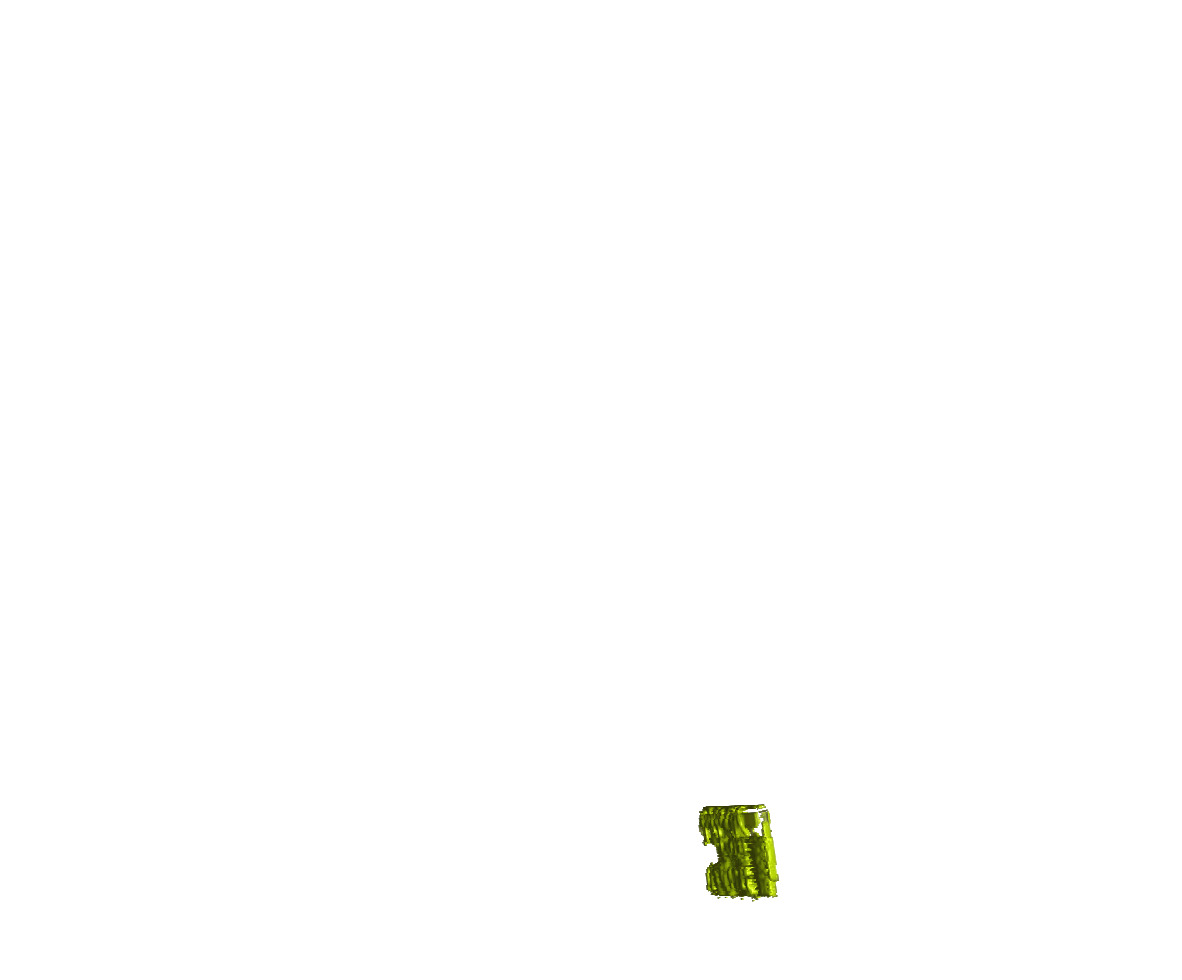} 
					& \includegraphics[width=\hsize]{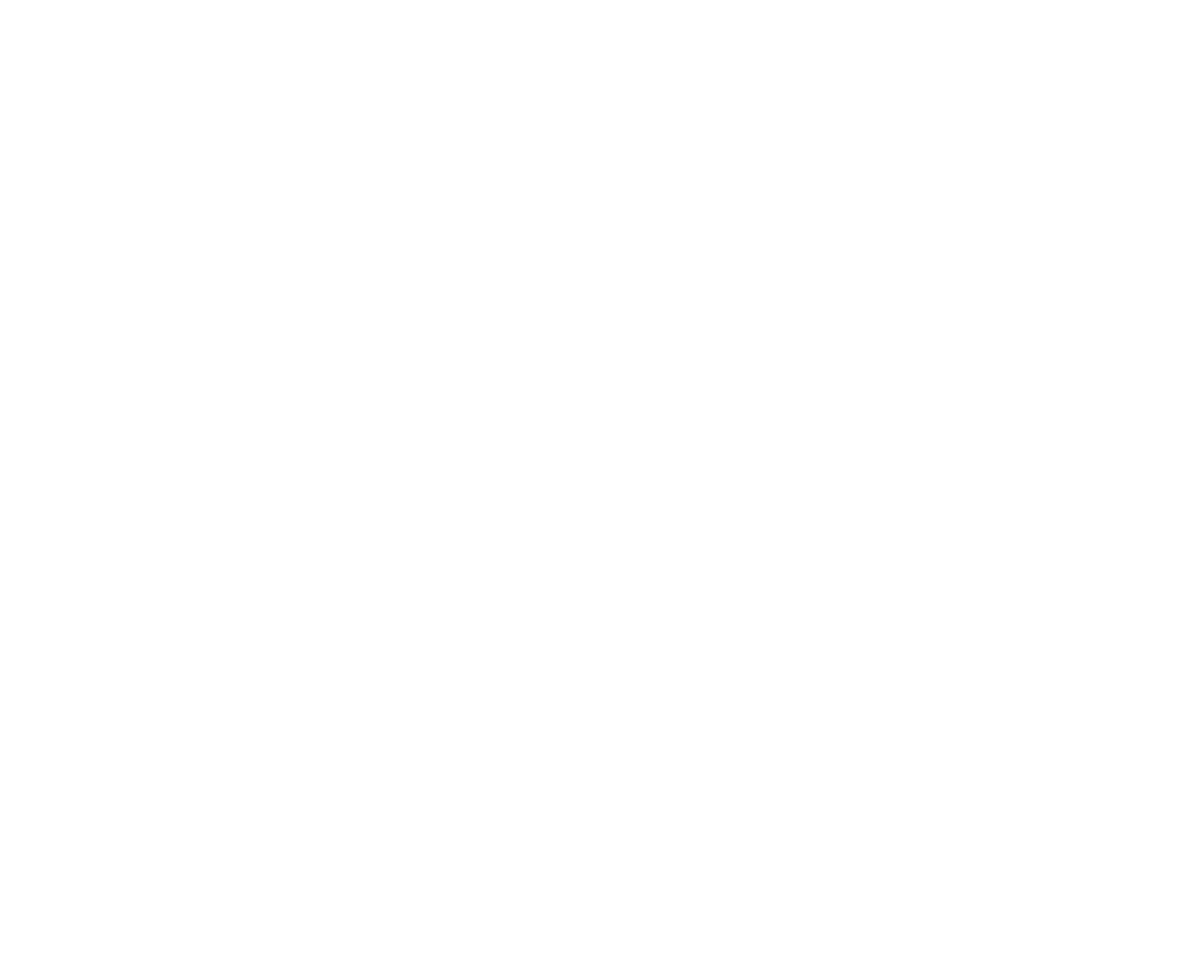} 
					& \includegraphics[width=\hsize]{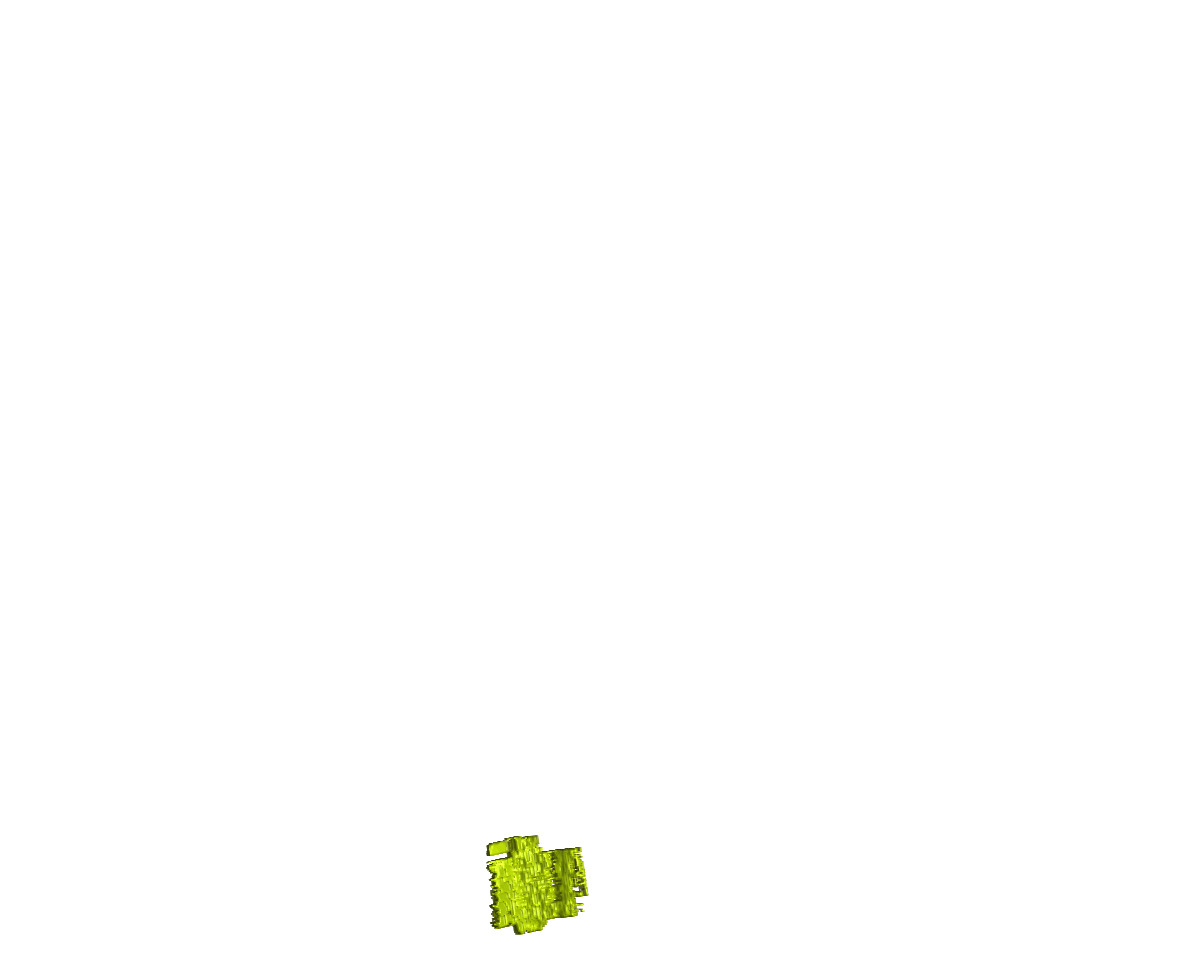} 
					\\ \addlinespace \hline \addlinespace	
					
					\multirow{2}{*}{\rotatebox[origin=c]{90}{vit\_b 1024\hspace{1.5em}}}
					\rotatebox[origin=c]{90}{pred}
					& \includegraphics[width=\hsize]{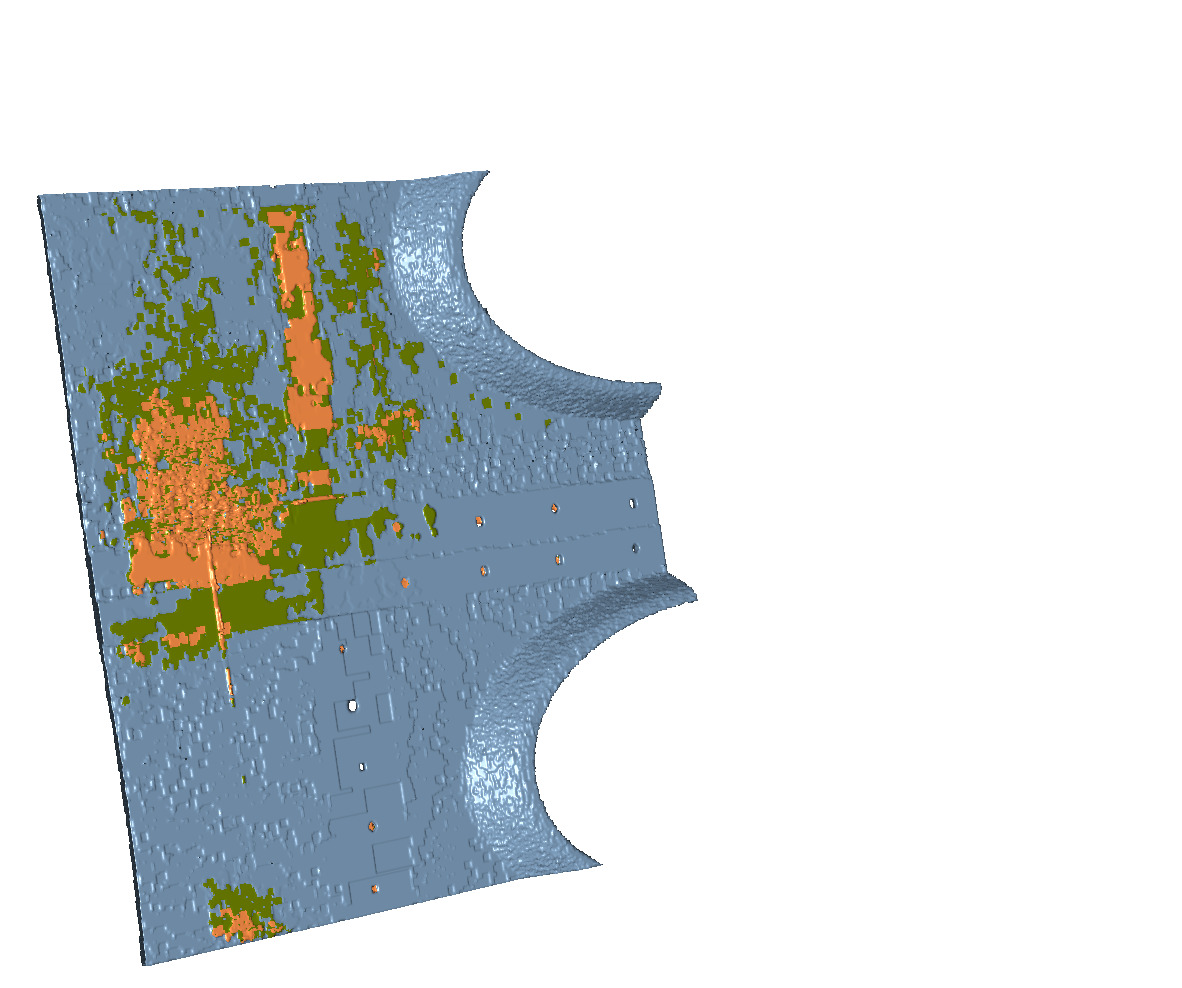} 
					& \includegraphics[width=\hsize]{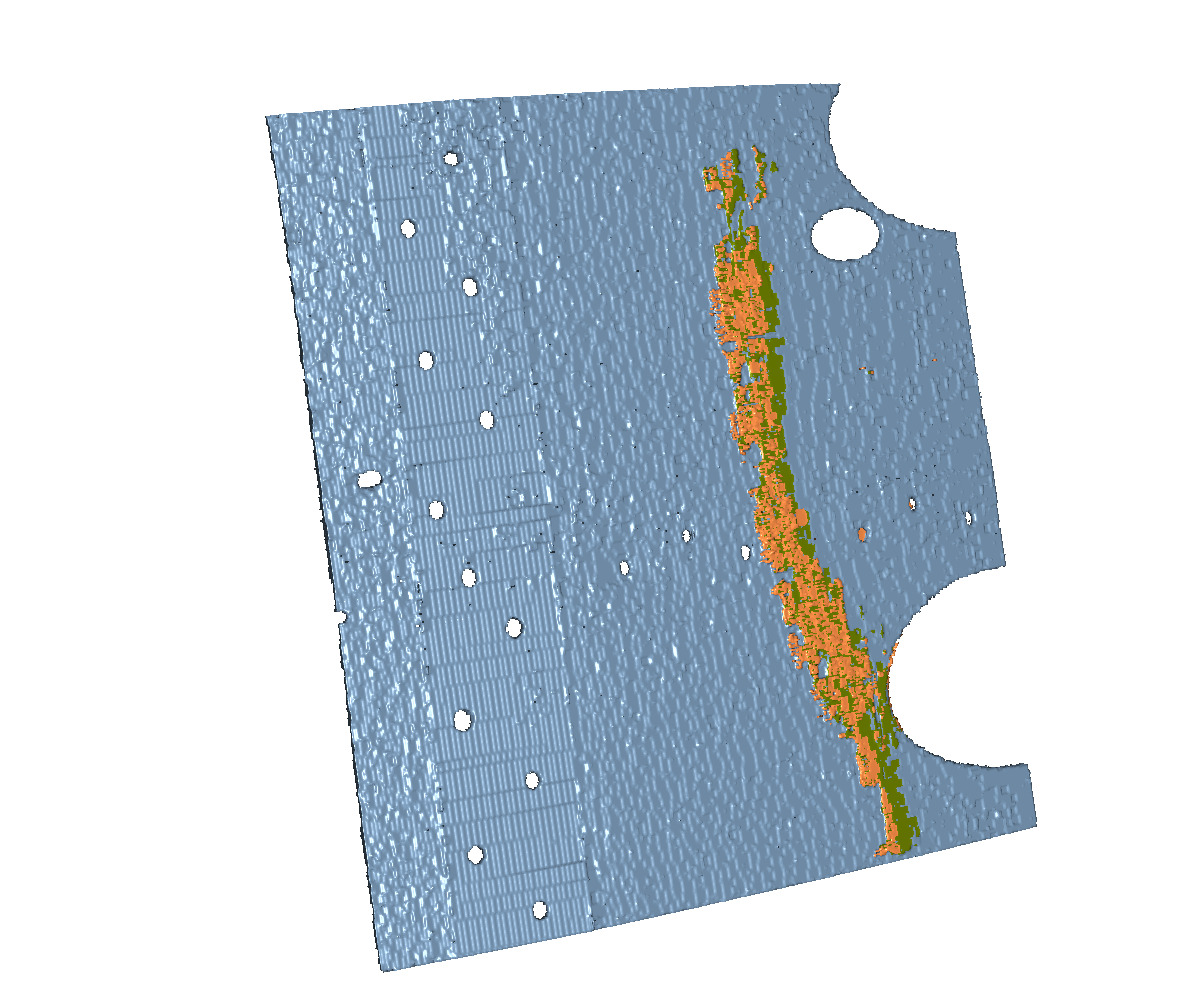} 
					& \includegraphics[width=\hsize]{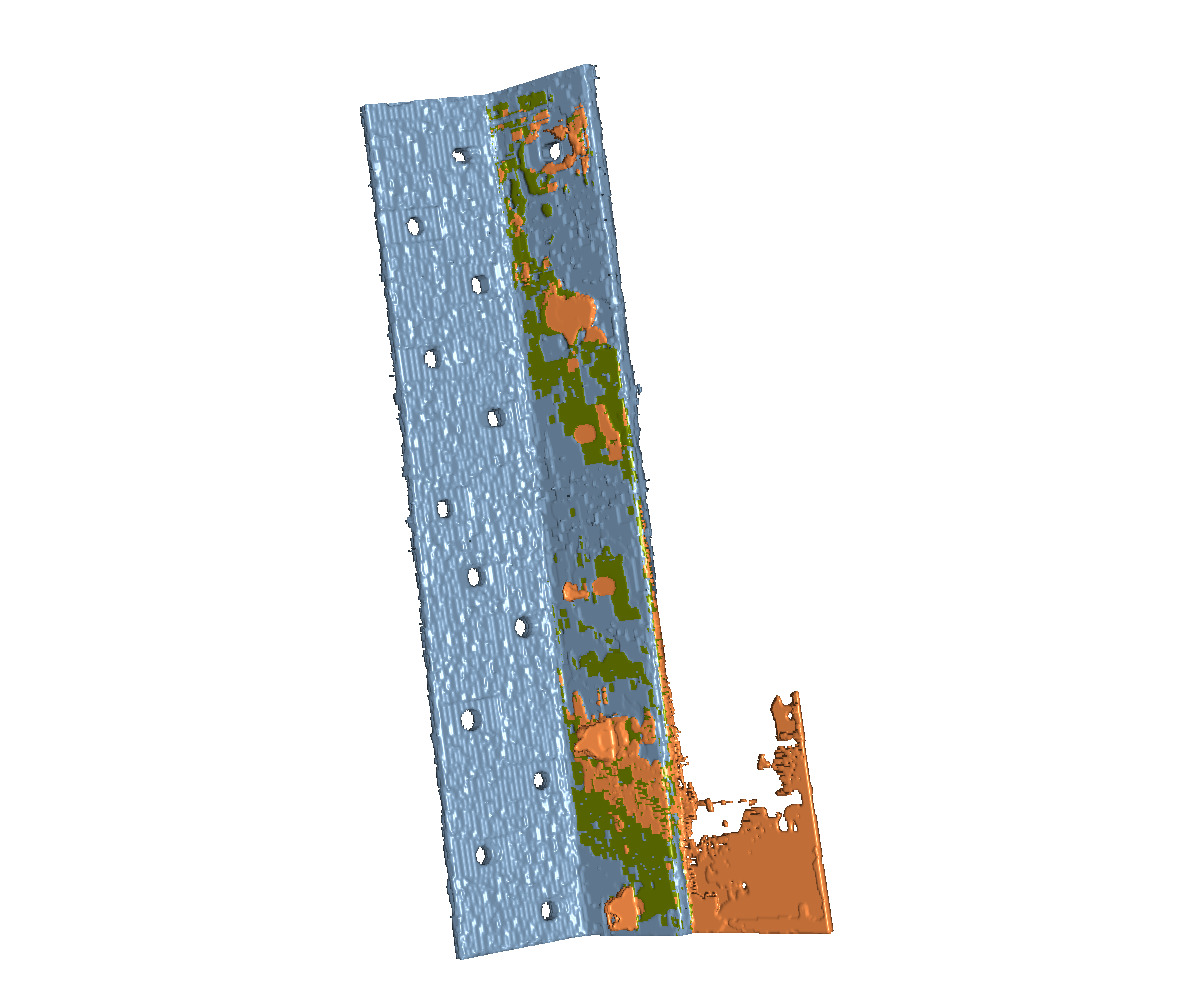} 
					& \includegraphics[width=\hsize]{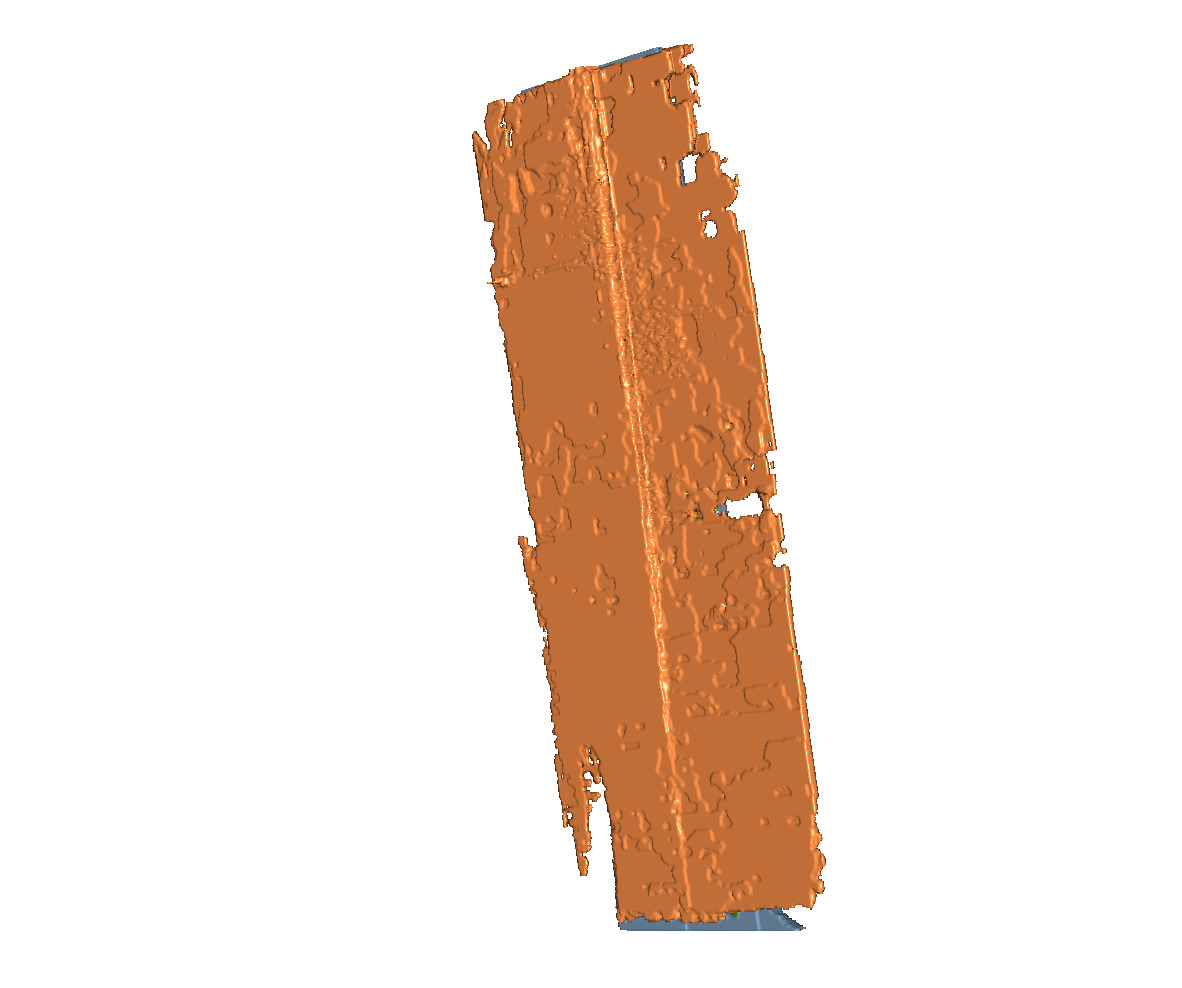} 
					& \includegraphics[width=\hsize]{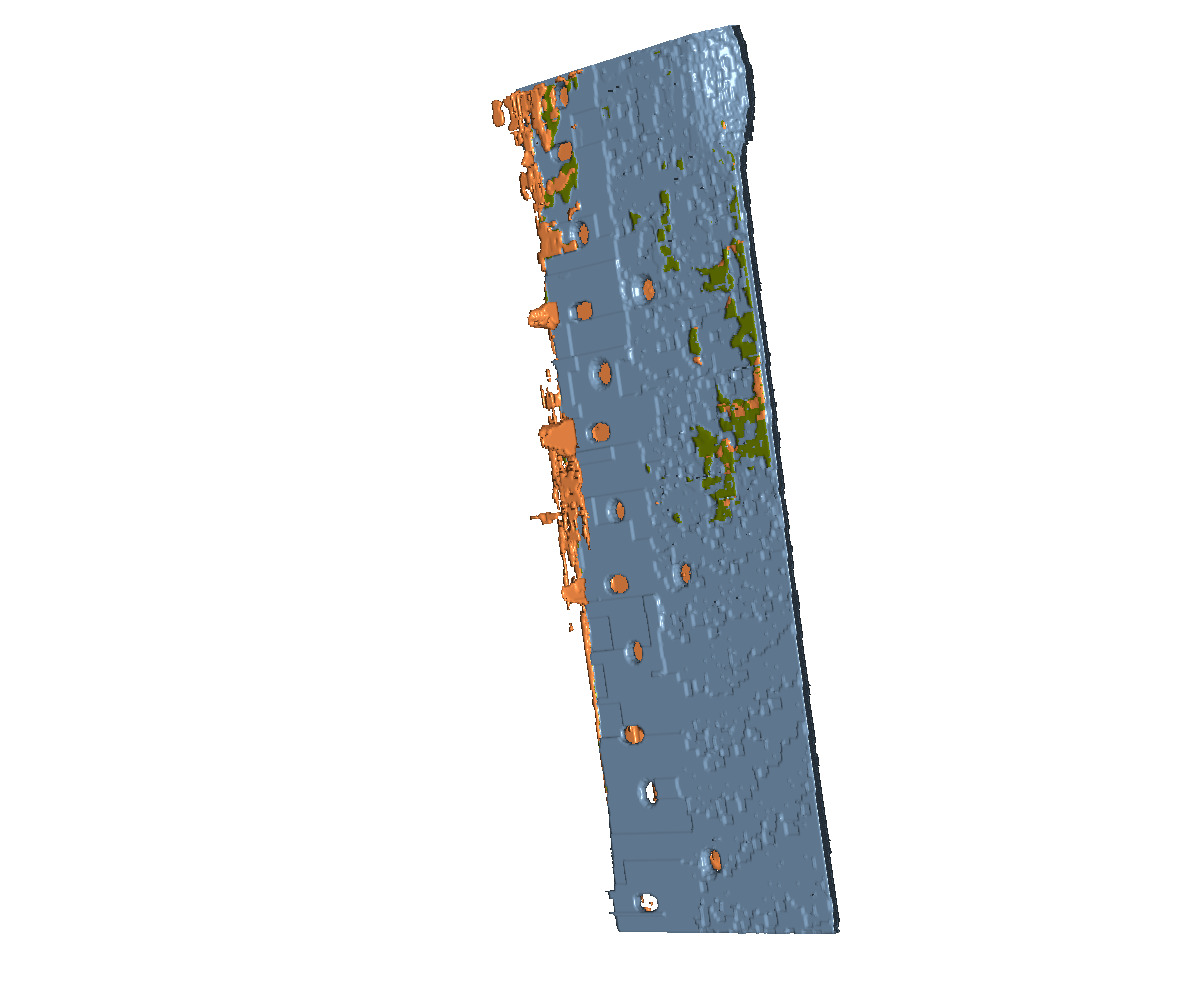} 
					& \includegraphics[width=\hsize]{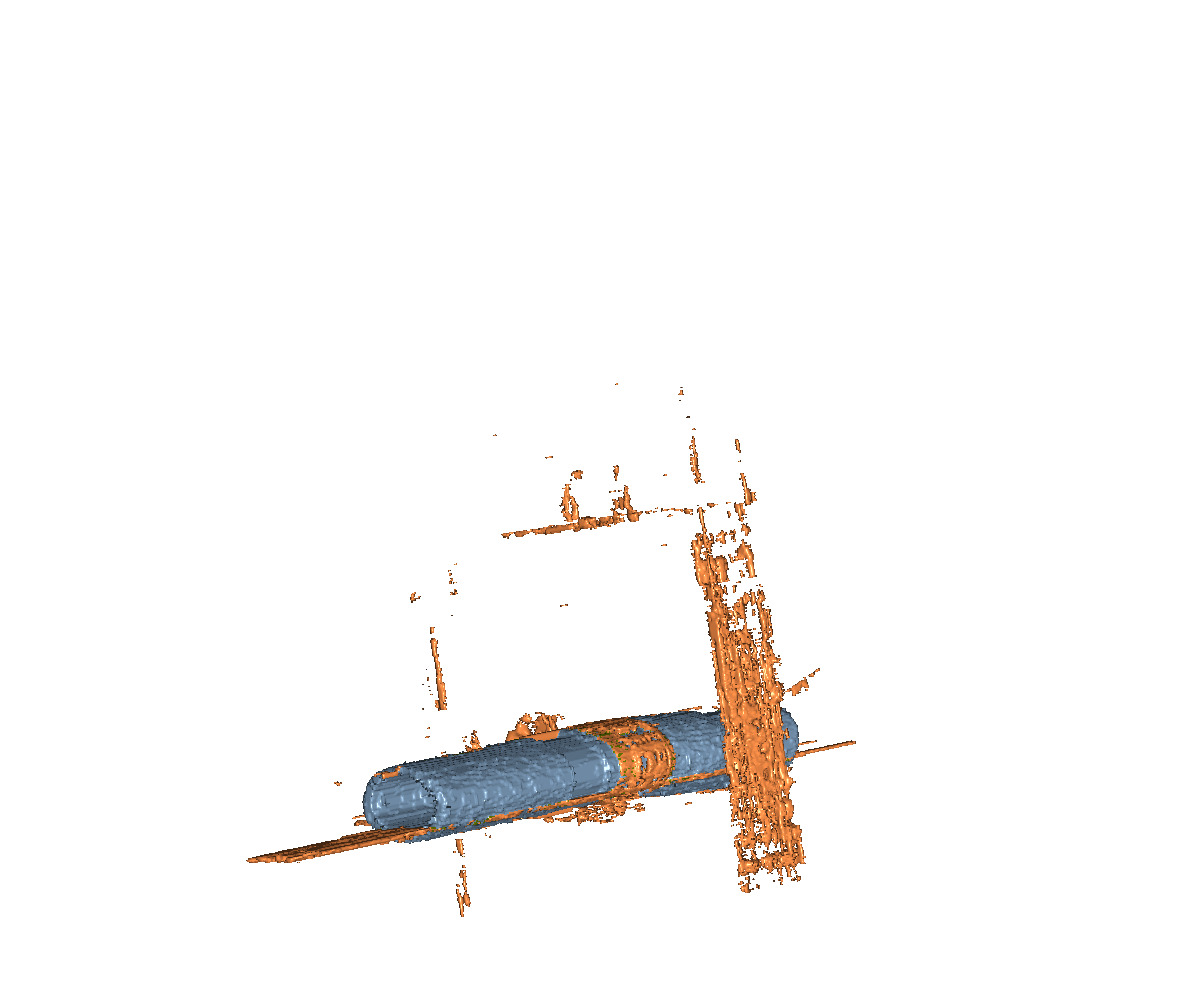} 
					& \includegraphics[width=\hsize]{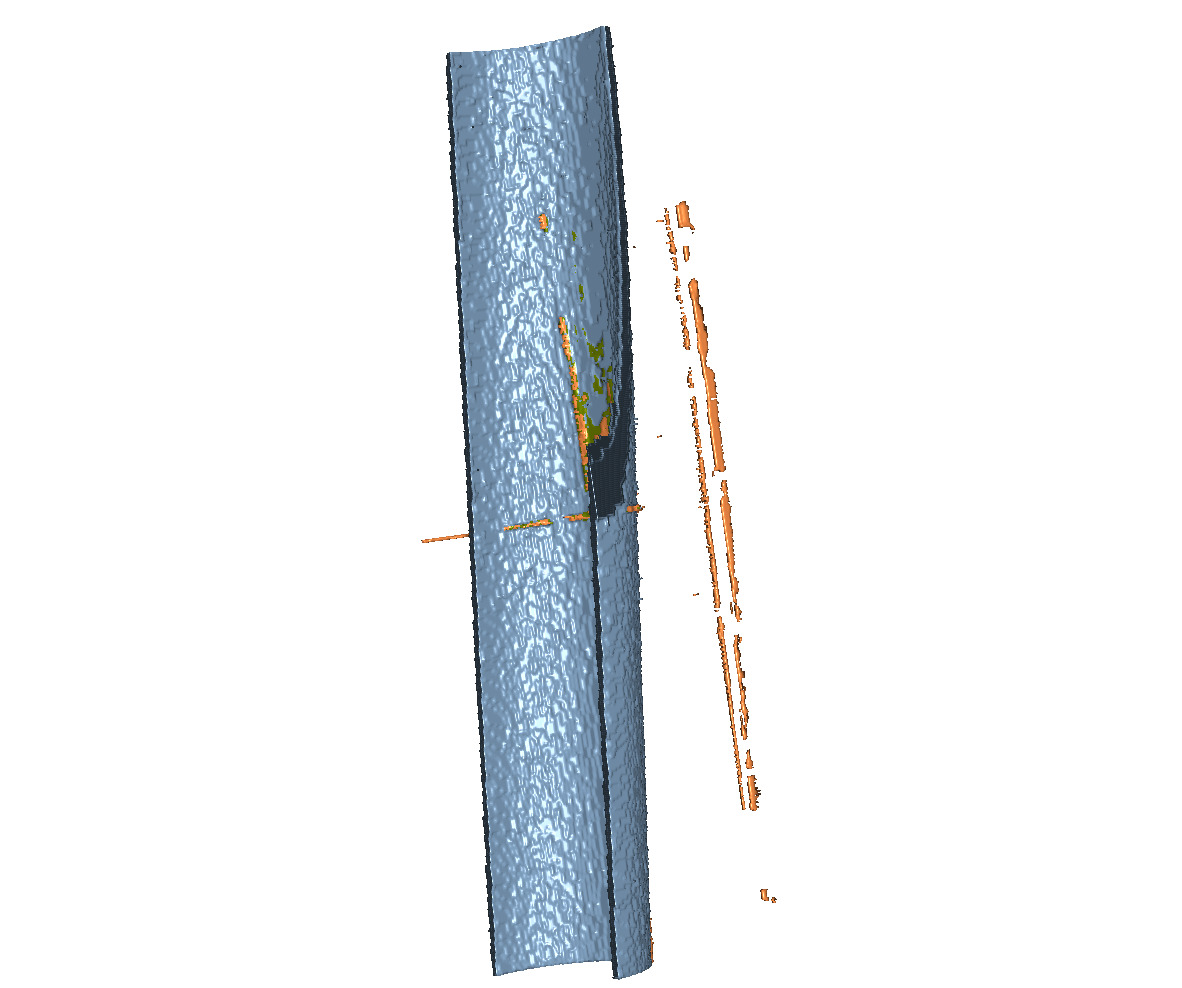} 
					\\ \addlinespace					
					\rotatebox[origin=c]{90}{TP}
					& \includegraphics[width=\hsize]{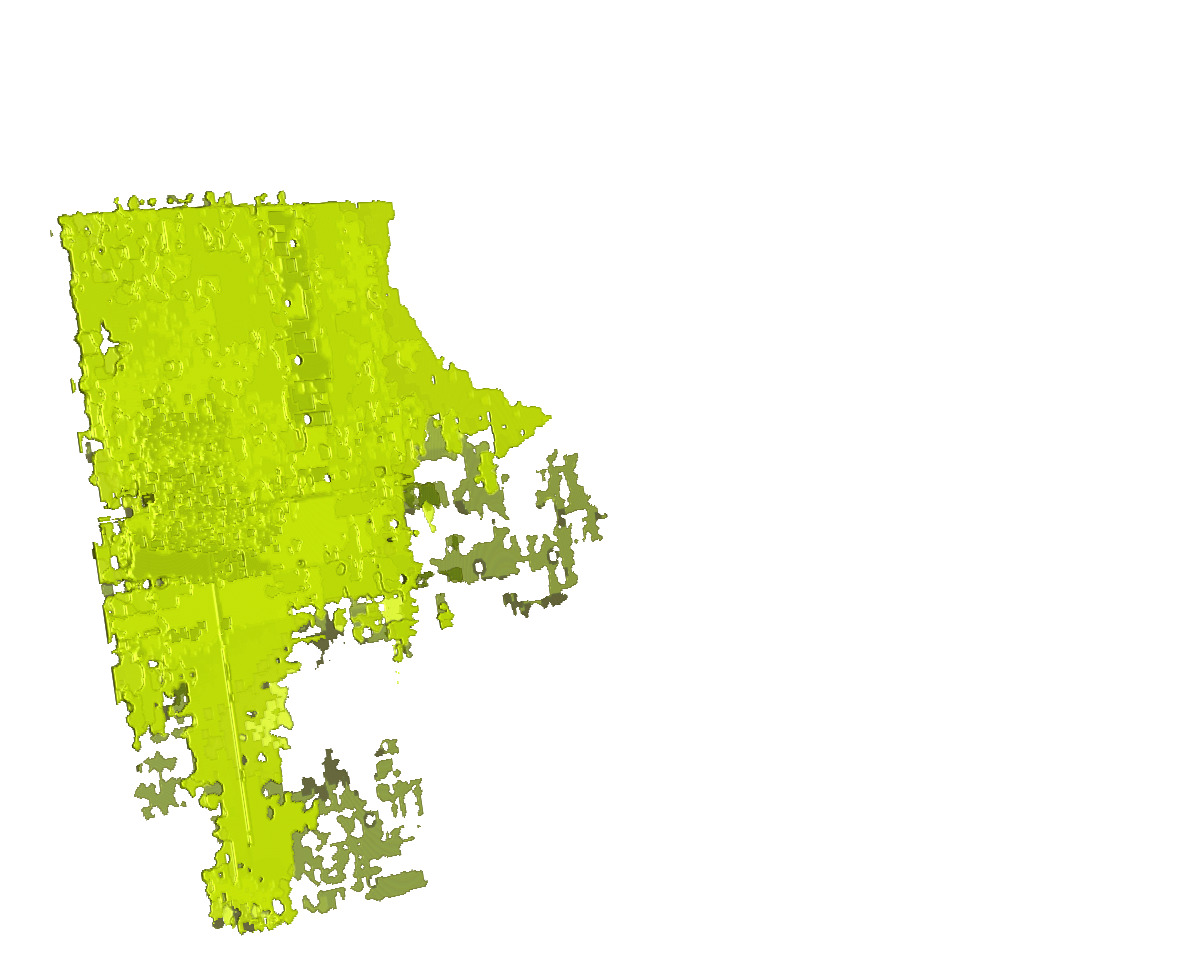} 
					& \includegraphics[width=\hsize]{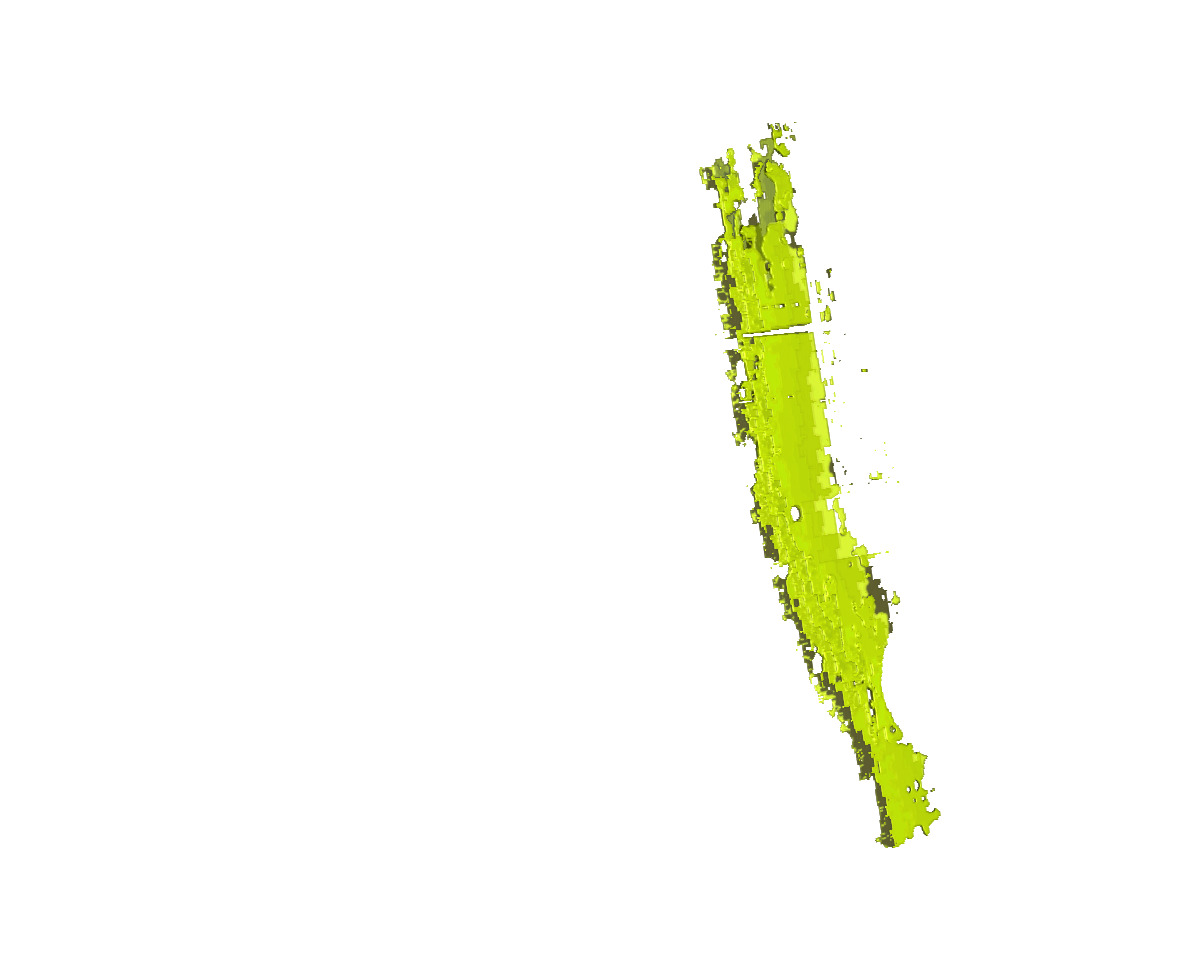} 
					& \includegraphics[width=\hsize]{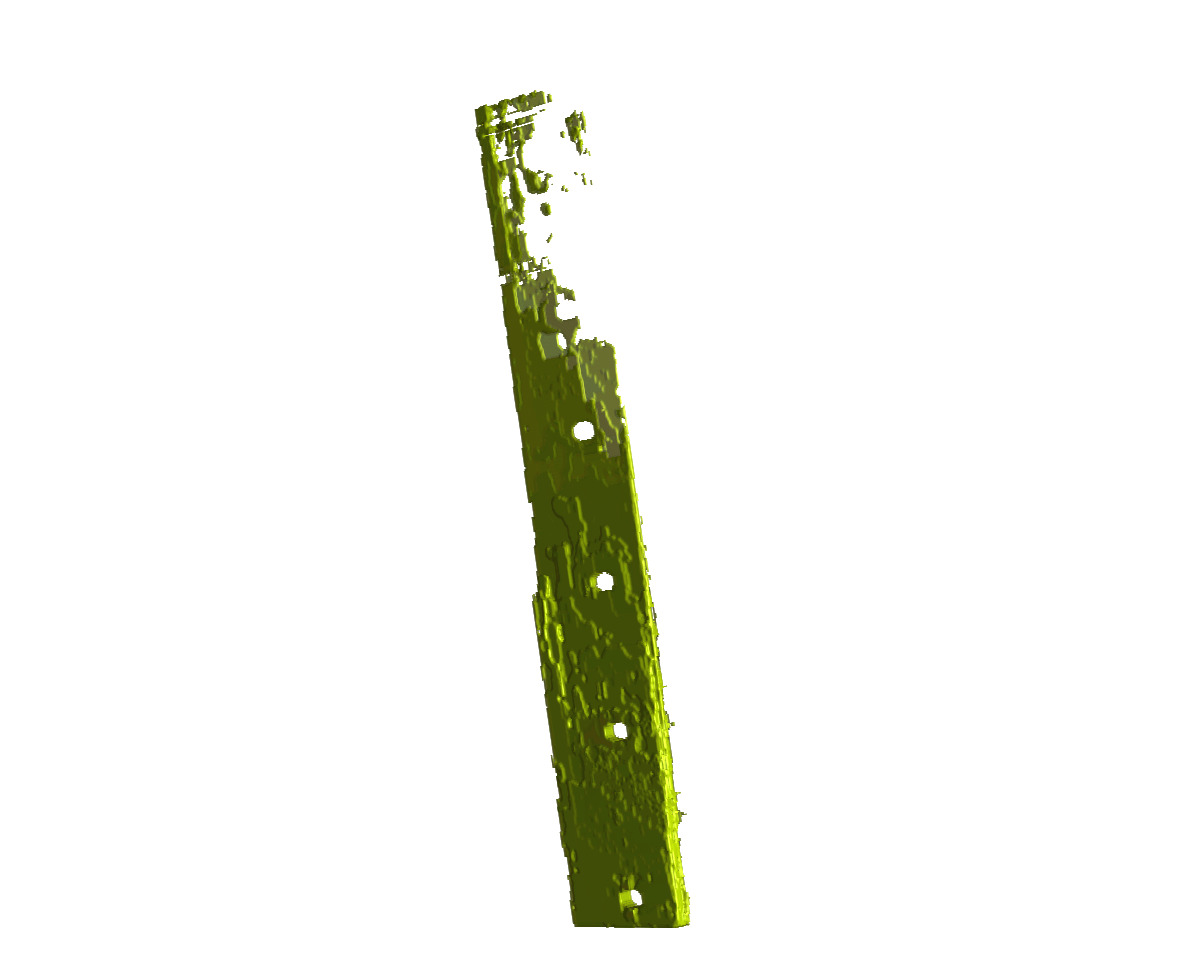} 
					& \includegraphics[width=\hsize]{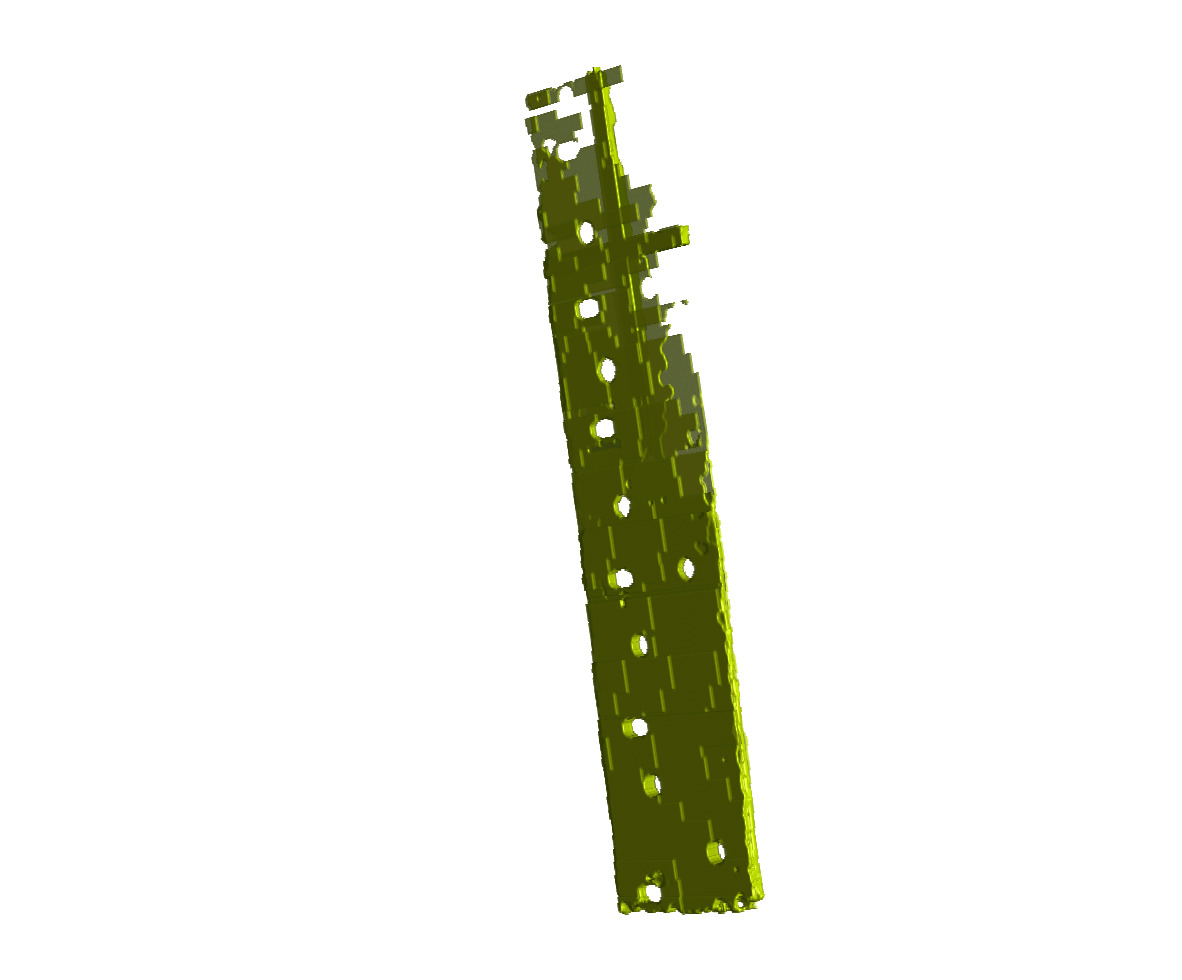} 
					& \includegraphics[width=\hsize]{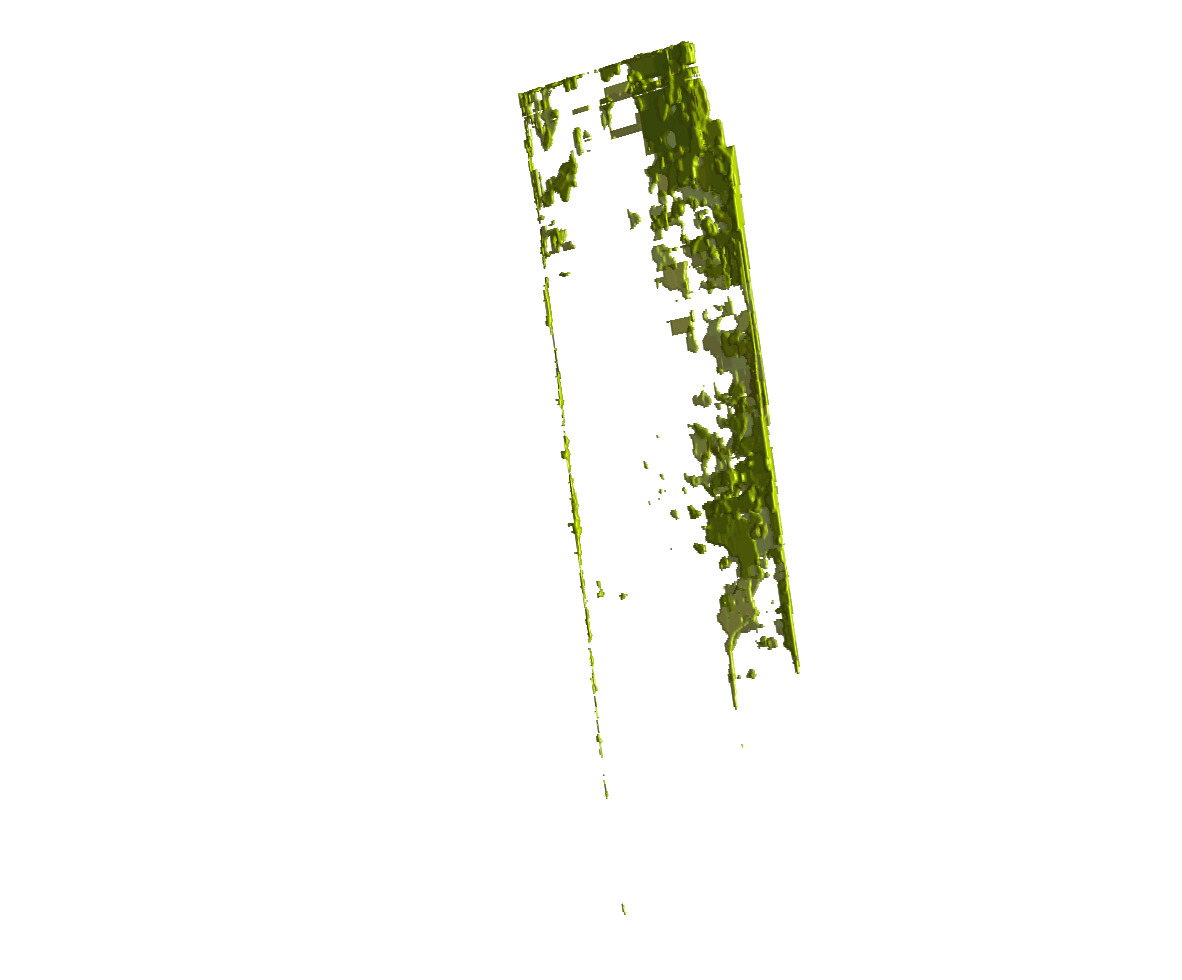} 
					& \includegraphics[width=\hsize]{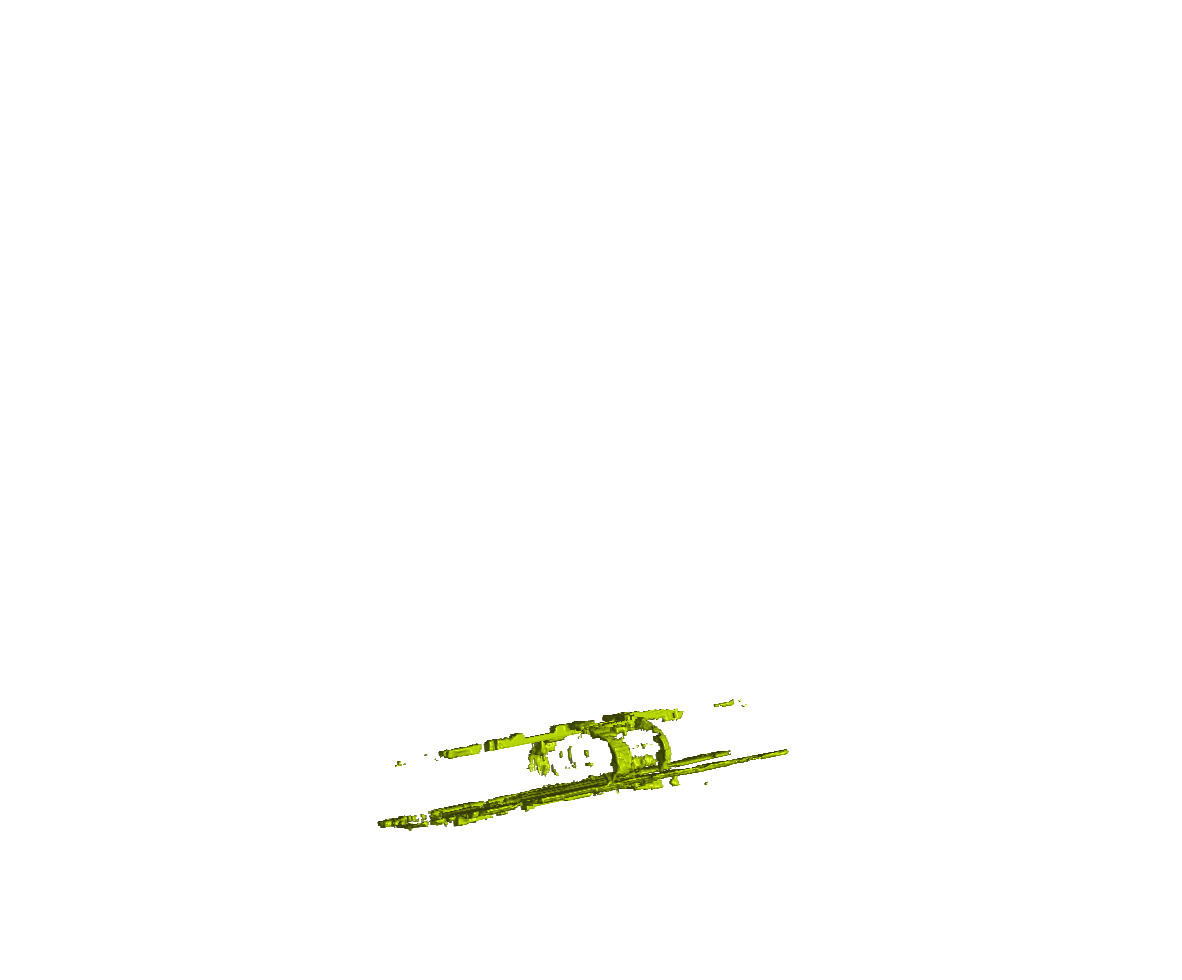} 
					& \includegraphics[width=\hsize]{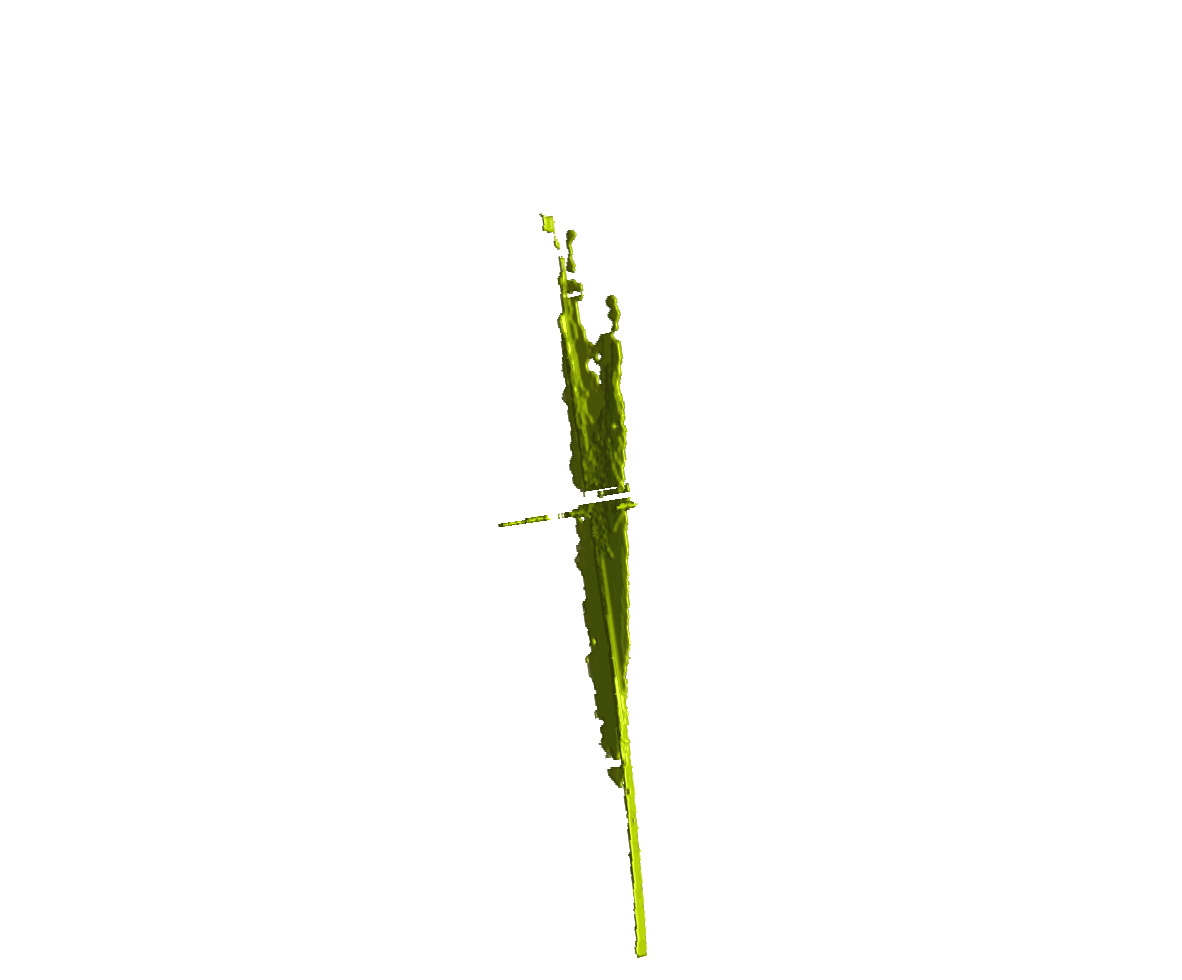} 
					\\ \addlinespace \hline \addlinespace
					
					\multirow{2}{*}{\rotatebox[origin=c]{90}{\fineTunedModel 1024\hspace{0.5em}}}
					\rotatebox[origin=c]{90}{pred}
					& \includegraphics[width=\hsize]{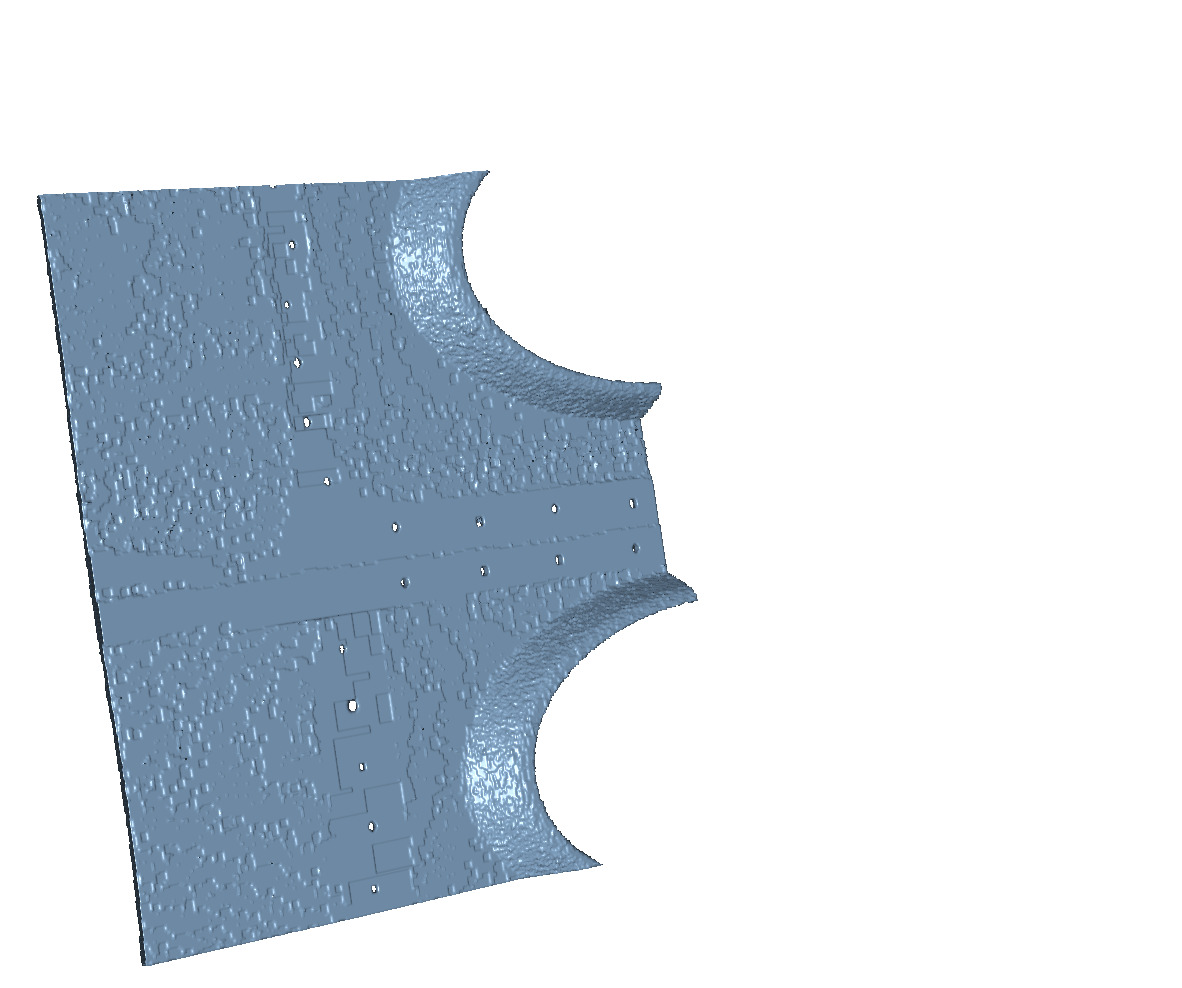} 
					& \includegraphics[width=\hsize]{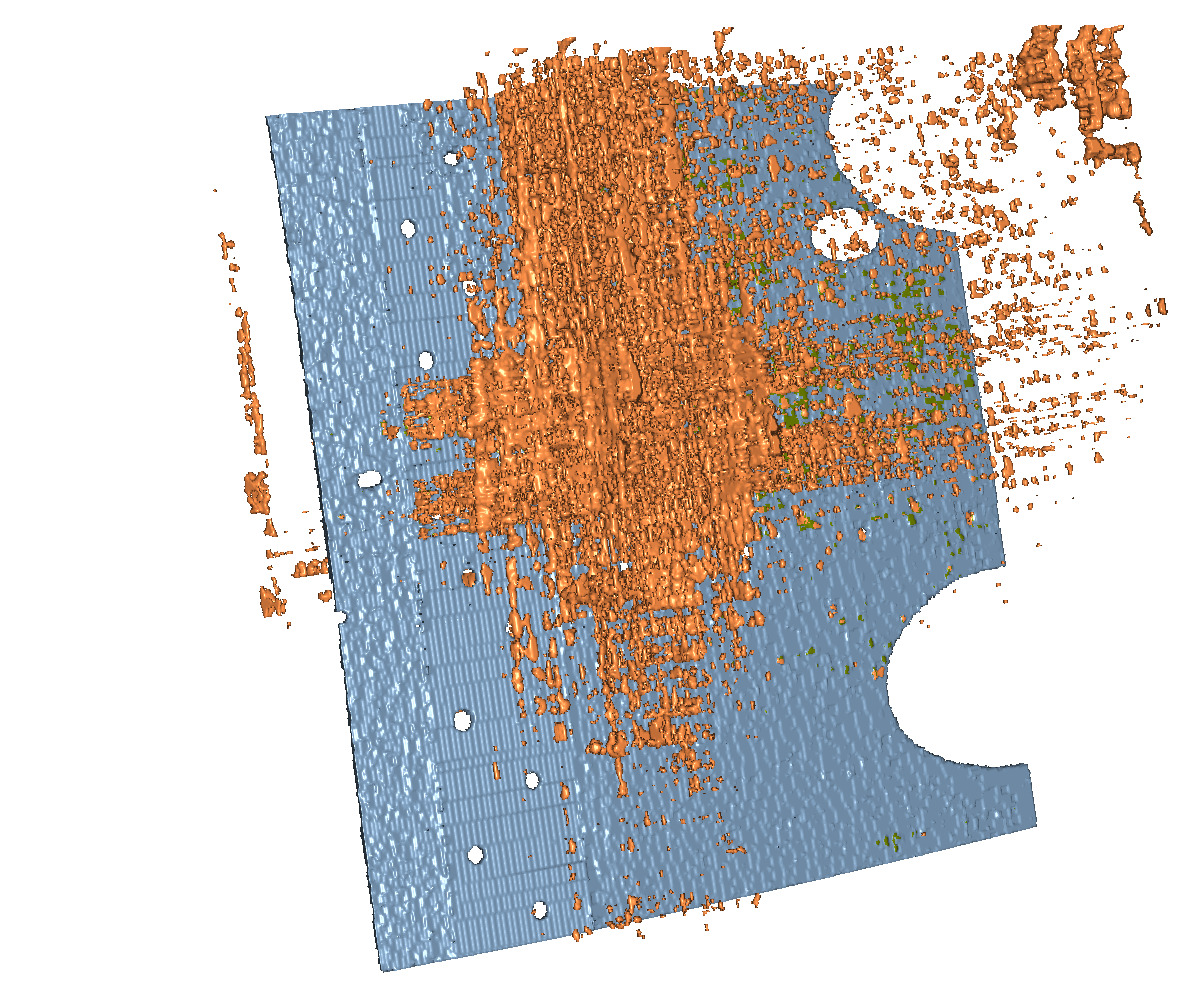} 
					& \includegraphics[width=\hsize]{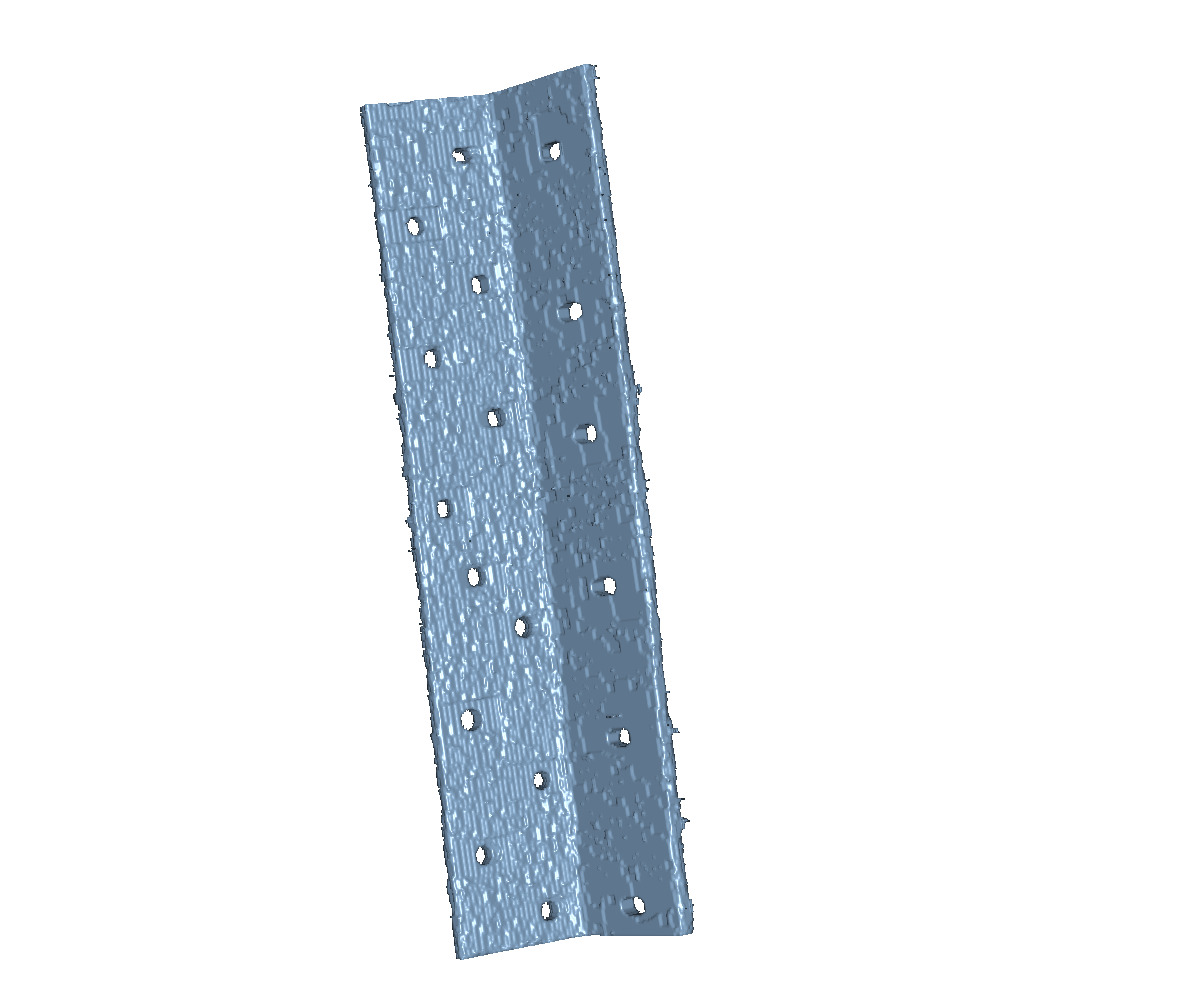} 
					& \includegraphics[width=\hsize]{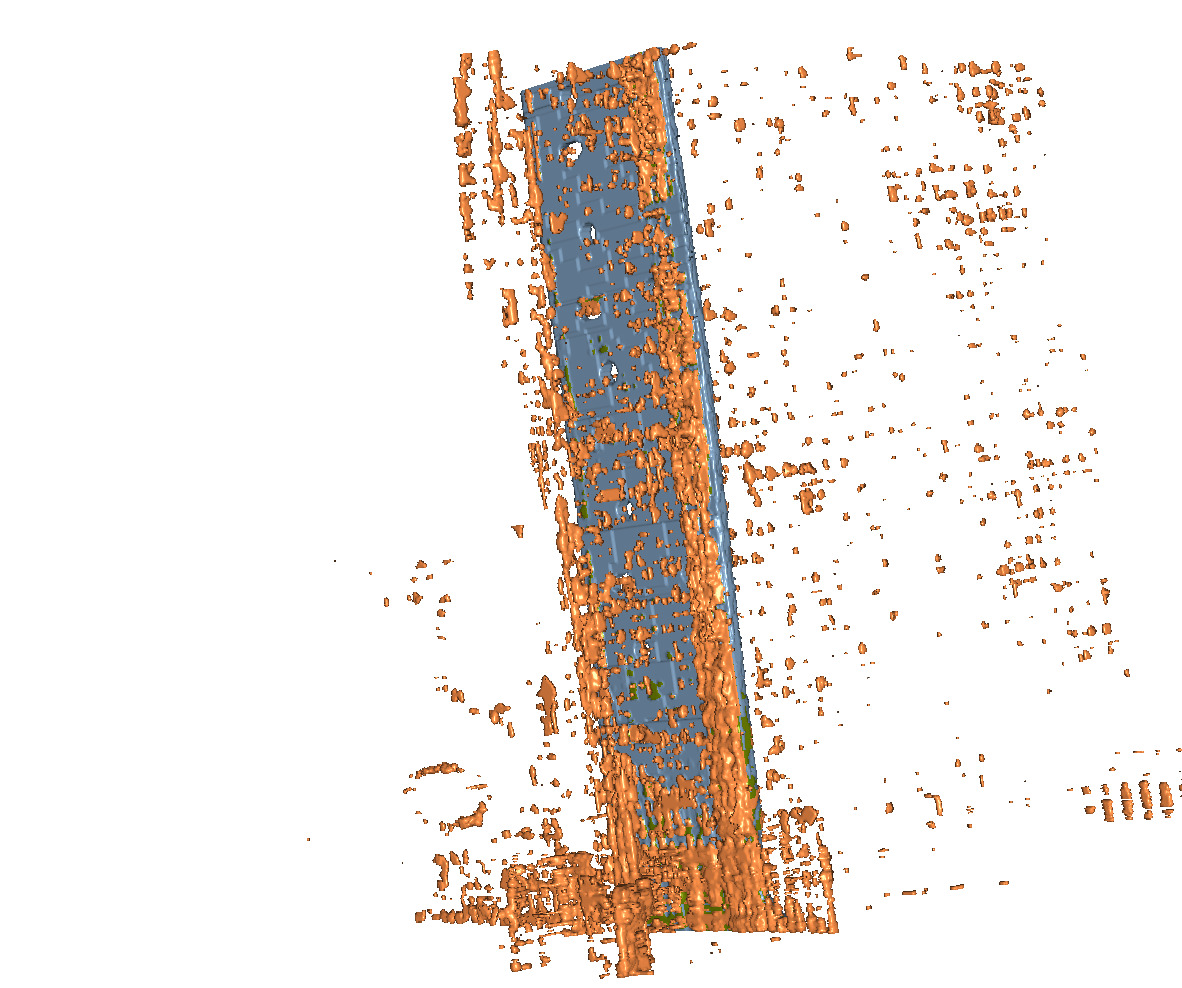} 
					& \includegraphics[width=\hsize]{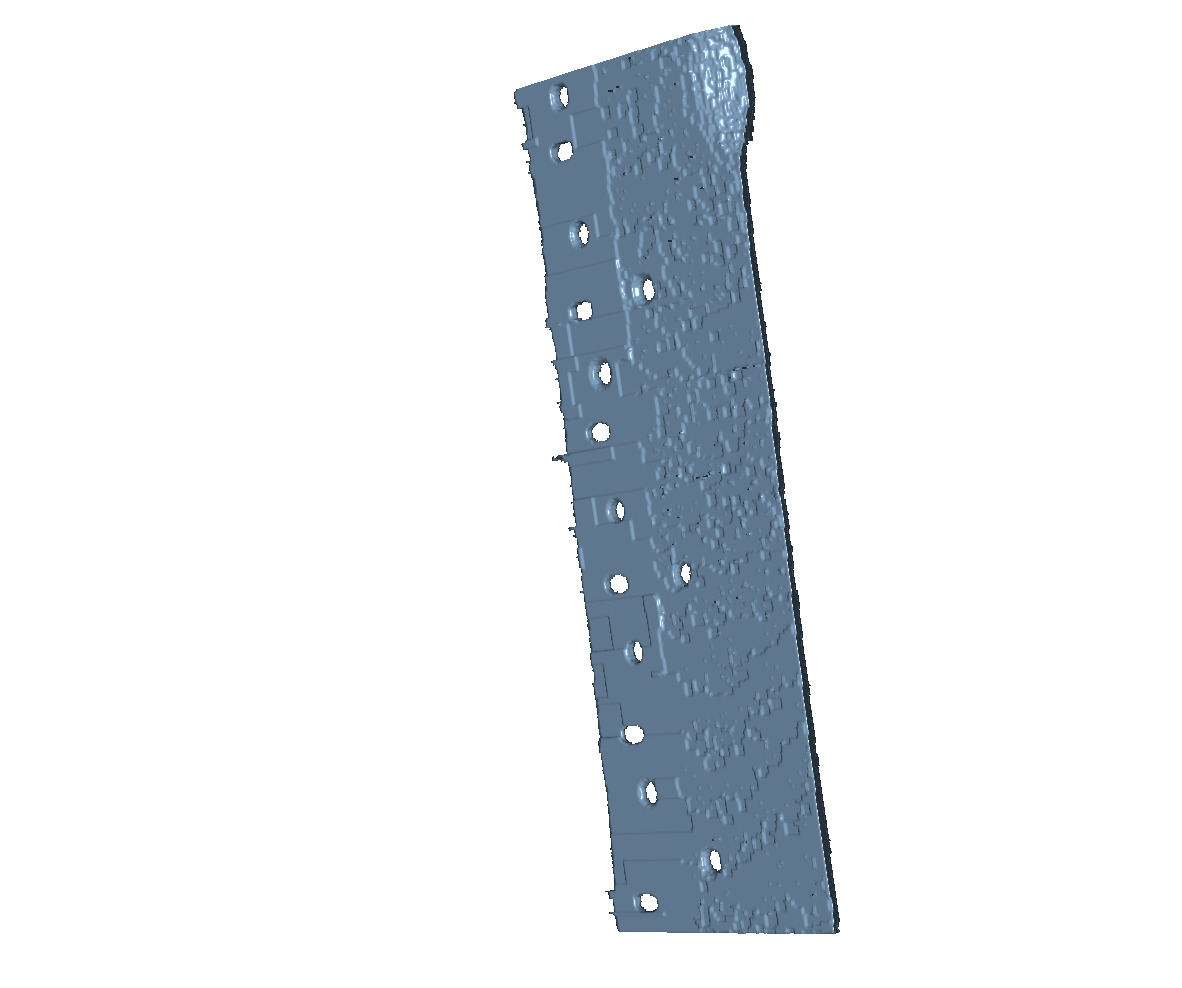} 
					& \includegraphics[width=\hsize]{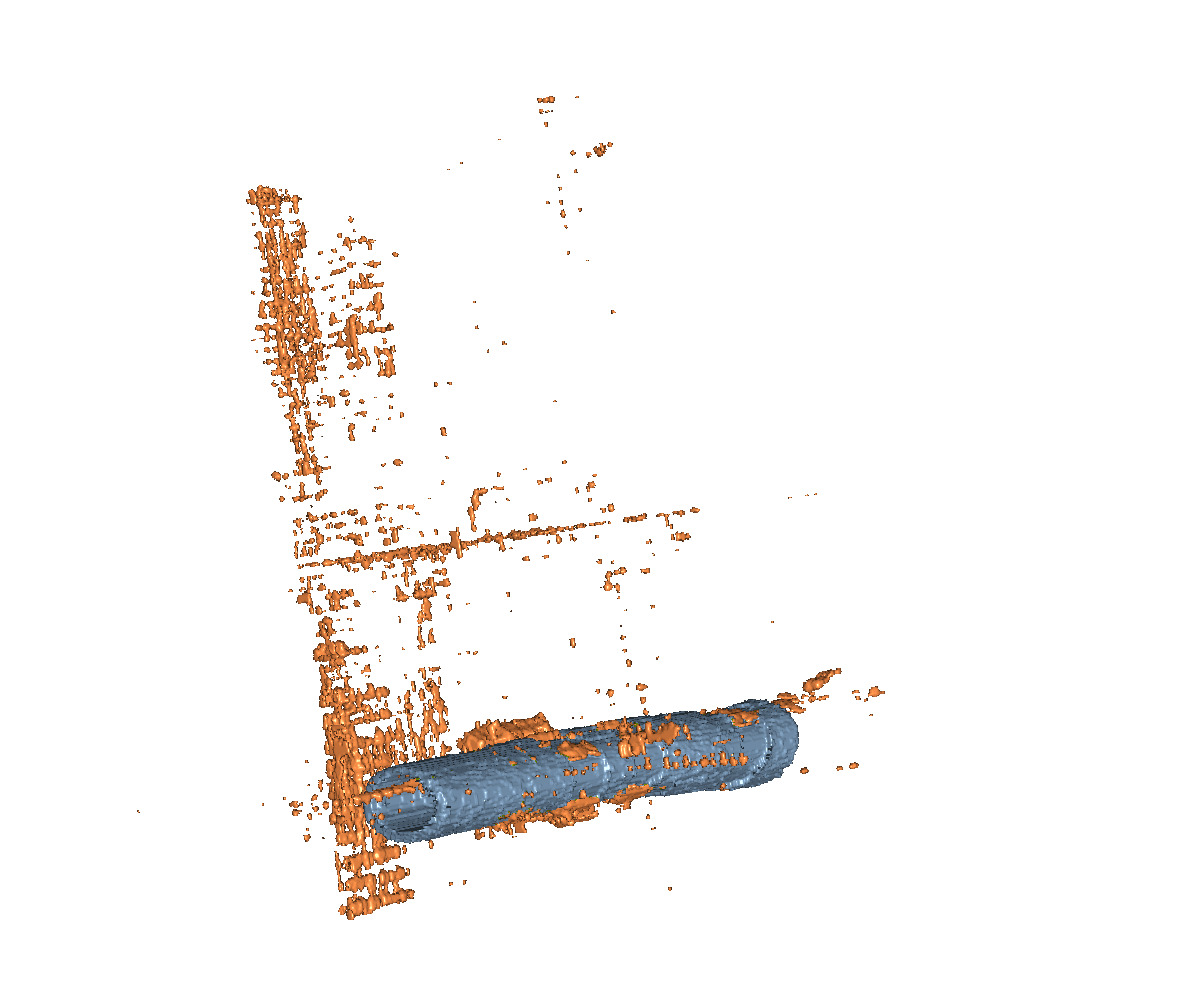} 
					& \includegraphics[width=\hsize]{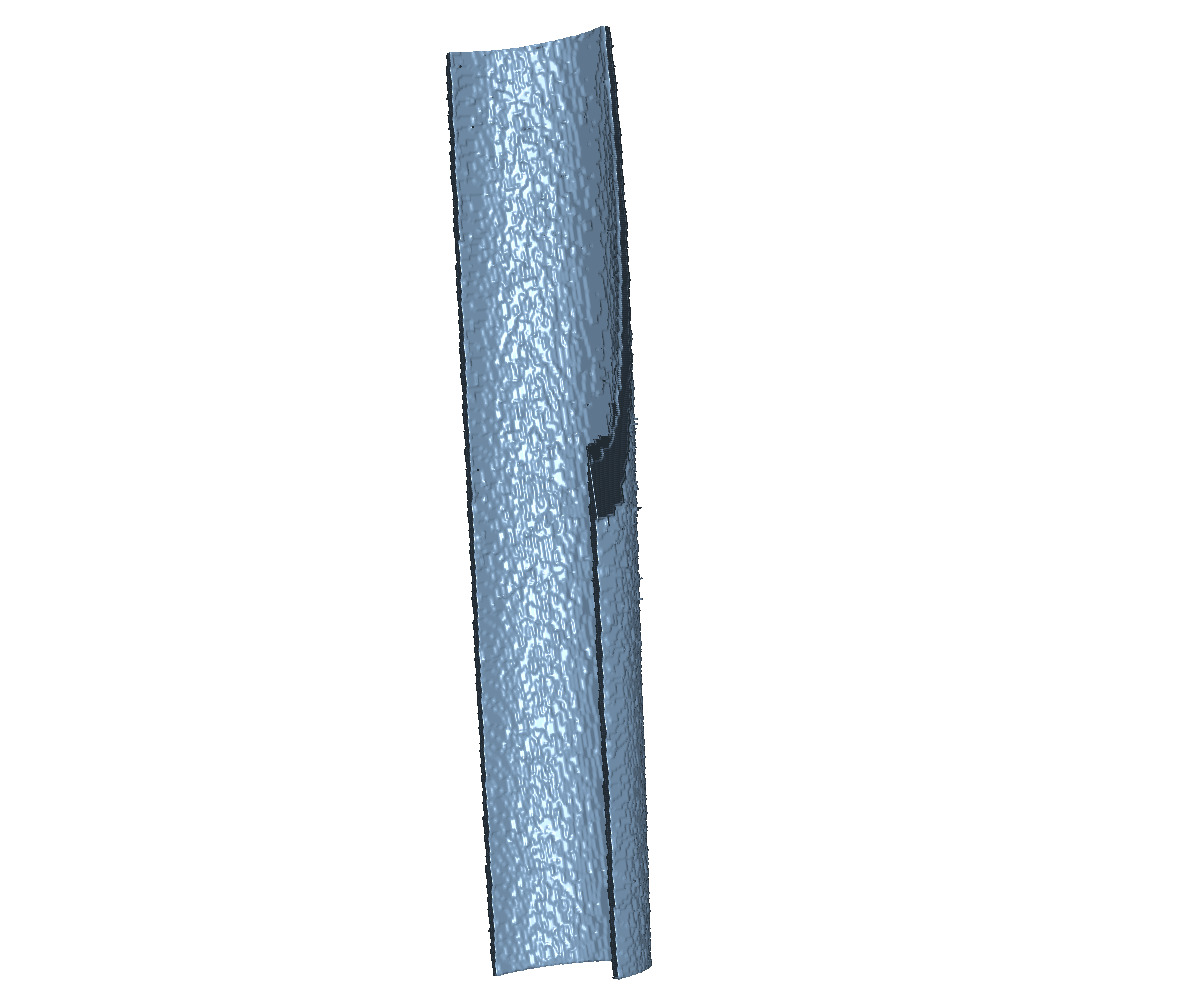} 
					\\ \addlinespace					
					\rotatebox[origin=c]{90}{TP}
					& \includegraphics[width=\hsize]{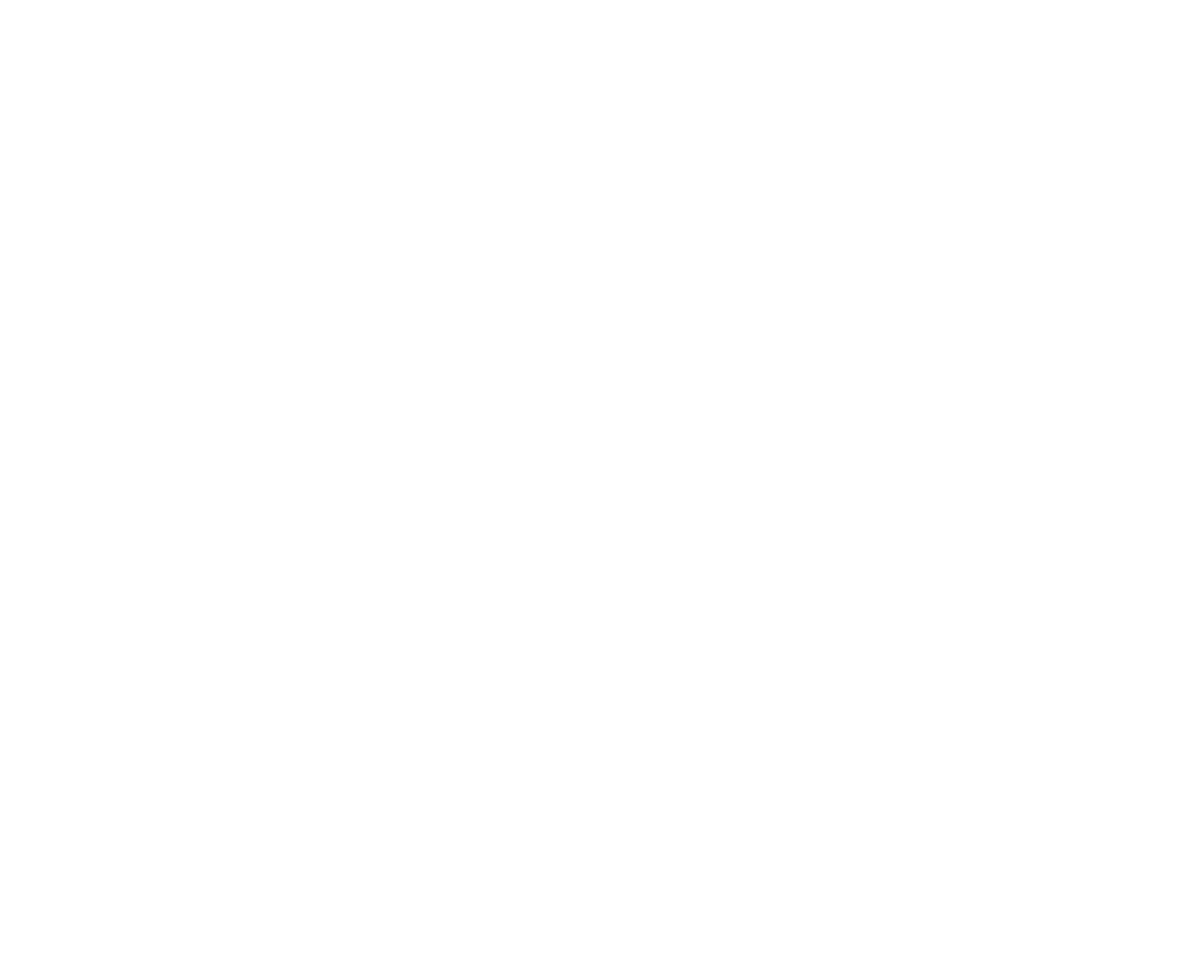} 
					& \includegraphics[width=\hsize]{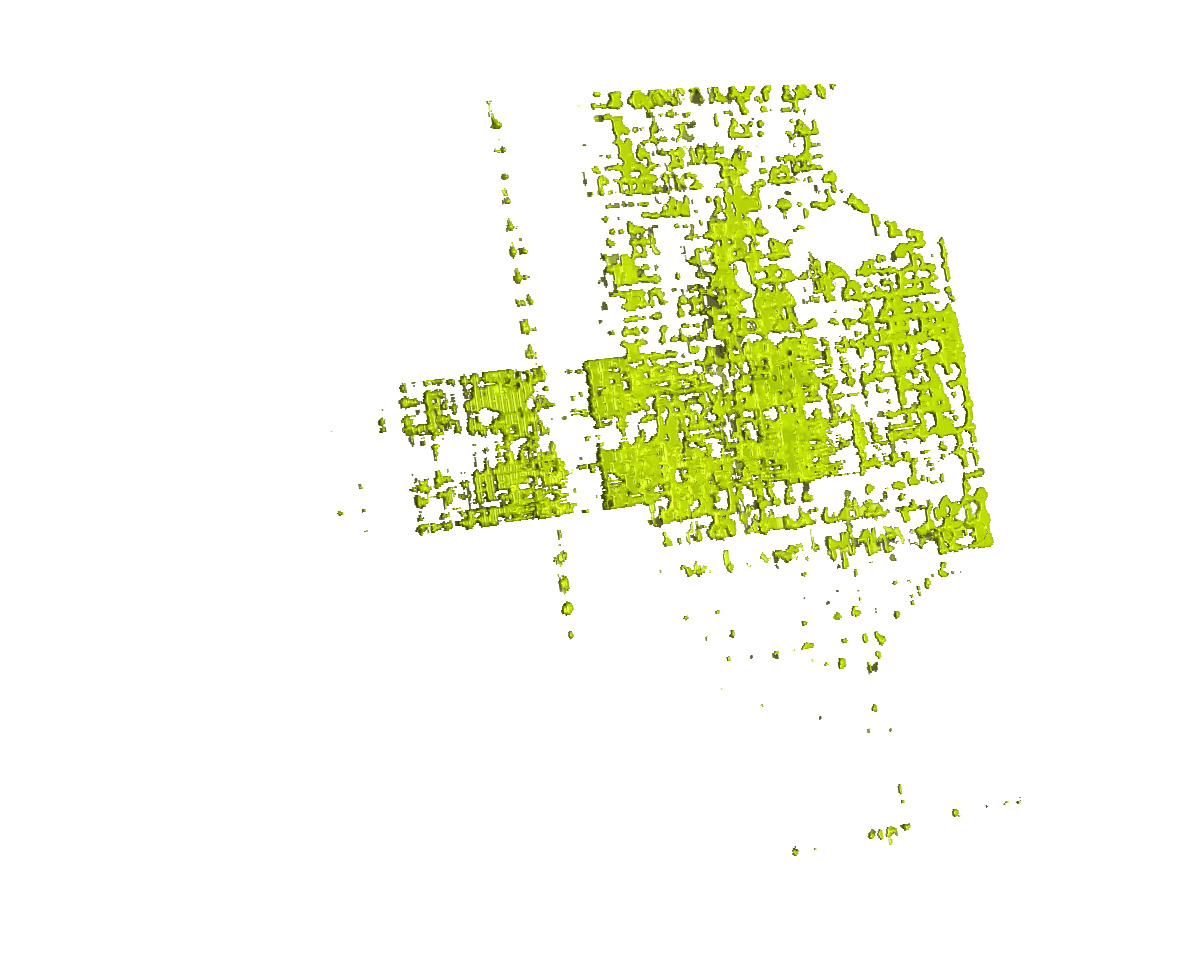} 
					& \includegraphics[width=\hsize]{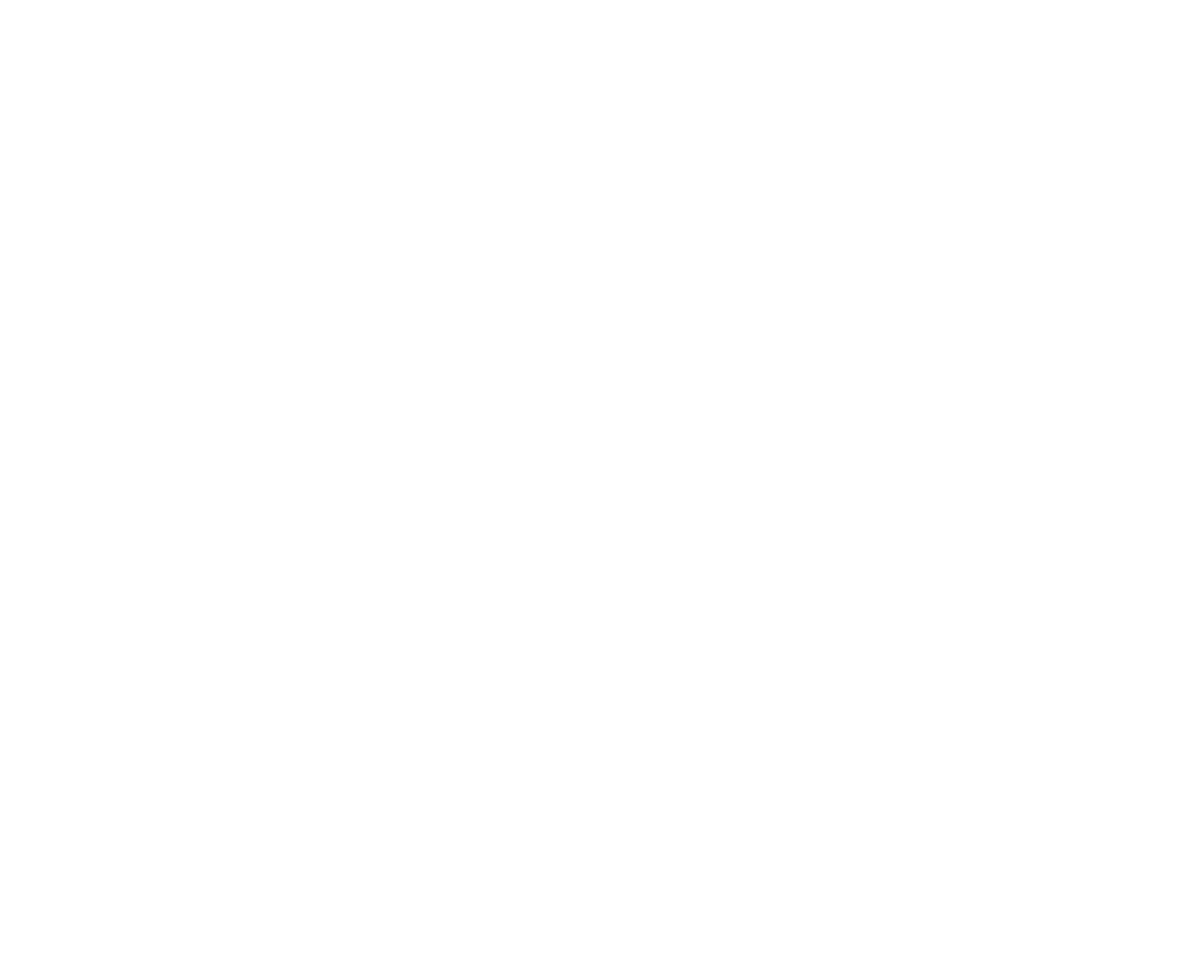} 
					& \includegraphics[width=\hsize]{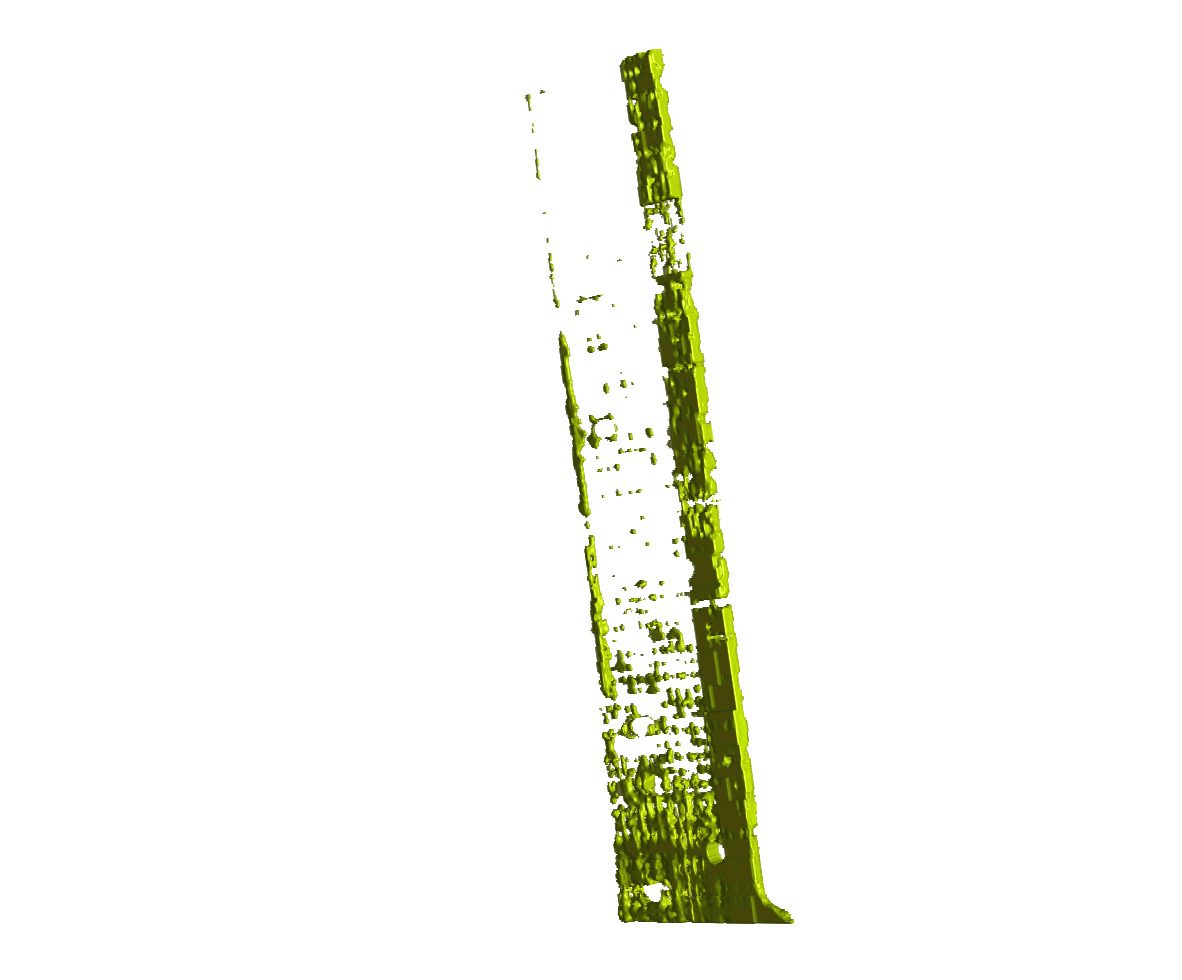} 
					& \includegraphics[width=\hsize]{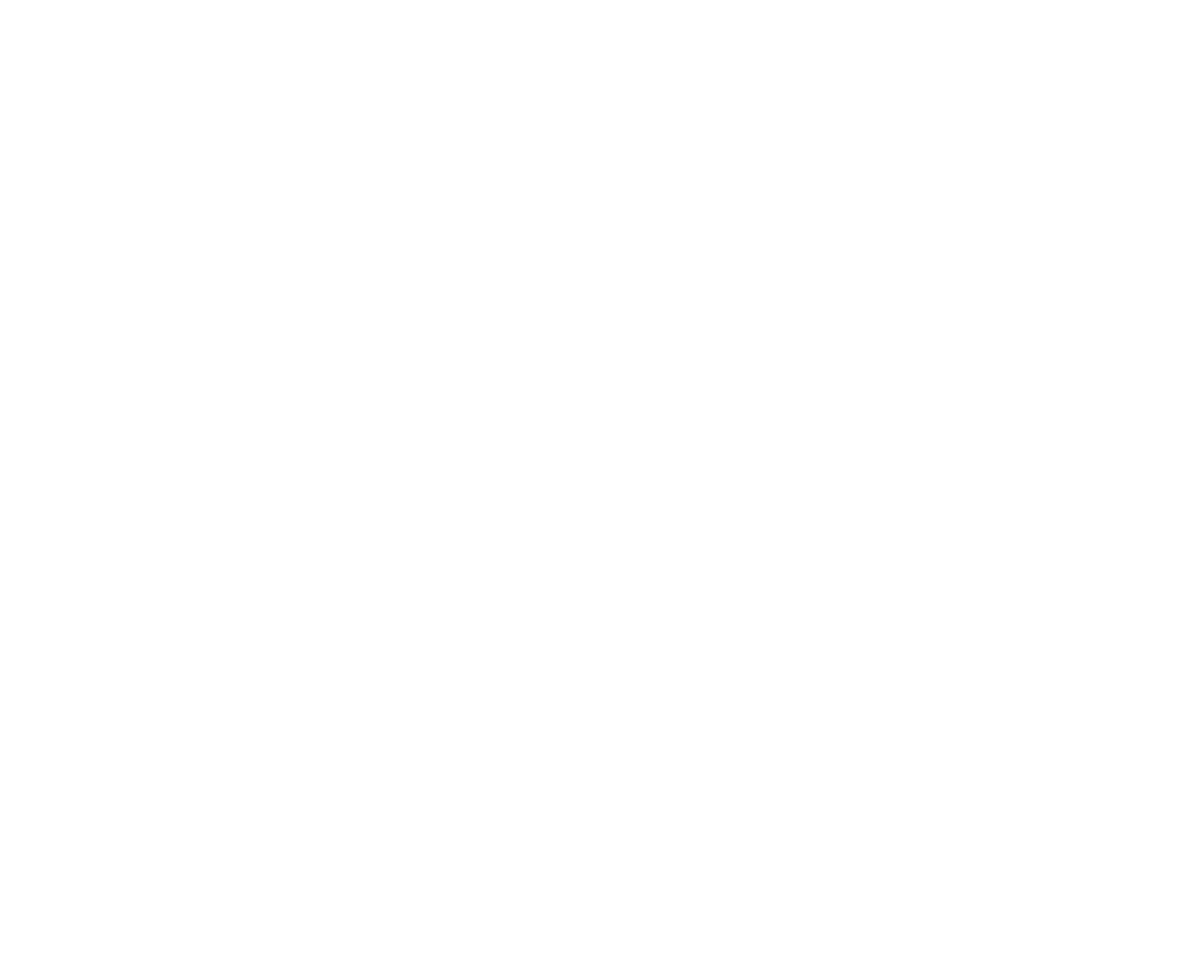} 
					& \includegraphics[width=\hsize]{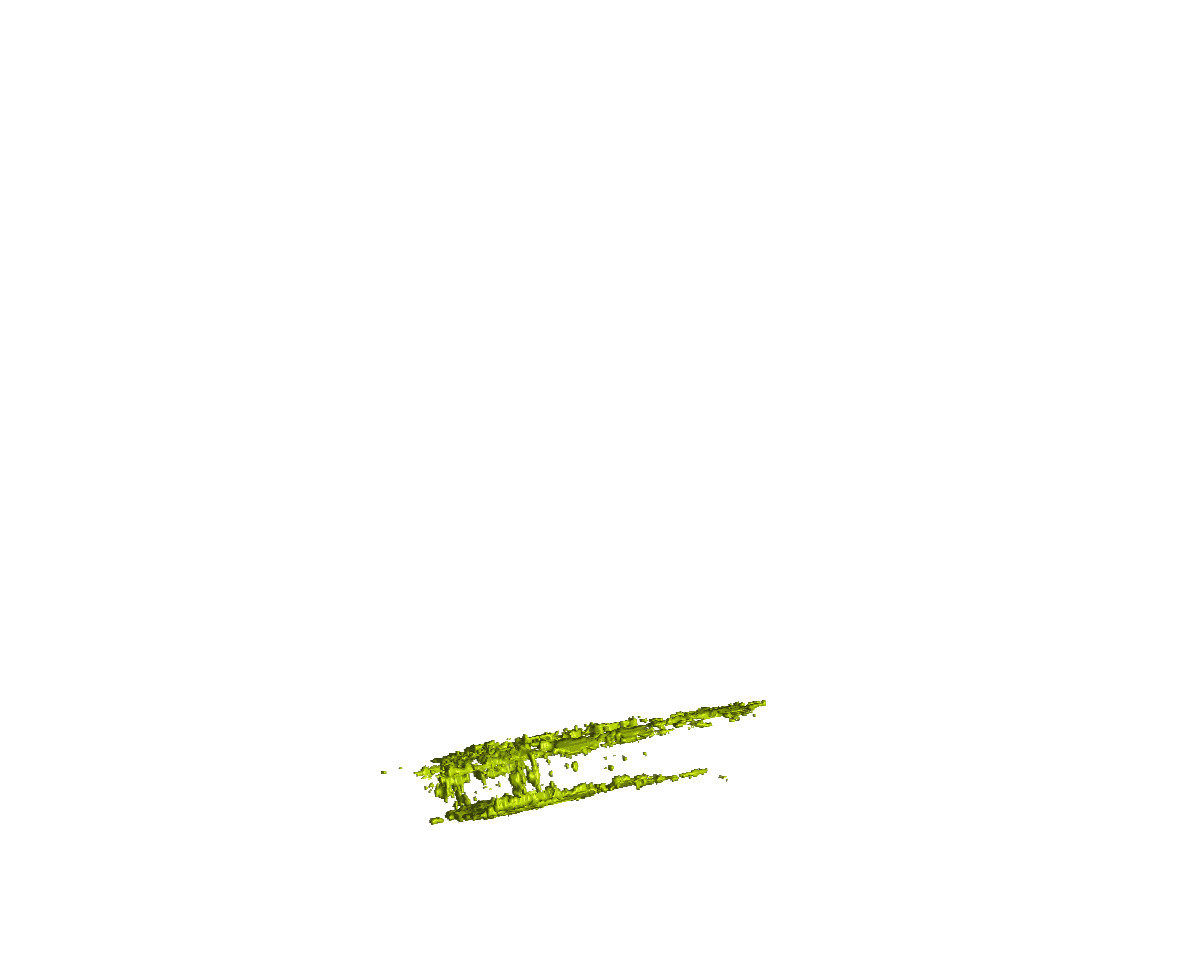} 
					& \includegraphics[width=\hsize]{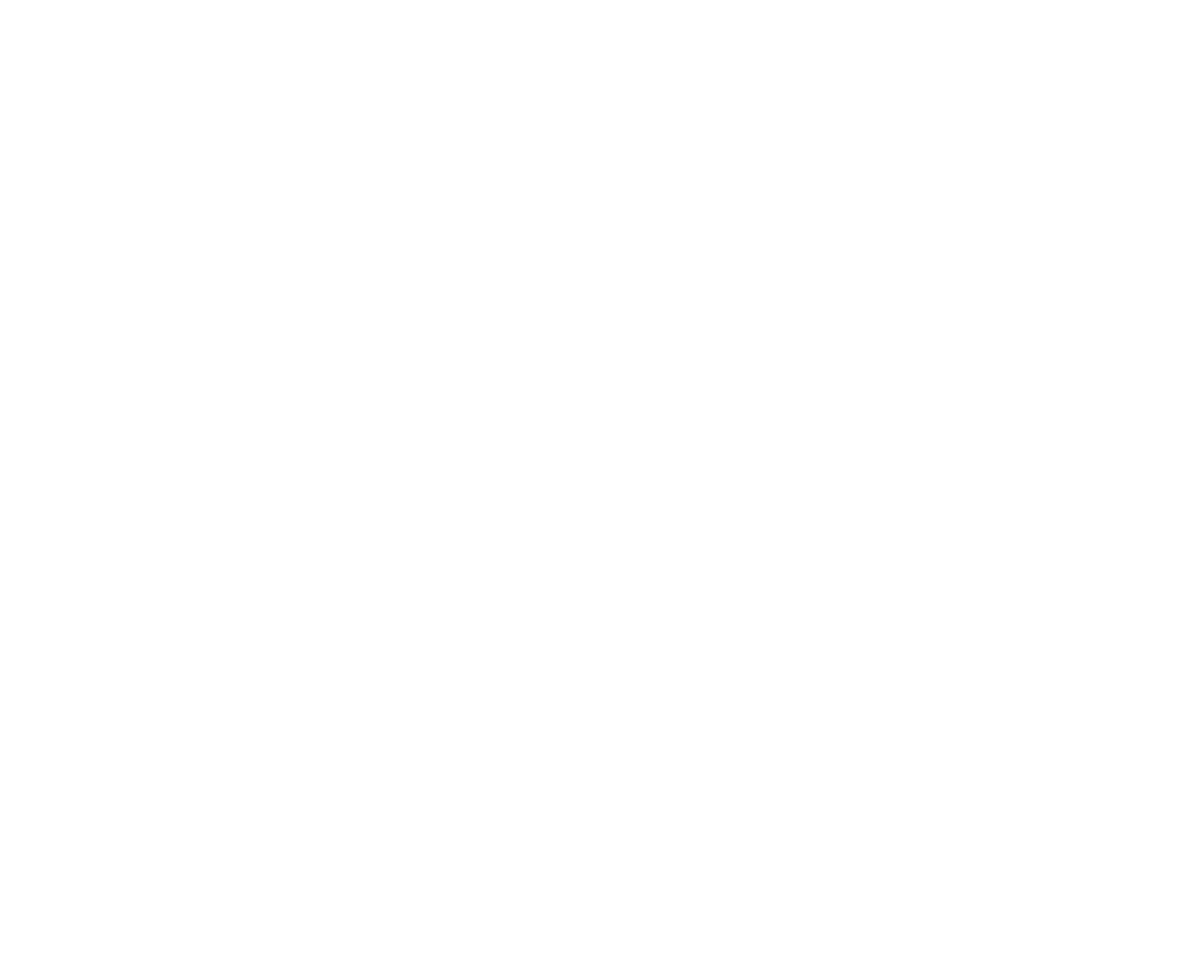} 
					\\ \addlinespace				
				\end{tabular}
				\caption{Renderings of the seven largest segments of the reference data-set and their corresponding predictions (pred) created with different snapshots of SAM and the volumetric algorithm. The colour coding is as follows: blue \tikzBoxFalseNegative\, reference segment, green \tikzBoxTruePositive\, true positives (TP), and orange \tikzBoxFalsePositive\, false positives.}
				\label{fig:result-me163-top}
			\end{figure*}
	
			Figure \ref{fig:result-me163-top} presents multiple renderings of the seven largest reference segments in the Me\,163 testing data-set, along with predictions generated by different SAM snapshots using the volumetric algorithm and fine-tuned parameters. The \emph{true positive} voxels are coloured green, while the eference segments are coloured blue and the \emph{false positive} voxels are coloured orange. It is evident that the volumetric segmentation of the data-sets using tiles of size $1024\times1024\times1024$\,voxels yields visually more appealing segments compared to using a tile size of $48\times48\times48$\,voxels.

\begin{figure}
				\centering			
				\begin{tabular}{@{} m{0.1\linewidth} @{\hspace{-5pt}} m{0.4\linewidth} m{0.4\linewidth} @{}}	
					& \makecell[c]{Reference} & \makecell[c]{Prediction}	\\		
					XY & \includegraphics[viewport=256 40 320 104, clip, width=\hsize, interpolate=false]{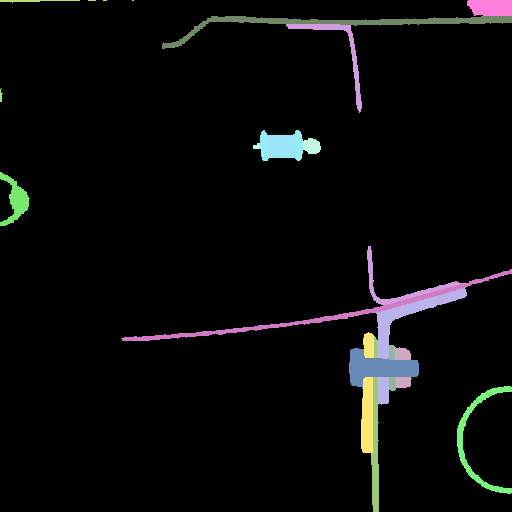}
					& \includegraphics[viewport=256 40 320 104, clip, width=\hsize, interpolate=false]{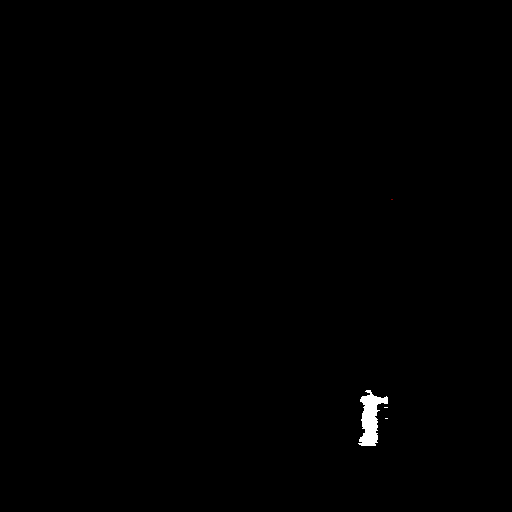}
					\\ \addlinespace					
					XZ & \includegraphics[viewport=256 264 320 328, clip, width=\hsize, interpolate=false]{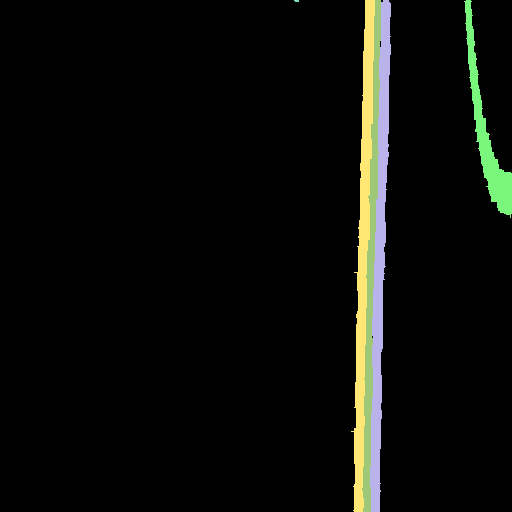}
					& \includegraphics[viewport=256 264 320 328, clip, width=\hsize, interpolate=false]{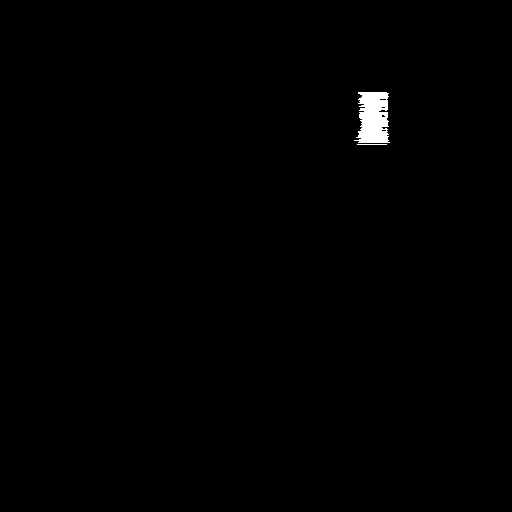}
				\end{tabular}
				
				\caption{\label{results-me163-top-undersegmentation} Slices obtained using the fine-tuned \fineTunedModel model and tile size of $48\times48\times48$ voxels. Due to under-segmentation the predicted segment erroneously intersects and merges multiple reference segments.}
			\end{figure}

			The predicted segmentation using the tile size of $48\times48\times48$\,voxels often appears \emph{empty}, as only a small count of voxels have been segmented correctly. This is because the segmentation quality of the algorithm is to poor to generate connected tiles, and so often only a limited amount of steps (see Section \ref{sec:methods-Tile-based-segmentation-for-datasets-of-arbitrary-size}) will be iterated for each segment. The segments are interrupted and only found in pieces. However, using a tile size of $48\times48\times48$\,voxels also often leads to under-segmentation, as demonstrated in Figure \ref{results-me163-top-undersegmentation}. Here, three adjacent segments were mistakenly connected by a single predicted segment.					

			But even the segmentation with a tile size of $1024\times1024\times1024$\,voxels is often insufficient, as both large-scale under-segmentations and over-segmentations occur, as can be seen from the correlation matrices in Figure \ref{fig:result-inference-volumetric-me163-mainDiagonal} and the cross-sectional images in Figure \ref{fig:results-volumetric-me163-slices-vit_b-tTF439-1024-mainDiagonal}.		
	
	\section{Discussion}		
		The transferability of the SAM model to instance segmentation of volumetric XXL-CT data-sets requires careful consideration. The presented results indicate that its two-dimensional image-based segmentation quality is insufficient for this specific problem domain. This limitation becomes particularly evident when dealing with the concatenation of numerous intertwined cross-sectional images in the volumetric case. The low contrast and high noise in these images pose challenges in accurately delineating individual segments. Additionally, using domain specific fine-tuning and improving slice-wise predictions did not yield substantial improvements for volumetric predictions.

		One potential source of error in the presented method might be the limited computational resources allocated for both fine-tuning and subsequent hyperparameter search. A more thorough optimization process could potentially improve the results. Furthermore, the availability of labelled training data-sets of sufficient quality in this problem domain was relatively limited for training the vision transformers included in SAM. Specifically, the absence of neighbouring voxels when adding the 512\,voxel wide border around the data-set for the Me\,163 data-set may have possibly contributed to a decrease in segmentation quality. 
			
		Additionally, considering improved algorithms for merging the slice-wise predictions could be an initial step in the further development process. Previous studies \cite{76-Xia2018, 77, 76} have demonstrated ample opportunities for the development of more sophisticated algorithms in this area. Implementing and embedding such algorithms into the processing pipeline has the potential to significantly enhance the segmentation quality.
	
	\section{Conclusion}
		The primary objective of this study was the exploration and possible applicability of the SAM algorithm for general image delineation to instance segmentation in XXL-CT volumetric data-sets. 
		
		In conclusion, our study highlights the potential of SAM for instance segmentation in XXL-CT volumetric data-sets, while acknowledging that there is still significant room for improvement. Furthermore, our research has contributed in the following areas: (1) the evaluation of SAM on volumetric NDT data-sets, (2) the exploration of various methods for integrating image-based SAM with volumetric data-sets, (3) the introduction of a tile-based approach for segmenting objects of arbitrary size, and (4) the utilization of dense prompts for tile combination using an accumulator. These contributions provide insights and establish a foundation for further advancements in this field.		
		
	\subsection*{Acknowledgement}
		This work was supported by the Bavarian Ministry of Economic Affairs, Regional Development and Energy through the Center for Analytics Data Applications (ADA-Center) within the framework of ,,BAYERN DIGITAL II`` (20-3410-2-9-8).	
	
	\section*{Author contributions statement} 
		\begin{description}
			\item [{RG}] – developed the algorithm and software, curated the data-set, carried out the experiments, conducted the analysis. He drafted and wrote the manuscript including the graphics.
			\item [{SR}] – provided the original training codebase and collaborated with RG to analyse the findings and provide guidance for subsequent experimental design and code modifications.
			\item [{TW}] - supervised the work, wrote the introduction and the setting, and did the final proofreading and editing.
		\end{description}
	
		All authors reviewed the manuscript.
	
	\section*{Competing interests} 
		The authors declare that they have no competing financial and/or non-financial interests in relation to the work described.			
	
	\bibliographystyle{plainurl}
	\bibliography{SAM}
\end{document}